# ON EFFECTIVE HUMAN-ROBOT INTERACTION BASED ON RECOGNITION AND ASSOCIATION

***DISSERTATION***

***Submitted by***

**Avinash Kumar Singh**

In Partial Fulfillment of the Requirements

For the Degree of

**DOCTOR OF PHILOSOPHY**

*Under the Supervision of*

**PROF. G.C. NANDI**

प्रज्ञानम् ब्रह्म

DEPARTMENT OF INFORMATION TECHNOLOGY

भारतीय सूचना प्रौद्योगिकी संस्थान

**INDIAN INSTITUTE OF INFORMATION TECHNOLOGY, ALLAHABAD**

***(A centre of excellence in IT, established by Ministry of HRD, Govt. of India)***

12 February 2016

# *CANDIDATE DECLARATION*

I, Avinash Kumar Singh**,** Roll No. **RS-110** certify that this thesis work entitled "***On Effective Human - Robot Interaction based on Recognition and Association"*** is submitted by me in partial fulfillment of the requirement of the Degree of **Doctor of Philosophy** in Department of **Information Technology**, **Indian Institute of Information Technology, Allahabad.**

I understand that plagiarism includes:

1. Reproducing someone else's work (fully or partially) or ideas and claiming it as one's own.
2. Reproducing someone else's work (Verbatim copying or paraphrasing) without crediting
3. Committing literary theft (copying some unique literary construct).

I have given due credit to the original authors/ sources through proper citation for all the words, ideas, diagrams, graphics, computer programs, experiments, results, websites, that are not my original contribution. I have used quotation marks to identify verbatim sentences and given credit to the original authors/sources. I affirm that no portion of my work is plagiarized. In the event of a complaint of plagiarism, I shall be fully responsible. I understand that my Supervisor may not be in a position to verify that this work is not plagiarized.

**Name: Avinash Kumar Singh** **Date: 12/02/16**

*Enrolment No: RS-110*

*Department of Information Technology, IIIT-Allahabad (U.P.)*

| | INDIAN INSTITUTE OF INFORMATION TECHNOLOGY ALLAHABAD (A Centre of Excellence in Information Technology Established by Govt. of India) |
|---|---|

# ***CERTIFICATE FROM SUPERVISOR***

I do hereby recommend that the thesis work done under my supervision by **Mr. Avinash Kumar Singh (RS 110)** entitled "***On Effective Human-Robot Interaction based on Recognition and Association***" be accepted in partial fulfillment of the requirement for the Degree of **Doctor of Philosophy** in Department of **Information Technology, Indian Institute of Information Technology, Allahabad.**

**Prof. G.C. Nandi**

**Supervisor, IIIT-Allahabad**

# Keywords

# Abstract

Robot cops or robot police is the new arena of robotics research. The agility in the vicinity of robotics, especially towards developing Social Robots, which can help police to identify criminals, diffuse bombs, rescuing hostages etc. has received more attention than any other. Therefore, in this thesis, we have devised four different techniques of criminal identification. Our hypothesis–"*it is possible to design a robust and secure criminal identification framework which can identify the criminal based on the visual perception of the eyewitness as well as by seeing them*". In order to establish the above hypothesis several challenges of perception acquisition, representation and modeling have been addressed while an anti-face spoofing system has been created to discard imposter attacks. NAO Humanoid Robot has been utilized as a testbed to test our hypothesis. In the process of establishment of our hypothesis, the first contribution *(chapter 3)* processes eyewitness imprecise knowledge to trace out the criminals. An eyewitness's visual perception about the criminal is first captured with the help of a dialogue based Humanoid Robot framework. The captured inferences are vague hence a linear transformation is applied to quantize this information. The quantized information is then modeled with the help of our proposed decision tree (query based) and rough set based approaches. Two case studies, viewed and forensic sketches have been considered to evaluate the strength of our proposed approach. A symbolic database has been created to perform a stringent analysis on 240 public faces (150 public Bollywood celebrities and Indian cricketers, 90 local faces (our dataset)) in case of query based approach, while the benchmark mugshot criminal database of 300 persons from Chinese University of Hong Kong, has been utilized to evaluate the rough set based modeling.

The building block of the human perception based criminal identification is laid down on the assumption that criminals must be enrolled in the system. The approach fails to produce the wanted results when the criminal information is not registered in the system (in case of international criminal or she/he is committing crime first time). Towards the solution of this problem, in *Chapter 4*, we have trained humanoid robot to perform sketch drawing of the criminals, based on the given perception about criminals. In order to mimic the drawing, two primitive challenges such as calibrating humanoid camera plane with respect to its body coordinate system and finding an inverse kinematic based solution, need to be addressed. The former problem has been solved by providing the three different mechanisms (a) 4 Point, (b) Fundamental Matrix and (c) Artificial neural network based Regression Analysis while a gradient descendent based inverse kinematic solution is utilized for the NAO Right Hand. The first two contributions empower NAO to identify and synthesis the criminal face based on eyewitness visual perception about the criminal.

In the third contribution of the thesis (*Chapter 5*), we have designed two face recognition algorithms based on the principle of facial symmetry and component based face recognition to identify criminals seeing the face images. The facial symmetry based face recognition utilizes either part of the face (left or right) to recognize the criminal. A Principal Component Analysis based experiment performed on the AT&T face database confirms that half faces could produce almost equal accuracy as the full faces, hence can be utilized for criminal identification. This reduces the time complexity as well as helps in partial information based criminal identification. Another study on 345 public faces confirms that 80% of the face discriminant features lie only within the eyes, nose and mouth region. The proposed component based face recognition localizes the face over these components and extracts the Scale Invariant Features Transform (SIFT) features from these regions. Later the classification is performed on the basis of weightage majority voting over these extracted features. The illumination effect is almost homogeneous on the face narrower regions, therefore the proposed approach can handle the variable illumination challenges as well recognize the criminal face covered with scarf or goggles. The component based face recognition can also solve the real time challenge like change in facial expression.

At last, in *Chapter 6*, we deployed the liveness detection mechanism over our face recognition system to detect and prevent face spoofing attacks. Face spoofing is an attack against a face recognition system to bypass the authentication system and to get unauthorized access to the system. The robot is an autonomous and re-programmable device therefore, its security must be assured. Liveness is a test to measure the life of the presented biometric in order to discard spoof attacks. Movement in facial macro features such as facial expressions and face 3D structure have been utilized to design the liveness test. The facial movement based face spoofing detection is based on the challenge and response mechanism while to evaluate the 3D structure, NAO stereo-vision has been exploited.

The human perception based criminal identification allows humanoid robots to identify criminals using the rough idea about the criminal physiological as well as facial descriptions. The sketch drawing capability empowers humanoids to synthesis the criminal face using the same facial description. The facial symmetry and component based face recognition facilitates them to recognize criminals even her/his face is occluded with scarf/googles or any other object. The addition of face liveness detection on top of face recognition secures the overall system to prevent from unauthorized access. The proposed framework has been verified theoretically as well as experimentally towards their scalability, universality and reliability.

# Table of Contents

# List of Tables

# List of Figures

# List of Abbreviations

| | |
|---|---|
| HRI | Human-Robot Interaction |
| WMV | Weighted Majority Voting |
| ASD | Autism Spectrum Disorder |
| RALL | Robotics Assisted Language Learning |
| CMSM | Constrained Mutual Subspace Method |
| SIFT | Scale Invariant Feature transform |
| PCA | Principal Component Analysis |
| LDA | Linear Discriminant Analysis |
| ICA | Independent Component Analysis |
| LBP | Local Binary Pattern |
| SURF | Speedup Robust Feature |
| EBGM | Elestic Bunch Graph Matching |
| AAM | Active Appearance Model |
| SW | Within Class Scatter matrix |
| SB | Between Class Scatter Matrixes |
| RCPR | Robust Cascaded Pose Regression |
| HOG | Histogram of Oriented Gradient |
| FDD | Frequency Dynamic Descriptor |
| FRR | False Rejection Rate |
| GMM | Gaussian Mixture Model |
| ANN | Artificial Neural Network |
| PID | Proportional Integral Derivative |
| MSE | Mean Square Error |
| ROI | Region of Interest |
| SVM | Support Vector Machine |
| DoG | Difference of Gaussian |

# Dissertation Publications

## *Journal Publications*

j1. Avinash Kumar Singh and G C Nandi, "Visual Perception based Criminal Identification- A Query based Approach", published in "Taylor & Francis International Journal of Experimental & Theoretical Artificial Intelligence", pp. 1-22. [**Expanded-SCI, Impact Factor: 1.703].**

j2. Avinash Kumar Singh and G C Nandi, "A Rough Set based Reasoning Approach for Criminal Identification", accepted for the publication in "Springer, International Journal of Machine Learning and Cybernetics". [**Expanded-SCI, Impact Factor: 1.110]**.

j3. Avinash Kumar Singh and G C Nandi**,** "NAO Humanoid Robot: Analysis of Calibration Techniques for Robot Sketch Drawing", published in "Elsevier International Journal of Robotics and Autonomous Systems", Vol. 79, pp. 108-121. [**Expanded-SCI, Impact Factor: 1.618].**

j4. Avinash Kumar Singh, Piyush Joshi and G C Nandi, "Face Liveliness Detection through Face Structure Analysis", published in Inderscience "International Journal of Applied Pattern Recognition", Vol.1, No.4, pp.338 – 360, [**Emerging-SCI**].

j5. Avinash Kumar Singh, G C Nandi, "Development of a Self-Reliant Humanoid Robot for Sketch Drawing", minor revision submitted in "Multimedia Tools and Applications", [**Expanded-SCI, Impact Factor: 1.331].**

# *Conference Publications*

c1. Avinash Kumar Singh, Neha Baranwal and G C Nandi, “Human Perception based Criminal Identification through Human Robot Interaction”, published in IEEE, Eighth International Conference on Contemporary Computing (IC3), 2015, pp.196-201, 20-22 August 2015.

c2. Avinash Kumar Singh, Pavan Chakraborty and G C Nandi, “Sketch drawing by Nao Humanoid Robot”, published in 35th IEEE International Conference TENCON 2015, pp.1-6, 1-4 Nov 2015.

c3. Avinash Kumar Singh and G.C Nandi, “Face recognition using facial symmetry”, published in 2nd International Conference on Computational Science, Engineering and Information Technology (CCSEIT-2012), Coimbatore, 2012, pp. 550-554, October, 2012.

c4. Avinash Kumar Singh, Arun Kumar, G C Nandi and Pavan Chakraborty, “Expression Invariant Fragmented Face Recognition”, published in IEEE, International Conference on Signal Propagation and Computer Technology (ICSPCT-2014), pp. 184-189 July 2014.

c5. Avinash Kumar Singh, Piyush Joshi, G C Nandi, “Face Recognition with Liveness Detection using Eye and Mouth Movement”, published in IEEE, International Conference on Signal Propagation and Computer Technology (ICSPCT-2014), pp. 592-597 July 2014.

c6. Avinash Kumar Singh, Piyush Joshi and G C Nandi, “Development of a Fuzzy Expert System based Liveliness Detection Scheme for Biometric Authentication” published in Elsevier, 7th International Conference on Image and Signal Processing (ICISP-2013), Vol. 4, pp. 96-103, October 2013.

# Acknowledgments

“**And, when you want something, all the universe conspires in helping you to achieve it**” **— Paulo Coelho**, **The Alchemist**

These words seem to be perfectly suitable on two august persons, without them this day would not be possible for me. The first person is my grandfather, who is a doctor by profession. He had a dream since my childhood that one day I would serve as a doctor. He inspired me to enroll in the Ph.D. degree and to pursue my research ability. Five years before, he harvested a seed, my supervisor, **Prof. G C Nandi** later irrigate that seed to become a plant and here I am. This thesis is the result of their utmost support, guidance, motivation, care and love.

First and foremost, I wish to thank my supervisor, **Prof. G C Nandi** to accept me as his research scholar. I felt myself very fortunate to be associated with his Robotics and Artificial Intelligence Laboratory, which I can say is the best place to nurture your ideas. In my supervisor’s words, “Research is not about finding the width, it’s about finding the depth”. His words too reflects in his guidance. His suggestions and insights went a long way to help me deal with numerous complexities of my research work. My skills as a researcher have been well trained by him.

I owe my most sincere gratitude to **Dr. Pavan Chakraborty**, Head of Ph.D. Cell, IIITA for his invaluable knowledge, he imparted to me and for teaching me how to present research work. His care and support felt me like home at IIITA. My special thanks to, **Dr. Rahul Kala**, and **Dr. Shashank Srivastava** Assistant Professor, IIIT-A and MNNIT-A for helping me in the proof reading of the thesis.

Rejection (rejection of research paper) is a phase which has to be faced by every research scholar. I am indebted to **Neha Baranwal**, who consoled me during the tough time of Ph.D. Her positive attitude towards looking the problem helped me to focus on my pitfalls. I would like to extend my thanks to my special friends **Dr. Anup Nandy**, **Seema Mishra**,

**Rajesh Doriya**, **Manisha Tiwari** for providing an enjoyable working environment in our robotics lab. I am grateful for their assistance, discussions and criticisms of my research.

My warmest thanks go to my Ph.D. batch mates **Bharat Singh**, **Himanshu**, **Ganjendra**, **Preetish Ranjan,** my seniors **Aditya Saxena**, **Shivangi Raman**, juniors, **Vijay Bhaskar, Manish Raj, Venkat Beri, Vishal Singh, Saurabh**. Their presence made this journey joyful to me. I am also profoundly grateful to **Piyush Joshi**, **Arun Tyagi**, **Akshay Singhai**, **Gopal, Ankit**, **Soumarko Mondal** for assisting me in my dissertation work.

Lastly and most importantly, I express my deep sense of reverence and gratitude to my parents **Mrs. Vibha Singh** and **Mr. L. P. Singh**, especially to my grandfather **Dr. Vashudev Joshi**, who made me what I am today. This life is the result of their blessings, unconditional love and sacrifices.  I, extend my heartiest thanks to two God gifts, my brother **Ankit** and sister **Anamika** for their immense love and care.

**Avinash Kumar Singh**

**12 February 2016**

# Chapter 1:

# Introduction

"*This chapter discusses the heritage of social robots, mediums of communication possible between the robot and human, and the application areas where these social robots have been applied to serve the humanity. The chapter also describes the Thesis objective, together with the problems addressed. Research contributions for each problem with their limitations are also summarized to give a brief idea about the work*".

## 1.1 Overview of Human-Robot Interaction

After years of existence only in science fiction movies, novels, and books now social robots are coming in practice in our day to day life [1-3]. They are employed in various fields, and are doing magnificent work like teaching social behavior to children [4-5], helping stroke patients for their physical rehabilitation exercises [6-8], in military [9-11], in rescue work [12-13], and in many more services. Human physiological and behavioral identities like gesture, posture, speech, and face are necessary for the efficient and safe interaction. It allows the robot to understand about the need of the users, and to generate an appropriate response. Human-Robot Interaction (HRI) [14-16] is the study of interactions (how people respond to the robot and vice-versa) between the human and robots. The interaction between the human and robot could be established using various tools [15-16] such as: 1. Giving commands from keyboard and mouse, 2. Through Vision based approach using Camera, 3. Through speech 4. Behavior-based approach, 5. Gesture and 6. Multimodal (combination of aforesaid tools). The vision-based approach is suitable for many applications, but it suffers from many computer vision and machine vision challenges. The speech based system has also many difficulties, including the stochastic nature of the human speech, noise presence, and complex method to recognize human speech and deploy it for HRI. The behavior-based approach is suitable for simple robots, but the designing of such robots is restricted to certain applications only. The Multimodal approach uses different approaches altogether so that all the issues are taken care of. All these tools need an appropriate learning mechanism for a robot and for controlling it. Robots need accurate commands when it works in the real time domain. It utilizes human-computer interaction methods to learn the task, control its behavior and perform the action. Seeing the limitation of the above said fields Human-Robot interaction has ample space for research.

Technology makes our life easier. Development of science opened the new doors for future. It has also changed the way a robot looks. Now robots are social and autonomous such as ASIMO [17], NAO [18], icub [19] and Actroid [20], which are ready to share the same workspace with a human. Rescue robots (Helios IX) [21] are primarily designed for the purpose of helping rescue workers and humans in danger. These robots [9-13] can be used in mining, urban disasters, hostage situations, and explosions. It can navigate itself in dangerous environments to look for surviving humans and also have the capability to neutralize any immediate danger and make the surrounding less perilous for human rescue workers. Service robots [6-7] are generally designed for providing assistance to humans, particularly to the people having disabilities. They serve by performing several jobs that are dirty, dull, distant, dangerous or repetitive, including household chores. Roomba [22] is a series of autonomous robotic vacuum cleaner. Under normal operating conditions, it is able to navigate living spaces and common obstacles, while vacuuming the floor. Military robots like BigDog[23] are the autonomous or remotely controlled robots which are specially designed to be used in the war field. They could be used for supplying the ammunitions and food. The main advantage of using them are: they don't feel tired, don't close their eyes, don't get ill and above all don't let any family to lose their loving ones. Entertainment and domestic are the other category of robots which are designed to entertain us and also to take care of old age people [24-26]. After a certain age, we need to have the care and company of our family and relatives. As the life is moving very fast and in this fast moving life we do not have time for our senior citizens. These robots are very helpful in assisting them and to look after them. Medical robotics added a milestone in the medical surgery field. Robots are performing surgery with a high precision and accuracy [27-28].

Seeing all these application areas and functionalities of robots, in this thesis we have added three new functionalities in humanoid robot which enable them to assist Police to identify criminals as well as helpful to recognize humans.

- The visual perception based criminal identification helps police to trace out the suspect. An eyewitness can communicate through speech with these robots; robot asks several questions about the physiological as well as facial structure of the suspect. Based on the given vague perception of eyewitness the robot searches for her/his existing criminal records inside the criminal database.
- We have trained these humanoid robots for creating criminal sketches using the perception captured through dialogue with them.
- The addition of face recognition capability on these humanoid robots facilitates them to identify any person/criminal just by looking at her/his face. It can identify one even by only seeing the half face or only using their eye, nose and mouth component. This increases the robustness of recognition as well as speed up the recognition process. On the other hand anti-face spoofing mechanisms help to achieve the reliability and ensure the robustness of recognition.

## 1.2 Motivation

The primary aim of this thesis is to add up functionalities inside humanoid robots so that they can interact with humans and help them to solve various real time issues and challenges. We have kept the domain of application as criminal identification [29]. Criminal Identification is the special case of person identification, having its own set of challenges. As the problem of criminal identification is divided into two sub problems each problem has its own motivational story.

The manual process of identifying the suspect in the criminal's photo album (every police station has a photo album of the existing criminal registered to that police station) is a very exhaustive process. The process becomes more arduous when the photograph is sent to all the police stations for searching the previous record of that suspect. This process should be faster and reliable to adapt the real time challenges such as change in facial look [30-31]. These two challenges, i.e. the recognition speed and reliability of recognition motivates us to look for the solution which can minimize the search time as well as increase the confidence of recognition. The solutions for both the problems have been presented in (a) facial symmetry based recognition [32] as well as (b) component based face recognition approach [33].

Sometimes it is not possible to have the suspect's photograph; in those circumstances the police sketch artist usually creates a manual sketch of the suspect. The process of creating sketches and then releasing them in newspapers, in public media is the crude way of criminal investigation. The whole process needs automation to produce faster and accurate results. These two issues serve as the catalyst to search for the other direction of solutions. We critically analyzed the process of sketch drawing and found that sketch is the embroidery of imprecise knowledge of the eyewitness on the piece of paper. Each facial attribute has its own category and type, a sketch artist can pick up any category and portray that facial attributes. Further, combination of these facial attributes represents a proper face. We have created an interface to capture the vague perception of the eyewitness. A humanoid robot communicates with the eyewitness and do dialogue about the physiological as well as facial characteristics of the suspect. Later these rough and vague perceptions have been processed with the help of query based and rough set based expert system to reduce the search space and to find the correct match.

It is not necessary that every time criminal record does exist in the existing database and hence sometime no match of record could be found for the given visual perception. It is possible that, the suspect has committed crime for the first time and it has no previous criminal record or the criminal is international. In those situations usually police takes help of sketch artist to create sketch of that suspect. The sketch image is further used to trace out the criminals, therefore, the quality of the sketch image is a big concern. These limitations and real time challenges inspired us to create a robotic framework which could extend the first solution and enable robots to draw any sketch. We have used NAO humanoid robot [18] to demonstrate the validity of the approach.

We have proposed four novel approaches which are helping robots to identify criminals whether in the presence or in absence of photograph. We also enable them to draw sketch of the criminals only on the basis of given description about the suspect. Having these qualities and skills it is very much possible that these humanoid robots will be the part of near future army and police. In those situations it would be challenging task to prevent them from unauthorized access. The more often and frequently used medium for person identification is face. But this medium can be forged and can be intruded as any malicious person can take advantage of it. These bad effects can be discarded if the face recognition system is spoof proof [34-35]. To make robot safe and to protect them from illegal access we have proposed an anti-face spoofing mechanism which can discard these kinds of attacks in their initial stage.

### 1.3 Objectives

The prime and foremost objective of this thesis is to make humanoid robot capable of recognizing criminals. In order to attain the said objective we have used face biometric as the media to recognize criminals. The accuracy of the recognition depends on the availability and quality of the face image. In most of the cases the availability of the criminal face is questioned while their quality is also mysterious.

In both the cases different research challenges will be met. Therefore, we have split the main objective into two sub objectives.

1. To recognize and to re-synthesis the face of a criminal based on the visual perception about its shape, physiological as well as behavioral characteristics.
2. To recognize a person/criminal reliably and accurately by the face image in real time as well as in state of change of expression and in the presence of partial occlusion.

In the first research objective the face is unavailable hence, the description about the facial and physiological structure of the criminal is captured with the help of an eyewitness. Whereas, the second research objective will be met when the face is available, but its quality is not desirable.

## 1.4 Research Challenges

There are two major research objectives and each one leads to different research domain. Therefore their research challenges are also different. The first research objective is to recognize a criminal when vague and rough visual perception about her/his is provided. Therefore, its research problem is to design a framework to capture and process the visual perception. On the other hand, the second research objective is the typical and classical problem of face recognition hence, their research problems are to recognize a person in the variable light conditions, change in expression, in partial or full occlusion, recognition in real time [31]. However, the reliability of the face recognition system is ensured by discarding the face spoofing attacks. Face spoofing is an attack where someone wants to bypass the authentication process (face recognition) just by showing photo/video/mask of the legitimate user in front of the camera [35]. The system should be capable to discard the face spoofing attacks to ensure the integrity of the face recognition system. Specifically, following four research problems have been solved.

1. To identify the criminal given the visual perception of the eyewitness about that criminal.
2. To resynthesize the criminal face by a humanoid robot through dialogue with the eyewitness.
3. To recognize a person in real time in the presence of partial occlusions and changes in facial expressions.
4. To detect liveness of the presented biometric traits (face) in order to discriminate between the real (live person) and spoof (imposters).

**Research challenges corresponding to the research problems mentioned above:**

1. In order to develop a framework to identify criminals using the visual perception as input, the biggest challenge is to design a knowledge acquisition system to extract information about the suspect's facial, physiological and behavioral characteristics from the eyewitness. The next research question is how to represent this information inside the system and how to transform and process this information. Therefore, a transformation function and a processing model is required to transform this imprecise knowledge up to recognition level to help in finding the most relevant matches from the existing database.

2. The above challenges raise the issue of accurate matching. But sometimes it is possible that the system gives false negatives or there is no suspect information inside the database. In all those cases, a robotic system draws the photo of that person having the imprecise knowledge about the suspect. In that respect the research challenge is how to define image points with respect to humanoid configuration space, and to provide the inverse kinematic solution.
3. Recognition of a person in real time leads to the research challenge of minimizing the computation cost of the face recognition systems. Recognizing a person in the presence of partial occlusion includes research challenges such as localization of facial features (landmarks), determining minimum number of face components to recognize persons, etc.
4. Liveness detection is a tool which could be used to distinguish between the real and the spoof. Therefore the research challenge involved is to design a liveness detection module which will be placed before the face recognition system to discard the face spoofing attack.

## 1.5 Major Contributions

This has been a topic of debate "*how to make humanoid robots social?*". Perhaps the best acceptable answer is to enrich them (humanoids) with several functionalities which will make them closure to human beings [48-49]. This Thesis extends its contribution in criminal identification. We are using face biometric as the medium of criminal identification which can be forged. This is a security breach which must be fixed if we are using these humanoid robots as an authenticator at any place. Therefore, we have provided anti-face spoofing system to discard these kinds of attacks at the very initial level. It is not always the case when the criminal face is available, in all these situations the previous face image based criminal identification will not help us to trace out the suspect. However, if any eyewitness is present, then their knowledge about the suspect could be utilized to identify criminals. In all these cases generally a police sketch artist interacts with the eyewitness and a sketch is prepared which acts as a tool to search the criminals. But this process is manual and requires a large amount of time and co-operation. Therefore, this Thesis provides a solution for automatically processing eyewitness views about the criminal and finding out her/his existing record. A framework has been proposed and demonstrated in which a humanoid robot NAO interacts with eyewitness and asks various questions about the suspects gender, age, height, its facial attributes such as face type, eye size, lip shape and nose type etc. Further, these informations are processed in order to find out the match from the criminal database. Based on the eyewitness perception robot provides top 30 results. Sometime it happens that the suspect has committed crime for the first time, hence there are no previous record, in all those cases the sketch drawing capability helps them to create the sketch of that suspect based on the given views. The process of sketch drawing is fully automatic and requires no user involvement. The descriptions of original contributions are listed below.

- Our first contribution is in the field of criminal identification. The human perception about the suspect has been quantized and acquisitioned with the help of our proposed Query and Rough Set based approaches. The proposed query based approach is based on a decision tree classifier which classifies the given person to the ‘n’ number of possible classes. Each class has some probability associated with each. A set of reduct set is figured out in case of rough set based approach. The decision table for each reduct set has been computed which classify the test vector. The results are drawn from the database, based on the belonging of the test vector to these classes. The results are summarized in terms of top 30 results.[255][256][263]
- There could be a situation where the techniques discussed in previous contribution will not work as the suspect is committing crime for the first time. Therefore, a police record will not have the details of this new suspect. In such a situation the face composite, or sketch drawing will generally be preferred to design the sketch of that suspect. In order to solve the issue of face drawing, a robotic framework has been proposed which is capable enough to draw your perception in the form of a sketch which could be later used by police to release in public. In this context a novel calibration mechanism has been proposed to define the relationship between humanoid camera plane and their body coordinate system. NAO humanoid robot has been used as a testbed to experimentally establish the proof of concept. An inverse kinematic solution for NAO right hand based on gradient descent approach has been utilized to perform the sketch drawing. [254][259][260]

- The third contribution of the thesis helps robot to identify persons as well as help to track and locate criminals. We have used face biometric to identify persons and criminals. In this regard two novel techniques (a) facial symmetry and (b) component based face recognition are proposed. In the facial symmetry based recognition we have utilized only half face of the person to recognize identity. A theoretical as well as experimental proof on the AT&T face database has been established to confirm that human faces have bilateral symmetry (in absence of any facial disease) and either half of the face can be utilized for the recognition. Component based face recognition suggests a new way to recognize criminals with most discriminative facial components such as: Eyes, Nose and Mouth. An experimental study has been performed on 345 different subjects which confirm that more than 80% features of the full face lies within these fragmented components. The framework intends to process each component independently in order to find its corresponding match score. Final score has been obtained by calculating weighted majority voting (WMV) of each component matched score.[258][262]
- A robot is an intelligent system which if misused could lead to devastating circumstances. As the working principle of face recognition does not bother about the liveliness of a person. The authentication can be easily bypassed by just placing a face spoofing attack against these systems. Therefore, in our forthcoming contribution we have proposed two novel techniques to safeguard against face spoofing attacks. The technique is based on the principle of challenge and response, where robot asks person to perform movement in their face such as eye blink, lip movement. In response to these challenges person has to perform the movements. Further these movements are analyzed to validate the liveness. As every robot has stereo vision which helps it to perceive the outer world. We have exploited the stereo vision and calculated the depth of the real face and impostor face. As impostor face is displayed in a device like laptop, mobile phone or on a printed paper, all will result the same disparity at every point. But in case of real face we will get different depths at each point of the face due to odd even surface of the face. This serves as a discrimination factor between the real and impostor classes. [257][261][264]

The contributions[1] of this thesis have been appreciated into some publications in international conferences and journals. The contributions related to first research objective "Human Perception based criminal identification - a query based approach" has been published in Taylor & Fransis Journal of Theoretical and Artificial Intelligence [j1] while the "Rough set based approach" has been accepted in Springer Journal of Machine Learning and Cybernetics [j2] and in IEEE conference [c1]. Resynthesizing criminal face "Sketch Drawing by NAO humanoid robot" has been published in TENCON 2015 [c2], while the solution of calibration problem in order to attain sketch drawing has been accepted in Elsevier Journal of Robotics and Automation System [j3]. The solution of real time face recognition has been presented in IEEE and ACM Conferences [c3-c4] while the anti-face spoofing contributions published in Iderscience Journal of Applied Pattern Recognition Journal [j4] and IEEE, Elsevier International Conferences [c5-c6].

## 1.6 Thesis Structure

The remaining Thesis is organized as follows. (The overall structure of the Thesis is shown in Figure 1.6.1), the effectiveness of the human-robot interaction has been insured due to its robustness against challenges such as recognition time and occlusion. The identity association has been delivered on the basis of perception and face image based criminal recognition. The limitation of the first contribution acted as a catalyst for the second contribution, while the limitation of the face recognition served the foundation of fourth contribution. The chapter description has been listed below.

---

[1] *For citation, please refer to dissertation publication section of the thesis*

In Chapter 2, Analysis of previous research provides an insight of the evaluation of social robot and human-robot interaction system. The global and local features based face recognition literature has been investigated with respect to the full face and component based approaches. The chapter also critically comments and summarizes the techniques for dealing the face spoofing attacks while the analysis on criminal identification techniques such as mugshot detection has setup the benchmark for our proposed system. The observations from literature review are listed up at the end of the chapter.

Chapter 3, particularly talks about three basic problems of criminal identification framework: (a) Knowledge acquisition (b) Knowledge representation and (c) Knowledge processing. There are two modes of solution introduced and discussed in this chapter. Query based approach and the Rough Set based modeling for criminal identification. The web based console and existing speech to text engine has been used to capture the imprecise knowledge of the eyewitness while a linear transformation has been devised to quantize the captured perception in terms of crisp values. It is believed and proved afterwards that a person can be categorized from its facial and physiological characteristics.

Chapter 4, presents sketch drawing capability of humanoid robot which can be used to entertain us, help us in assisting artistic work or to create sketches of the criminals. Two basic problems such as estimation of the calibration matrix and inverse kinematic solution are discussed in this chapter. Three different calibration techniques are proposed to establish a relation between the Humanoid camera plane and its end-effector position. While, a numerical and closed form inverse kinematic solutions are developed to help Humanoid robot to reach to those Cartesian points (end-effector position).

Chapter 5, briefs about the two component based approaches of face recognition. The face recognition functionality has been added inside humanoid robot to trace out the criminals as well as to recognize persons. A facial symmetry and fragment face recognition techniques are presented here. Facial symmetry helps in fast recognition. The proposed solution is well justified when the recognition rate is very much important and the database is really big in size. The fragmented face recognition approach works well when only partial information of the face is available.

Chapter 6 provides solution to protect these humanoid robots to get illegal access and to use them as an authenticator. It is very much required that the presented face recognition approaches are reliable against face spoofing attacks. This chapter adds intelligence into the proposed face recognition system which can discard the imposter from the actual in order to insure the reliability; liveness test has been proposed and described in this chapter.

Conclusion and future work has been summarized in Chapter 7, which depicts the possible ideas and scope for future work. It also offers an insight into the limitations of entire criminal identification process. It suggests exploring feasible aspects with untapped potential for the enhancement of the identification process.

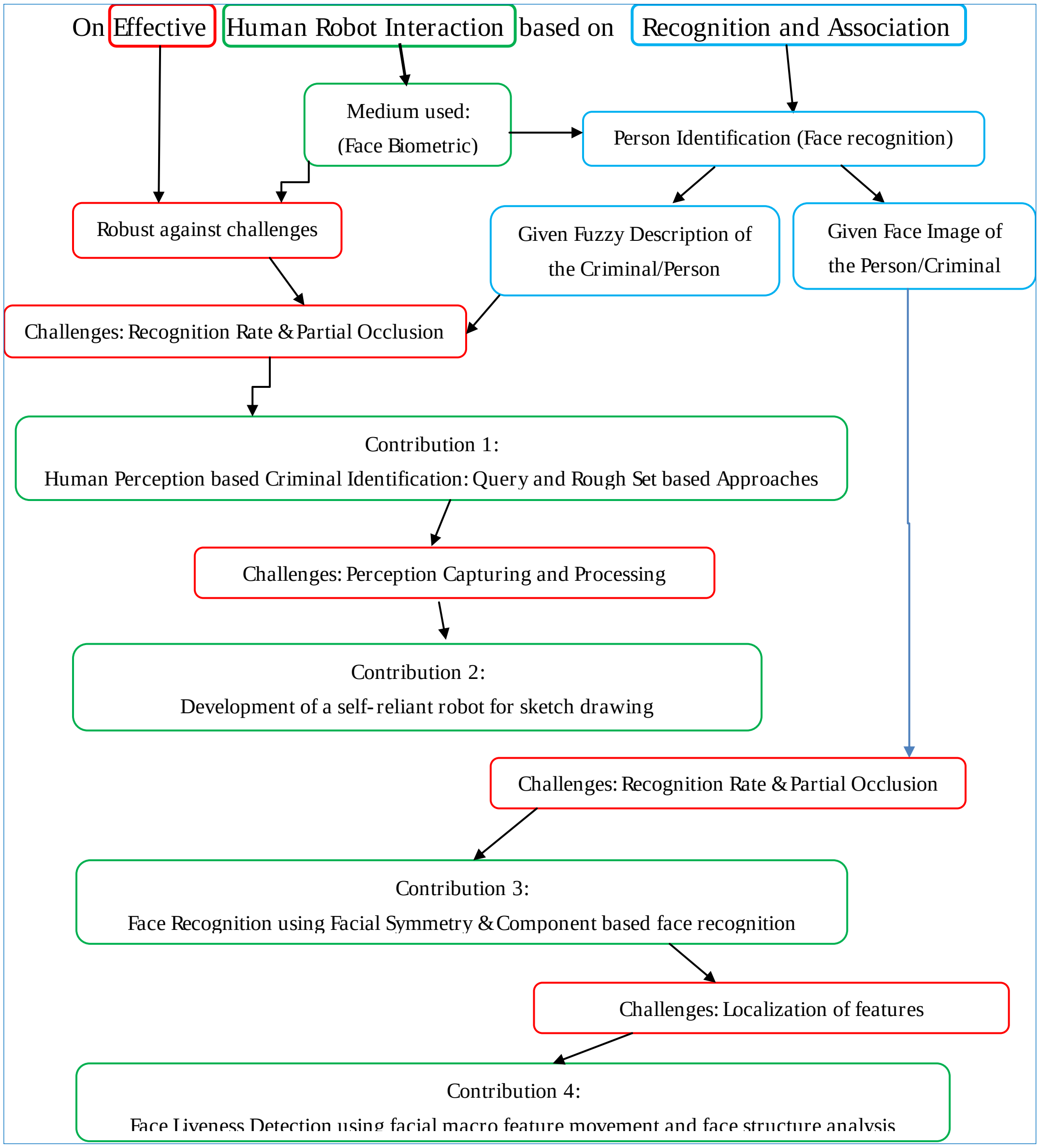

**Figure 1.6.1 Thesis Structure Overview**

# Chapter 2:

# Analysis of Previous Research

*This chapter laid down the foundation stone of this thesis, we have summarized the literature by first discussing the development and evolution of social robots, followed by their use in our society and in our day to day life. Their presence and capability in the field of person and criminal identification is discussed in further sections. The literature is divided over human perception based criminal identification, criminal identification through facial images and face spoofing detection.*

## 2.1 Development of Social Robot

The earlier developments of social robots [35] were inspired by the nature and the biology of insects. Researchers [36-37] were interested to know the stigmergy between the agents. Therefore, in the initial phase of development ant like robots came in the picture [38-40]. Researchers [41-43] were also keen to know the principles behind their self-organization and production of complex behaviours. Later the study about these insect societies and their synergism lay down the framework of social robotics. The first definition proposed by Dautenhahn and Billard [44] about the social robot was:

***"Social robots are embodied agents that are part of a heterogeneous group: a society of robots or humans. They are able to recognize each other and engage in social interactions, they possess histories (perceive and interpret the world in terms of their own experience), and they explicitly communicate with and learn from each other."***

The term recognition and interaction gave birth to the object recognition and human robot interaction in social robot. The object recognition [45] helps the robots to identify and discriminate between the objects while the human robot interaction [46][240-251] helps them to interact with humans. The interaction between robot to robot or between human to robot is the most required and mandatory behaviour to be called as social robots [47]. The domain and the field of study about the social behaviour and interactions of robots are termed as social robotics. There is a very thin line between the social robotics and societal robotics is defined by [48]. Social robotics is the integration of robot into our society by adding intelligence and human assistive features, while the societal robotics social empowerment of robots permitting opportunistic goal solution with fellow agents [48]. The robot having the capability to assist humans is known as social assistive robots, however, there is no clear definition for this [49]. A survey of social assistive robots is summarized by Feil-Seifer et al. [49] while the challenges are listed by Tapus [50]. The interaction between the human and robot could be established by various interactive mediums such as vision, speech, or by providing system commands (keyboard), etc. The field of Human Robot interaction in multidisciplinary, hence a guideline and evaluation criteria for robotic developer is presented by [51] to maintain the homogeneity of research under this domain. The proposed toolkit gives a way to analyze the research, provide the lists of do's and don'ts in order to design a social assistive experiment. It also measures the user's acceptance of the socially assistive robots [51]. The design process of social robots is specific to the purpose. For example, Bram et al. [52] have designed a story teller robot "probo" which helps in robot assisted therapy (RAT). Provo tells stories to the autistic children. Probo is also capable to produce facial expression and blink their eyes [53]. The other humanoid robot NAO's has [8] specially designed program for Autism Spectrum Disorder (ASD) children to help them in having interaction with them and to find other ways to deal with [54]. NAO has been used in education to assist their supervisor. Robotics Assisted Language Learning (RALL) [5][56] is such a program designed and developed by Minoo et al. [5] to provide help to junior high school students in learning the Iranian English. They are becoming a good housekeeper, especially in assisting physically disabled persons [7] [22]. They can play various sports like soccer, tic-tac- toe, etc., keeping their interest in sports a Robocup [13][56] is being organized every year. The other humanoid robots such as

Enon [57] is a fully automatic robot which can assists human by bringing objects/things to them. It can communicate through speech due to its inbuilt speech recognition system. ASIMO is a most advanced humanoid robot developed by Honda [58].

It is equipped with vision and speech system, it can recognize and do dialogue with humans. The rapid development in the field of robotics also changed the look and feel of these humanoid robots. Actroid [59] and Han [60] are the examples of those.

## 2.2 Person/Criminal Identification and Humanoids

A social robot is incomplete without the facility of person identification. As the face is the most frequently available medium for person recognition, most of the researchers have utilized the face biometric for recognizing identities. The face recognition literature is wide and mature [31] [61] yet the robotic vision based face recognition is challenging and require more attention. The mobile nature of robot and due to its more degree of freedom, the camera source is not fixed. It induces the problem of pose variation which creates the problem in face recognition. The viewing angle of the robot vision (camera) is also not aligned with the person. In order to solve these issues, Fukui and Osamu [62] have proposed view invariant face recognition. The proposed face recognition system can recognize the identity from the different view angle and also in the presence of profile faces. A constrained mutual subspace method (CMSM) is used to process multi-viewpoint face pattern for recognition. Kai-Tai and Wen-Jun [63] proposed a face recognition system for a small family (working for a small dataset), they have created Radial basis functional Neural Network for each family member in order to recognize them. Their robot can also track multiple faces using the colour based tracking (face and hair colour is utilized here). The system is showing the accuracy of 94%, however the scalability of the system is not assured. Cruz et al. [64] have used only single image to recognize identities. They have divided the face over three small components left eye, right eye, nose and mouth. Further the Scale Invariant Feature transform (SIFT) [65] is used as feature extraction while naive Bayes [66] is used as a classifier. The reliability of the proposed system is dependent on the landmark localization while the system fails to explain how it localizes the facial components in the face. The solution offered for solving the face recognition problem is based on supervised learning, where we first trained the system using the face images of the enrolled persons. Contrary to this Aryananda [67] proposed an unsupervised face recognition module for the Kismet [68], whenever a new person come whose pattern is different from the previous pattern it assigns a new class label to it. The integration of face recognition and other specific module over the robot is a challenging issue, Sakagami et al. [58] and Kuroki et al. [69] have given a demonstration on ASIMO and SDR-4X, how to add new functionality over it. The step by step addition of several functionalities such as: gesture recognition, vision and audio based human interaction, access to online databases and using the internet. SDr-4X is also trained to perform dance on some specific songs.

Face tracking is equally important as recognizing the person [70] [72]. The face tracking features enable them to maintain the eye contact as well as to localize the face among the other objects. Brethes et al. [70] has utilized watershed algorithm [71] to segment skin colour which helps in tracking. As the skin colour and watershed algorithms both have a drastic effect due to variable lighting condition. The 3D structure of the face helps in localization as well as in tracking, therefore, based on stereoscopic vision a 3D face tracking is proposed by Matsumoto and Zelinsky [72]. The proposed system robustly tracks the head movement as well as the eye gaze in real time. [73][74] have designed expression and sentiment analysis for these robots. The expression recognition enables robots in getting closer to humans. The face recognition module in robots helps them to recognize persons as well to identify criminals. Due to the erroneous real time challenges such as pose, illumination and occlusion, more robust algorithms are required [31].

The presence of these robots is not only limited to entertainment, identification or for household use, they have also marked their presence in the homeland security. Now, many countries such as USA, UK, Japan, China, South Africa, South Korea, Australia, and Canada have joined them in their police force [12]. These robots are called police robots. They are used for doing the dull and dirty job, which a human does not want to undertake. Some of them are equipped with the lethal and nonlethal weapons. They can help in surveillance, hostage release, bomb defusing and saves police's lives. Mostly they are semi-automated and manual. ANDROS Remotec [12] is a wheel robot designed by the Remotec Company is a part of SWAT (Special Weapons and Tactics) of US Police. ANDROS is a bomb diffusing robot which could perform variety of dangerous task such as breaking down doors, approaching hostage, trashing windows, localizing a building, etc. In 2007 on the demand of US army a nonlethal robot is created with the combination of TASER robot [9] and the PackBot Robot [10]. A non-lethal is attached to this robot having compressed nitrogen to shoot darts which can hit a human from the distance of 10.6 meters. South Korea commissioned Samsung has built a border security robot which can find a person in the demilitarized zone between North and South Korea. The robot has the capability of automatically shoot that person within the range of 4 km. In 2008, China has also included police robots (Dragon Guard 3, RAPTOR) to their team during the occasion of Bejing Olympics [13]. Japan has a variety of security robots like Roberg-Q. It can communicate with us and can lift us. It is used in several shopping malls for patrolling as well as to prevent the unauthorized access. UK has included the bomb defusing robot many times before in his army, but in recent year, two of its states Liverpool and Glasgow are using micro drones for the surveillance [75-76]. Seeing the presence of these humanoid robots in the army, police, home, and office, it has become an integral part that they should recognize people and criminals. The rest of the literature discussed in the following sections discussed the progress made in these fields so far. The literature is divided over three main issues, the criminal identification approach based on human perception, face image and anti-face spoofing to avoid the unauthorized access.

## 2.3 Human Perception based Criminal Identification

A person can be categorized from their behavioural, physiological and facial attributes or the combination of all three [77]. A person’s behaviour can be captured in terms of their gait patterns and the way of speaking, their physic can be represented in the form of age, gender and height, while the face description can be summarized as face category, face tone, eyebrow’s type, eye shape, nose shape and lip size. The working envelope of forensic consists of two major blocks. Evidence collection and later the reliable matching of these evidences [78-79]. The face can be treated as one of the kernel evidences therefore its proper matching is required. But it is a known fact that if someone is committing a crime, then she or he will never want to disclose her/his identity. In such scenarios we would be having only partial information of the faces or maybe no face information at all. A detailed list of these challenges was addressed by A K Jain [79] Due to parallel processing of the human brain and its logical capability, it is easier for us to recognize any person with the given partial information. But this is not true in the case of machines. Please refer [80] to know more about how human perception and machine react towards the face recognition problem and their synergism too.

We started our journey of criminal identification following the work of Gary L Wells [81] According to Wells, investigations and prosecutions of criminal cases heavily depend on the eyewitness testimony. Eyewitness testimony plays a supportive role in order to achieve criminal justice. Generally an eyewitness has to follow a lineup to identify criminal in the bunch of suspects. In such scenarios, those who are not culprit are known as fillers. But, sometime the impreciseness and vague perception about the criminal could raise conditions like (a) target absent and (b) target present. Target absent is when the culprit is added into fillers and target present is when the culprit is excluded away from fillers. The whole process of identification could be processed following two different methodologies. Simultaneous lineup where all suspicious members are present in front of the eyewitness, while the sequential lineup is when suspects are shown one by one. The literature also categorizes the eyewitness into two categories, genuine and mock witness. Genuine witness are those who have seen the event live while mock witnesses are people who have actually not seen the event or were not present at the site based on the eyewitness verbal description; they are asked to pick a person in a lineup. There are other factors which can influence the identification accuracy and can affect the criminal justice. Factors which can affect the justice process are known as system variables while others are known as estimator variables. The system variables are the collection of set of protocols such as the instruction given prior to viewing the lineup, lineup size, etc. while the estimator variable includes the evaluation of current witnessing situation and the relation between the culprit and the eyewitness. In most of the cases the identification is performed by seeing the face and other physiological as well as behavioural characteristics of the suspect. Therefore, we have quantified face and physiological characteristics to identify criminals. As face falls under the system variable category it is an integral part of jurisdiction. It is widely useful in such scenario where we don't have any idea about the criminal. Face recognition is a tool which could be helpful to trace out the victim by crawling the existing criminal database and to find the match. But it is applicable only when we have a photograph of the criminal. When we don't have photographs of the criminal an eyewitness who has seen the guilty can help us. Usually it helps sketch artist to prepare the sketch of the criminal. These sketches are the embroidery of eyewitness imagination and perception on the piece of paper. In most of the cases, these sketches are similar to the criminal face. Later these sketch images are published publicly by the police department to find the guilty. But the problems of manual matching of sketches to photographs are ambiguous and need automation.

The existing literature defines automatic way of matching these sketches to their respective photo images as mugshot detection. The first step towards the mugshot detection started in 1996 with Roberg et al. [82]. In his experiment they demonstrated the sketch to photo matching successfully on a limited database. Later the significant contribution in this direction has been carried out by [83-85]. Unfortunately, this problem has not received proper attention and requires further research in this direction.

## 2.4 Face Image based Criminal Identification (Face Recognition)

Face Recognition has been an interesting research area over the past 30 years, but still it has very limited commercial use (11.4% according to Andrea F.[86]). According to the A.F abate et al. [86] a good recognition ration has been achieved by several researchers over some global face databases, but none of them are able to cover the problems of pose variations, occlusion, change in illumination change in expressions, etc. The real time implementation of face recognition system is still a challenging task and require a great deal of attention. The face recognition literature has different approaches in order to extract unique features from the face images. The most frequently used approaches are global, local and geometrical [30-31]. Global approaches used overall population of the face in order to capture the most significant patterns. Principal Component Analysis (PCA) [87-89], Linear Discriminant Analysis (LDA) [90-92], Independent Component Analysis (ICA) [93], etc., are some widely used techniques. They focused on finding the number of independent variables to express the population by finding the best possible projection direction. Samples projected over the feature subspace are linearly independent and can help in classification. But these features are sensitive to rotation, variable light condition, orientation, etc. While the local approaches used particular sample to discover the unique features out of it. These features are detected by analyzing their neighbourhood pixels. Therefore, they are invariant to rotation. They can even handle variable light effects because the effect of light is almost homogeneous if we consider a very narrow region. Local Binary Pattern (LBP) [94], Scale Invariant Feature Transform (SIFT) [65] and Speedup Robust Feature (SURF) [95] are widely used local feature techniques. The geometrical features localized the presence of eyes, nose, mouth, and other components of the face in the face. Based on their shape, size and the distance between them, a unique representation is found. This unique representation is further used for the classification, Elestic Bunch Graph Matching (EBGM)[96], Active Appearance Model (AAM)[97] and an Active Shape Model (ASM)[98], are the famous methods falling under this category. They are typically the machine vision techniques using features like edges, corners, etc. We have studied all the approaches and decided to use Principal Component Analysis (PCA)[87], Linear Discriminant Analysis (LDA)[90] as global feature extraction techniques and Scale Invariant Feature Transform(SIFT)[65] recognition as local feature extraction and Geometrical features are used for the component based face recognition. Their application in face recognition technology has been briefly described in below sections.

### 2.4.1 Principal Component Analysis: Eigenfaces Approach

The objective of PCA is to reduce the dimension by analyzing the overall population. If we have a population which consists of "p" samples and each sample is represented in terms of n variables (Features/Dimensions). Not all features are important to us. There are some dependent and some independent variables out of these n variables. PCA will help us to find out k independent variables out of these n variables which can be helpful to express the overall population. Thus, by leaving those n-k variable, we can be able to reduce the dimension while preserving the best features. These k variables are the k principal components. The first principal component shows that maximum variance, Second component shows the second highest variance and so on. These principal components are also orthogonal to each other ensuring no dependency in the projected feature space.

The approach was first proposed by Karl Pearson in 1901[88] and later extended by Hotelling [99] in 1933 respectively. It has been brought to solve the problem of face recognition by Kirby & Sirvich [100] in 1990, he introduced the concept of Eigen picture. Eigen pictures are the projected pictures of the overall population. Any face can be represented in the combination of these Eigen pictures. They have also suggested to process only specific modules of the face as they have bilateral symmetry. Next year Turk and Pentland [87] came up with the idea of Eigen faces, a modified one from the previous approach of kirby & sirovich [100]. If face is composed of "$mn$" variable where "$m$" represents number of rows and "$n$" represents the number of columns and if we calculate the correlation between these variables, we will end up with $mn \times mn$ computation. It is very hard to compute the correlation between these variables. They suggested to compute this correlation but on an abstract layer. If the correlation between one face to another face is computed, it can give you the same measurement. It is also because, if we will calculate the characteristic roots of the covariance matrix, there will be only p roots that are meaningful rest of the roots are zero. Hence, rather than finding the co-variance between pixels, we can compute the same over images. Later by finding the principal components of co-variance matrix, we can form the projection matrix which is used for generating the Eigen faces and for creating the signature of each individual face.

### 2.4.2 Linear Discriminant Analysis: Fisherfaces Approach

Linear Discriminant Analysis is a supervised feature extraction technique where for each sample a predefined class label is assigned. There are two parameters that we have to estimate and optimize in order to get the optimal projection direction. They are a) Within Class Scatter matrix (SW) and b) Between Class Scatter Matrices (SB). Within class scatter matrix described how the data is distributed along the meanwhile between class scatter matrix shows how mean of one class differ from the other classes. In order to get the best projection these SW should be minimized while the SB should be maximized. In 1990 Sir

Ronald Aylmer Fisher suggested a criterion function which can optimize these two parameters: (J = SB/SW). It is applied on the face recognition problem by [90], the faces generated here are well known as Fisherfaces [90]. The problem in this criterion function is when we apply this on high dimensional data the inverse of SW become singular. Hence we cannot calculate the inverse of SW. This is because the small sample size of the high dimensional data. Therefore, various techniques exist in the literature to solve this issue [101]. They suggested to use the total scatter matrix S=SW+SB in place of SW in order to calculate J. The best directions are chosen by selecting the best Eigen Vector of J. A modular approach has been also investigated by him. Another approach suggested by W. Zao et al. [102] is to first reduce the dimension by using PCA and then apply LDA on it. PCA helps in throwing the null spaces of SW which causes a singularity in SW. But later it has been analyzed by [103-104] that null spaces of SW also contain discriminative information. Hence [103] has proposed a new LDA method which first extracts the feature of the face only on the basis of mean value of eye, nose and mouth regions. Later these features are used to calculate the SB and SW, He has modified the existing fisher criterion function (J = SB/SW+SB). SW+SB is known as the total scatter matrix. The best direction vectors are composed of the most discriminant Eigen Vectors of SB, while SB is computed with the help of SW which consists of only null spaces. Chen et al. [103] used only null space information in order to generate the between class scatter matrix.

His solution works best when the SW is singular, but not about the cases where they are not singular. Therefore H. yung [104] has proposed a direct LDA (D-LDA). According to Yung they first diagonalized the SB and then finding the best Eigen vector of it they diagonalize SW. Diagonalization help in throwing null spaces of SB while preserving the null spaces of SW. This approach is later extended by Juwei Lu et al. [105] They first applied D-LDA in order to reduce the dimension, then additional fractional step is added to get the most discriminant vectors. In place of diagnalize the SW they diagnalize the overall scatter matrix.

### 2.4.3 Scale Invariant Feature Transform (SIFT)

Scale Invariant Feature Transform basically used in object detection originally proposed by D. Glove [65]. The features extracted are scale Independent, rotation invariance face images and even they can handle the illumination condition up to some extent. But the use of the SIFT algorithm on face recognition has limited efforts but very deep impact. It was started on the face recognition domain by Manuele et al. [106]. We get 128 dimensional feature descriptor, location, orientation and scale for each feature after applying SIFT. But how these features are useful for face recognition, therefore[106] has suggested 3 differentiate matching techniques (a) Minimum Pair Distance: Only those distance pairs will be selected which have the minimum distance between the test and gallery points.(b) Matching Eye & Mouth: Match is found only for eye to eye region and mouth to mouth region in train and text points.(c) Matching a regular Grid: In this image is divided into several sub images (grids) and then for each grid minimum pair distance is calculated between train and test grid. In the same year, another effort has been made by Mohand Aly [107], he compared the effectiveness of SIFT algorithm over the Eigenfaces [87] and Fisherfaces [90]. Minimum distance classifier has been used as classifier with Euclidean, City Block and cosine distance measurement technique. Another feature matching is suggested by Jun et al. [108] in 2007 respectively, He followed the concepts of [106] and divided the features over grids and calculated the local and global similarity. Each grid will produce a local matching value multiplied with a particular weight, while the global method used these local methods in order to compute the overall matching. A graph based technique is also proposed by Daskshim et al. [109] based on the graph matching. Fisher linear discriminant method based feature selection technique is proposed in [110]. A modified of the SIFT algorithm for face recognition is given by Cong Geng et al. [107]. The first approach is Key Point is preserving SIFT (KP-SIFT), which suggests that removing of feature point in step 2 & 3 of SIFT [65] algorithm discard those feature of the face which can create the most discrimination. These features could be face wrinkles, mouth corners etc., therefore in his modified version, they followed only the first and last step of SIFT. Another modified version is Partial Descriptor based SIFT (PD-SIFT) which preserves the face boundary information by selecting a minimum grid of size 16X16, in comparison to original i.e. 256X256 which can discard some significant information of the face.

### 2.4.4 Face Geometrical (Landmark) Localizations

Face detection is the primary and the initial step if we do perform face recognition in real time. The face detection is also important from the robotics points of views, as it helps Robot to discriminate between the human and other objects. There are different techniques to localize a face in an image frame. These techniques are grouped among (a) Skin colour based face detection [111-113], human skin has a particular range over the RGB/HSV/YCbCr colour space [114]. Fixing up the valid range for the face skin colour we can segment the face from other objects. Since human face has other macro component associated with it, eyes, mustache (for men), eyebrows which has different tones from the face skin. They create black holes, therefore some morphological image processing tools [115] are used to label the face images. As the localization of face is dependent on skin colour, and skin colour is decedent on person's geographical location. We need to fine tune the face window every time. The second and the very critical issue which measurably affects the quality of detect is variable light effect. Due to these challenges, skin based face detection could not get much attention. The second group of research utilized image features such edge, corner and line features for face detection. In this category they used Haar Classifier [116], Hausdorff Distance [117], Edge-Orientation Matching [118], etc. to detect faces in images. Principal Component Analysis [119] and Artificial Neural Network [120] are also being utilized for detecting faces. The research in the domain of face detection has a significant accuracy [116-120], whereas the detection of facial macro features is a challenging issue.

Active Shape Model [98] and Active Appearance Model [97] is the most widely used methods for localizing the facial features. In both of the techniques a default model of the face is initialized and then based on the human face geometry it is reiterated to fit on the face. Support Vector Machine and Regression analysis is being performed to fit the model. But these techniques are not sufficient enough to handle the occlusion and the pose variations. Also, if the shape differs more from the initial guess, the process fails to relocate the facial landmarks. In order to handle the shape variability and occlusion effect Cascade Pose Regression model [121] is proposed in 2011. The successive growth of CPR model can be seen in [121-124].

- CPR is constructed by a cascade of X –regressors $R^{1..X}$ that begins from a rough initial mould guess $S^0$ and proceeding towards better estimation, producing final shape estimation$S^T$.
- Mould S is defined as a series of P landmark locations $S_p = [X_p, Y_p]$, $p \epsilon 1..P$. In every iteration, regressors $R^t$ generate a new update ϭS, then after this ϭS combine with last iteration estimate $S^{t-1}$ to produce a new module.
- In Training, every regressor $R^i$ is trained to find some way to minimize the difference between the real mould and the mould estimate of the last iteration $S^{i-1}$.

The present features are dependent on the current mould estimate so the update in required in every iteration. These kinds of features are known as the pose indexed features. They lead to CPR, depend on the evaluation of robust pose indexed features and training, the regressors would lead to progressive decrease in the estimated error in every iteration. In the real world scenario with different lightening conditions (varied illumination), occlusions, expressions and pose variations are too common. Robust Cascaded Pose Regression (RCPR) [125] is a method through which we can improve robustness in all these challenging conditions.

The previous approaches [121-124] struggle with occlusion because they do not take it in the decent way. RCPR [125] incorporated the occlusion directly at the time of training for the betterment of shape estimation. This method need ground – reality annotations specifically for occlusions under the training set. This type of information can be incorporated with lesser worth during the annotation process, incorporating a flag with each landmark and encoding its visibility. The Cascade Pose Regression [121] conventionally denoted shape S as a chain of P part locations $S_p = [X_p, Y_p]$, $p \epsilon 1..P$. In RCPR [125] they continued the same definition, adding an additional visibility part $S_p$ = = $[X_p, Y_p, V_p]$, where $V_p \epsilon \{0, 1\}$. The locations are randomly initialized with re-projecting learning modules into the rough guesses of object locations. After that all three dimensions are learnt parallel with the use of cascaded regression. Provided an Image, the face (whose coordinates are provided by a face detector) is segmented into a 3x3 grid. In every iteration t, the mass of occlusion and shape variations can be estimated through projecting the latest estimate $S^{t-1} = [X_{1..P}, Y_{1..P}, V_{1..P}]$ in image. After that in place of training a single regessor $R^t$ in each iteration t, they suggest to train $S_{tot}$ regressors $R^t_{1..S_{tot}}$. In the end every regressor's suggested output updates $\sigma S_{1..tot}$ are merged through a weighted mean voting. In which weight is reciprocal to the whole amount of occlusion available in the zones from where the regressor draw features. The real dynamic environmental faces maintain a huge variability in pose and expressions. Pose indexed features are invariant with poses and face size. They are the main lead to shape evaluation successfulness within these conditions. [65] Suggested calculating the matching transformation for normalizing the current pose to mean pose and point pixel by its native coordinates $\sigma_{x,}, \sigma_y$ with regards to its nearest landmark. However, the evaluated features are still not good enough in comparison to big shape variation and shape deformation. If the count of landmarks in a particular region is low, then it is assumed that most of the features are behaving like outliers in the evaluated landmarks. For reducing this problem, RCPR [125] suggest to point features by continuous interpolation between 2 landmarks. These newly created features are very good to pose variations. The newly created features improve whole performance and manage to greatly decrease failure cases. Because there is no further need to find or estimate the closest landmark in the current estimate of the pose for each feature, evaluation is comparably faster. The convergence of Cascaded Pose Regression depends on the initial mould points. Both [121] [123] suggest to execute the algorithm number of times using dissimilar initializations and use median of all predictions as the final output. Given an image, every restarts, run independently, avoiding evaluations till we reach out the very end. This method influence a principal fact: Provided an image and a number of dissimilar initializations, in the first 10% of initializations the cascade is used to everyone. After that the variance across corresponding predictions is checked. If that variance is lesser than a certain threshold, then the remaining 90% of the cascade is used as usual. If somehow the discrepancy (variance) is greater than the threshold, the procedure is again started using a different set of initializations.

## 2.5 Analysis of Face Spoofing Attacks

The multimodal biometric provides a more secure and reliable way of person identification. Recently the multi modal biometric is also being used to detect and prevent the face spoofing attacks.

### 2.5.1 Image Quality Assessment for Liveness Detection

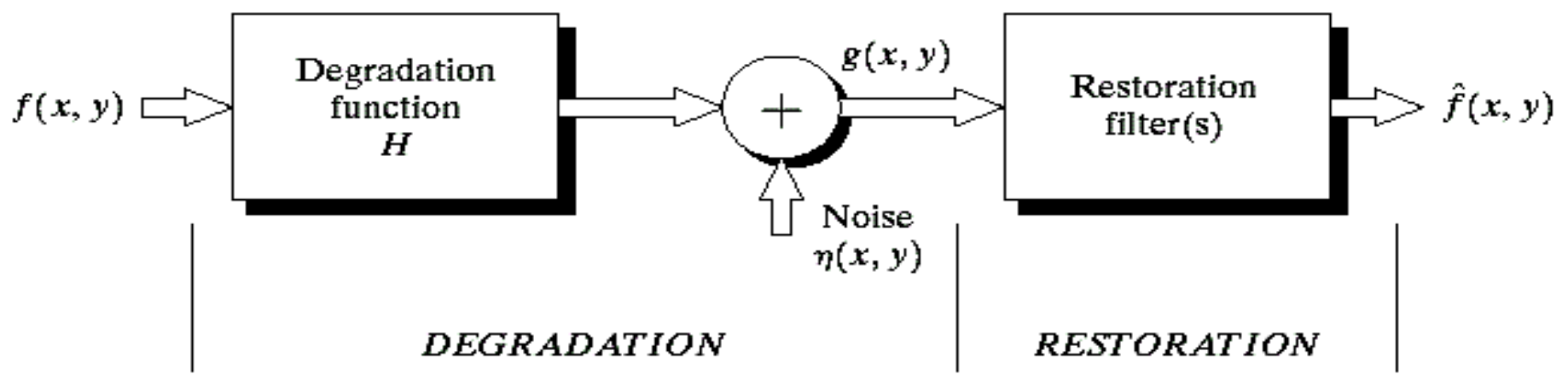

**Figure 2.5.1. The process of digital image formation [126]**

Any captured image $g(x,y)$ is the combination of actual image $f(x,y)$ medium of degradation $h(x,y)$ and noise $\eta(x,y)$[126]. The process of forming a digital image is summarized in Figure 2.5.1. If we can estimate the degradation function $H$ and the noise $\eta$ we can somehow get back to the captured image. The expression of getting the transformed image is presented in equation (2.1).

$$g(x,y) = h(x,y) \times f(x,y) + \eta(x,y) \qquad (2.1)$$

$h(x,y)$ is the convolution operator performed over captured image $f(x,y)$ while $\eta(x,y)$ is the additive noise in this case. The degradation function H could be estimated by using observations, experimentation or mathematical modeling, while the noise factor could be compensated using anti-noise filtering. In general image restoration we tried to minimize the effect of H and $\eta$ to get the captured image. In other words, we try to minimize the difference between $f(x,y)$ and$\hat{f}(x,y)$. If we pass again the digital image $\hat{f}(x,y)$ to this process of image formation, the medium of degradation and noise will be added twice, as shown in equation (2.2).

$$\hat{\hat{f}}(x,y) = h(x,y) \times \hat{f}(x,y) + \eta(x,y) \qquad (2.2)$$

If the difference between the $\hat{f}$ and $\hat{\hat{f}}$ is sufficiently large, this could serve as a discriminator between the real and fake face. Through image quality researchers have tried to find the discrimination between these two classes through maximizing this difference.

In order to estimate the image quality of both the $\hat{f}$ and $\hat{\hat{f}}$, most of the previous research has performed texture analysis. Face texture analysis is the mechanism to distinguish between these two classes based on the surface roughness (texture), in this regard a liveness detection scheme is put forward by Peixoto et al. [127] under bad illumination condition. Spoof is detected by measuring the boundary value of the image taken by the camera. They assumed that if the photograph or LCD is used in spoofing, images taken by the camera will result a blurring effect at the borders owing to the pixels with higher values brightening their neighbourhood. The real face does not have this type of effect that leads to the discriminative factor. However, this approach does not give any idea about how it will work in the case of real faces. They might report False Rejection Rate (FRR) owing to the illumination change (in case of brighter contrast) the same person looks like an imposter, having a brighter region than the natural real face. Gahyun Kim et al. [128] proposed single image-based liveness detection scheme which is able to discriminate photograph of the live face. They measured frequency by power spectrum and texture information by Local Binary Pattern (LBP) [94] of a given image. Frequency information is extracted in order to transform given facial image to frequency domain. LBP is used to assign a code to each pixel by calculating the relative intensity difference of a pixel and its neighbours in order to achieve texture-based information. They fail to explain shape statistics of the facial images which would have been more prominent to fuse these features. Since the research has been carried out on single face databases, the results cannot be generalized to a large number of other existing databases. Xiaoyang Tan et al. [129] have also used real-time method for liveness detection using a single image. They analyzed surface properties of photographs and live human face by Lambertian model [130]. They extended standard sparse logistic regression classifier both non-linearly and spatially. It improves performance of anti-photo spoof and a sparse low rank bilinear logistic regression model achieved by spatial extension controls complexity of the model. However, they overlooked the problem of bad illumination. J.Matta et al. [131] have used image texture and gradient structure for detecting the spoof attack. They have designed feature space by learning the fine difference between fake and real faces. Facial microscopic information is extracted by two powerful texture features 1) Local Binary Pattern (LBP) [94] and 2) Gabor Wavelets [133]. In addition to these two features, a local shape description is also extracted through Histogram of Oriented Gradients (HOG) [132]. A nonlinear Support Vector Machine is utilized to design a classifier for real and fake faces. The nonlinear SVM transforms each low-level descriptor into compact linear representation. The experimental study is performed on two benchmark datasets NUAA imposter database [129] and on YALE recaptured database [127].

A low level descriptor and partial least square methods are utilized by William at el [134]. For robust liveness detection, distinctive characteristics can also be extracted through spatial and temporal information. They have extracted low-level features and assigned weights to each feature defined by partial least square [134]. The feature descriptor combined for anti-spoofing are histogram of oriented gradient (HOG), color frequency (CF), gray level co-occurrence matrix (GLCM) and histogram of shearlet coeffcient(HSC)[20]. They have tested their efficiency on NUAA [129] and FSA dataset [135]. Jiangweili at el [136] used lambertian to define the discrimination features, the lambertian model given as:

$$I(x, y) = \rho(x, y)n(x, y)^T \qquad (2.3)$$

$I(x, y)$ is input image,$S$ is light source,$\rho$ is albedo (surface texture),$n(x, y)$ is surface normal. They have suggested that for 2D planner surface (fake face image) $n(x, y)$ will be constant, real face image depends on $\rho(x, y)$and $n(x, y)$ and for fake image it depends on only $\rho(x, y)$. To differentiate between fake and real face they have used Fourier spectra. Fake face has found much less “high frequency” component than real one. They calculated high frequency descriptor (HFD), which is the ratio of the high energy frequency component and overall frequency component of the image. To build a robust technique, they also considered the motion in images to extract temporal and spatial information. For liveness detection through motion images, they have used following steps: 1) Images are extracted from video and 2) Energy value of each frame is computed. They used frequency dynamic descriptor (FDD) to measure temporal changes in the face, so FDD for the fake face should be zero. FDD is more successful to discover fraudulence as motion in high resolution and in big size fake photographs. Gahynkim et al. [137] exploited frequency and texture information obtained by power spectrum and local binary pattern (LBP) [10] respectively. They have trained power spectrum, LBP and performed fusion of two. Xiaoyang [41] investigated the nature of image variability for real and fake face image based on lambertian model. They show that illumination invariant face recognition methods can be modified to extract latent samples. Latent samples are given by two methods 1) variational retinex-based method 2) difference Gaussian. These latent samples are used to discriminate surface properties of real and fake face image.

Seeing the performance of LBP [131] [137] in face liveness detection, Benlamoudi et al. [138] introduced overlapping block LBP operator while to reduce the feature space fisher score is utilized. The classification is performed using the Support Vector Machine (SVM). The local binary pattern is a good tool to extract and represent the face texture information. There are other variations of Local Binary pattern such as: Local Ternary Pattern, Local Tetra Pattern, but the best result so far is reported on Local Binary pattern (LBP) [131]. However the LBP code for the face is not able to handle the dynamicity of the expression. The movements in facial macro features are the positive addition in making the large difference between the two classes (Real and Imposters). Therefore, in order to use the dynamicity as well as to preserve the maximum discrimination Tiagoet et al. [139] and Jukka et al. [131] has proposed spatiotemporal (dynamic texture) analysis to detect the face spoofing. The training is imparted to detect the facial structure and the dynamicity involve in full faces while the reliability of the approach is assured on the two benchmark datasets (Replay-Attack Database[135] and CASIA Face Anti-Spoofing Database [140]).The texture analysis of both real and imposter face images are playing a vital role to create discriminations between these two classes. However the variable light/illumination condition is still a challenging issue. In order to normalize the illumination condition, most of the researchers have used only the chrominance part of the face image to perform the texture analysis, nevertheless Colour information is also an integral part of the face image, which defines the life and geography of the person.

Therefore, Boulkenafet et al. [141] proposed a Colour based texture analysis called as Colour local binary pattern CLBP. In the proposed CLBP they have utilized both the luminance and chrominance part of the image. The experiment of the in house database depicts the promising results than the state of the art. As discussed above the LBP [131] and LBP-TOP [139] are not able to furcate these two classes due to illumination variance and imaging quality. These feature extraction techniques fail miserably due to the light change. Hence, Yang et al. [142] have utilized Convolutional Neural Network to automatically extract features as well as to create an auto representation of the given faces.

The analysis of facial local and macro features were in big attention among researchers to design the novel tools and techniques to deal with spoofing attacks. For example Housam et al. [143] improved local graph structure (ILGS) for the better representation of the facial macro features. First the whole face image is divided into several small blocks and then the local graph structure is calculated for each block. The concatenated histogram of these blocks are used as a feature. Benlamoud et al. [144] designed a novel Multi-Level Local Phase Quantization (ML-LPQ) to extract features of the face region of interest while the libSVM is used as a classifier. On the other hand Dong et al. [145] have proposed multi spectral face recognition system to prevent the face spoofing attacks and Wen et al. [146] introduced a novel image distortion analysis (IDA) to detect face spoofing attacks. Dong et al. [145] utilized visible and infra-red spectrum region for the analysis with the analysis factors such as type of imaging device used for the attack generation, tools used for face landmark extraction, feature extraction and feature matching. Whereas Wen et al. [146] composed their feature as the combination of specular reflection, blurriness, chromatic moment, and color diversity.

### 2.5.2 Challenge and Response based Face Liveness Detection

In the challenge and response method, usually the system throws some challenges to the user and calculates the response in the account of the given challenges. If the person fulfills the challenges, the system understands that the person is live and if users fail to generate the responses, the system treats him/her as imposter. Most of the researchers utilized eye blinking for testing the liveness. Gang Pan et al. [147] presented an approach for detecting spoofing attacks, by using the eye closity feature. Eye closity is defined in terms of three parameters ($\alpha$) open, ($\beta$) ambiguous and ($\gamma$) close. The typical eye blinking may involve $\alpha \rightarrow \beta \rightarrow \gamma \rightarrow \beta \rightarrow \alpha$. Eye closity $\Omega(I)$ is computed by using discriminative measurement derived from the adaptive boosting algorithm and calculated by a series of weak binary classifier. The problem with Gang's [147] approach is that it detects the liveness on the basis of eye blinking which can be easily forged by playing the recorded video of the genuine user. This problem was also observed by Westeyn et al. [148]. They used the same concept of eye blinking to detect the liveness with the addition of a constraint challenge and response. A song is chosen by the user at the time of their enrollment, a particular eye blink sequence, then generated which is stored in the database together with the face data. At the time of verification, users have to choose the song and blink the eye in the same way as it is stored in the database. Although this approach is good to detect the liveness, the user has to remind their blink patterns and if they fail to do so, the system will not grant access. Therefore, it could increase the false rejection rate (FRR). A good technical report has been discussed by Nixon et al. [149], who have investigated different biometric traits which can be spoofed.

Hyung-KemJee et al [150] proposed a method for liveliness detection based on eye movements. They stored 5 sequential images to detect variation in the eye region. They have used Viloas and Johns [116] eye detector for detecting and localizing eyes in face, while the illumination effect is normalized by using SQI- Self Quotient ( a kind of high pass filter [151]). The binarized eye region is passed to compute the score of liveness. The score is compute based on Hamming Distance which is computed on the number of pixels who have not same value in two consecutive images of eyes. The score of real eye is high because of blinking and pupil movements and in case of fake eyes, this is low. AhdreAnjos et al [152] proposed motion based liveness detection which finds the correlation between head movements and scene background. They have used optical flow technique for motion detection. They measured relative movement intensity between user's head movement and scene background in sequence of images. There is high-correlation between fake head movements and scene background ROIs (Region of Interest).

The previous literature is grouped over two categories on the basis of the challenges performed to evaluate the liveness of the user. In the first category mostly researchers have analyzed the lip reading, challenges are thrown to repeat pass phrases or alphabets. While the second category analyze movement in the face, such as blinking eyes, change in facial expression and lip movements. These challenges can be performed by placing video imposter attack. Therefore the major challenge is to discard/ detect the video imposter attack. [153] have proposed motion magnification and multi feature aggregation to encounter these kinds of attack. The change in facial expression is magnified using the Euclidean motion magnification. They have combined both texture features (LBP [131]) and motion features (optical flow [154]) and used them in Support Vector Machine (SVM) for classifying the test input to one of the classes (Real and Imposter). [155] have used a light dependent resistor (LDR) for detecting the reflected light intensity while the eye blink pattern is used as a password. The combination of two are serving a test for liveness. Eye blinking password is the union of eye blink and chin movement. The tracking of the eye gaze and facial movement in the random environment is a complex job, therefore they kept the light source as structured.

The incremental growth of display and imaging devices, make the attack more accurate. The high definition display helps attackers to generate the replay attack in a much cleaner way. [156] have proposed a regression and incremental learning based eye gaze estimation which predict the eye gaze position with respect to screen. The difference between the predicted gaze location and the actual is treated as a fake score. More less the fake score, more chances to be treated as a real user and vice versa. [157] have proposed a novel spatial, temporal facial movement based authentication mechanism to detect the liveness. A 3D camera is used to decode the facial expression, change in facial expression is the result of movement in facial muscles. [158] kept eye blinking as a password. An ultrasonic range sensing device is used to detect the eye blinking while the addition of local facial features such as chin movement pattern is adding more complexity and making the system more secure. [159] proposed Dynamic Mode Decomposition (DMD) as a generalized feature to capture the movement either in the eye or face. DMD and LBP are used as the feature extraction while SVM serves as the classifier. The ability of DMD to represent temporal information of a video with a single image, saves the computation as well as the memory. The approach is well suited for the mobile devices having less memory and computation unit.

There is an inverse relation between the zoom and camera focal length, if the focal length is wide the zoom is less and vice versa. [160] have utilized the variable focal length of the sequence of face images to extract the essence of liveness. Focus, power histogram, gradient location and orientation histogram are used as features. All these features are fused together with the help of feature level fusion approach. The photo and video imposter attack, in both cases the material is represented/reprinted in front of the camera. If we assume (lambda) is the transformation noise present in transforming the real scene in the digital photo/video, then the recapturing of this digital photo/video will increase the noise again. This kind of noise is known as additive noise and serve as the discriminative factor between the real and the imposter. [161] have exploited and evaluated this quality difference and designed a content-independent noise signatures to test the liveness. Fourier transformation is used to transform the images over the frequency domain and perform the analysis.

### 2.5.3 Multimodal based Face Liveness Detection

Multimodal biometrics could be the other way to stop these attacks [127-129] [147-149], the permission is granted on behalf of the resultant score. For liveness verification, Chetty and Wagner [162] proposed multimodal fusion framework based on face-voice fusion techniques. Their framework consists of bi-modal feature fusion, cross-modal fusion, and 3D shape and texture fusion techniques. They verified their approach against imposter attack (photo and pre-recorded audio) and (CG animated video from pre-recorded audio and photo). Faces are detected by using face skin colour information and then the hue and saturation threshold are applied for finding lip position. As every human face has different complexion this threshold varies from person to person. Hence a proper tuning of threshold is required. Apart from this, although their approach reported good accuracy on the given dataset, it could create a problem when some objects appear in the background with the same colour as human skin. One limitation of their proposed paper is, it does not provide a generic mathematical model to determine fusion rules with the proposed features. In 2007, Faraj and Bigun [163] used lip-motion and speech for identity authentication. For biometric identification system, they presented motion estimation technique; they described normal image velocity widely known as a normal flow estimation technique used to extract features for audio-visual interaction. They used structure tensor to compute a dense optical flow. This method uses the number of frames to find out the velocity of moving points and lines, so that they can find out the movement and able to distinguish between the real and imposter. For the visual features, they have used lip region intensity. This paper fails to discuss the computation cost incurred in the experimental results analysis. In 2009, Shah et al. [164] proposed another multi-model fusion technique in which correlation between audio-video features are used during speech. This system utilized Gaussian mixture model (GMM) classifier and concatenates features to correlate between audio and visual data. They presented two other classifiers which have superior performance than conventional GMM and which have boosted feature-level fusion performance. The designed system is a real time and demands a low amount of memory. The only demerit of their approach is, if attacker comes with an authenticated person face image and his recorded voice then the attacker can access over the system. Lip-motion and speech are also utilized by Mycel-Isac et al [163], for liveliness detection. Lip motion is measured by a computing velocity component of the structure tensor by 1D processing. This technique does not depend on lip-contours for velocity computation therefore more robust features are obtained. For speech analysis, Mel Frequency features are extracted.

Other group of researcher used fingerprint, iris and palm print with face to block face spoofing attack. Gragnaniello et al. [165] and Akhtar et al. [166] used fingerprint, face and iris all together to identify the identity as well as to discard the face spoofing attack. Gragnaniello et al. [165] independently extracted features from all the biometric while the decision is aggregated. Akhtar et al. [167] have used Locally Uniform Comparison Image Descriptor (LUCID) as a feature extraction. Only one image is used for detecting the liveness as well as for face recognition. The only difficulty faced in the multimodal biometric is to perform the fusion of decision. The decision process is twofold, either the fusion is placed at the feature level or it is placed at the decision level. In [168] Telgad et al. have used fingerprint and face as multimodal biometric and defines the decision level fusion. Principal Component Analysis (PCA) is used as the feature extraction for both the biometrics while the minimum distance classifier is used for the classification. Although the matching score is normalized yet the fusion is biased and could serve negative in the presence of large intra class variation such as change in facial look and expression. In order to solve the issue of decision level fusion, generally a weighing factor is assigned to each biometric. The weight factor impact the final decision, hence the proper weight selection is a challenging task. Menotti et al. [169] utilized Artificial Neural Network (ANN) to determine the proper weight factor. The convolution network is used as the feature extraction while nine biometric spoofing benchmarks are used to test the reliability and efficiency of the system. [170] analyses image quality for fingerprint, iris and face images for the liveness detection.

The other group of researchers utilized [171], movement image quality as well as speech recognition to increase the accuracy of recognition as well as to discard the spoofing attack. The liveness of the user is determined based on the synchrony between the audio stream and the lip movement of the user. A pass phrase is spoken by the user and in account of this the system will track the movement of lips. ANN is used for the training while the Bayes classifier to classify the lip movement to their respective classes. [172] used palm print, face and iris together to prevent the spoofing attack with score level fusion. The score/decision level fusion is a complex job, therefore [173] they have designed a rule base system to perform the score level fusion. Equal Error Rate (EER) and Genuine Acceptance Rate (GAR) are the testing parameters to evaluate the experimental results. Profile faces are also sufficient for face recognition, although the recognition is challenging. [174] used profile faces and ears to detect the liveness as well as to recognize the person. Block-based Local Binary Pattern (LBP) is used as feature for both the biometrics. The feature level fusion of these biometric could lead the sparsity in the feature space. Therefore [175] proposed a multimodal sparse representation for representing the test samples in terms of the sparse linear combination of training samples. Due to the high intra class variation in biometrics, kernel functions are used to handle the non-linearity preset in the features.

## 2.6 Observation from Literature Review

- Criminal identification is the special instance of person identification. It is different from the classification face recognition due to its challenges and the application area.
- Although the face recognition based on multiple views is proposed, but none of the listed techniques has received satisfactory results.
- The face symmetry information is not utilized so far either for dealing these challenges or directly used this for recognition.
- Only few literature is listed for component based real time face recognition, moreover the localization of facial landmark is still a challenge. The question is also not answered why the face is divided to only eyes, nose and mouth why not to other components.
- Criminal identification based on criminal sketch images is overlooked, only few literature is available.
- The design framework for Sketch drawing humanoid robot is not available, only one effort is listed on the literature. The relation between the humanoid image plane and their body coordinate system is missing in the literature.
- The depth perception of the face is not properly utilized for detecting the liveness of the person.

# Chapter 3:

# Human Perception based Criminal Identification

*This chapter proposes a human-robot interaction based framework to solve the criminal identification problem. This new approach captures each eyewitness's visual perception by doing dialogue with eyewitnesses. It induces a criminal's physiological and facial characteristics which help in identification of the criminals. A web based knowledge capturing module has been added to capture the imprecise and vague perception of the eyewitness. A Decision Tree and Rough Set Theory based modeling is performed to predict the criminal only on the basis of rough description about the criminal.*

## 3.1 Introduction

Although the problem of criminal identification can be solved using the web based interface (demonstrated in Figure 3.3.1), but we have added a humanoid robot to make the system more user friendly. It is an accepted truth that an eyewitness is exhausted during the time of Police interrogation. A robot police (humanoid robot) would be the better replacement of the real police, which could acquire the information about the criminal and process them to find the possible matches through interactions. The proposed framework is experimented with one of the humanoid robot "NAO" (shown in Figure 3.3.2 ) in the real time over the students of Robotic and Artificial Intelligence Laboratory, Indian Institute of Information technology, Allahabad, India. According to the majority of students, interaction with NAO has been a pleasant experience in comparison to answering to any human or filling data in the web based interface.

A person can be categorized from their behavioural, physiological and facial attributes or the combination of all three [77]. The person's behaviour can be captured in terms of their gait patterns and way of speaking, their physic can be represented in the form of age, gender and height, while the face description can be summarized as face category, face tone, eyebrow's type, eye shape, nose shape and lip size. In this chapter, we have utilized all these attributes and talks about its contribution over forensic domain more precisely on criminal identification. The working envelope of forensic consists of two major blocks, evidence collection and later the reliable matching of these evidences [78-79]. The evidences collected in the form of the above said attributes specially the facial information about the suspect plays a vital role in criminal identification. The past and previous summary of that suspect can be filtered out just by matching the photograph collected from the digital sources from the existing criminal or citizen face database. Sometimes due to the absence of the digital media such as CCTV or camera, the face information about the suspect is not available. In all those cases, if any eyewitness exists, who has seen the crime could provide some description about the suspect. In all these cases, a police sketch artist communicates with the eyewitness and draw a portrait of the suspect [77-78]. The problem of manual matching of sketches to photographs is trivial and requires a large time.

The existing literature defines an automated way of matching these sketches to their respective photo images as mugshot detection. The first step towards the sketch to photo matching has been put by Roberg el al. in 1996 [82]. Although the experiment is performed on small database of 7 sketches and 17 real photographs, but they managed to prove that this process can be further enhanced. Later the significant contribution in this direction has been carried out by [83-84] [176]. Unfortunately, this problem has not received proper attention and requires further research in this direction. The existing solutions help only for constraint environment where the gap between the sketch and photo is less. The existing literature is not utilizing the other two features (behavioural and physiological) to trace out the victim. The addition of these two additional attributes can improve the recognition accuracy. These issues motivated us to look for other options which could be useful. If we closely analyze the process of sketch embroidery, we find that these sketches are the representation of imprecise knowledge of eyewitnesses. The same can also be recorded by verbal communication with eyewitnesses. Here, we have presented a prototype of each facial attribute which is helpful to capture the perception about a suspect. There are basically three different categories of these facial features discussed in the literature [177]. They are: (a) visual features which define the outline and shape of each facial attribute, (b) anthropomorphic features are the soft tissues on the surface of the face, mainly explored by plastic surgery surgeons, (c) cephalometric features and identifying features such as mole and cut marks are mainly studied by forensic experts. The prototype library has been set up after analyzing these categories and features.

We have proposed two expert systems based on the decision tree and rough set based theory to predict the suspect based on the given prototype descriptions. In order to mimic eyewitness we have invited students of Indian Institute of Information technology to volunteer in this project. We have also set up two different person's database to treat them as criminal database. The first database is made of Bollywood, Hollywood celebrities, Indian cricketers, politicians and students of Indian Institute of Information Technology, Allahabad (IIITA) while in the second we have used the benchmark datasets used in mugshot detection. In total we have collected 190 pictures for in-house database and collected 300 faces from the CUHK (Chinese University of Hong Kong), AR face database and from FERET dataset. We have invited 105 volunteers to give the perception about these faces. A web based interface and speech recognition based human robot interaction system have been developed to assist these volunteers to fill up the values of each prototype. We have considered two test cases to evaluate the performance of the proposed system. In Test case-I, we have shown the suspect's face in the web based interface and ask to pick the appropriate values for different facial and physiological features whereas in the second case, the photograph of the suspect is not visible. Only caption is available, in all these cases volunteers (eyewitness) have to recall his/her memory to give the description about all these prototypes. Since the processing of these features is a challenging task, we have proposed and experimented with two different approaches. Decision tree based classification and rough set based prediction. Both the approaches are giving the encouraging results in our in house dataset as well as the benchmark datasets. These systems could be used before mugshot module and avoid the process of sketch creation.

The rest of the chapter is summarized as follows: Section 3.2 presents the problem definition, followed by section 3.3 which describes the design issues of the system. This section also summarizes the background of decision based and Rough set based reasoning with their implementation on these issues. At the end of section 3.3, we present a proof of our hypothesis in terms of results and validations. Finally, section 3.4 and 3.5 concludes the chapter with advantages, limitation and future scope of this new way of criminal identification.

## 3.2 Problem Definition

In continuation of the mugshot detection, a visual perception based criminal identification has been proposed in this chapter. The proposed system has following challenges that need to be addressed.

- **Knowledge Acquisition**

  The first and foremost challenge faced to deliver these kinds of system is, how to extract the imprecise knowledge of eyewitness. What are the components which could be useful to contain this knowledge?
- **Knowledge Representation**

  The second challenge is to give a concise representation of the extracted data. The representation should reflect the knowledge and the transformation should be lossless.
- **Knowledge Processing**

  Once the knowledge is represented, the last question is how to model the system, so that we can get expected results. The model should derive and reflect the data given to it.

## 3.3 Proposed Methodology

The proposed system works on the rough and vague perception of eyewitness about the suspect. The main factors involved in this process are, how to capture the visual perception of eyewitness and how to represent the captured knowledge, so that it can be processed further. The solutions to these questions are discussed below.

### 3.3.1 Knowledge Acquisition

Knowledge acquisition is the first step towards modeling this problem. In order to extract valuable knowledge from the user we have created a web interface and a human robot interaction based communication setup. The web interface demands, values to be filled up by the eyewitness about the criminal’s physical and facial attributes. A snap shot of the designed system is shown here in Figure 3.3.1 and Figure 3.3.2.

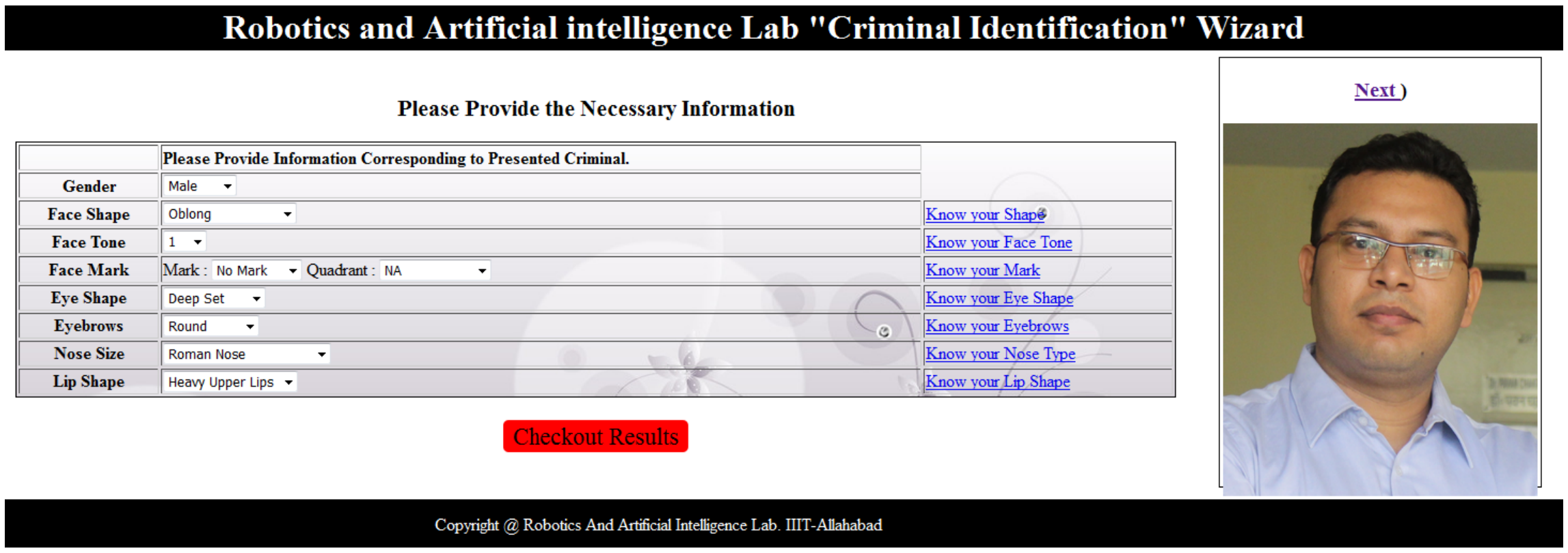

**Figure 3.3.1 Interface of the proposed system**

Figure 3.3.1 shows the experimental setup which has been used to collect information using a dialogue system. The system has three major parts, (a) system where the prototype of each facial feature is displayed, (b) Robot having speech recognition module which ask category/type of the feature and last (c) eyewitness who have the understanding of English language. The category of each facial feature prototype is represented by a numeric value ranging from one to the number of categories. The maximum number of categories for face mark is 13, and others have less than this. Therefore, in total we have 13 speech class, each class is representing the one category like 1, 2, 3…13. We have used the NAO’s speech to text conversion API [178] to convert the user input speech category (1, 2, 3...13) to the text.

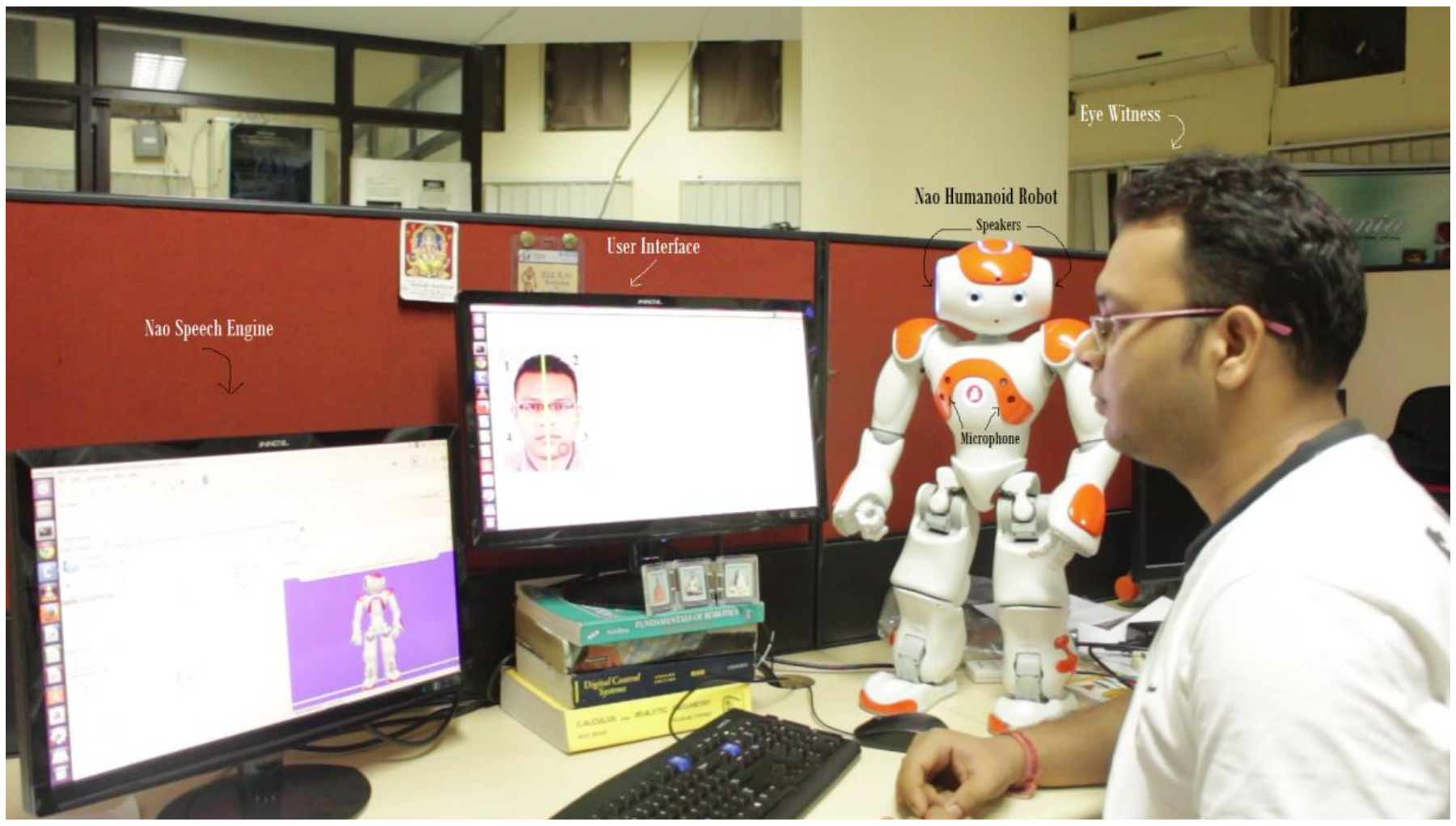

**Figure 3.3.2 Human-Robot Interaction based interface to capture eyewitness information**

NAO, the humanoid robot has microphones which captures and records the user input, it then sends the recorded speech file to its server and wait for the response. In response to this the server gives the text conversion, if the text conversion lies between 1 and 13. NAO, moves for the next feature specification otherwise ask the same feature description. The NAO speech to text conversion API is speaker invariant, hence it makes the framework generalize and effective. The accuracy of the API is also satisfactory which helps in further modeling the system.

Although every human face has facial attributes like eyes, nose, mouth, etc, it is also possible that two persons can have the same kind of nose or eyes (generally this is genetically transferred from father or mother to their children). But their variations and their combination with reference to each person, make the human face unique, thereby we can differentiate one from the other. Based on these attributes, features type they can be categorized into four different sets [177]. (a) Visual Features: set of features like eye, nose, facial outline. (b) Anthropometric Landmarks: Special points and location of the human face, which are considered by doctors for their study and experiments. (c) Cephalometric Landmarks: Shape and size of skull. (d) Special Features: They are the special marks or cut on the face, which usually does not match with the other person's face marks. We have tried to extract all four categories of feature set through our web interface. The feature description is presented one by one here.

A. **Face Type:** Based on the face anthropometric structure [179-180] a face can be categorized over one of the classes shown in Figure 3.3.3, based on its outer boundary.

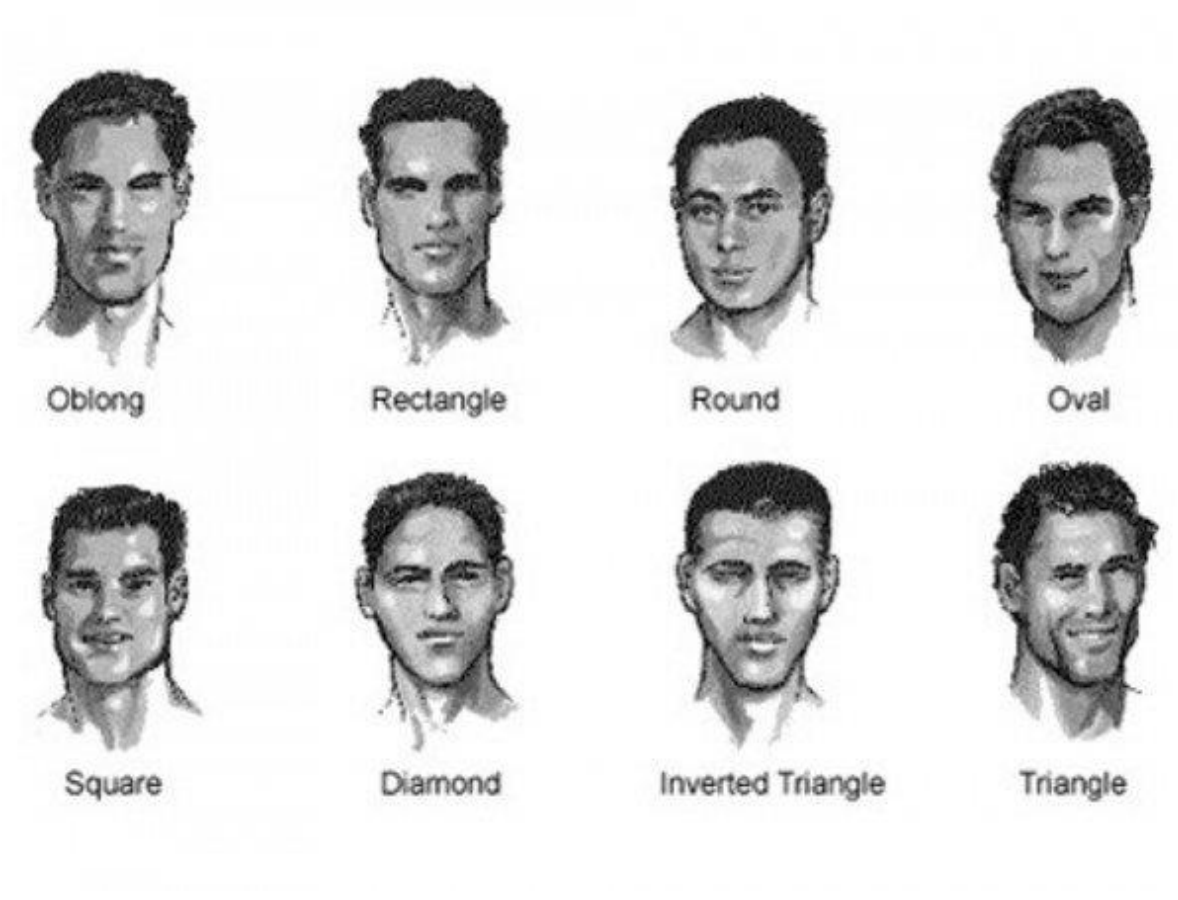

**Figure 3.3.3 Different Shapes of Face [180]**

B. **Face Tone:** Face tone speaks several things about one's origin, your geographical origin etc. [181] can be described by a face tone parameter presented below in Figure 3.3.4.

| | | | | | | | |
|---|---|---|---|---|---|---|---|
| | 1 | 10 | | | 19 | 28 | |
| | 2 | 11 | | | 20 | 29 | |
| | 3 | 12 | | | 21 | 30 | |
| | 4 | 13 | | | 22 | 31 | |
| | 5 | 14 | | | 23 | 32 | |
| | 6 | 15 | | | 24 | 33 | |
| | 7 | 16 | | | 25 | 34 | |
| | 8 | 17 | | | 26 | 35 | |
| | 9 | 18 | | | 27 | 36 | |

**Figure 3.3.4 Face Skin Tone [181]**

C. **Eye Shape:** As like the face tone and shape type Eye has also different kinds of shape, described below in Figure 3.3.5 [182].

*Deep Set Eyes:* Deep set eyes are usually bigger in size and set as deeper into the skull cause more salient brow bone.

*Monolid Eyes:* Monolid kind of eyes represents flat with respect to facial anatomy. No stronger crease and not very much prominent brow bones visibility.

Hooded Eyes: Brow is covered with an extra layer of skin resultant upper eye lids look smaller.

*Protruding Eyes:* Eyelids seem to be bulging from the eye socket. Such eye lids make eye attractive.

*Upturned Eyes:* They look like an almond shape eye with the kick at outer corner.

*Downturned Eyes:* They are same in shape as of upturned eyes only a drop in the outer corner.

*Close Set Eyes:* When the distance between two eyes are less between the eyeballs.

*Wide Set Eyes:* Just opposite the close set eyes, here the distance is more between eye balls.

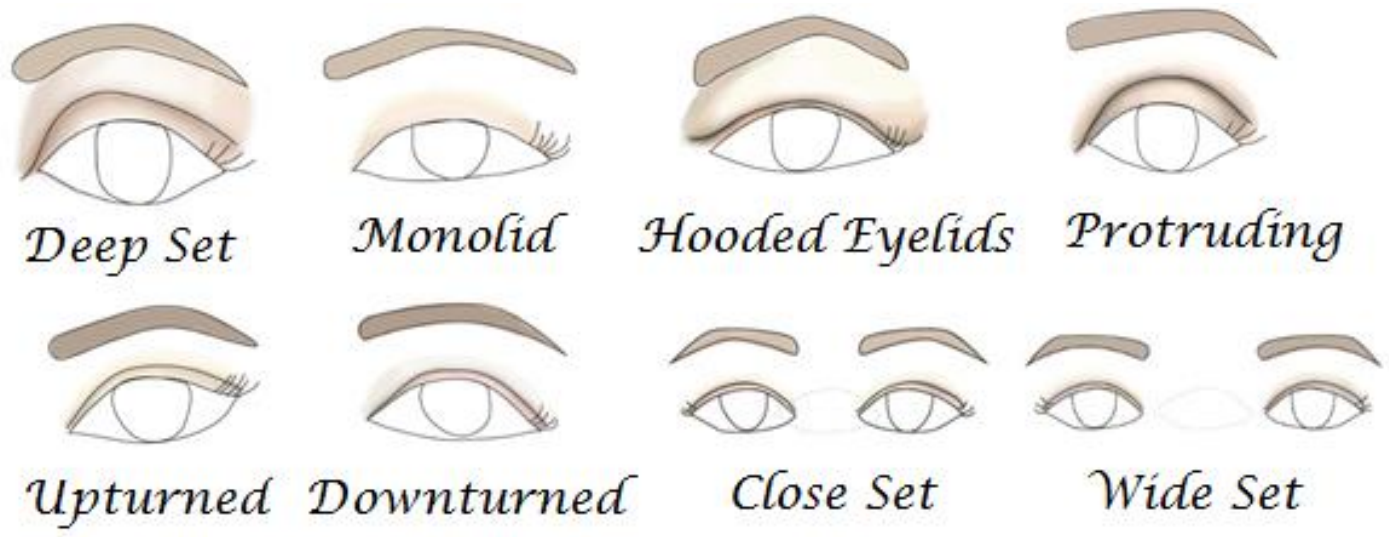

**Figure 3.3.5 Possible Eye Types [182]**

**D. Eyebrows Type:** There are nearly 5 types of category shown in Figure 3.3.6 It is possible that each category consists of several sub categories. But for the sake of simplicity, we are concerned about these main category descriptions and representations provided below [183].

*Rounded:* Rounded eyebrows are typically like a graph with respect to time, which monotonically grows up to a level and slow down after some time. The end point of the graph is always above the height of the starting position.

*Hard Angled:* This represents the equal change in x and y which symbolize a line. This line grows up to a limit and fall in the same proportion. The width of the curve decreases from its slope from where the curve goes down.

*Soft Angled:* They are same like hard angled but the first half is more in width than the hard angled.

*“S” Shaped:* The width is almost identical to the soft angle, but the shape represents a curve. This curve is having similarity with English alphabet “S”.

*Flat:* This is same like a flat curve with almost no changes. It gets affected only when we make any expression.

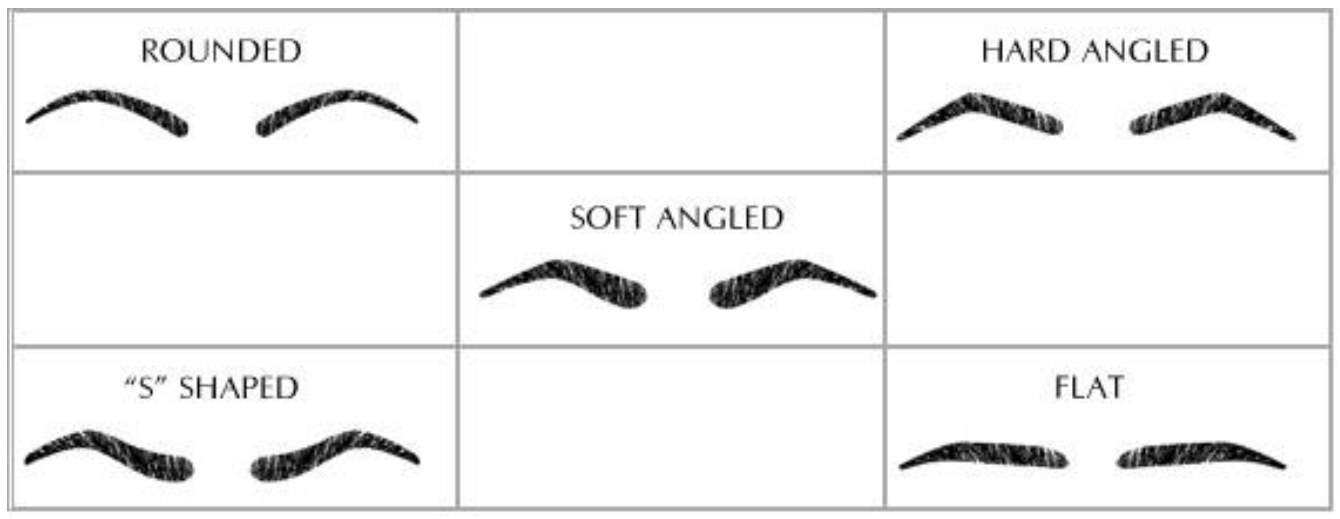

**Figure 3.3.6 Eyebrow Types [183]**

**E. Lips Type:** A complete list of Lip types are presented in Figure 3.3.7 [184]. There are total 6 categories of lips. These categories are formed based on the lip geometry as well as physical property.

*Heavy Upper Lips & Lower Lip:* The lower lip is dominated by upper lip and vice versa. It could be possible that they would be of varying size, but the main factor would always be present.

*Thin Lips:* As the name suggests the width of both the upper and lower lips are very less. They can be easily classified.

*Full & Wide Lips:* The full lips are the by default type, if any, property is dominated on this then the dominated category is assigned to the lip otherwise it is considered a full lips. If the lip width is more then it will be treated as wide lips.

*Round Lips:* They are compact in size. They form a round shape.

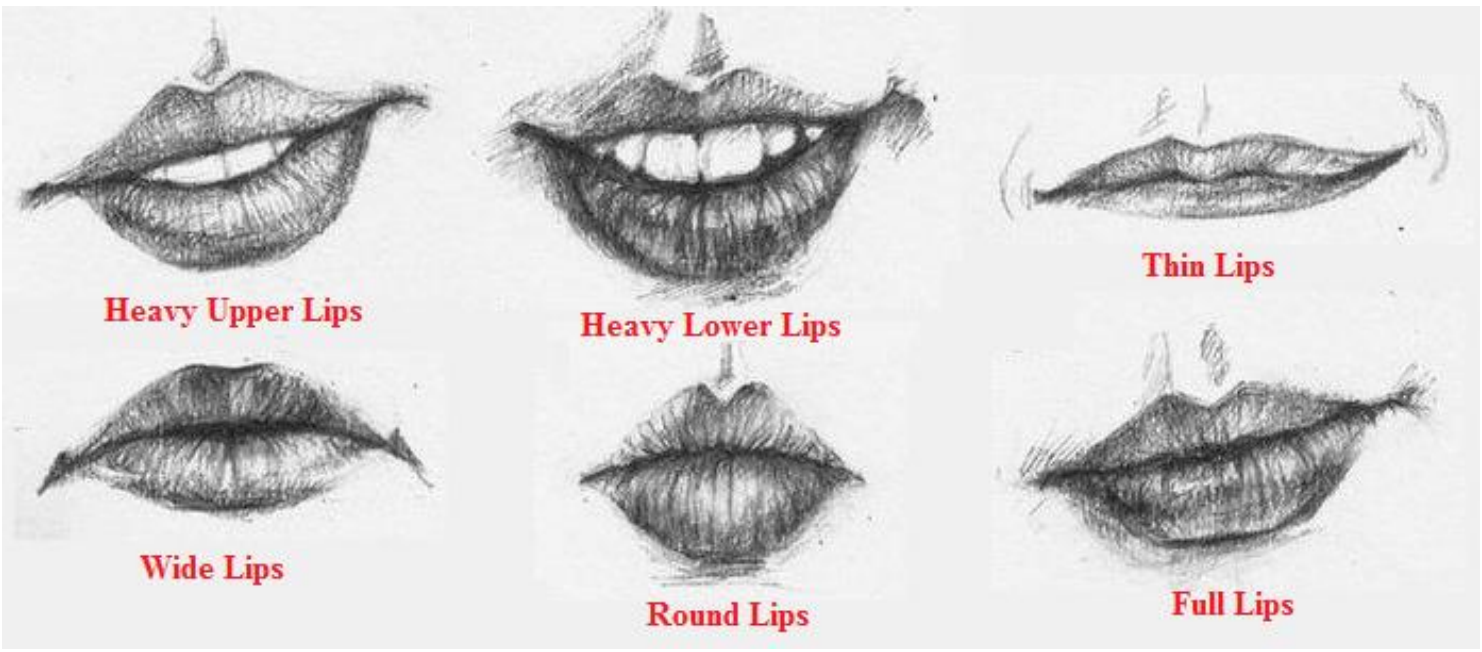

**Figure 3.3.7 Lip Shapes [184]**

**F. Nose Type:** Different kinds of nose shapes are possible. The shape of the nose depends on climate. Their shape and size varies in order to adapt the climate change. The other reason responsible for the variability is due to the different size of the ethmoid and maxillary sinuses in nose. These are the pockets below the eyes on both sides. In total we have explored 9 types of nose, presented in Figure 3.3.8 [183].

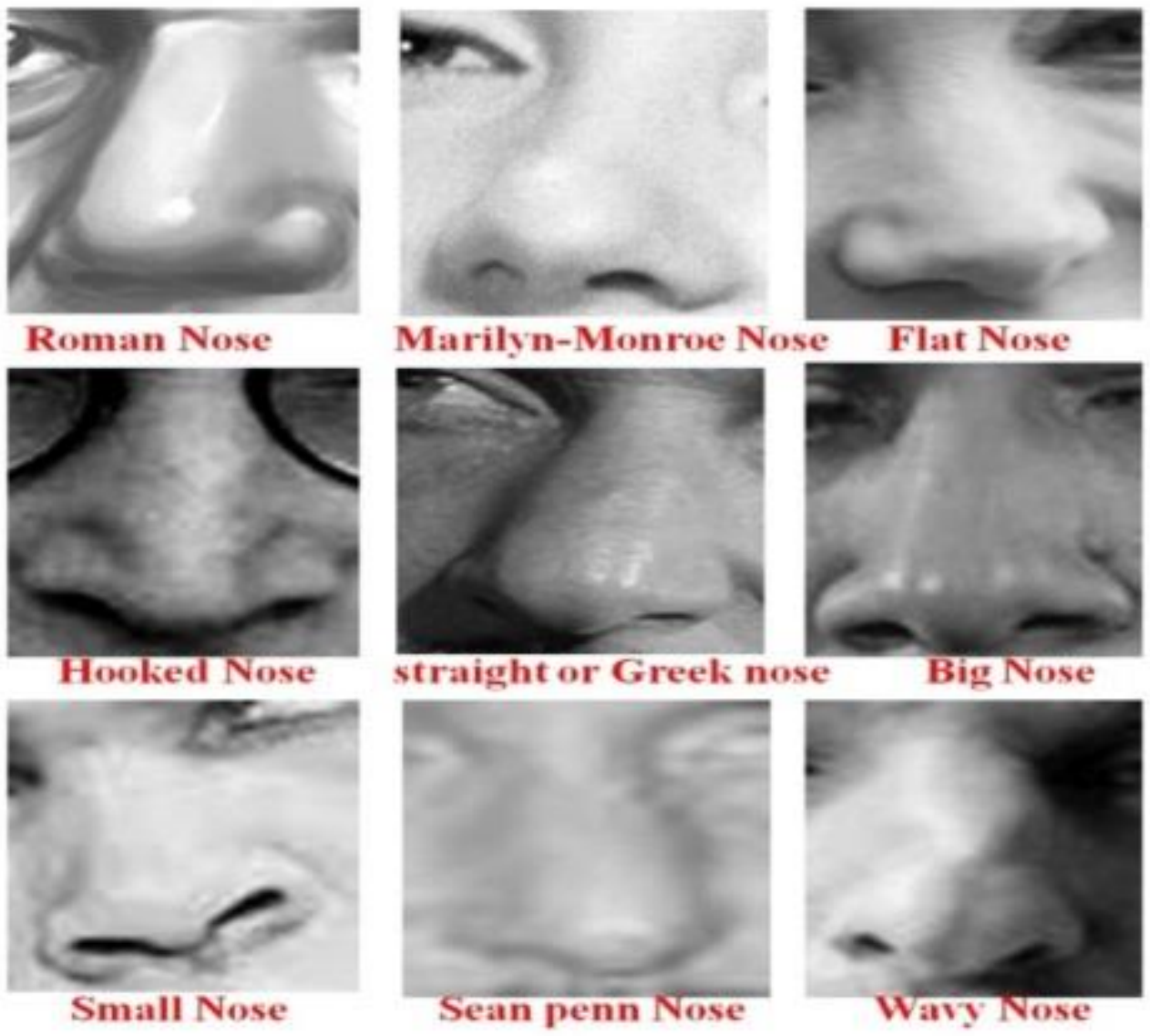

**Figure 3.3.8 Nose Types [183]**

**G. Special Marks:** Special marks could be of any cut or mole or any birthmark on the face. Sometimes these marks have special meaning based on the division of the face where they are located. In the realm of face reading the human face can be divided among 130 individual macro physical features or three micro features [185]. These micro features are named as celestial region (upper zone), self-will region (middle zone) and earthly region (lower zone). The shape and size of each zone are varying from person to person. Each zone has its unique property like upper zone explains about the imagination power of a person, middle zone states about the memory and lower zone describe about the observation quality of the person [185]. The fragmentation mechanism discussed in face reading realm motivated us to quantify the problem by dividing the face over the four regions. Here we are not claiming that face can only be divided over the four regions. We have divided it over four regions to deduce the complexity as well as to keep the maximum discriminative power. These are recorded in the form of their location, type of special mark and the size of mark. An example is shown in Figure 3.3.9.

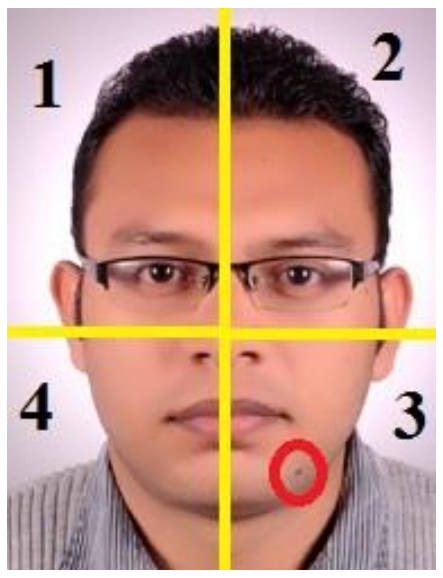

**Figure 3.3.9 Special Marks**

### 3.3.2 Knowledge Representation

All features and their values extracted in the previous unit of knowledge acquisition are raw and cannot be used directly in system modeling. Therefore, we need to transform these values over the set of simplified (Symbolic) values. These transformations help in setting the domain for each attribute set. Symbolic feature transformation for each feature set is discussed below.

**A. Gender Type:** Gender is key attributes which can reduce our search space. If the criminal is male we will search for only male criminals and vice versa. The transformation is shown below in Table 3.3.1.

**Table 3.3.1 Transformation for Gender Type**

| Attribute Type | Transformed Value |
|---|---|
| Male | M |
| Female | F |
| Domain size [1-2] | |

**B. Gender and Age Transformation:** We have imposed some constraint over age and height to represent this information in symbolic form. Age is represented in years while height is represented in foot and inches. The Transformation is shown in Table 3.3.2.

**Table 3.3.2 Age and Height Transformation**

| Attribute | Transformed Value | | | | | | | | |
|---|---|---|---|---|---|---|---|---|---|
| Age | 1 | 2 | 3 | 4 | 5 | 6 | 7 | 8 | 9 |
| Constraint | 15-20 | 21 - 25 | 26-30 | 31-35 | 36-40 | 41-45 | 46-50 | 51-55 | 56-60 |
| Height | 4.5-4.8 | >4.8 - 5.1 | >5.1 -5.4 | >5.4 - 5.7 | >5.7 - 6.0 | >6.0 - 6.3 | >6.3 -6.6 | | |

**C. Face Type:** There are 8 possible categories for face. Each category is symbolized by a numeric value shown in Table 3.3.3. In this case the domain for face attribute will be (1-8).

**Table 3.3.3 Transformation of face type**

| Attribute Type | Transformed Value |
|---|---|
| Oblong | 1 |
| Rectangle | 2 |
| Round | 3 |
| Oval | 4 |
| Square | 5 |
| Diamond | 6 |
| Inverted Triangle | 7 |
| Triangle | 8 |
| Domain size [1-8] | |

**D. Face Tone:** A wide variety of face tone is possible. It is not logical to assign particular values for each type. Therefore, with the help of a beauty expert we have categorized the face tone into several classes based on Von Luschan's chromatic scale [181] shown in Table 3.3.4.

**Table 3.3.4 Transformation for Face Tone using Von Luschan's chromatic scale**

| Attribute Type | Probable Class | Transformed Value |
|---|---|---|
| 1-5 | Very Light or White | 1 |
| 6-10 | Light | 2 |
| 11-15 | Light Intermediate | 3 |
| 16-21 | Dark Intermediate or Olive Skin | 4 |
| 22-28 | Dark or Brown Type | 5 |
| 29-36 | Very Dark or Black Type | 6 |
| Domain size [1-6] | | |

E. **Eye shape, Nose Type, Lip Type and Eyebrow Type:** As like the face type we can apply the linear transformation to Eye shape, Nose type, Lip type and Eyebrow type to decode Transformed values. The transformation is described Table 3.3.6, Table 3.3.5, Table 3.3.7 and Table 3.3.8 respectively.

**Table 3.3.5 Eye Shape Transformation**

| Attribute Type | Transformed Value |
|---|---|
| Deep Set | 1 |
| Monolid | 2 |
| Hooded | 3 |
| Protruding | 4 |
| Upturned | 5 |
| Downturned | 6 |
| Close Set | 7 |
| Wide Set | 8 |
| Domain size [1-9] | |

**Table 3.3.6 Nose Type Transformation**

| Attribute Type | Transformed Value |
|---|---|
| Roman Nose | 1 |
| Monroe Nose | 2 |
| Flat Nose | 3 |
| Hooked Nose | 4 |
| Greek Nose | 5 |
| Big Nose | 6 |
| Small Nose | 7 |
| Sean penn Nose | 8 |
| Wavy Nose | 9 |
| Domain size [1-9] | |

**Table 3.3.8 Lip Transformation**

| Attribute Type | Transformed Value |
|---|---|
| Heavy Upper Lips | 1 |
| Heavy Lower Lips | 2 |
| Thin Lips | 3 |
| Wide Lips | 4 |
| Round Lips | 5 |
| Full Lips | 6 |
| Domain size [1-6] | |

**Table 3.3.7 Eyebrows Transformation**

| Attribute Type | Transformed Value |
|---|---|
| Round | 1 |
| Hard Angled | 2 |
| Soft Angled | 3 |
| "S" Shaped | 4 |
| Flat | 5 |
| Domain size [1-5] | |

**F. Special marks:** Special marks also play an important role in identification. We have used three kinds of special marks; Birthmark, Mole and Cut mark for the sake of representation. They can be present anywhere on the face, hence we have divided the face into 4 quadrants. The quadrants are represented by Symbol 1/2/3/4 and marks are represented by Mole (M), Cut(C) and Birthmark (B) respectively. The transformation design is summarized in Table 3.3.9.

**Table 3.3.9 Special mark Transformation**

| Attribute Type | Transformed Value | Attribute Type | Transformed Value |
|---|---|---|---|
| No Mark | 1 | 3M | 8 |
| 1M | 2 | 3C | 9 |
| 1C | 3 | 3B | 10 |
| 1B | 4 | 4M | 11 |
| 2M | 5 | 4C | 12 |
| 2C | 6 | 4B | 13 |
| 2B | 7 | | |
| Domain size [1-13] | | | |

**Table 3.3.10 Attributes, Labels and Domains: a summarized representation**

| Features | Attribute | Domain |
|---|---|---|
| $f_1$ | Gender | [1-2] |
| $f_2$ | Face Type | [1-8] |
| $f_3$ | Face Tone | [1-6] |
| $f_4$ | Eye Shape | [1-9] |
| $f_5$ | Eyebrow Type | [1-5] |
| $f_6$ | Lip Type | [1-6] |
| $f_7$ | Nose Type | [1-9] |
| $f_8$ | Special Mark | [1-13] |
| $f_9$ | Age Type | [1-9] |
| $f_{10}$ | Height Type | [1-7] |
| $Class$: 240 classes Each class represent one criminal | | |

In a nutshell, we have used 8 types of attribute to define a person. One physical attribute and, 7 facial attribute. Each attribute has its own domain size and way of representation. But for making the whole system simple, we have generally used a linear transformation. Table 3.3.10 shows the summarized representation of these attributes. This table helps us to recall the attribute definition and domain.

### 3.3.3 Knowledge Processing

There are two major modules in the proposed system shown in Figure 3.3.10. The first module (training module) labels each criminal face with the help of some symbolic transformations. The criminal face database is labeled with the help of a beauty expert. The database is labeled on the basis of criminal's facial features such as face shape, face colour, face tone, eyebrow type, eye type, nose type, lip tone, lip type and anthropomorphic features such as gender, height, age. At first these facial attributes are filled up by the beauty expert after analyzing criminal photos while the anthropomorphic attribute values are already known (generally the anthropomorphic attributes are measured by the Police when a criminal is arrested before). These values are further transformed by the knowledge representation module. These transformed values are known as the feature vector. For each criminal face one feature vector is stored in the symbolic criminal database. Both the criminal photo database and the criminal symbolic database are connected with the help of reference key. The second module (testing module) also captures the imprecise knowledge of the eyewitness with the help of our designed web interface shown in Figure 3.3.10. The interface asks different queries about suspect physical appearance and facial attribute. We believe that a person can be categorized and differentiated based on these attributes [77-79]. Each attribute has different types associated with it which is required to represent any subject. Eyewitness can select the appropriate attribute type. The collected information is modeled using a query based system to find the best possible matches. As a result the system shows top 10, top 20 and top 30 results from the photo database. A testing framework is designed which consists of 240 so called criminal faces which includes 110 Indian Bollywood celebrities dataset, 40 Indian Cricket team dataset and 90 from Robotics and Artificial Intelligence Lab dataset.

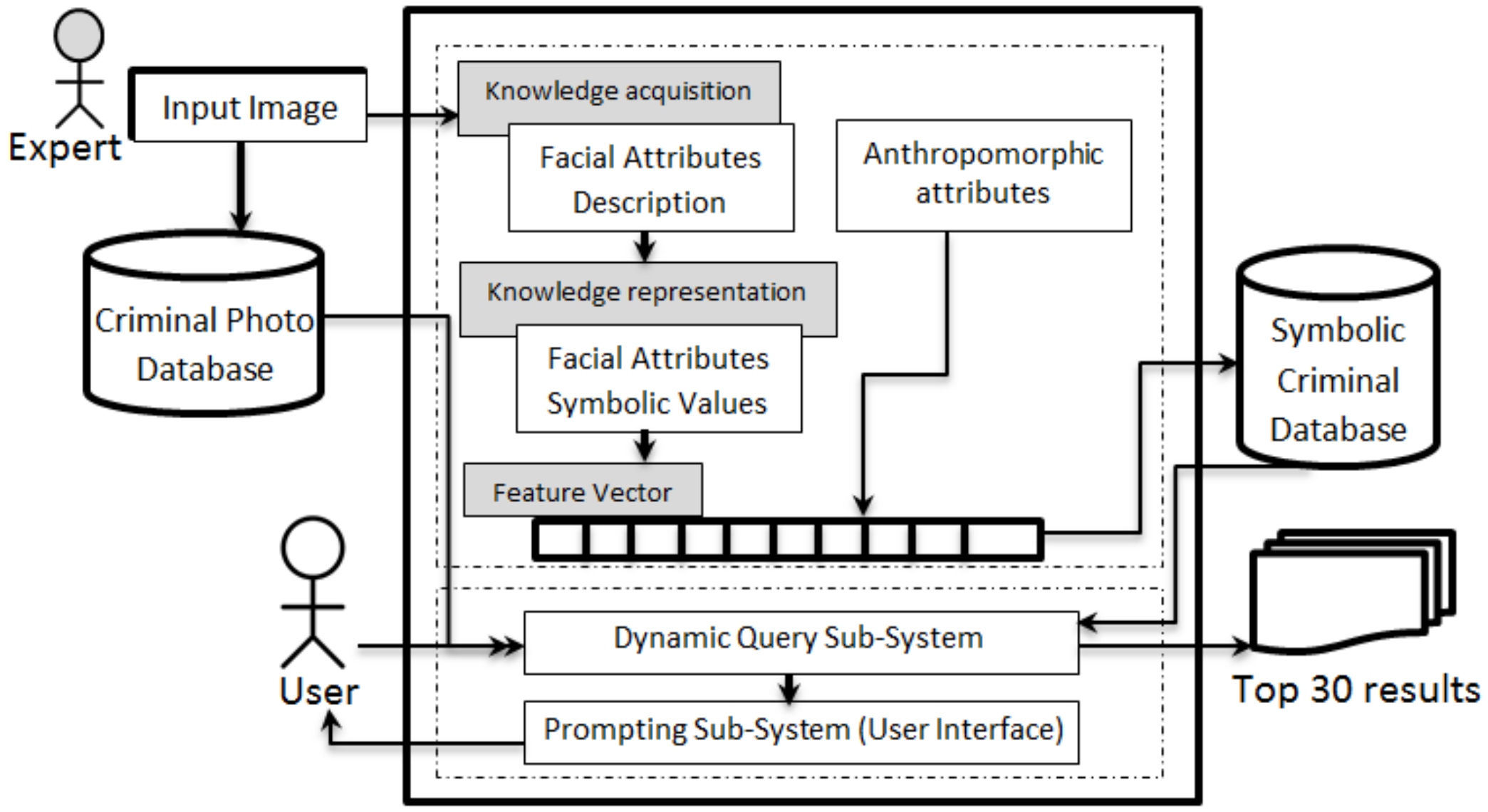

**Figure 3.3.10 Proposed Query based Expert System**

#### 3.3.3.1 Justification of the proposed framework

The proposed framework shown in Figure 3.3.10, has three key pillars. Knowledge acquisition, knowledge representation and knowledge processing which are analogous to the feature extraction, feature selection and classifier design of existing mugshot detection approach [84-85]. The knowledge acquisition module helps to setup the feature library. But, before setting up the feature library number of features should be known. Majority of researchers have used eye, nose and mouth, as kernel features to recognize the person while the other facial features such as face tone, face shape are used as additional set of features. The above said features have been reported as the discriminative features, therefore we have utilized all these feature set to form our feature library. Facial and physiological both type of features are used to create the feature vector. A feature vector is the collection of attribute values which can represent any person. A transformation has been applied on these features to represent the vague inference in terms of crisp values. We have used linear transform to decode these values, while a dynamic decision tree has been formed for each query. The decision tree clusters the population on the basis of precedence of attributes, while the matching is performed using the Hamming Distance.

In some cases when the desired results are not received, we can further change the attribute values of some particular feature for the betterment of the result. The proposed architecture directly utilizes the vague perception, therefore sometimes due to the perception biasness, it could lead to the unwanted results. Therefore, in further solution we have applied rough set to handle the uncertainty present in the human perception.

#### 3.3.3.2 Database Creation and labelling

This is the kernel step of our proposed experimental setup. The database generation is divided into two modules. The first module is for collecting the public and local faces and their anatomical structure specification. We have used 110 Bollywood celebrities including 53 males and 57 females + 40 Indian cricketers + 90 students of Indian Institute of Information Technology, Allahabad to represent the database. A glimpse of the database is shown in Figure 3.3.12. Only those photographs are selected from the internet sources which have clear visibility of all the facial macro features. As we can see the first row of Figure 3.3.12 represents the female population of Bollywood celebrities, the second row shows the male community. Cricketers and our in home database are getting reflected in third and fourth rows respectively. The second module is to label the designed database using symbolic notations discussed in the previous section. The guidance of the expert and some student volunteers has been used for database labeling. For example, given a picture of a lady shown in Figure 3.3.11, the expert can make a guess about her facial and anatomical attributes shown below in Table 3.3.11. We also invited students of Indian institute of information Technology, Allahabad to contribute their views (appearances) for the same. We are treating these volunteers as eye witnesses. Later the performance will be tested on the views collected from these eyewitnesses. A description about the labeled database is presented in Table 3.3.12. The first group separated by a line in Table 3.3.12 symbolizes Bollywood celebrities. The second group pointed to Indian cricketer's database and the third group shows records from our in-house database.

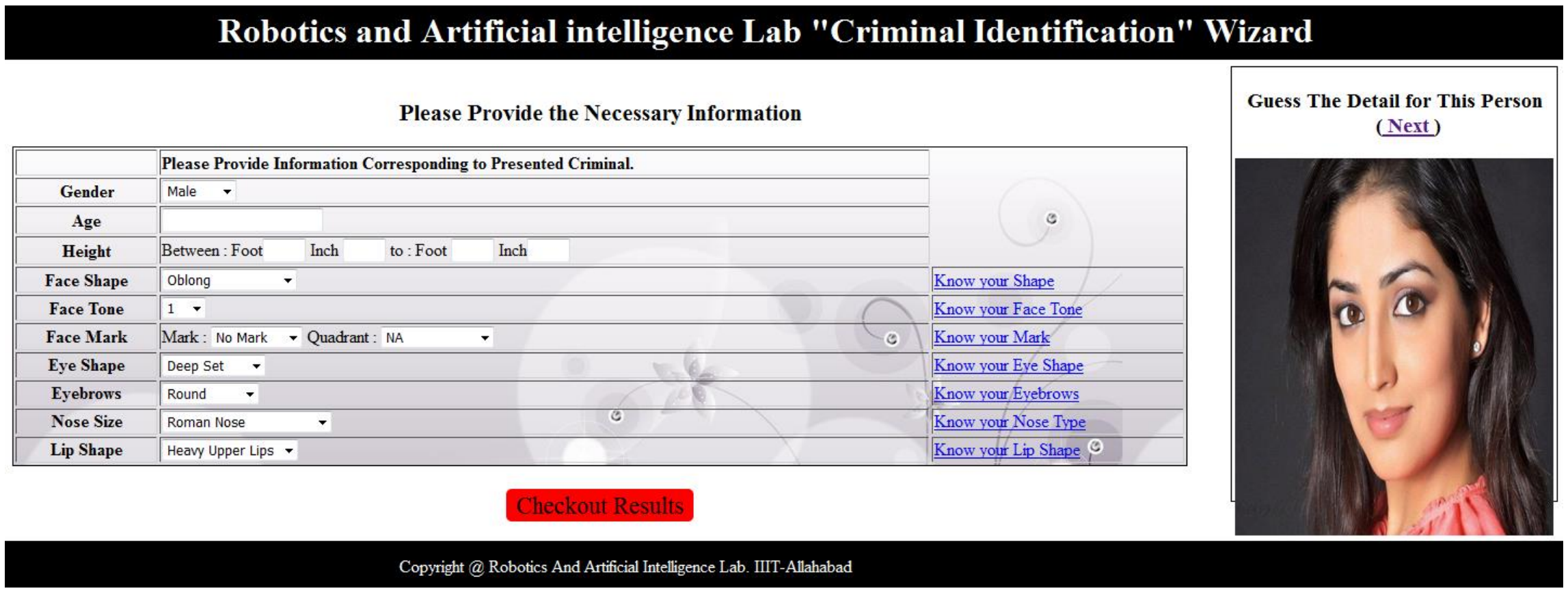

**Figure 3.3.11 Proposed System with additional attributes**

**Table 3.3.11 A sample from the symbolic database w.r.t. Figure 3.10**

| **Attributes** | Gender | Age years | Height cm | Face Type | Face Tone | Eye Shape | Eye-brows | Lip Type | Nose Type | Special Mark |
|---|---|---|---|---|---|---|---|---|---|---|
| **Values** | Female | 28 | 162 | Oval | 7 | Hooded | Round | Full | Roman | No |
| **Symbols** | F | 28 | 162 | 4 | 2 | 2 | 1 | 6 | 1 | 1 |

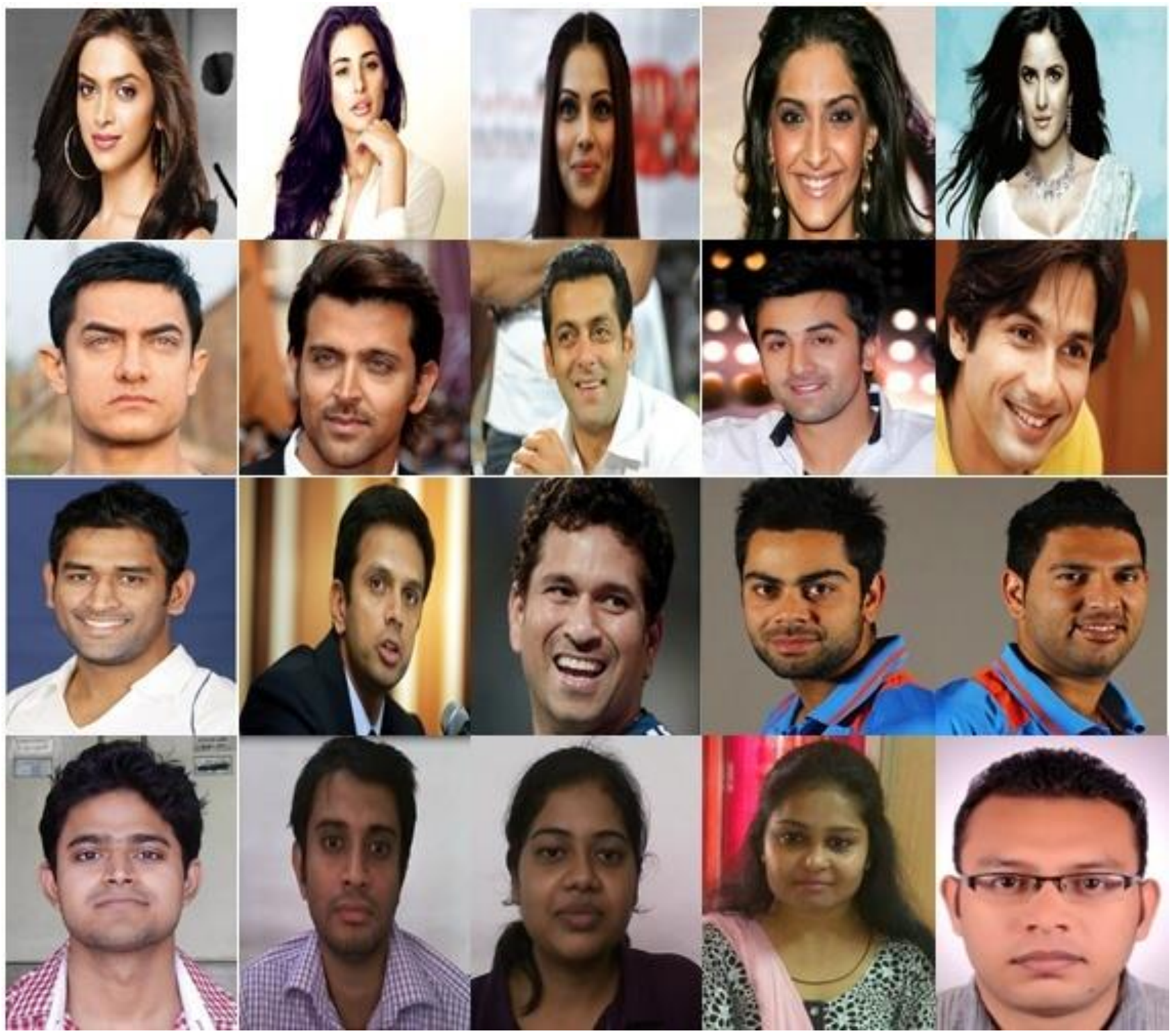

**Figure 3.3.12 Sample of In-House Database**

### 3.3.3.3 Query Design and Optimization

When the criminal database increases in size it is required that only concerned records should be displayed to the user. Therefore the query for extracting these records should be designed in such a way that it should give the minimal matched records [186]. It has been analyzed and seen during several time testing on this project that all 10 features (please see Table 3.3.10 for feature specification) are not having the same weightage. Some special features have to be set which optimizes the query and search results [187]. Here we have considered gender, age and height of the suspect as key features. Most of the records are filtered out by gender discrimination. The rest of these are further optimized by placing the age and height constraints. The fetched results depend on the window of age and height. More accurate the guess more accurate search results. The search query will work as shown in Figure 3.3.13.

**Table 3.3.12 Sample of our in house, symbolic database after transformation**

| Subjects | Sex | Age years | Height (cm) | Face Type | Face Tone | Face Mark | Eye Type | Eye brows | Nose Type | Lip Shape |
|---|---|---|---|---|---|---|---|---|---|---|
| S1 | F | 33 | 162.56 | 1 | 3 | 1 | 4 | 3 | 5 | 2 |
| S2 | F | 28 | 170.18 | 6 | 4 | 1 | 3 | 1 | 5 | 2 |
| S3 | F | 26 | 175.26 | 4 | 3 | 1 | 4 | 2 | 9 | 4 |
| S4 | F | 46 | 162.56 | 4 | 3 | 1 | 4 | 2 | 5 | 4 |
| S5 | F | 21 | 165.10 | 1 | 3 | 1 | 1 | 4 | 9 | 4 |
| S6 | M | 47 | 172.72 | 4 | 4 | 1 | 2 | 4 | 2 | 3 |
| S7 | M | 32 | 172.72 | 6 | 3 | 1 | 2 | 1 | 2 | 2 |
| S8 | M | 56 | 175.26 | 4 | 4 | 1 | 4 | 5 | 8 | 2 |
| S9 | M | 70 | 199.64 | 4 | 4 | 1 | 6 | 4 | 5 | 4 |
| S10 | M | 31 | 182.88 | 4 | 3 | 1 | 4 | 4 | 5 | 2 |
| S11 | M | 26 | 175.00 | 3 | 3 | 1 | 3 | 2 | 3 | 3 |
| S12 | M | 27 | 188.00 | 5 | 4 | 1 | 1 | 3 | 8 | 2 |
| S13 | M | 41 | 165.00 | 4 | 4 | 1 | 1 | 4 | 4 | 1 |
| S14 | M | 41 | 180.00 | 4 | 3 | 1 | 3 | 1 | 1 | 4 |
| S15 | M | 27 | 182.88 | 6 | 4 | 1 | 2 | 4 | 2 | 3 |
| S16 | M | 22 | 160.20 | 8 | 4 | 1 | 1 | 5 | 2 | 5 |
| S17 | M | 22 | 180.34 | 2 | 4 | 1 | 2 | 3 | 8 | 3 |
| S18 | F | 21 | 177.80 | 4 | 5 | 8 | 8 | 6 | 5 | 6 |
| S19 | F | 23 | 165.10 | 5 | 3 | 3 | 5 | 4 | 2 | 4 |
| S20 | F | 26 | 157.48 | 5 | 5 | 1 | 7 | 3 | 4 | 6 |

The lower 7 features $\{f_2, f_3, f_4, f_5, f_6, f_7, f_8\}$ are considered as the vague features and therefore we have assigned them lower significance. The matching score is calculated based on the hamming distance between the train and test features. Two symbolic databases have been created to show the efficiency of the proposed system. The first database which is also known as the training database is established with the help of a beauty expert, while the second database is the collection of views of the general users. The query is crafted in such a way so that only the concerned record will come from the training database. Usually, the training database is crawled based on the test views about the criminal's gender, age, and height attributes. The rest of the features such as face type ($f_2$), face tone ($f_3$), face mark ($f_4$), eye type ($f_5$), eyebrows ($f_2$), nose type ($f_2$) and lip type ($f_2$) are projected as a result of these three pioneer features. The query used for extracting only concerned records are presented below.

$$\pi_{f_2,f_3,f_4,f_5,f_6,f_7,f_8}\left(\sigma_{\substack{gender=t_{gender}\ and\ age\geq t_{age-5}\quad age\leq t_{age+5} \\ and\ height\geq t_{height-5}\ heigh\leq t_{height+5}}}(DB)\right)$$

Here $t_{gender}, t_{age}, t_{height}$are the gender, age and height for the test views. We have created window of 5 years and 5 centimeters for age and height, because most of the time the approximation about the criminal's age and height is not accurate for test views. The above query will extract "$n$" number of records with respect to the given test features.

**Matching Technique:** The matching or similarity between the train and test perception about the criminal is performed with the help of hamming distance. Hamming distance is applied to the fixed vector length and it gives the minimum mismatch between two same length vectors. Since the length of the vector is fixed, we have applied it to find the minimum mismatch between the test vector and existing vector library. The captured perception totally relies on eyewitness's memory recall, therefore, it is vague. The hamming distance is calculated between the presented test samples and extracted "n" records from the training database based on the above query. If the test feature vector is symbolize by $\{ft_2, ft_3, ft_4, ft_5, ft_5, ft_6, ft_7, ft_8\}$ and the train features are represented by $\{f_2, f_3, f_4, f_5, f_6, f_7, f_8\}^i$ where i reflect the record number and it belongs to$i \in [1, n]$. Let's take an example how hamming distance will help to find a match. Whoever has the highest matching will be considered as the probable match case.

$$FTest = \{ft_2, ft_3, ft_4, ft_5, ft_5, ft_6, ft_7, ft_8\} \qquad (3.1)$$

For instance, let the test features set have the values.

$$FTest = \{1,3,1,4,3,5,2\} \qquad (3.2)$$

Similarly Training set have:

$$FTrain = \{f_2, f_3, f_4, f_5, f_6, f_7, f_8\} \qquad (3.3)$$

For instance, let the train features set have the values.

$$\text{FTrain} = \{1,3,1,1,4,9,4\} \qquad (3.4)$$

The hamming distance between (3.2) and (3.4) is basically the number of places these two vectors differ.

$$\text{FTest} = \{1, 3, 1, 4, 3, 5, 2\}$$

$$\text{FTrain} = \{1, 3, 1, 1, 4, 9, 4\}$$

Hamming Distance (HD) = 4

Match Score= $\frac{7-4}{7} * 100$ // Matching Score = $\frac{\text{FeatureLength} - \text{HD}}{\text{FeatureLength}} * 100$

Which is 3/7*100= 23.34%. The match score shows the possibility of a match with the inside database on the basis of given test input. We can put a threshold here to eliminate some of the unwanted results. It will display only those records whose Matching Score>=threshold (Here the threshold is 14.28).

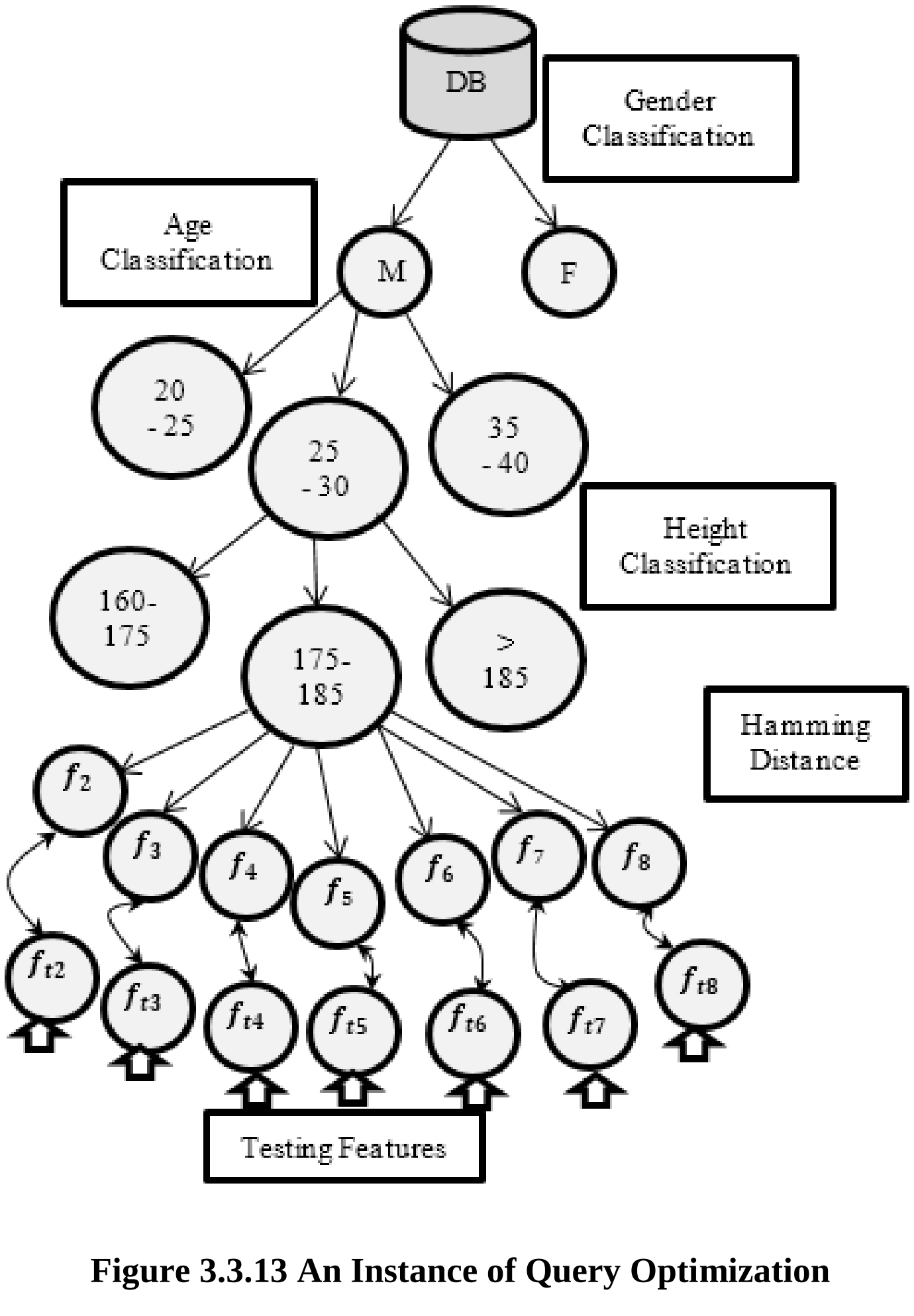

**Figure 3.3.13 An Instance of Query Optimization**

The proposed system is modelled on the basis of 7 facial and 3 anthropomorphic features. The anthropomorphic feature set consists of features like gender, age and height while facial feature set includes values such as face type, face tone, face mark, eye type, eyebrow type, nose type and lip shape. The facial features are considered to be vague and hence they have lower significance. The similarity score is computed using only these 7 facial attributes. So, if all the attributes are matched the accuracy would be 100% and if only one is matched the minimum accuracy would be 14.28%. We can define any value between 14.28 and 100 which will act as a threshold.

The figure 3.3.13 represents an instance of query execution. Relating to the above figure 3.3.13 let's have a situation where we have to search a Bollywood actress presented in Figure 3.3.11. The query is optimized by first discriminating male from female. In our database we have total 97 females (57 female celebrities and 40 female students) out of 240 populations. As we can see the first attribute reduced the search space from 100% to 40%. The next parameter age, again, reduces this search span. We have 63% population of the class female which have the average age between 20 and 30. After applying the age filter, we were left with only 15% search space.

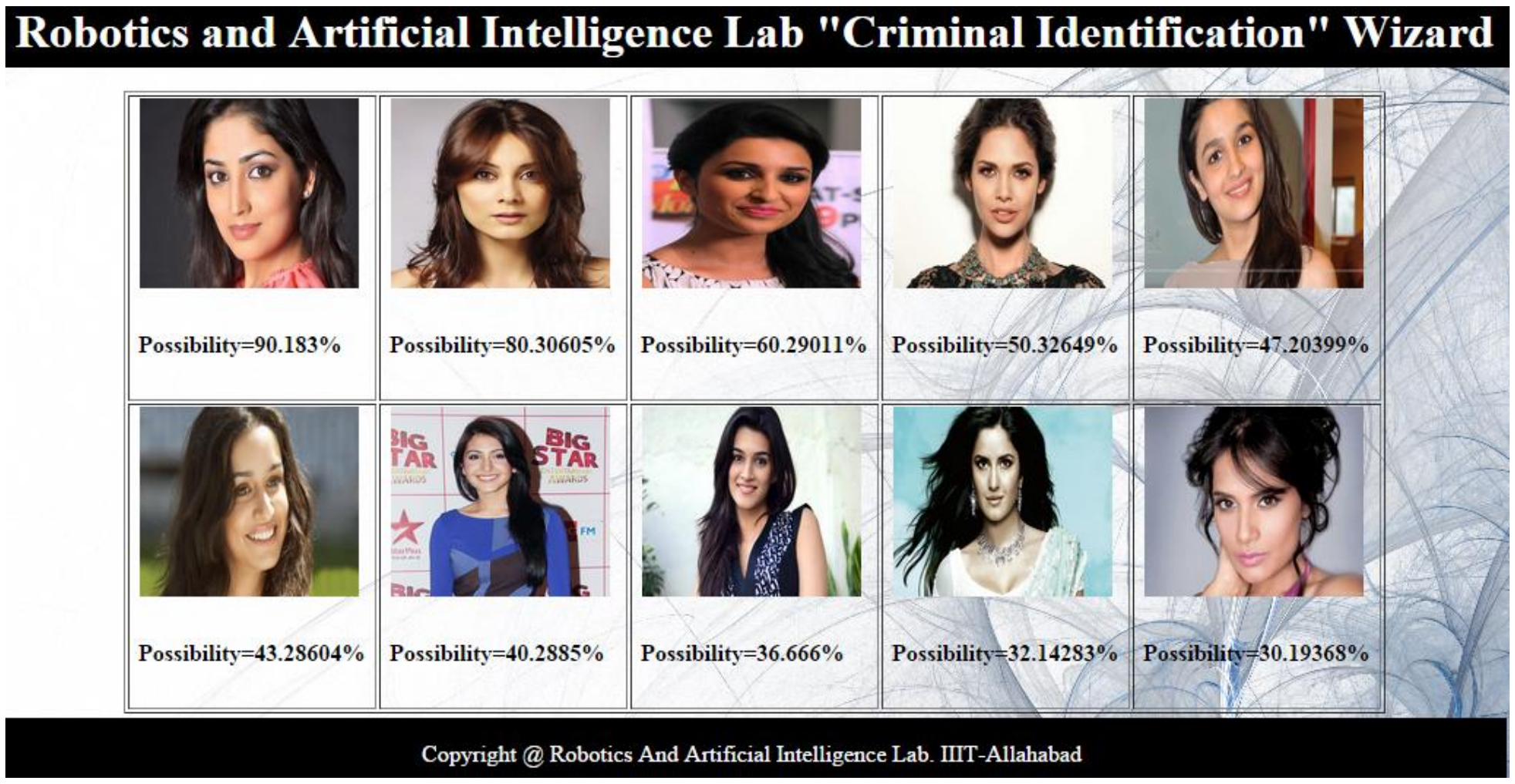

**Figure 3.3.14 Matched results from the labeled database**

The height factor further reduces this search span by putting height constraint. At last whatever the population, we had, we apply hamming distance between the train and test samples to find the match score. The records were then sorted based on their match score. We have presented the records in the form of top 10, top 20 and top 30. The result of the same query and for the given test face (please refer to Figure 3.3.11) is shown in Figure 3.3.14. It could be also possible that during the process of knowledge capturing, eyewitness is not sure about some of the facial as well as anatomical features of the criminal. In order to overcome this, we have introduced the notion of trusted features. Trusted features are those features in which eyewitness is more confident than other features. In such circumstances the decision tree is generated based on only the trusted features. Suppose, the eyewitness is only sure about criminal's gender, age, height, face and cut mark features, then the tree is constructed only using these features.

**Result and Discussion**

We have designed two test cases analogous to the viewed and forensic sketches [85]. In the first test case the photograph of the guilty is visible to the eyewitness and she/he can describe the attributes on the basis of presented photograph. In the second test case (presented in Figure 3.3.15) the photograph of the guilty is absent and shown one day before the test. Based on these test cases the matching performance and the validity of the proposed system have been tested. We took help from our Robotics and Artificial Intelligence Laboratory's students to act like eyewitness and to provide the description of these pseudo criminals. We have invited 5 views per criminal for both the test cases (viewed and forensic). Total 80 students participated in our pen test and each has contributed the views for 15 criminals. We have created a database of 1200 views per test case. The aim of introducing query optimization is to reduce the search space. We have considered three testing parameters (a) Size of search space (b) Match score and (d) False Negative (FN) to evaluate the performance of our proposed system. It is desired that the search span is minimum while the system shows the maximum match with the respective face in the database. False Negative are those cases where the criminal is present inside the database but the system is not including it inside the search space. This shows the failure of the system. Here we are not concerned about the false positive (FP), as these results are discarded by the eyewitness. The results are decomposed in the form of top 30 matches ranked with respect to their possibility (please refer to Figure 3.3.14).

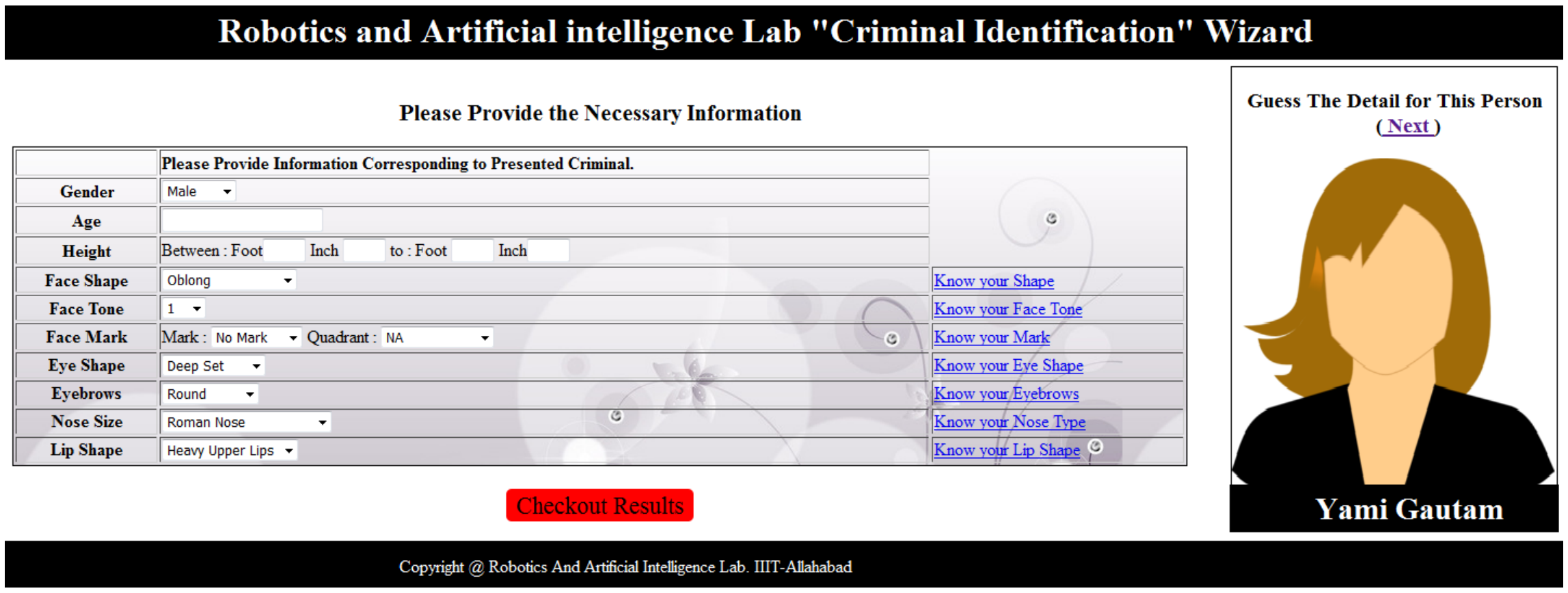

**Figure 3.3.15 Test Case-II (Photograph of the person is absent only caption is available)**

The top 30 identities are covered to minimize the chance of false negatives. The results in both the test cases are summarized in Figure 3.3.16. There are three abbreviations used in figure 3.15, Top 10, Top 20 and Top 30. The meaning of the top 10 is that the matched face fall into the first 10 records displayed to eyewitness. Top 20 and top 30 represents the next 10 records after the top 10 and top 20 respectively. A false negative (FN) of 2.33% and 8.42 % has also been reported in test case-I and test case-II.

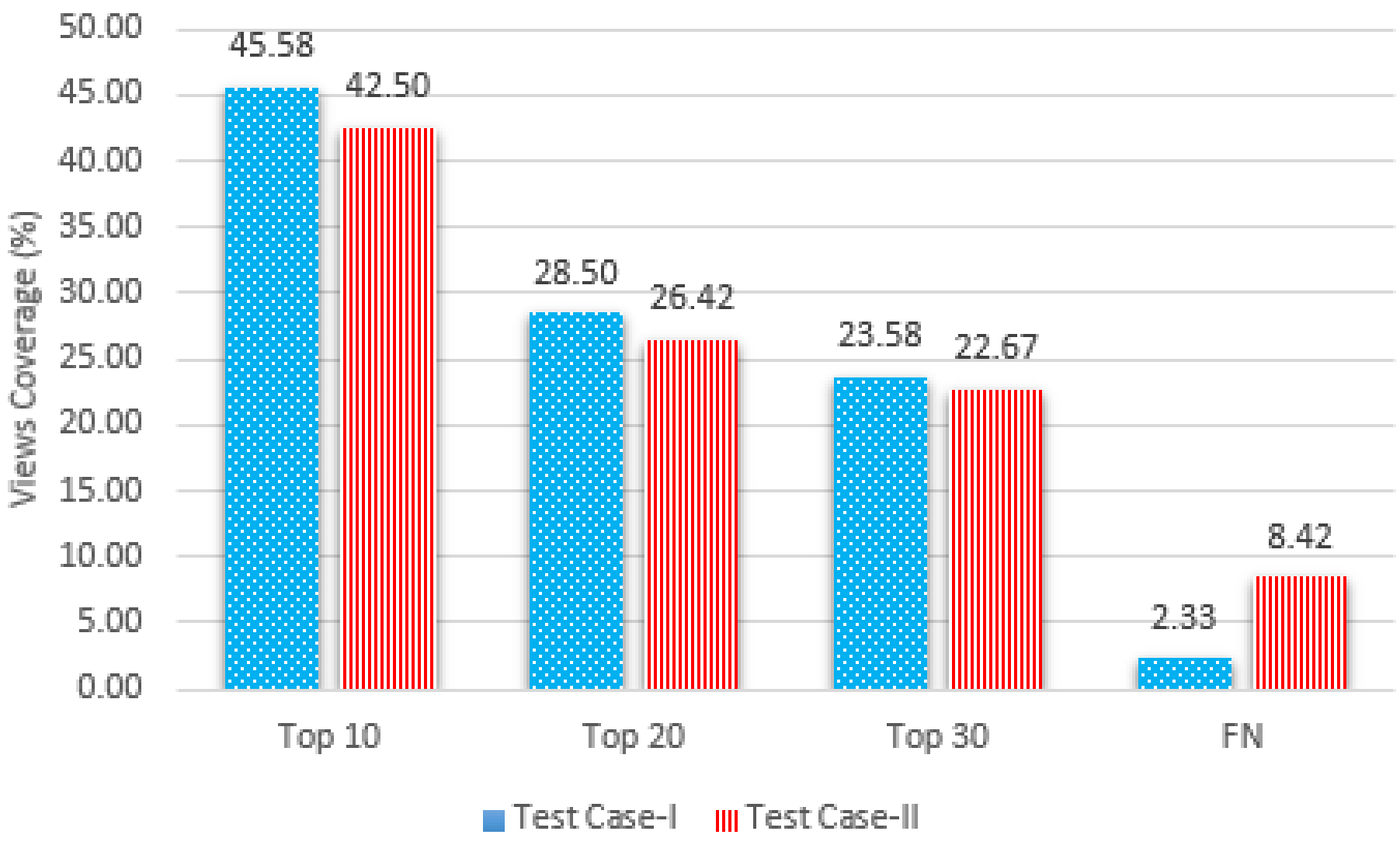

**Figure 3.3.16 Performance evaluation under both the test cases**

These results are further refined by two additional factors. The first factor is the measurement of approximation for age and height, the second factor introduces the concept of trusted features. The close approximation about age and height helps in setting up the hierarchy of records while the trusted features give you the freedom to work on only certain feature set. The approximation is evaluated by measuring the gap between the actual age and height presented by the eyewitness and the extracted age and height from the labelled database. We have defined the range of $\pm 5$year for age and $\pm 7.62$cm for the height attribute. The value of the approximation is computed by:

$$Age_{Diff} = abs(Age^{actual} - Age^{retrived}) \tag{3.5}$$

$$Height_{Diff} = abs(Height^{actual} - Height^{retrived}) \tag{3.6}$$

$$Age_{Approximation} = \frac{Age_{Diff}}{Age_{Intervl}} \tag{3.7}$$

$$Height_{Approximation} = \frac{Height_{Diff}}{Height_{Intervl}} \tag{3.8}$$

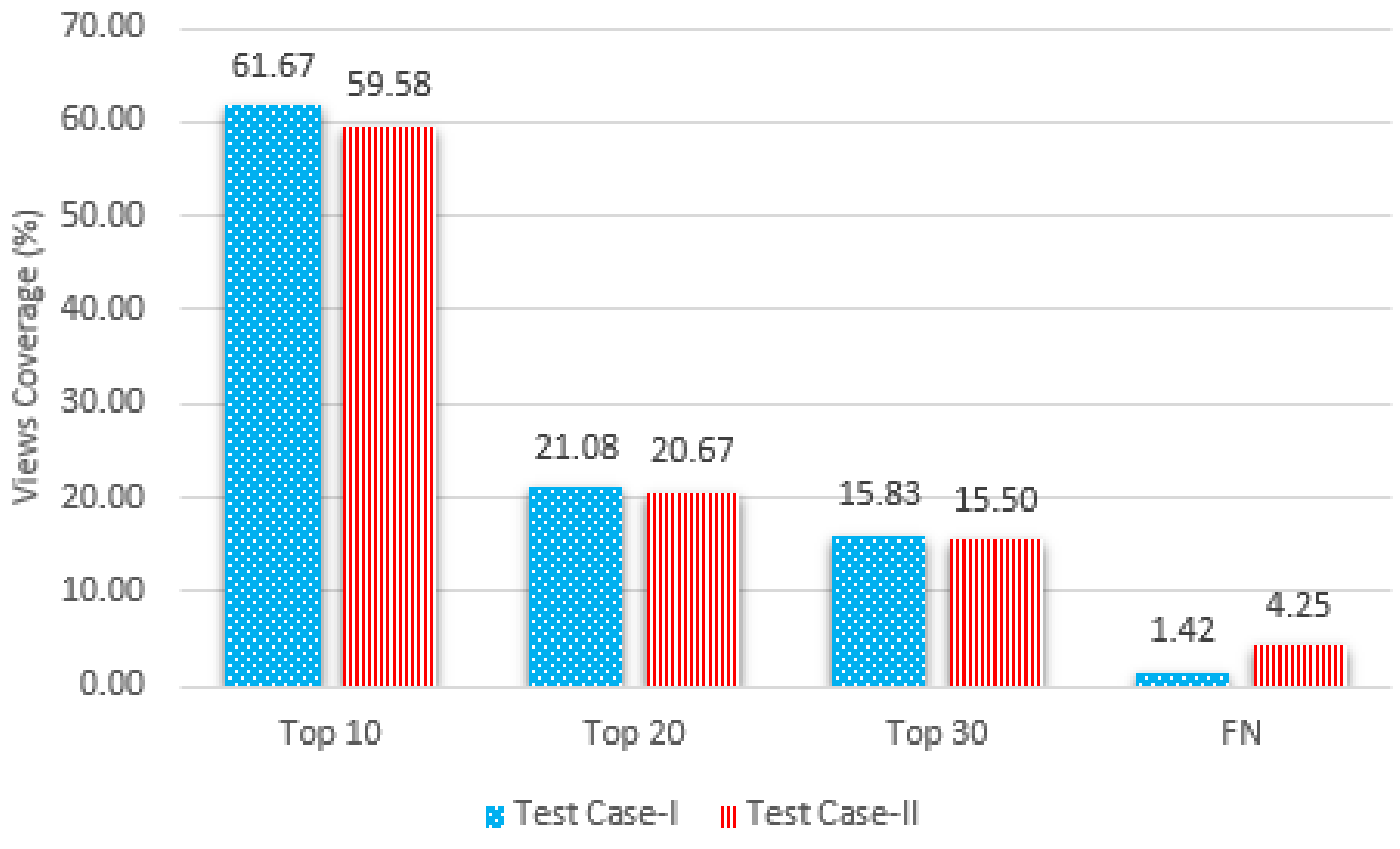

**Figure 3.3.17 Results after including two additional attributes**

The fetched records are arranged on the basis of their age and height approximation. Further the matching is performed only on the selected trusted features. Trusted features are those features on which eyewitness is sure. Like if he/she agrees on face type, face tone and eye shape, then the matching will be performed only on these features. The effects of using two features are summarized in Figure 3.3.17. The results are very surprising after introducing the additional two features. The false negative is minimized and reached up to 1.42% and 4.25 % with respect to both the test cases. We have presented a spread sheet analysis of the results obtained from our experimental evaluation in Table 3.3.13. The entire population of 1200 views is divided separately over 285 Bollywood female views, 265 male views, 200 Indian cricketer's views, 250 and 200 male and female population views. It has been found that all those views which belong to Bollywood male and female population, Indian cricketer's population are classified with the accuracy of 100%. Their contributions towards the total population views are 23.75%, 22.08% and 16.67% respectively. In total they cover 62.50% of the total population and total accuracy. We have achieved these kinds of fine results because the number of samples for each individual age group is less than the coverage ratio (top 30 results). The results are depicted in terms of top 10, top 20 and top 30 retrieved faces. If we observe the age distribution for each population (please refer to Table 3.3.13), we find that none of the age window consist more than 30 people. Since the results are depicted in terms of 30 search results, hence, if the eyewitness select the wrong prototype of any attribute, the system would give you all the 30 people in result. Therefore, in these cases, we have observed the classification rate of 100%, while in other population such as IIITA where there are 90 persons registered in the database, we are receiving 35.20 % and 33.80% of accuracy with respect to both test cases.

The fall of accuracy is due to the large number of population belonging to the same age group 20-30 and average height of 5 foot 5 inch. The improvement in accuracy is noticed after the addition of two new parameters (a) closure approximation and (b) trusted feature concept. The classification accuracy of male students was improved from 17.16% to 17.25% for test case-I and 18% to 18.33% for test case-II. The same happened with female population where the classification accuracy reached up to 19.25% from 17.67% with respect to test case-I, while suffers little bit in test case-II. If none of the feature matched with the existing labeled database, it is very hard to produce accurate results. The problem becomes more severe when the majority of people belong to the same partition. As happened in our in house database case.

**Table 3.3.13 Spreadsheet analysis of the result obtained**

| Bollywood Male: Total population 53 | | | |
|---|---|---|---|
| | Age (20-30) | Age (31-40) | Age(41-50) |
| Samples | 12 | 10 | 31 |
| Views | 60 | 50 | 155 |
| Total Views: 265, Contribution: 22.08%, Classification: 100% | | | |

| Bollywood Female: Total population 57 | | | |
|---|---|---|---|
| | Age (20-30) | Age (31-40) | Age(41-50) |
| Samples | 21 | 26 | 10 |
| Views | 105 | 130 | 50 |
| Total Views: 285, Contribution: 23.75% Classification: 100% | | | |

| Indian Cricketers: Total population 40 | | |
|---|---|---|
| | Age (20-30) | Age (31-40) |
| Samples | 22 | 18 |
| Views | 110 | 90 |
| Total Views: 200, Contribution: 16.67% Classification: 100% | | |

| IIITA Database: Total population 90 | | | | |
|---|---|---|---|---|
| | Female | Age (20-30) | Male | Age (20-30) |
| | Samples | 40 | Samples | 50 |
| | Views | 200 | Views | 250 |
| | Total Views: 200, Contribution: 16.67% | | Total Views: 250, Contribution: 20.84% | |
| Test Case-I | Classification: 17.20% | | Classification: 18% | |
| Test Case-II | Classification: 17.70% | | Classification: 16.10% | |

| | | |
|---|---|---|
| | Total Accuracy: 35.20% | Total Accuracy: 33.80% |
| | After introducing additional attribute | |
| Test Case-I | Classification: 17.25% | Classification: 18.84% |
| Test Case-II | Classification: 19.25% | Classification: 16% |
| | Total Accuracy: 36.09% | Total Accuracy: 35.25% |

#### 3.3.3.4 Solution II: Rough Set based Approach

The knowledge extracted from the eyewitness is imprecise. Therefore, we have used several attributes to define proper feature type. We have used benchmark datasets to model the system depicted in Figure 3.3.18.

**Database Generation**

There are three different datasets we have used to demonstrate the effectiveness of roughest based modeling, over mug shot detection. The glimpse of the dataset is given in Figure 3.3.18. We have randomly selected 50 colour photographs of 50 females and 50 males from the FERET dataset. The RGB images are taken because there we can define the face skin tone, one of the attributes of information system. We have selected only one image per person. The similar kind of strategy, we followed to select face images from the AR database while we have used all 100 photo faces from the CUHK dataset. In total we have 300 criminal faces in our database. This work is totally based on human perception. Therefore we have invited several views from different volunteers to create our dataset. Views are collected with the help of our designed web interface shown in Figure 3.3.1. The interface displays photograph of the guilty and capture user response through input fields. Prototype of each facial attributes (like what kind of face type you have or eye shape) is added to assist volunteer's views. We have represented each criminal to one class; therefore we have 300 classes and per class 5training samples. Similarly, we have 2 test samples per class in testing database. In total we have used 105 volunteers to mark the criminal's identity in a symbolic way. Each volunteer has given the description about 20 criminals. We have created 15 groups, each group have7 members. Hence, for each group we have 140 views. Each criminal has 5 training views and test views. The training dataset is filled up with 1500 objects (criminal descriptions) each one can be expressed with the help of 8 attributes. Similarly the test set consists of 600 views of the same population. We have used different volunteer views for training and testing. The volunteers used here to generate training and testing views are the students of Indian Institute of Information Technology, Allahabad, India.

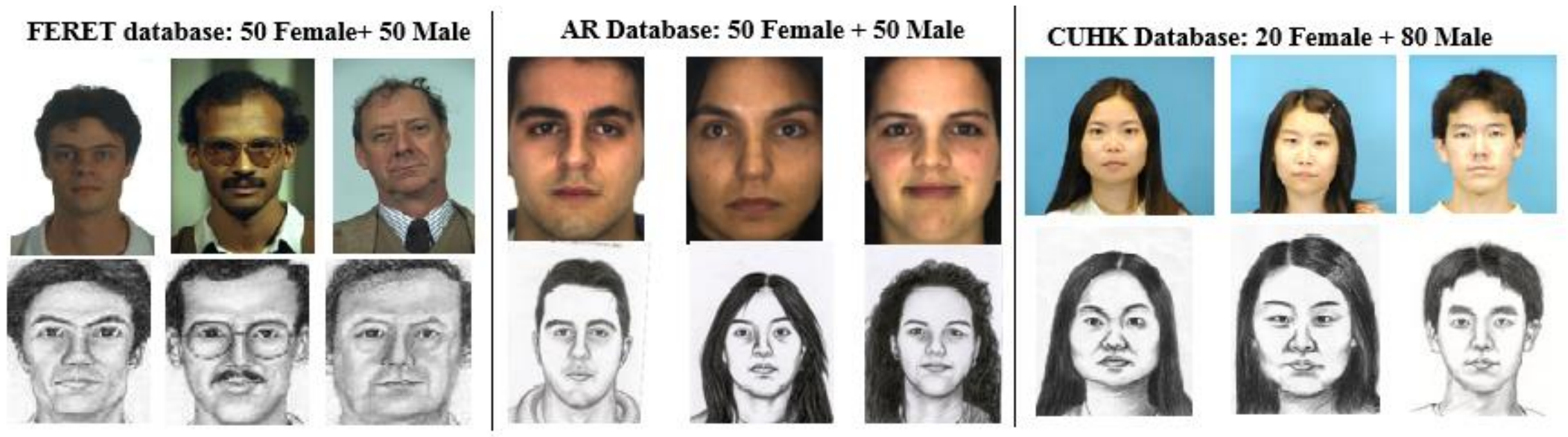

**Figure 3.3.18 Samples from Mugshot Database**

### Rough Set based Modeling

The knowledge we captured is vague, which could be processed by rough set theory [188-190]. RST is sufficient enough to handle the uncertainty and imprecision [191-192]. There are four basic steps involved in rough set modeling [192-195] (1) Construction of a Decision Information System (2) calculating reducts and cores for the given information system (3) Generation of rules (4) Testing the validity of predictions, error analysis. All of these modules and key terms are discussed below.

1. **Construction of Information System**

   The information system (IS) is composed with the universe of discourse U and the attribute set A. It is represented by:

   $$IS = (U, A)$$

   The universe of discourse "U" consists of all the objects (U= $\{o_1, o_2, o_3, o_4, \ldots\ o_n\}$, here n is the number of objects $\epsilon$ U) and the attribute set is the set of all attributes required to represent these objects (A= $\{a_1, a_2, a_3, a_4 \ldots, a_m\}$, here m is the number of attributes). Each attribute could have several possible values which can be mapped by an attribute function($f_a : U \rightarrow V_a$), here $V_a$represents the value of a particular attribute. In our model the Information system consists of 190 classes and for each class 5 training views and 2 testing views exist. Therefore, we have 950 objects and 10 attributes to describe them. These 10 attributes, description were presented in Table 3.3.10. A transformation is required in order to encode these values so that it can be processed using the rough set theory. The transformation we used is shown in Table 3.3.10. For demonstrating our hypothesis, we have randomly selected only 10 objects to represent the IS as shown in Table 3.3.14 (please refer to Table 3.3.10 for transformed values).

**Table 3.3.14 An Information System Instance**

| Objects | $Conditional\ Attributes\ Set$ | | | | | | | | | | $Decision\ Attribute\ (d)$ |
|---|---|---|---|---|---|---|---|---|---|---|---|
| U | $a_1$ | $a_2$ | $a_3$ | $a_4$ | $a_5$ | $a_6$ | $a_7$ | $a_8$ | $a_9$ | $a_{10}$ | D |
| $o_1$ | M | 2 | 4 | 3 | 1 | 1 | 3 | 5 | 1 | 1 | 2 |
| $o_2$ | M | 2 | 4 | 2 | 1 | 2 | 3 | 5 | 1 | 1 | 2 |
| $o_3$ | F | 3 | 3 | 1 | 1 | 1 | 3 | 6 | 3 | 8 | 3 |
| $o_4$ | M | 2 | 4 | 2 | 3 | 1 | 2 | 4 | 2 | 1 | 1 |
| $o_5$ | F | 3 | 3 | 4 | 1 | 1 | 4 | 7 | 3 | 8 | 3 |
| $o_6$ | M | 2 | 4 | 2 | 1 | 2 | 3 | 5 | 1 | 1 | 2 |
| $o_7$ | M | 2 | 4 | 2 | 3 | 1 | 2 | 4 | 2 | 1 | 1 |
| $o_8$ | F | 3 | 3 | 1 | 1 | 1 | 3 | 6 | 3 | 8 | 3 |
| $o_9$ | M | 2 | 4 | 4 | 3 | 1 | 2 | 5 | 2 | 1 | 1 |
| $o_{10}$ | F | 3 | 3 | 1 | 1 | 1 | 3 | 6 | 3 | 8 | 3 |
| (950 $Samples\ from\ the\ Criminal\ Dataset,$ 190 $classes$) | | | | | | | | | | | |

The kernel of RST is to discover the indiscernible relation out of the given information system. The indiscernible relation is defined on two objects $o_i\ and\ o_j\ where\ i \neq j$ as: $a(o_i) = a(o_j)$ for every $a \in A$. An object that belongs to same indiscernible relation is represented as $IND(A)$ known as elementary sets and it is denoted by$[o_i]_{IND_{(A)}}$. It may so happen that we are interested about the only specific attribute set. Then the elementary set description will be different. Let $B \subset A$ and it consists of attribute set $\{a_1, a_2, a_3, a_9\}$. In this case the elementary set created on attribute set A and B will be different as shown in Table 3.3.15 and Table 3.3.16 respectively.

**Table 3.3.15 An Elementary Set Description over attribute set A**

| U/A | $a_1$ | $a_2$ | $a_3$ | $a_4$ | $a_5$ | $a_6$ | $a_7$ | $a_8$ | $a_9$ | $a_{10}$ | d |
|---|---|---|---|---|---|---|---|---|---|---|---|
| $\{o_3, o_8, o_{10}\}$ | F | 3 | 3 | 1 | 1 | 1 | 3 | 6 | 3 | 8 | 3 |
| $\{o_4, o_7\}$ | M | 2 | 4 | 2 | 3 | 1 | 2 | 4 | 2 | 1 | 1 |
| $\{o_2, o_6\}$ | M | 2 | 4 | 2 | 1 | 2 | 3 | 5 | 1 | 1 | 2 |
| $\{o_1\}$ | M | 2 | 4 | 3 | 1 | 1 | 3 | 5 | 1 | 1 | 2 |
| $\{o_5\}$ | F | 3 | 3 | 4 | 1 | 1 | 4 | 7 | 3 | 8 | 3 |
| $\{o_9\}$ | M | 2 | 4 | 4 | 3 | 1 | 2 | 5 | 2 | 1 | 1 |

**Table 3.3.16 An Elementary Set Description over attribute set B**

| U/B | $a_1$ | $a_2$ | $a_3$ | $a_9$ | D |
|---|---|---|---|---|---|
| $\{o_3, o_5, o_8, o_{10}\}$ | F | 3 | 3 | 3 | 3 |
| $\{o_2, o_6\}$ | M | 2 | 4 | 2 | 2 |
| $\{o_4, o_7\}$ | M | 2 | 4 | 1 | 1 |
| $\{o_1\}$ | M | 2 | 4 | 1 | 2 |
| $\{o_9\}$ | M | 2 | 4 | 2 | 1 |

**2. Finding Core and Reducts:**

The attribute used to represent the system is further processed and analyzed in this section. The motive of finding core and reducts of the "IS" is to find the independent set of attributes. Reducts of the attribute set are the minimal subset which leads to the same partitions (elementary sets) as we achieved on the full set [196-197]. These attributes will later help in setup the rules library. Core shows the presence of a single attribute which discerns from other attribute of the other object. But before proceeding for core and reducts some key ingredients which will be useful in this section should be discussed.

**Lower, Upper and Boundary Approximation:**

There could be three conditions arising for predicting the data over one of the elementary set which will be expressed by these three categories [195][198].

*Lower Approximation: Elements possibly those belong to the elementary set.*

*Upper Approximation: Elements those certainly belong to the elementary set.*

*Boundary Approximation: Elements those belong to the boundary sets.*

Let O be the subset of U ($B \subset U$) then the lower approximation of O is defined by the union of all the elementary sets which fall into the domain of O. The lower approximation is denoted by $\underline{\text{BO}}$.

$$\underline{BO} = \{ o_i \epsilon U \mid [o_i]_{IND_{(B)}} \subset O \} \qquad (3.9)$$

Similarly the upper approximation of B is defined by $\overline{BO}$:

$$\overline{BO} = \{ o_i \epsilon U \mid [o_i]_{IND_{(B)}} \cap O \neq \emptyset \} \qquad (3.10)$$

Any object which belongs to the lower set can also be belongs to the upper set.

The boundary approximation is defined by the intersection of lower and upper approximation. It can be represented as: $BNO = \overline{BO} - \underline{BO}$

Example: Let O be a group of some objects O= $\{o_2, o_3, o_4, o_6, o_7\}$ and over the attribute set A. (Please refer to Table 3.3.15)

Then the lower approximation set will be defined as:

Elementary set: $\{ o_2, o_6\}$ and $\{o_4, o_7\}$ then $\underline{BO} = \{ o_2, o_4, o_6, o_7 \}$

The upper approximation set will consider the common attribute of $\underline{BO}$ also.

Elementary set: $\{ o_2, o_6\}$, $\{o_{4,} o_7\}$ and $\{o_3, o_8, o_{10}\}$then

$$\overline{BO} = \{ o_2, o_3, o_4, o_6, o_7, o_8, o_{10} \}$$

The boundary approximation will be calculated as:

$$BNO = \{o_2, o_3, o_4, o_6, o_7, o_8, o_{10}\} - \{ o_2, o_4, o_6, o_7 \}$$

$$= \{ o_3, o_8, o_{10} \} \quad (3.11)$$

**Approximation Accuracy:**

The accuracy of these approximations is also represented by rough membership and it is calculated using:

$$\mu_B(O) = cardinality(\underline{BO})/cradinality(\overline{BO}) \quad (3.12)$$

Here $cardinality(\underline{BO})$ represents the cardinality of lower approximation which is basically the number of elements present in the lower approximation elementary set. Similarly the $cardinality(\overline{BO})$ represents the elements present in the upper approximation elementary set.

Here $card(\underline{BO}) = 4$ and $cardinality(\overline{BO}) = 7$

Then $Accuracy : \mu_B(O) = \frac{4}{7} = 0.57$

**Independence of attributes:** The attributes used here for representing the Information system has a dependency on others or they may be independent. If the attribute is dependent we can remove them, therefore, minimize the attribute set. The dependent attributes can be discovered by removing it from the system and calculating the number of elementary sets. If the number of elementary sets are identical as before (in the absence of a particular attribute). This attribute is called superfluous attribute. An example is shown below in Table 3.3.17.

**Table 3.3.17 Attribute dependency and elementary set**

| | Removed Attributes | | | | | | | | | | |
|---|---|---|---|---|---|---|---|---|---|---|---|
| | None | $a_1$ | $a_2$ | $a_3$ | $a_4$ | $a_5$ | $a_6$ | $a_7$ | $a_8$ | $a_9$ | $a_{10}$ |
| Number of Elementary Sets | 6 | 6 | 6 | 6 | 6 | 6 | 6 | 6 | 6 | 6 | 6 |

| | Removed Attributes | | | |
|---|---|---|---|---|
| | None | $a_1a_2$ | $a_4a_6$ | $a_4a_8$ |
| Number of Elementary Sets | 6 | 6 | 5 | 5 |

| | Removed Attributes | | | |
|---|---|---|---|---|
| | None | $a_1a_2a_3$ | $a_4a_7a_8$ | $a_4a_6a_8$ |
| | 6 | 6 | 4 | 4 |

From the above table 3.3.17, we can see that each attribute is superfluous with respect to the whole attribute set. So we can remove any of them. But when we remove the combination of two only $a_1 a_2$remain superfluous, rest becomes dispensable. Similarly, by removing three attribute $a_1 a_2 a_3$act as superfluous.

**Computing Core and Reducts:**

The concept of dispensable attribute gives rise to the concept of reducts. Reducts are the set of attributes which give out the same level of classification as well as preserves the partitions. Reducts (RED) is computed with the help of discernibility matrix. The discernibility matrix is a square matrix ($n \times n$) where n is the number of elementary sets. The discernible relation of two objects $o_i$and $o_j$shows how they discern each other based on their attribute set. Core is the common attribute/attributes among the reduct sets. We represent core as: $CORE(B) = \cap RED(B)$

An example for calculating the core and reducts is demonstrated in Table 3.3.18. Here the sets are same as we have already calculated over attribute set A. (Refer to Table 3.3.15)

**Table 3.3.18 Discernibility Matrix**

| | $S_1$ | $S_2$ | $S_3$ | $S_4$ | $S_5$ | $S_6$ |
|---|---|---|---|---|---|---|
| $S_1$ | - | $a_1, a_2, a_3, a_4$ $a_5, a_7, a_8,$ $a_9, a_{10}$ | $a_1, a_2, a_3, a_4$ $a_6, a_8, a_9, a_1$ | $a_1, a_2, a_3, a_4$ $, a_8, a_9, a_{10}$ | $a_4, a_7, a_8$ | $a_1, a_2, a_3, a_4$ $a_5, a_7, a_8,$ $a_9, a_{10}$ |
| $S_2$ | | - | $a_5, a_6, a_7,$ $a_8, a_9$ | $a_5, a_4, a_7,$ $a_8, a_9$ | $a_1, a_2, a_3, a_4$ $a_5, a_7, a_8,$ $a_9, a_{10}$ | $a_8, a_{10}$ |
| $S_3$ | | | - | $a_4$ | $a_1, a_2, a_3, a_4$ $a_6, a_7, a_8,$ $a_9, a_{10}$ | $a_4, a_5, a_7$ $a_6, a_9$ |
| $S_4$ | | | | - | $a_1, a_2, a_3, a_4$ | $a_4, a_5, a_7$ $a_9$ |
| | | | | | $a_7, a_8, a_9,$ $a_{10}$ | |
| $S_5$ | | | | | - | $a_1, a_2, a_3$ $a_5, a_7, a_8,$ $a_9, a_{10}$ |
| $S_6$ | | | | | | - |

The set description we used in above table 3.3.318 is as follows:

Set 1($S_1$): $\{o_3, o_8, o_{10}\}$. Set 2 ($S_2$):$\{o_2, o_6\}$. Set 3 ($S_3$):$\{ o_4, o_7\}$. Set 4 ($S_4$):$\{ o_1 \}$. Set 5 ($S_5$):$\{ o_5 \}$ and Set 6 ($S_6$): $\{ o_9 \}$

The discernibility function $f(A)$ will be applied to construct the reducts for the calculated discernible matrix. Here we have used 5 discernibility functions for each elementary set. Like $f_1(A)$take care only about how the elementary set 1 differs from 2, 3, 4, 5 and 6.

$f_1(A) =$

$(a_1 + a_2 + a_4 + a_3 + a_5 + a_7 + a_8 + a_9 + a_{10})X(a_1 + a_2 + a_4 + a_3 + a_6 + a_7 + a_8 + a_9 + a_{10})X(a_1 + a_2 + a_3 + a_5 + a_8 + a_9 + a_{10})X(a_4 + a_7 + a_8)X(a_1 + a_2 + a_4 + a_3 + a_5 + a_7 + a_8 + a_9 + a_{10})$

$= (a_4 + a_7 + a_8)$*(Corresponding to 1st row of Table 3.3.18)* (3.13)

$f_2(A) =$

$(a_5 + a_6 + a_7 + a_8 + a_9)X(a_5 + a_4 + a_7 + a_8 + a_9)X(a_1 + a_2 + a_4 + a_3 + a_5 + a_7 + a_8 + a_9 + a_{10})X(a_8)$

$= (a_5 + a_7 + a_8 + a_9)\ X(a_8 + a_{10})$ (3.14)

$f_3(A) =$

$(a_4)\ X(a_1 + a_2 + a_4 + a_3 + a_6 + a_7 + a_8 + a_9 + a_{10})X(a_5 + a_6 + a_7 + a_4 + a_9)$

$= (a_4)\ X\ (\ a_4 + a_6 + a_7 + a_9)$

$= a_4$ (3.15)

$f_4(A) = (a_1 + a_2 + a_4 + a_3 + a_7 + a_8 + a_9 + a_{10})X(a_5 + a_4 + a_7 + a_9)$

$= (a_4 + a_7 + a_9)$ (3.16)

$f_5(A) = (a_1 + a_2 + a_5 + a_3 + a_7 + a_8 + a_9 + a_{10})$

The core of the set is calculated by:

$$f(A) = f_1(A)Xf_2(A)Xf_3(A)Xf_4(A)Xf_5(A) \quad (3.17)$$

$$f(A) = (a_4 + a_7 + a_8)\,X(a_5 + a_7 + a_8 + a_9)X(a_8 + a_{10})Xa_4X(a_4 + a_7 + a_9)X(a_1 + a_2 + a_5 + a_3 + a_7 + a_8 + a_9 + a_{10})$$

$$= a_4X(a_8 + a_{10})X(a_4 + a_7 + a_8)\,X(a_4 + a_7 + a_9)X(a_5 + a_7 + a_8 + a_9)X(a_1 + a_2 + a_5 + a_3 + a_7 + a_8 + a_9 + a_{10})$$

$$= a_4X(a_8 + a_{10})X(a_4 + a_7 + a_8)\,X(a_4 + a_7 + a_9)X(a_5 + a_7 + a_8 + a_9)$$

$$= a_4X(a_8 + a_{10})X(a_4 + a_7 + a_8)\,X(a_7 + a_9)$$

$$= a_4X(a_8 + a_{10})Xa_7 \text{ OR } a_4Xa_8X(a_7 + a_9)$$

$$= a_4X\,(a_7a_8 + a_7a_{10}) \text{ OR } a_4a_8\,X(a_7 + a_9)$$

$$= a_4a_7a_8 + a_4a_7a_{10} \text{ OR } a_4a_7a_8 + a_4a_7a_9 \quad (3.18)$$

There are 3 different reducts we achieved from each of the above discernibility matrix $\{a_4a_7a_8\,, a_4a_7a_9\,, a_4a_7a_{10}\}$. These are the set of optimal attribute which will result in the same partition. Table 3.3.19 shows an instance of this. (C=$a_4a_7a_8$ $and D = a_4a_7a_{10}$)

**Table 3.3.19 Elementary set description over attribute set C and D**

| U/C | $a_4$ | $a_7$ | $a_8$ | d |
|---|---|---|---|---|
| $\{o_3, o_8, o_{10}\}$ | 1 | 3 | 6 | 3 |
| $\{o_4, o_7\}$ | 2 | 2 | 4 | 1 |
| $\{o_2, o_6\}$ | 2 | 3 | 5 | 2 |
| $\{o_1\}$ | 3 | 3 | 5 | 2 |
| $\{o_5\}$ | 4 | 4 | 7 | 3 |
| $\{o_9\}$ | 4 | 2 | 5 | 1 |

| U/D | $a_4$ | $a_7$ | $a_{10}$ | d |
|---|---|---|---|---|
| $\{o_3, o_8, o_{10}\}$ | 1 | 3 | 8 | 3 |
| $\{o_4, o_7\}$ | 2 | 2 | 1 | 1 |
| $\{o_2, o_6\}$ | 2 | 3 | 1 | 2 |
| $\{o_1\}$ | 3 | 3 | 1 | 2 |
| $\{o_5\}$ | 4 | 4 | 8 | 3 |
| $\{o_9\}$ | 4 | 2 | 1 | 1 |

**Classification:** Reducts help to find out the minimal attribute set which a help to represent the whole system. These reducts also help in classification. The entire rule library is based on reducts, so before proceeding to generate rules for particular reduct. It is required to check that accuracy on the given classes [199-200]. Let we have 3 classes as defined earlier in this section.

Class 1: $\{o_3,\ o_5, o_8, o_{10}\}$

Class 2: { $o_2, o_6, o_1$}

Class 3: {{ $o_4, o_7,\ o_9$} and reduct is C={ $a_4 a_7 a_8$ }

The classification accuracy of each class over the attribute set $a_4 a_7 a_8$ is presented in table Table 3.3.20. Here all the classes are showing the confidence of 100% classification which means we can proceed with this reduct set. Further, we have evaluated the accuracy of these classes after removing each individual attribute from this given attribute set. There is only one case as shown in Table 3.3.21 where classification accuracy of class 2 and class 3 decreases. Apart from this rest of the classes are showing the same confidence as they have shown in Table 3.3.20.

**Table 3.3.20 Classification Accuracy over { $\boldsymbol{a_4 a_7 a_8}$ }**

| Class Number | Number of Objects | Lower Approximation | Upper Approximation | Accuracy |
|---|---|---|---|---|
| 1 | 4 | 4 | 4 | 1 |
| 2 | 3 | 3 | 3 | 1 |
| 3 | 3 | 3 | 3 | 1 |

**Table 3.3.21 Classification Accuracy After removal of individual attribute**

| Class Number | Removing One Attribute | | | |
|---|---|---|---|---|
| | None | $a_4$ | $a_7$ | $a_8$ |
| 1 | 1.0 | 1.0 | 1.0 | 1.0 |
| 2 | 1.0 | 0.4 | 1.0 | 1.0 |
| 3 | 1.0 | 0.25 | 1.0 | 1.0 |

### 3. Construction of Decision Table:

Decision rules are constructed based on the decision table. A decision table is based on two types of attributes [199-201. The first group consist of those objects that agree on certain attribute and their values and other group includes all those objects who discern from the first group on that particular attribute and values. Let's take an example to understand this fact. We want to create a decision table for criminal #1. { $\boldsymbol{o_4, o_7, o_9}$} Belonging to the first group as they agree on decision attribute d, while the second group consists of { $\boldsymbol{o_1, o_2, o_3, o_5, o_6, o_8, o_{10}}$}. The decision table shown in Table 3.3.22 presents all attributes which discern from each other over attribute set { $\boldsymbol{a_4 a_7 a_{10}}$}.

**Table 3.3.22 Classification Accuracy After removal of individual attribute**

| | $o_1$ | $o_2$ | $o_3$ | $o_5$ | $o_6$ | $o_8$ | $o_{10}$ |
|---|---|---|---|---|---|---|---|
| $o_4$ | $a^2{}_4, a^2{}_7, a^1{}_{10}$ | $a^2{}_7$ | $a^2{}_4, a^2{}_7, a^1{}_{10}$ | $a^2{}_4, a^2{}_7, a^1{}_{10}$ | $a^2{}_7$ | $a^2{}_4, a^2{}_7, a^1{}_{10}$ | $a^2{}_4, a^2{}_7, a^1{}_{10}$ |
| $o_7$ | $a^2{}_4, a^2{}_7$ | $a^2{}_7$ | $a^2{}_4, a^2{}_7, a^1{}_{10}$ | $a^2{}_4, a^2{}_7, a^1{}_{10}$ | $a^2{}_7$ | $a^2{}_4, a^2{}_7, a^1{}_{10}$ | $a^2{}_4, a^2{}_7, a^1{}_{10}$ |
| $o_9$ | $a^4{}_4, a^2{}_7$ | $a^4{}_4, a^2{}_7$ | $a^4{}_4, a^2{}_7, a^1{}_{10}$ | $a^2{}_7, a^1{}_{10}$ | $a^4{}_4, a^2{}_7$ | $a^4{}_4, a^2{}_7, a^1{}_{10}$ | $a^4{}_4, a^2{}_7, a^1{}_{10}$ |

From the table 3.3.22, we can see that object $o_4$differ from $o_1$over the attribute $a_4a_7a_{10}$, we can also notice that this difference is due to the values of $a_4 = 2,\ a_7 = 2\ and\ a_{10} = 1$. The rules formed in this table can be expressed using the conjunction operator (∨), disjunction operator (∧) and implication operator (⇒).

$$\begin{cases} (a^2{}_4 \vee a^2{}_7 \vee a^1{}_{10}) \wedge (a^2{}_7) \wedge (a^2{}_4 \vee a^2{}_7 \vee a^1{}_{10}) \wedge (a^2{}_4 \vee a^2{}_7 \vee a^1{}_{10}) \\ \wedge (a^2{}_7) \wedge (a^2{}_4 \vee a^2{}_7 \vee a^1{}_{10}) \wedge (a^2{}_4 \vee a^2{}_7 \vee a^1{}_{10}) \ldots\ldots\ldots\ldots\ldots\ldots\ldots\ldots\ldots\ldots (Rule-1) \\ (a^2{}_4 \vee a^2{}_7 \vee a^1{}_{10}) \wedge (a^2{}_7) \wedge (a^2{}_4 \vee a^2{}_7 \vee a^1{}_{10}) \wedge (a^2{}_4 \vee a^2{}_7 \vee a^1{}_{10}) \\ \wedge (a^2{}_7) \wedge (a^2{}_4 \vee a^2{}_7 \vee a^1{}_{10}) \wedge (a^2{}_4 \vee a^2{}_7 \vee a^1{}_{10}) \ldots\ldots\ldots\ldots\ldots\ldots\ldots\ldots\ldots (Rule-2) \\ (a^4{}_4 \vee a^2{}_7) \wedge (a^4{}_4 \vee a^2{}_7) \wedge (a^4{}_4 \vee a^2{}_7 \vee a^1{}_{10}) \wedge (a^2{}_7 \vee a^1{}_{10}) \\ \wedge (a^4{}_4 \vee a^2{}_7) \wedge (a^4{}_4 \vee a^2{}_7 \vee a^1{}_{10}) \wedge (a^4{}_4 \vee a^2{}_7 \vee a^1{}_{10}) \ldots\ldots\ldots\ldots\ldots\ldots (Rule-3) \end{cases}$$

The rules are interpreted as:

Criminal #1 is identified when all the conditions are satisfied as mentioned below:

a) Either $a_4$must have value 2, or $a_7$must have value 2, or $a_{10}$ must have value 1, or all three hold true
b) $a_7$ must have value 2
c) Either $a_4$must have value 2, or $a_7$must have value 2, or $a_{10}$ must have value 1, or all three hold true
d) Either $a_4$must have value 2, or $a_7$must have value 2, or $a_{10}$ must have value 1, or all three hold true
e) $a_7$ must have value 2
f) Either $a_4$must have value 2, or $a_7$must have value 2, or $a_{10}$ must have value 1, or all three hold true

g) Either $a_4$ must have value 2, or $a_7$ must have value 2, or $a_{10}$ must have value 1, or all three hold true

The problem with this rule library is that it consists lots of redundancy. This redundancy can be minimized using traditional Boolean algebra. The expression:

$$(a^2{}_4 \vee a^2{}_7 \vee a^1{}_{10}) \wedge (a^2{}_7) \wedge (a^2{}_4 \vee a^2{}_7 \vee a^1{}_{10}) \wedge (a^2{}_4 \vee a^2{}_7 \vee a^1{}_{10})$$
$$\wedge (a^2{}_7) \wedge (a^2{}_4 \vee a^2{}_7 \vee a^1{}_{10}) \wedge (a^2{}_4 \vee a^2{}_7 \vee a^1{}_{10})$$

Corresponds to object $o_4$ and $o_7$. This rule makes an implication that if (a) and (b) hold true then the criminal is identified as criminal #1.

$$\begin{cases} (a_4 = 2) \wedge (a_7 = 2) \Rightarrow C_1 \\ (a_{10} = 1) \wedge (a_7 = 2) \Rightarrow C_1 \\ (a_7 = 2) \Rightarrow C_1 \end{cases} \text{(New Rule-1)}$$

Similarly The expression:

$$(a^4{}_4 \vee a^2{}_7) \wedge (a^4{}_4 \vee a^2{}_7) \wedge (a^4{}_4 \vee a^2{}_7 \vee a^1{}_{10}) \wedge (a^2{}_7 \vee a^1{}_{10})$$
$$\wedge (a^4{}_4 \vee a^2{}_7) \wedge (a^4{}_4 \vee a^2{}_7 \vee a^1{}_{10}) \wedge (a^4{}_4 \vee a^2{}_7 \vee a^1{}_{10})$$

Will be summarized as

$$= (a^4{}_4 \vee a^2{}_7) \wedge (a^2{}_7 \vee a^1{}_{10}) \wedge (a^4{}_4 \vee a^2{}_7 \vee a^1{}_{10})$$
$$= (a^4{}_4 \vee a^2{}_7) \wedge \{(a^2{}_7 \vee a^1{}_{10}) \wedge (a^4{}_4 \vee a^2{}_7 \vee a^1{}_{10})\}$$
$$= (a^4{}_4 \vee a^2{}_7) \wedge [a^2{}_7 \vee \{a^1{}_{10} \wedge (a^4{}_4 \vee a^1{}_{10})\}]$$
$$= (a^4{}_4 \vee a^2{}_7) \wedge \left[a^2{}_7 \vee \left\{(a^1{}_{10} \wedge a^4{}_4) \vee (a^1{}_{10})\right\}\right]$$
$$= (a^4{}_4 \vee a^2{}_7) \wedge \left[a^2{}_7 \vee (a^1{}_{10} \wedge a^4{}_4)\right]$$
$$= (a^4{}_4 \vee a^2{}_7) \wedge \left[(a^2{}_7 \vee a^1{}_{10}) \wedge (a^2{}_7 \vee a^4{}_4)\right]$$
$$= (a^4{}_4 \vee a^2{}_7) \wedge (a^2{}_7 \vee a^1{}_{10})$$

The following rule is obtained from this deduced expression:

$$\begin{cases} (a_4 = 4) \wedge (a_{10} = 1) \Rightarrow C_1 \\ (a_7 = 2) \Rightarrow C_1 \end{cases} \text{(New Rule-2)}$$

According to the new formed rules, the criminal #1 is identified as:

$$\begin{cases} (a_4 = 2) \wedge (a_7 = 2) \Rightarrow C_1 \\ (a_{10} = 1) \wedge (a_7 = 2) \Rightarrow C_1 \\ (a_4 = 4) \wedge (a_{10} = 1) \Rightarrow C_1 \\ (a_7 = 2) \Rightarrow C_1 \end{cases}$$

These rules can be further explained as:

1) $if\ face\ is\ RECTANGLE\ and\ eyebrow\ is\ HARDANGLED\ then\ criminal\ is$ #1.
2) $if\ face\ is\ OVAL\ and\ face\ cut\ mark\ is\ NOMARKthen\ criminal\ is$ #1
3) $if\ eyebrow\ is\ HARDANGLED\ and\ face\ cutmark\ is\ NOMARKthen\ criminal\ is$ #1
4) $if\ eyebrow\ is\ HARDANGLED\ then\ criminal\ is$ #1

4. **Rules Verification/Validation:**

There are four basic parameters suggested by [200-201] to check the validity of each rule. These four parameters are: (a) Support of the rule (b) Certainty factor (c) Strength of rules and (d) coverage.

**a)** Support: Generally the rule is defined as $C \rightarrow_o D$. Here C is the set of conditional attribute and D is the decision attribute. If we have defined a rule say $(a_4 = 2) \wedge (a_7 = 2) \Rightarrow C_1$ then support of this rule include all objects (o) which belongs to Universe of discourse U, follow this rule. The expression for calculating support of a rule is expressed as: $SUPP_o(C, D) = |C(o) \cap D(o)|$ .
Example: The support for each rule is presented in Table 3.3.23.

**b)** Strength: Strength of the rule is calculated based on total objects and support of the rule. It is expressed as:$\sigma_o(C, D) = \frac{SUPP_o(C,D)}{|U|}$ .|U| is the cardinality of objects. The example is shown in Table 3.3.23.

**c)** Certainty: The certainty factor $cer_o(C, D)$ is defined as the ratio between support of the rule and |C|. |C| represents the number of objects grouped over the conditional attribute (C). Certainty factor is calculated as follows: $cer_o(C, D) = \frac{|C(o) \cap D(o)|}{|C(o)|} = \frac{SUPP_o(C,D)}{|C(o)|}$. Example is presented in Table 3.3.23.

**d)** Coverage: Coverage of the rule is computed over the decision attribute D. It is the ratio between the support of the rule and the Decision attribute D. If a class consist of n samples/objects. Then the coverage of the rule is considered as the ratio between support and sample per class. Expression used in calculation is as follows: $\mathrm{cov}_o(C, D) = \frac{|C(o) \cap D(o)|}{|D(o)|} = \frac{\mathrm{SUPP}_o(C,D)}{|D(o)|}$. Example is presented in Table 3.3.23.

**Table 3.3.23 Classification Accuracy After removal of individual attribute**

| Rules | Support | Strength | Certainty | Coverage |
|---|---|---|---|---|
| 1 | 2 | 0.2 | 1.0 | 0.67 |
| 2 | 3 | 0.3 | 1.0 | 1.0 |
| 3 | 1 | 0.1 | 1.0 | 0.33 |
| 4 | 3 | 0.3 | 1.0 | 1.0 |

*Scores are computed based on Table 3.3.15

## Experimental setup

The experimental setup consists of several modules. The layout of our experiment is presented in Figure 3.3.19. The training module is involved to generate different reduct and rule sets. These, reducts set are further filtered based on their classification accuracy. A rule library for each reduct has been set in the next module. As we can see the rule library has been defined for each individual reduct. We have lots of rules; therefore a filter is also placed to remove unwanted rules. Certainty and coverage of these two factors are used to evaluate each individual rule. Only those rules who have coverage and certainty factor greater than the threshold qualified as valid rules. These rules are saved to use later for projecting the criminal identity. The second part of this module is testing. A test vector is supplied to the entire rule library. Based on the attribute values different rules from the different rule library classify the test vector to different classes. For each rule library, we are keeping status of which rule classified the test vector to which class in the form of a classification table. Each library has its own classification table.

These rule libraries are further analyzed as shown in Figure 3.3.19 to get the actual level of the test vector. The actual class label is based on the most occurring label from all classification tables. Let's assume that we have 3 rule libraries L1, L2 and L3 and 3 classification tables T1, T2 and T3 respectively. Also assume that criminal #1 is the majority voted name from T1 and T3, while T2 shows the most occurred name as criminal #2. In this case the final class label is given as criminal #1, as two classification table support this.

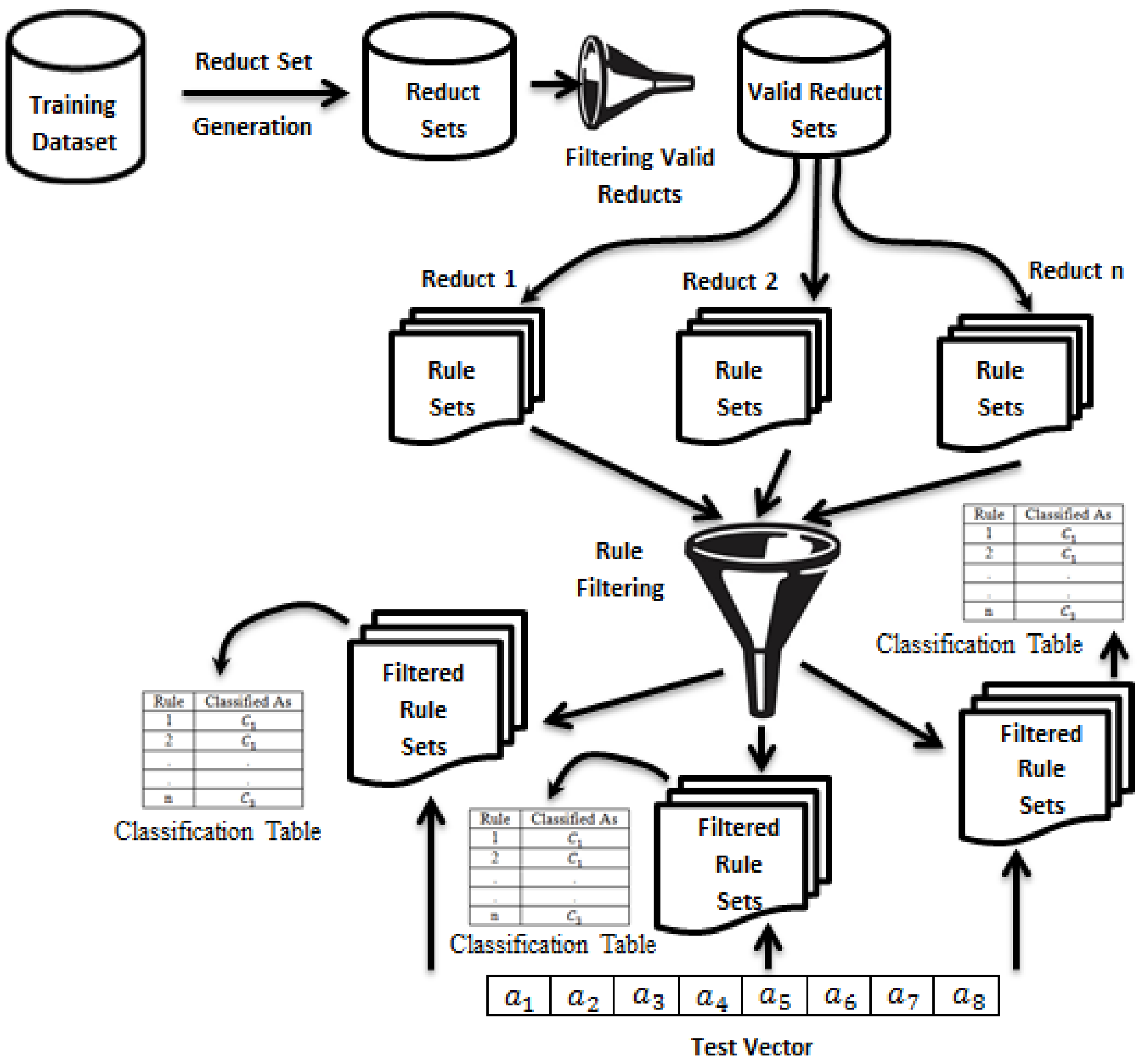

**Figure 3.3.19 Experimental Setup for RST based Criminal Identification**

Moreover a prediction table shown in Figure 3.3.20 is created which consists of top ten nearest matches based on all classification tables. The rank is assigned based on the votes given to the particular criminal in the respected classification table. Let's take a scenario when classification table 1 has 75 rules. Out of 75 rules 50 rules classified the test identity to criminal #1. This means decision table 1 has 50/75=66.37% votes in favour of criminal #1. Similarly votes from all the classification table is generated and analyzed to find top ten votes and concerned identities. Later these results are summarized in a prediction table shown in Figure 3.3.20. The rest modules of this experiment are briefly discussed in subsections.

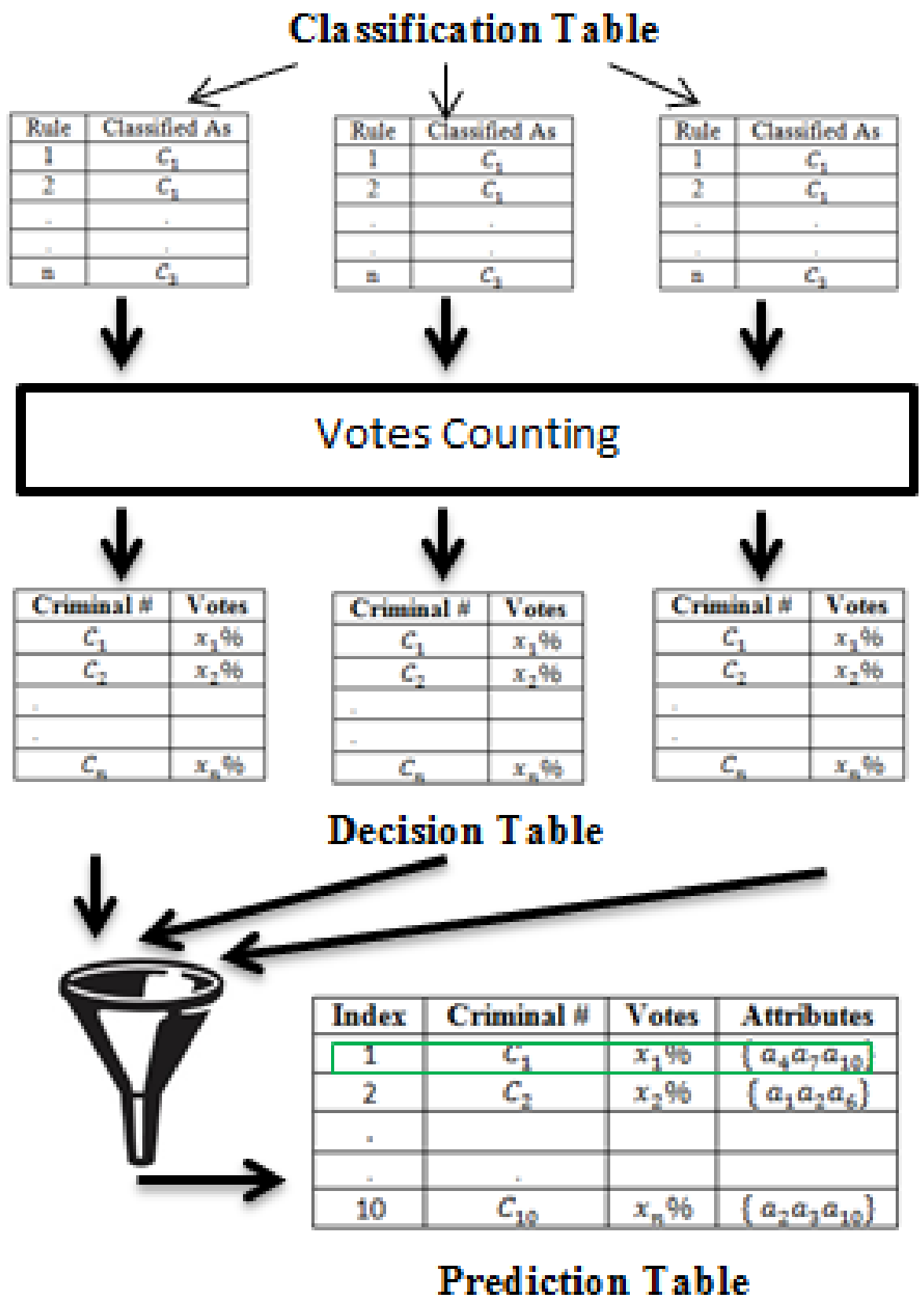

**Figure 3.3.20 Classification and Prediction Setup**

1. **Reduct Sets and Rule Generation:** The reducts set generated on 1200 objects are presented in Table 3.3.24. There are 240 classes (unique decision attributes) exist and each class consist of 5 samples. There are 87 reduct sets. Most of the reducts set are having average attribute value of 3, moreover, they don't have any core. Each reduct set is later filtered out to remove some unwanted reducts. We have kept only those reduct who have a classification accuracy of 100%. After applying the filtering process we ended up with only 34reducts. The reduct set helps in generating rule library. The number of rules attached with each rule set is presented in Table 3.3.24 (a & b). The numbers of rules are huge. Many of the rules are redundant and need simplification. The simplified version of these rules is shown in
2. Table 3.3.25. Each rule is tested against certainty and coverage factor. If the certainty and coverage factor are greater than 90%, then we consider all those rules.

**Table 3.3.24 (a) Reduct set with their accuracy (b) Valid reducts with respected number of rules**

| Reduct | Classification (%) |
|---|---|
| $(a_4, a_7, a_8)$ | 100 |
| $(a_5, a_7, a_8, a_8)$ | 0.67 |
| $a_8, a_{10}$ | 0.73 |
| $(a_4, a_6, a_7, a_9)$ | 0.50 |
| $(a_2, a_5, a_9)$ | 100 |
| $(a_1, a_7, a_{10})$ | 100 |
| $(a_1, a_2, a_3)$ | 0.63 |
| $(a_4, a_7, a_9)$ | 0.80 |
| $(a_3, a_5, a_7)$ | 100 |
| $(a_3, a_6, a_7)$ | 100 |
| $(a_2, a_4, a_7, a_{10}$ | 100 |

| Reduct | Rules |
|---|---|
| $(a_4, a_7, a_8)$ | 725 |
| $(a_2, a_5, a_9)$ | 685 |
| $(a_1, a_7, a_{10})$ | 615 |
| $(a_3, a_5, a_7)$ | 580 |
| $(a_3, a_6, a_7)$ | 470 |
| $(a_2, a_4, a_7, a_{10})$ | 410 |

**Table 3.3.25 Reduced rule sets for each reduct**

| Reduct | Filtered Rules |
|---|---|
| $(a_4, a_7, a_8)$ | 489 |
| $(a_2, a_5, a_9)$ | 457 |
| $(a_1, a_7, a_{10})$ | 433 |
| $(a_3, a_5, a_7)$ | 418 |
| $(a_3, a_6, a_7)$ | 395 |
| $(a_2, a_4, a_7, a_{10})$ | 380 |

3. **Classification of Test Vector:** There are 480 test views having two test views per class. The test vector is supplied to each filtered rule library. Each rule in the rule library classifies the test vector to one of the classes. We see which class has a maximum number of rule support. The ratio between the number of rules and number of rules supported to a particular class is known by vote percentage. Whoever has the highest vote percentage will be considered as the output of the particular rule library. The final result is summarized in the table in the form of top ten results. These results are based on a hierarchy of maximum votes. Example is shown in Figure 3.3.21.

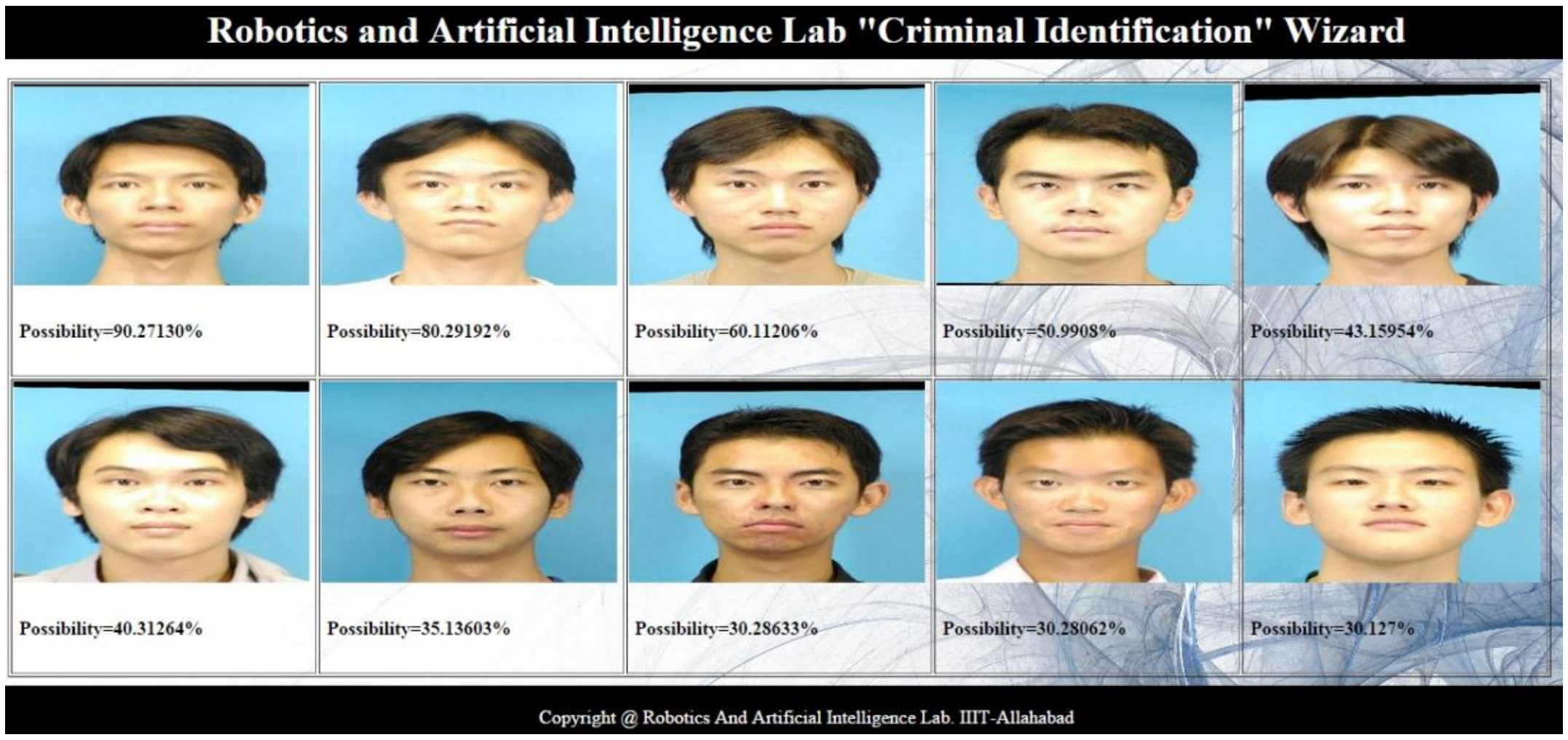

**Figure 3.3.21 Top 10 results of criminals**

## Results and discussions

The results are analyzed for number of test cases taken and how many of them were classified correctly. We have tested the existing system over three different conditions. In first condition, we show a face from the training database and ask person to give her/his views about the person. So, in this scenario we have liberty to see and decide the presented criminal attribute like face type, eye shape etc. In the second test condition, we have made these photos disappear from the interface as shown in Figure 3.3.22. We have shown the photograph of each 10 train faces to each person. We have also kept an unknown identity photo and placed a caption of that identity (please refer Figure 3.3.22). So, they have to make a guess about captioned person. In both the cases we have reported different classification results. In test case one, we have achieved 92.29% accuracy. This shows that out of 480 test cases 443 test cases are classified correctly. In other word we can say that the predicted criminal comes into top 10 results. The results are summarized in Figure 3.3.21.

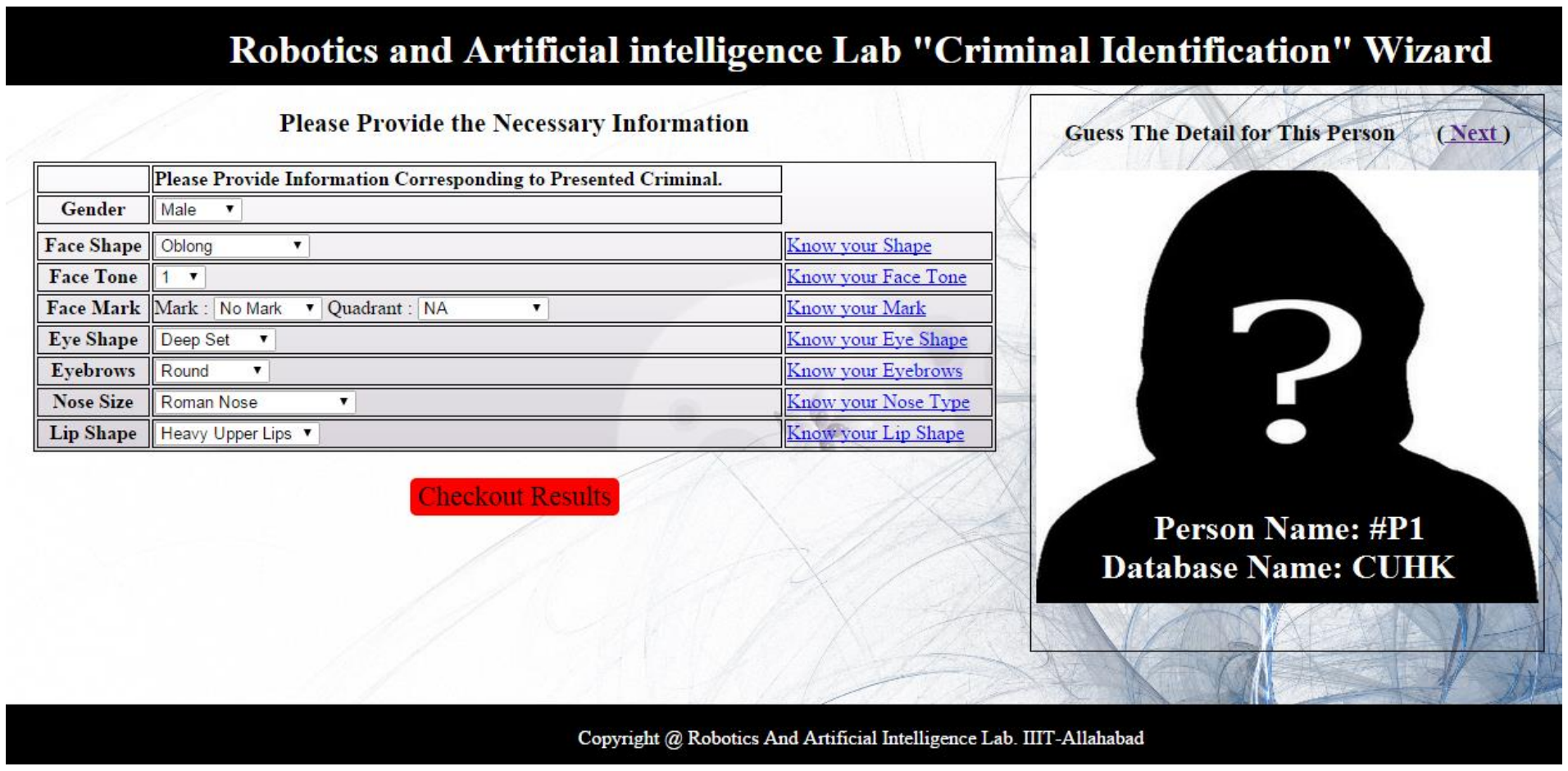

**Figure 3.3.22 Interface without Criminal face**

Figure 3.3.23 and Figure 3.3.24 depicted how many test cases are used and how many of them are misclassified.  The classification has the value one and the misclassification has value 0. We have used bins to represent the misclassified points. We can observe from both the test scenarios that for certain sample points the misclassification is same.

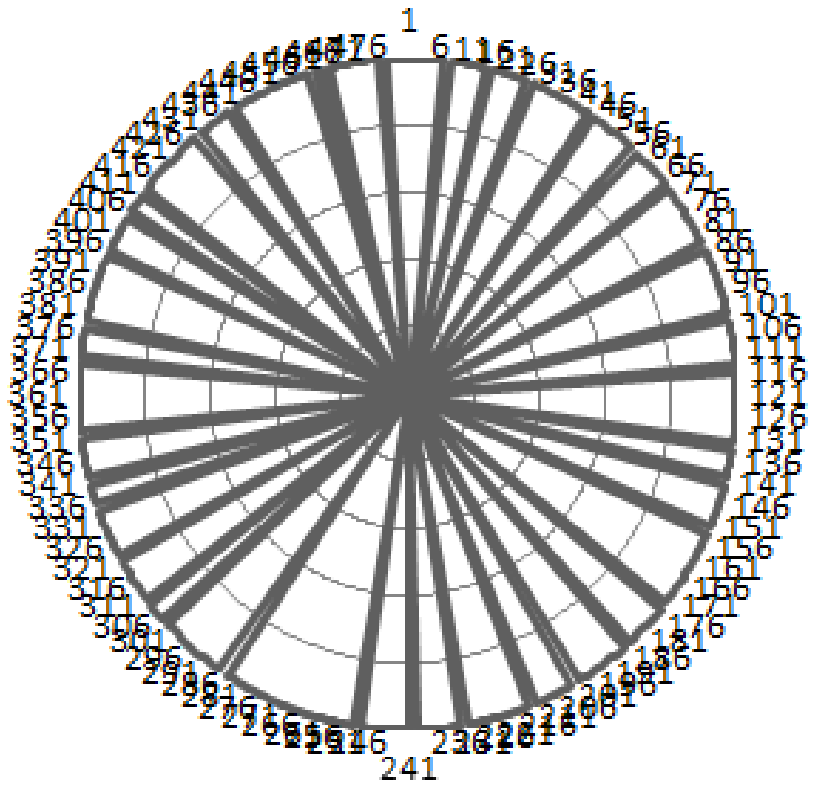

**Figure 3.3.23 Misclassification on test case-I**

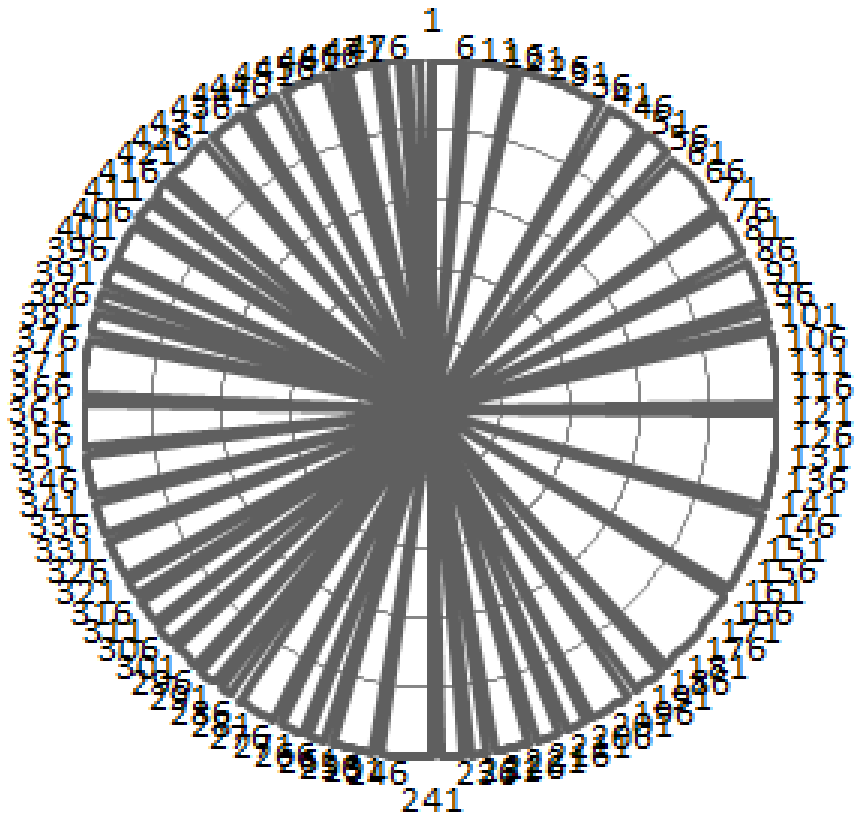

**Figure 3.3.24 Misclassification on test case-II**

These points are called the biased perception influenced points. In total we have 24 biases sample points. We are getting the biased perception because the same person has given his views for both the test scenarios. We have recorded a fall in classification accuracy in our second test case. The accuracy has decreased by 2.92% and reached up to 89.36%. Moreover, in both the cases the top 1 result which is rated as highest voted criminal turned to be true with the accuracy as 60.32% and 47.28 %.

## 3.4 Comparative Analysis between Query based Vs Rough Set based Vs Mugshot Detection Approach

Technically, all the approaches have different working principle, but as all are applied in the same domain, their analysis is possible up to some extent. The mug shot detection approaches are categorized over two wings (a) viewed sketches and the (b) forensic sketches. In order to create the same schema, we have also created two instances in our experiment Test Case-I and Test Case-II analogous to viewed and forensic sketches. The query based approach is demonstrated on our in house dataset. The test is conducted over 240 persons of different categories. On contrary to sketches of mugshot detection we have used views of the user. The database is labeled with the help of an expert knowledge. Here views are similar of the sketches drawn in mugshot detection. In total query based approach has reported 98.58%accuracy in test case-I while 95.75% in test case-II. The time complexity of the system is directly dependent on the size of the database. As we are using the linear search, its time complexity is O(n). On the other hand rough set based approach is an expert system which produces its result on the basis of number of rules fired and their majority. To give a tight competition with mugshot approaches, we have tested our rough set approach on some benchmark datasets. A quick comparison between the rough set approach and mugshot approaches is presented in below sections. Both the proposed approaches have sufficient strength which made them a supplementary of the mugshot detection approach. Both could be applied at the initial level of investigation. The only constraint involved in the query based approach is the presence of an expert which is required to label the database, while the rough set based approach utilizes general/common man views to train the system.

### 3.4.1 Query based Vs Mugshot Approach

If we closely observe the process of mug shot detection, it consists of two major steps. The first step is the pictorial representation of eyewitness perception and the second is to design an algorithm which maps these sketch images to the corresponding photo images of the criminal. To design a sketch similar to the actual criminal is a challenging task which needs proper guidance from the eyewitness. On the other hand the algorithm designed for matching the sketch images to the existing database should have the caliber and capacity to handle the non-linearity in sketch images. This non linearity is due to the different kinds of noise (absence/presence of beard, different hair style, shading to show 3D face structure) present in the sketch images. In comparison to the mug shot detection the proposed approach captures the same perceptual information of criminal's face and physiological structure in the form of symbols and attributes. Further, these symbols are processed with the help of a query processing module. The results depicted in Figure 3.4.23 and Figure 3.4.24 clearly shows the strength of this approach. The proposed system covers around 98.6% and 97.8% in both the test cases analogous to view and forensic sketches [77]. In comparison to the mug shot detection approaches presented in [83] [129] [126] the proposed approach performs well in those scenarios where the eyewitness perception is mature. Searching cost of the proposed system is also very low. It reduces the search space to 2.5% (top 30) of the entire search space. The total process takes an average time of 2 to 4 minutes, which emphasis to use this module at the very initial stage of investigation.

### 3.4.2 Rough Set based Approach Vs Mugshot Approach

A comparative study between the mug shot detection approaches and proposed a rough set based criminal identification is presented in this section. An analysis has been taken separately for each individual database. The result is summarized in Table 3.4.1 for both the test cases. We have achieved encouraging classification accuracy in test case-I while satisfactory in test case-II. Perhaps the proposed rough set approach is not superseding the existing mugshot detection approaches, but they save lots of the efforts and time to create sketches.

**Table 3.4.1 A comparative analysis between mugshot and proposed approach**

| Author | Train/Test Size | Test Case-I Accuracy | Test Case-II Accuracy |
|---|---|---|---|
| Xiaoo Tang et al [83] | 188 sketch and photo pair (Train :88 + Test:100) | 96% | |
| Qingshan Liu et al [129] | 606 sketch and photo pair (306 Train+ 300 Test) | 97% | |
| Xinbo Gao et al [176] | 606 sketch and photo pair (306 Train+ 300 Test) | 98.89% | |
| Proposed Approach | 500 Training Views+ 200 Testing Views | 94.50% | 87.50% |
| Proposed Approach | 500 Training Views+ 200 Testing Views | 96% | 79% |
| Proposed Approach | 500 Training Views+ 200 Testing Views | 89% | 73% |

## 3.5 Conclusion

A new technique has been introduced in this chapter in order to minimize the manual search cost of criminal identification. Two different approaches (a) Query based approach and (b) Rough set based approach based on two different ideologies has been proposed. Two case studies have been considered to test the effectiveness of the proposed approaches. In the first case, the photograph of the criminal is available to the eyewitness at the time of knowledge acquisition, whereas, in the second case, these are absent (shown one day before). The performance of the Query based approach is tested on 1200 views taken from both test cases. In first case study, we have achieved encouraging results which describe that 55.2%of the entire population falls into top 10 matches whereas 26.5% and 16% of the population belongs to top 20 and top 30 matches respectively. This study infers that the search space has been reduced up to only 2.5%. The results obtained from second case study demonstrates the reliability of this approach with the accuracy level of 42%, 34.8% and 19.5% of the entire population on top 10, top 20 and top 30 matches. We have also reported the false negative (FN) of 2.3% and 3.8% with respect to first and second case study. Further, these results are tuned by introducing the concept of closure approximation and trusted attributes. The closure approximation helps to index the searched records while trusted attributes help to compute the match score only on these attribute set. These two additional attributes minimize the FN to 1.4% and 2.2% in both the test cases and increases the level of accuracy up to 61.6% and 56.2% for top 10 matches.

In rough set theory based modeling, we have generated 32 reducts. Classification accuracy is considered as the metric to select valid reducts. Several performance metrics such as strength, confidence and coverage are also estimated to deduce the rule library associated with each reduct. The strength of this approach is evaluated in the light of 2100 samples (1500 training + 600 testing) of our domestic symbolic database. The domestic database is created on the basis of 300 pseudo criminals (CUHK dataset). Each criminal is treated as one class and for each class 5 training and 2 testing views exists. In both test cases the accuracy is reported around 92.67% and 73.83%. The difference in accuracy is due to the different levels of perception maturity of the viewer. Different level of maturity is due to the different memory structure of the brain. If the criminal face is stored in eyewitness long term memory, then this will give better result in comparison to short term memory.

The future work includes use of additional attributes like weight, hair style, eyes colour etc. to produce better matches. As a future scope this module could be applied prior to using mugshot detection module. The strength of this proposed work is that only eyewitness notions are required to identify the criminal which avoids the complexity of making the sketch.

# Chapter 4:

# Development of a Humanoid Robot Application for Criminal's Sketch Drawing

*The preceding chapter has established two novel solutions to identify criminals using the imprecise facial and physiological knowledge about them. However, if the criminal has no background information stored in the criminal database, the proposed approach fails to produce desired results. Therefore, this chapter discuss the tools and techniques to synthesis the face using the same facial knowledge used in the previous chapter. This chapter portrays the design and implementation challenges encountered during the process of sketch drawing performed by humanoid robots. There are two challenges that have been addressed in this chapter, the calibration problem which helps the NAO humanoid robot to perceive each point of the image with respect to its body coordinate system. The second problem is to provide the inverse kinematic solution, the gradient descent based numerical approach is used to solve the second issue while the approach is also extended to get the close form inverse kinematic solution for NAO right hand.*

## 4.1 Introduction

The previous chapter has established its presence in the criminal identification. The foundation stone of the previous chapter was laid on the constraint that the criminal information must have been stored in the criminal database. This is perhaps the limitation of the previous chapter. In this chapter, we create the sketch of that criminal using the same knowledge of eyewitness captured as in the previous chapter. In this chapter we have designed an automated module which makes the NAO Humanoid Robot self-reliant to draw face sketches. The prototype discussed in the previous chapter for each facial attributes is already stored in the database. As soon as the user select the particular prototype, the boundary points of that prototype is picked up to create the face composite.

An example of the same is illustrated in Figure 4.1.1. The flow of the diagram shows, how the addition of each facial prototype gives a better representation of the identity. The identity of the person is Indian cricket captain Mahendra Singh Dhoni. Although, in absence of facial texture, it is quite hard to recognize the identity, yet it gives believe and direction that it can further be extended and improved. Once the composite sketch is represented in image. This is supplied as input to the proposed system which is portrayed by the NAO humanoid robot on the sketch board. The shading effect has not been added in the current state of work which could further improve the sketch quality. One of the sketch drawing is shown in Figure 4.1.2.

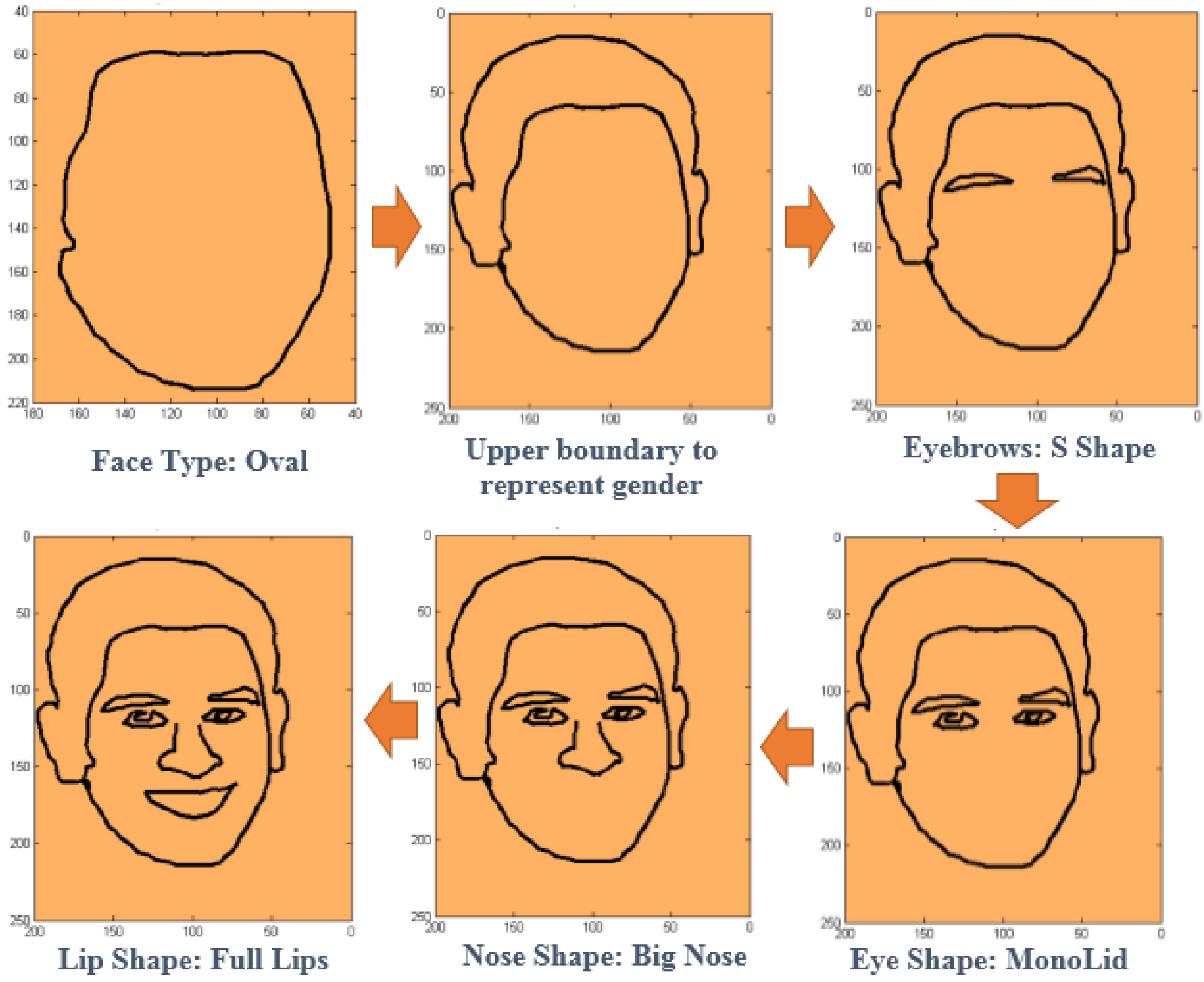

**Figure 4.1.1 The illustration of face composite drawing**

The designed system enriching NAO, not only to create portraits based on the eyewitness description, it can also draw sketches what it sees through its camera. All of the approaches discussed in the following subsections are developed to create human portraits, however, some of them were designed to perform accurate painting for industrial purpose. The first work in this direction of portrait drawing has been noticed in the year of 1988. A program designed by Harold Cohen named as AARON [202] to a pictorial representation of visual scenes. AARON was in the continuous development phase and it has learnt the morphological representation of different objects like plants, animals and persons. There are a small number of robots specified to create drawing in an open loop without having any feedback system. The other advancement in this field is categorized in two wings (a) the first wing discusses the development with respect to manipulators while the (b) second wing focuses more on humanoid robot drawing.

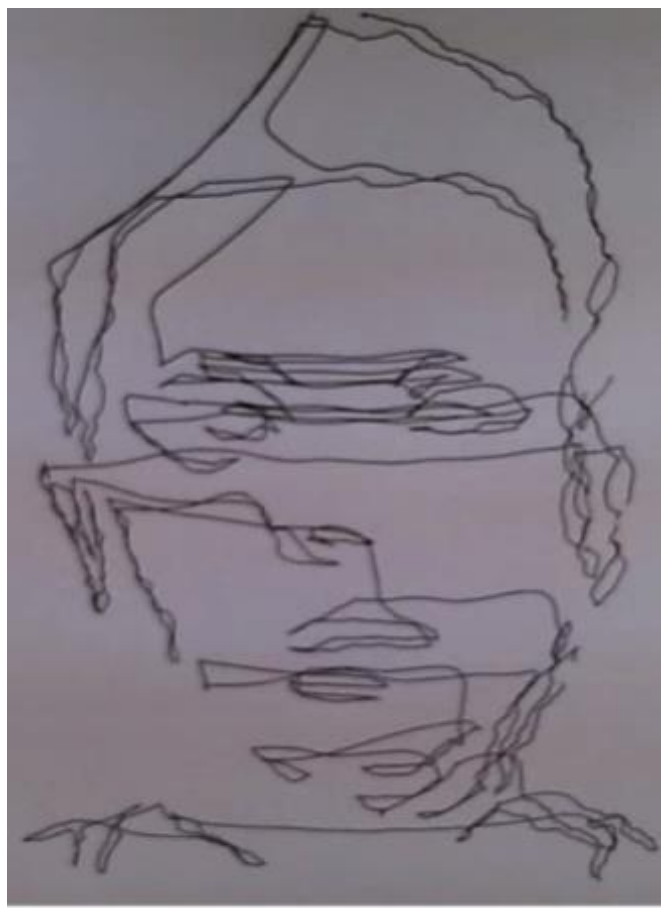

**Figure 4.1.2 Face Sketch of the Suspect based on Eyewitness Description**

### 4.1.1 Painting/Sketch, Drawing by Manipulators

There are several inspiring as well as successfully running robotic machines which prove that robots can do better drawing and painting if they are trained properly. Haeberli [203] has developed different techniques such as sampling geometry using ray-painting, approximating images by relaxation, etc. to create static and animated images which are the abstract representation of the photograph and synthetic scenes. The type of brush used in the painting has great impact on the quality; therefore Hertzmann [204] varied the brush size to paint rough approximation of given images. The painting process is divided into several layers and for each layer different size of brush is used to keep the gradient of image into consideration. PUMA paint [205] was such an online painting project started in late 1998. The PUMA robot was connected to the internet which allows one to draw anything with the help of its Java based interface. The PUMA paint has accurate drawing capability; however, it is not a fully automated tool. The previous semi-automated techniques require human involvement. Later, several fully automated systems came into picture which can create sketches in a fully automated manner. Collomosse [207] defined an automatic system to control brush stroke and sizes based on the image salience and gradient. The other issue associated with these systems was to accumulate feedback system. Therefore the vision based system has been taken into use. The vision guided system helps in providing the feedback to the system as well as helping in achieving the full automation [207-209] Rectification of the input image and proper gripping are the two major challenges in sketch drawing/painting. The 3 step solution has been discussed by Shunsuke et al. [210] to discuss these challenges. The first step used to create the 3D model of the object in order to recognize it; the second module used to extract the edge, line, etc. like features from the transformed silhouette images and the third module discusses how multi fingered hand paints the geometry. Paul a robotic hand-eye system designed by Tresset et al. [211-212] does the aesthetic human face sketching more accurate than all the previous approaches. It has 4 degrees of freedom with the camera placed on different bodies. It has efficient feedback control and facial feature extraction modules. The other kind of robotic manipulator based painting has also been developed in order to paint cars and other objects [213-214].

### 4.1.2 Painting/Sketch Drawing by Humanoids

The first fully automated humanoid system which could be tried as sketch artist was developed at EPFL taking help of HOAP-2 robot [215]. The robot can draw the sketch of any person who is sitting in front of him. The primitive techniques of face detection, boundary and edge extraction with the trajectory planning are used in this application. The other effort has been made by Srikaew et al.[216] to create artistic portraits using the ISAC robot. The beauty of using ISAC as a sketch artist is due to their soft hands who can mitigate the drawing as fine as humans. They have McKibben artificial muscle, which is an excellent actuator while the stereo vision helps it in a 3D representation of the object. After analyzing the existing literature it has been observed that development in the field of painting/sketch drawing by manipulators has gained much attention in comparison to humanoids. One of the possible reasons could be the extra and redundant degrees of freedom associated with robotic hands. Moreover reachable work space of humanoid robot and solving the inverse kinematics for at least 5 degrees of freedom also act as an inhibitor. We have provided a solution (calibration) which defines the image plane with respect to an NAO end effector. Once NAO define each point of the image plane with respect to its body coordinate system, the only thing left is to calculate the feasible inverse kinematic solution.

The rest of this chapter is summarized as: in section 4.2, we define the problem faced in sketch drawing, followed by section 4.3 where we have discussed the proposed solutions. Section 4.3 includes the proposed calibration matrix which helps to transform image points with respect to NAO body coordinate system and the inverse kinematic solution for its right hand. In section 4.4, we evaluate the performance of each proposed technique on the benchmark parameters of computational complexity and error analysis. In the end, we conclude this chapter with its contributions and future scope described in section 4.5.

## 4.2 Problem Definition

There are two problems which need to be solved in order to design the system. The first problem is to estimate the calibration matrix which defines the NAO image/camera plane to its body coordinate system. Given the point $(x_i, y_j)^I$corresponds to image plane where i ∈imageHeight, j∈imageWidth, find out the transformation matrix "T" such that it will define each point of image plane with respect to NAO end effector position. The problem is also described in Figure 4.2.1.

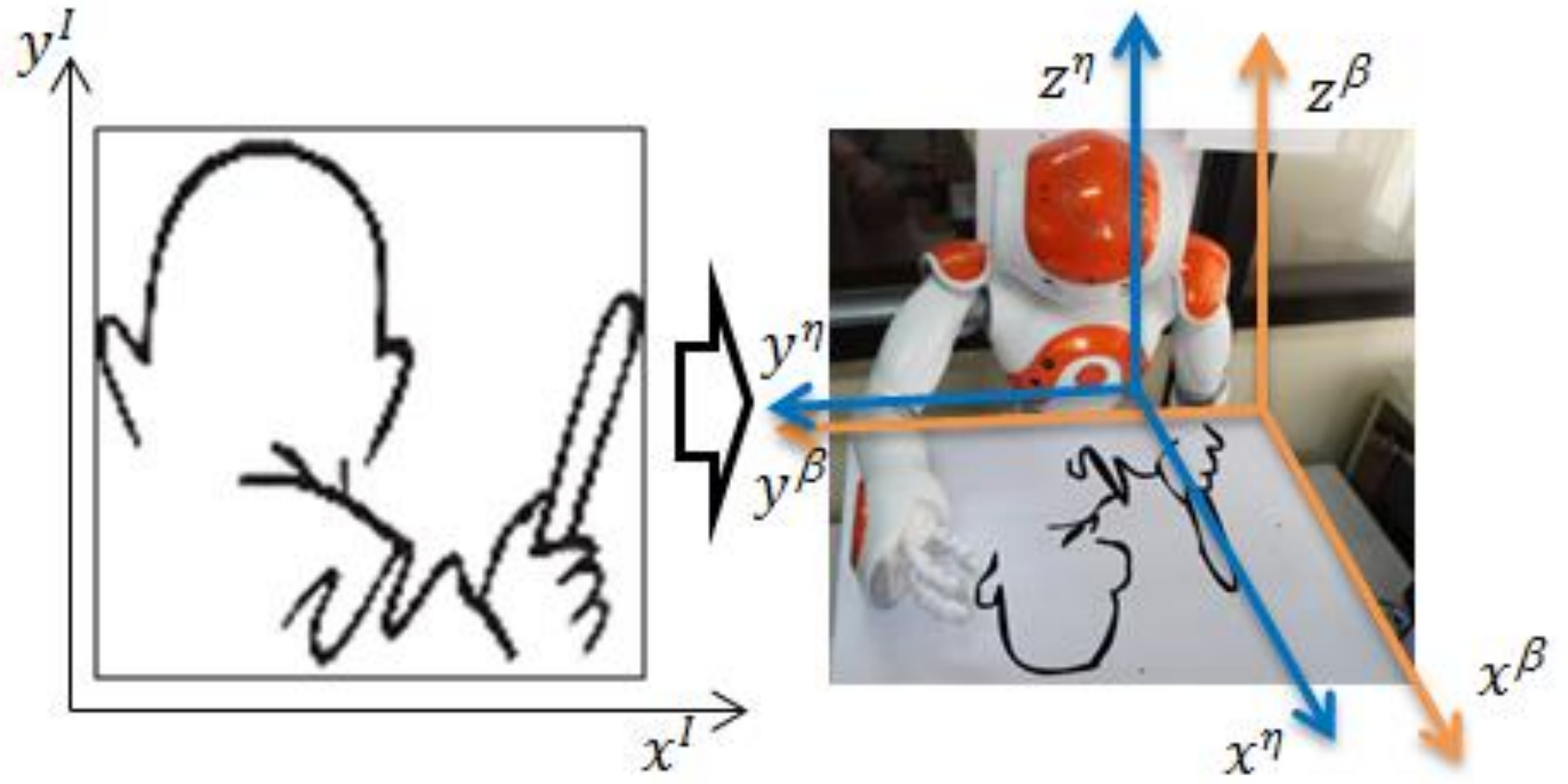

**Figure 4.2.1 Problem Definition (Transformation from image plane to NAO end effector position)**

The first image shows its presence in the image plane while the second image represents the same drawn image by NAO. There are two coordinate system defined in the later image. The yellow coordinate system defines the board/table where NAO has to draw the object. The third blue coordinates system is representing the NAO body coordinate system lying on its torso position. There are two transformation matrixes, the first transformation defines an image in terms of board coordinate system and the second transformation represent the board values in terms of NAO body coordinate system. If $(x', y', z')^{(\eta)}$are the points corresponding to NAO End effector and $(x, y)^I$are the points corresponding to Image then the transformation matrix would be:

$$\begin{bmatrix} x' \\ y' \\ z' \end{bmatrix}^\eta = [T] \times \begin{bmatrix} x \\ y \end{bmatrix}^I \qquad (4.1)$$

The direct transformation from image plane to NAO end effector is decomposed into two sub problems. First we can transform the point of image with respect to board coordinate system (NAO is supposed to draw image points on the board) and then we can define the second transformation which we will transform each point of board with respect to NAO end effector. If we introduce the $(x, y, z)^\beta$with respect to board the whole problem will reduce as given below.

$$\begin{bmatrix} x \\ y \\ z \end{bmatrix}^\beta = [T_1] \times \begin{bmatrix} x \\ y \\ 1 \end{bmatrix}^I \qquad (4.2)$$

$$\begin{bmatrix} x' \\ y' \\ z' \\ 1 \end{bmatrix}^{\eta} = [T_2] \times \begin{bmatrix} x \\ y \\ z \\ 1 \end{bmatrix}^{\beta} \quad (4.3)$$

## 4.3 Proposed Methodologies

The solution of both the problems is addressed here. The calibration problem is solved by estimating the value of transformation matrices T1 and T2. The inverse kinematic solution is presented using the numerical methods.

### 4.3.1 Solution-I: Calibrating between Humanoid Camera Plain and End-effector

Determining the first transformation matrix T1 required a linear transformation, as the Z axis of image is absent and for the board there is no such role of Z axis. Usually the Z axis of the table is fixed. An image is a two dimensional representation of objects and its representation drawn on the board is also a two dimensional image. Therefore, adding a constant z axis will not affect the solution. The second transformation matrix T2 is estimated by defining the nonlinear mapping between the board coordinate system and NAO body coordinate system. This mapping is nonlinear as the board z axis is fixed, while the NAO has almost varying Z axis. There are three techniques four point calibration, fundamental matrix and neural network based regression is utilized to define the second mapping T2. The error analysis is also performed based on the known points to give an analysis for all proposed techniques.

### 4.3.2 Calibration between Camera Plain and Board

In order to map points from image plane to NAO end effector, it is required to first transform the point with respect to board coordinate and then transform the point in terms of NAO end effector position. In order to do this, first we physically estimate what is the reachable region where NAO end effector can reach with respect to the board origin. Once the reachable region's coordinate are retrieved with respect to board origin, the next task is to map every pixel of image to this reachable region. Let the four reachable corner points defined after the physical measurement of board with respect to board origin are $(x_1, y_1)^\beta, (x_2, y_2)^\beta, (x_3, y_3)^\beta, (x_4, y_4)^\beta$. Similarly, we can define the image with four corner points $(x_1, y_1)^I, (x_2, y_2)^I, (x_3, y_3)^I, (x_4, y_4)^I$. Both image and board's four corner and their mapping is presented in Figure 4.3.1. A mapping from image plane to board coordinate system is shown in Figure 4.3.2. The relationship between these coordinate systems can be derived with the linear transformation given in equation (4.4) & (4.5).

$$\beta_X = \min(\beta_X) + \frac{X - \min(I_X)}{\max(I_X) - \min(I_X)} * (\max(\beta_X) - \min(\beta_X)) \quad (4.4)$$

$$\beta_Y = \min(\beta_Y) + \frac{Y - \min(I_Y)}{\max(I_Y) - \min(I_Y)} * (\max(\beta_Y) - \min(\beta_Y)) \quad (4.5)$$

where $\theta^\beta = \{\max(x_1, x_2, x_3, x_4)^\beta\}\}$, $\ominus^\beta = \{\min(x_1, x_2, x_3, x_4)^\beta\}\}$,

$\theta^I = \{\max(x_1, x_2, x_3, x_4)^I\}\}$, $\ominus^I = \{\min(x_1, x_2, x_3, x_4)^I\}\}$ $\rho^\beta = \{\max(y_1, y_2, y_3, y_4)^\beta\}\}$, $\sigma^\beta = \{\min(y_1, y_2, y_3, y_4)^\beta\}\}$ $\rho^I = \{\max(y_1, y_2, y_3, y_4)^I\}\}$, $\sigma^I = \{\min(x_1, x_2, x_3, x_4)^I\}\}$

$\theta^\beta$, $\ominus^\beta$, $\rho^\beta$, $\sigma^\beta$ are the maximum and minimums along the x and y coordinates with respect to board while $\theta^I$, $\ominus^I$, $\rho^I$, $\sigma^I$ are the maximum and minimums along the x and y axis with respect to Image coordinate system.

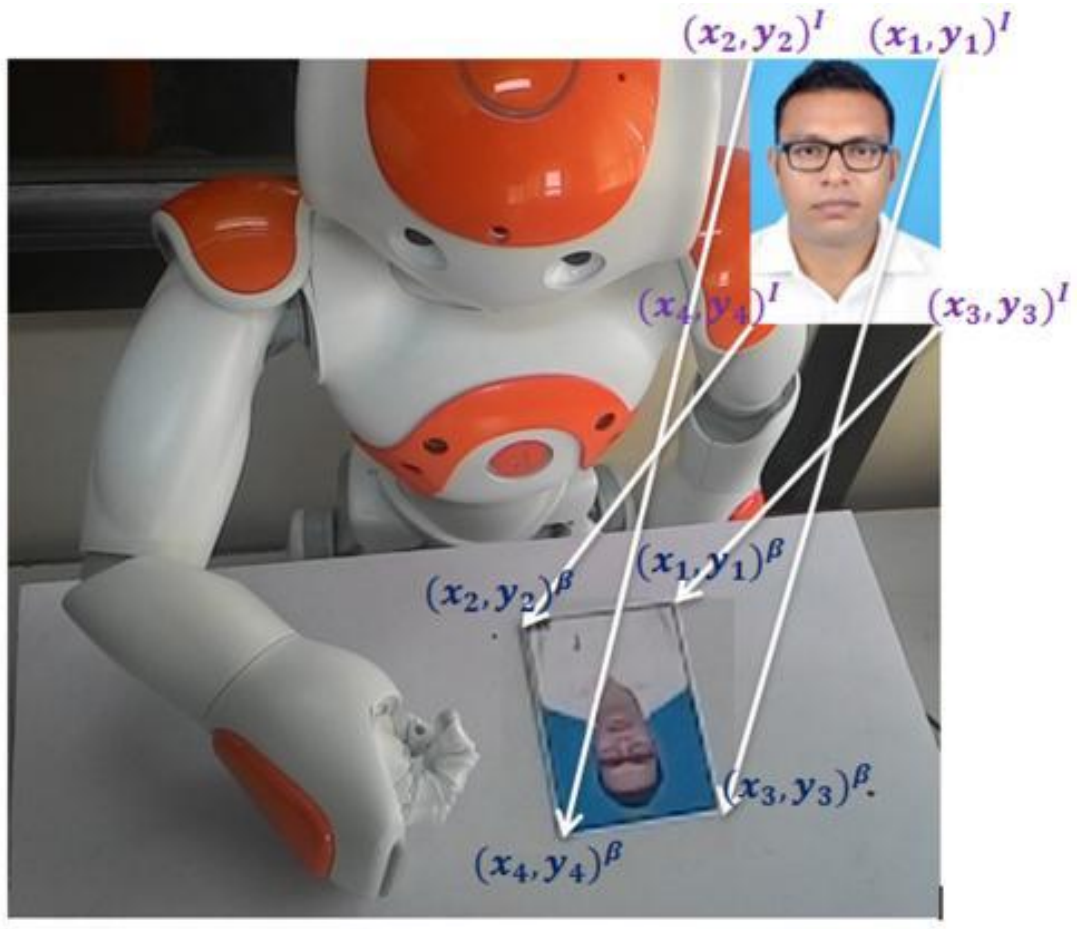

**Figure 4.3.1 Image and Board Coordinate System**

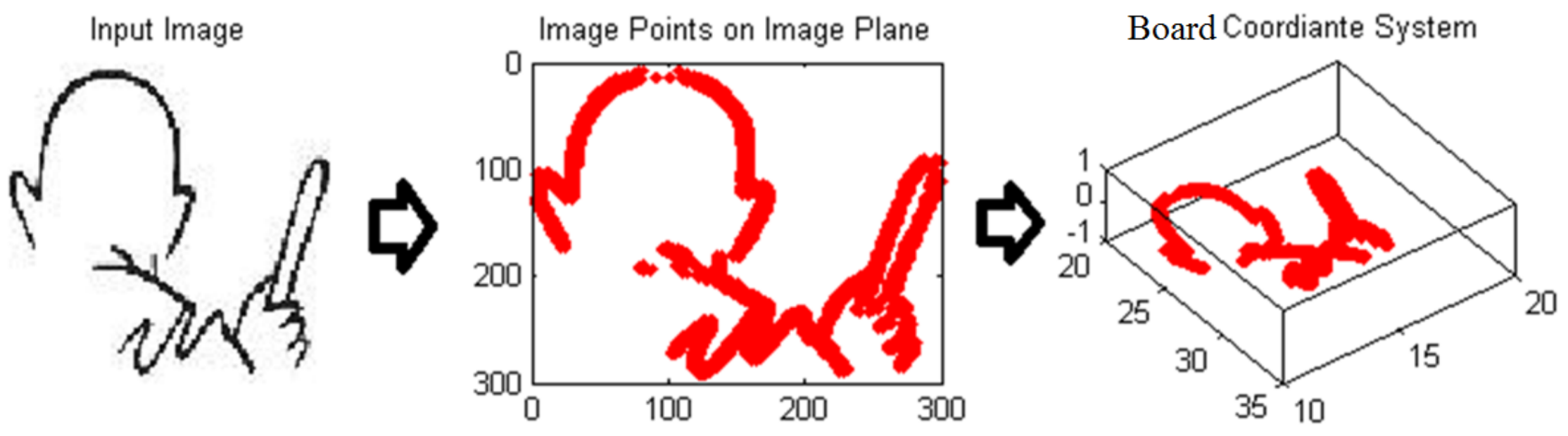

**Figure 4.3.2 Transformation from image Plane to board coordinate system (a) Raw Input image (b) Boundary points (x,y) extracted with canny edge detection (c) points corresponding to board coordinate system (after transformation)**

### 4.3.3 Calibration between Board and NAO End-Effector

The first step towards solving the inverse kinematic problem for NAO right hand is to establish a relation between the points lying on the Board surface with respect to NAO body coordinate. Both coordinate systems can vary with respect to each other in terms of scaling, translation and rotation. Hence, in order to make a proper alignment with these three parameters should be estimated first. As the Board coordinate and NAO coordinate both are represented in the same unit (meters) we assume that scaling parameter will not play any role here. Hence, it is worthy to estimate only the remaining two parameters. The calibration matrix which would transform the points from one coordinate frame to and other will be made up of rotation and translation parameters. In the given circumstances, this is the simple homogeneous transformation matrix given in equation (4.6)

$$(x'y'z')^{(\eta)} = H \times (x,y,z)^{\beta} \tag{4.6}$$

Where, $(x'y'z')^{(\eta)}$ represents points on the NAO coordinate frame. H is the homogeneous transformation matrix and $(x,y,z)^{\beta}$ is the point represented with respect to Board coordinate frame.

$$\begin{bmatrix} x' \\ y' \\ z' \\ 1 \end{bmatrix} = \begin{bmatrix} r11 & r12 & r13 & T_x \\ r21 & r22 & r23 & T_y \\ r31 & r32 & r33 & T_z \\ 0 & 0 & 0 & 1 \end{bmatrix} \times \begin{bmatrix} x \\ y \\ z \\ 1 \end{bmatrix} \tag{4.7}$$

The component of the calibration matrix is in the form of $H = \begin{bmatrix} R_{3\times3} & T_{3\times1} \\ 0_{1\times3} & 1 \end{bmatrix}$. R presents the rotation along x, y and z axis whereas T represents the translation along these axis. The combined rotation of all three axes is given in equation (4.8) while the translation is shown in equation (4.9):

$$R = \begin{bmatrix} \cos\theta_y\cos\theta_z & -\cos\theta_x\sin\theta_z + \sin\theta_x\sin\theta_y\cos\theta_z & \sin\theta_x\sin\theta_z + \cos\theta_x\sin\theta_y\cos\theta_z \\ \cos\theta_y\sin\theta_z & \cos\theta_x\cos\theta_z + \sin\theta_x\sin\theta_y\sin\theta_z & -\sin\theta_x\cos\theta_z + \cos\theta_x\sin\theta_y\sin\theta_z \\ -\sin\theta_y & \sin\theta_x\cos\theta_y & \cos\theta_x\cos\theta_y \end{bmatrix} \quad (4.8)$$

$$\text{Similarly } T = \begin{bmatrix} T_x \\ T_y \\ T_z \end{bmatrix} \quad (4.9)$$

Three different approaches fundamental matrix, 4 point calibration and artificial neural network based regression have been introduced to estimate the content of the homogeneous transformation matrix H. In order to solve the above equation, we have collected 60 points from the Board coordinate frame and corresponding points represented with respect to NAO end effector. We have randomly selected 15 points from both the coordinate system presented in Table 4.3.1. K is a constant and its value could be anything depending on Board coordinate system.

**Table 4.3.1 Points with respect to NAO and Board Coordinate**

| S.No | NAO Coordinate End Effector Position (cm) | | | Points on Board Surface(cm) | | |
|---|---|---|---|---|---|---|
| | $x'$ | $y'$ | $z'$ | $x$ | $y$ | $z$ |
| 1 | 13.43 | -6.05 | 31.73 | 21 | 10 | K |
| 2 | 13.41 | -7.11 | 31.69 | 22 | 10 | K |
| 3 | 13.26 | -8.13 | 31.73 | 23 | 10 | K |
| 4 | 13.08 | -9.11 | 31.72 | 24 | 10 | K |
| 5 | 13.16 | -10.24 | 31.67 | 25 | 10 | K |
| 5 | 12.97 | -17.02 | 31.65 | 32 | 11 | K |
| 6 | 13.19 | -16.05 | 31.65 | 31 | 11 | K |
| 7 | 13.39 | -15.28 | 31.65 | 30 | 11 | K |
| 8 | 13.49 | -14.33 | 31.64 | 29 | 11 | K |
| 9 | 13.69 | -13.33 | 31.61 | 28 | 11 | K |
| 10 | 15.95 | -6.77 | 31.39 | 21 | 12 | K |
| 11 | 15.79 | -7.76 | 31.50 | 22 | 12 | K |
| 12 | 15.55 | -8.90 | 31.54 | 23 | 12 | K |
| 13 | 15.38 | -9.80 | 31.56 | 24 | 12 | K |
| 14 | 15.26 | -10.72 | 31.57 | 25 | 12 | K |
| 15 | 18.53 | -7.13 | 31.70 | 21 | 15 | K |

### 4.3.4 Fundamental Matrix Approach

The problem of estimating the value of r11, r12, r13 ,….,r33 expressed in equation (4.9) is analogous to the estimation of the fundamental matrix in stereovision calibration. Fundamental matrix has the great impact in finding the correct epipolar line with respect to the given point either from left frame or right frame [217]. In the same way estimation of fundamental matrix will help us to find the representation of Board coordinate frame with respect to NAO coordinate frame or vice versa. To do, this it is mandatory that both the coordinate frame satisfy the equation of fundamental matrix given in equation (4.10).

$$(P^{\eta})_{4\times1} = (F)_{4\times4} \times (P^{\beta})_{4\times1} \tag{4.10}$$

$$P^{\eta} \times F \times P^{\beta} = 0 \tag{4.11}$$

Here $P^{\eta}$ represents the point of the same object corresponding the NAO end effector position, F is the fundamental matrix shown in equation (4.12) while $P^{\beta}$ represents the points corresponding to Board coordinate frame.

$$F = \begin{bmatrix} f1 & f2 & f3 & f4 \\ f5 & f6 & f7 & f8 \\ f9 & f10 & f11 & f12 \\ f13 & f14 & f15 & f16 \end{bmatrix} \tag{4.12}$$

Let the point $(x'y'z') \in \eta$ and $(x, y, z) \in \beta$ then putting these points in equation (4.13) we will get:

$$\begin{bmatrix} x' \\ y' \\ z' \\ 1 \end{bmatrix}^T \times \begin{bmatrix} f1 & f2 & f3 & f4 \\ f5 & f6 & f7 & f8 \\ f9 & f10 & f11 & f12 \\ f13 & f14 & f15 & f16 \end{bmatrix} \times \begin{bmatrix} x \\ y \\ z \\ 1 \end{bmatrix} = 0 \qquad (4.13)$$

The above equation can be further simplified and represented as given in equation (4.14)

$$f1xx^1 + f2yx^1 + f3zx^1 + f4x^1 + f5xy^1 + f6yy^1 + f7zy^1 + f8y^1 + f9xz^1 + f10yz^1 + f11zz^1 + f12z^1 + f13x + f14y + f15z + f16 = 0 \qquad (4.14)$$

This is the linear equation of 16 unknown; it can be solved by applying Hartley 8 point calibration [217] described in Algorithm 1 while the description of the symbol is presented in Table 4.3.2.

**Table 4.3.2 The description of the symbols used in algorithm 1**

| Symbol | Meaning |
|---|---|
| $(P')^{\eta}$ | Point corresponds to NAO end effector |
| $(P)^{\beta}$ | Point on Board Coordinate |
| N, $R^4$ | Number of points here (N=60), Real Space (4 dimensional) |
| $\mu'_x, \mu'_y, \mu'_z$ | Mean corresponding to x,y and z axis with respect to NAO coordinate frame |
| $\mu_x, \mu_{y,}\mu_z$ | Mean corresponding to x,y and z axis with respect to Board coordinate frame |
| $(\overline{\mu_1})^{\eta}, (\overline{\mu_2})^{\beta}$ | Mean corresponds to x,y and z axis for both NAO and Board coordinate frame |
| $T^{\eta}$, $T^{\beta}$ | Scale Translation matrix for both NAO and Board coordinate frame |

---

**Algorithm 1: Pseudo code for the Estimation of Fundamental Matrix**

---

**Input:** $(P'_i)^{\eta}$ and $(P_i)^{\beta}$ where $i \in [1, N], (P', P) \in \Re^4$ *and* $\forall (P', P) \exists (x, y, z, 1)$

**Output:** *Fundamental Matrix* : $F_{4\times4}$.

**Begin:**

***Step1:*** Calculate $\mu'_x \leftarrow \frac{1}{N}\sum_{i=1}^{N} x_i$ , $\mu'_y \leftarrow \frac{1}{N}\sum_{i=1}^{N} y_i$ , $\mu'_z \leftarrow \frac{1}{N}\sum_{i=1}^{N} z_i \; \forall \; (P')^{\eta}$,

similarly calculate $\mu_x, \mu_{y,}\mu_z \; \forall \; (P)^{\beta}$

***Step2:*** Calculate $(P^{1'}{}_{i})^{\eta}$: $(P^{1'}{}_{i}(x) \leftarrow P'_{i}(x) - \mu'_{x}, P^{1'}{}_{i}(y) \leftarrow P'_{i}(y) - \mu'_{y}, P^{1'}{}_{i}(z) \leftarrow P'_{i}(z) - \mu'_{z})$, similarly calculate $(P^{1}{}_{i})^{\beta}$.

***Step3:*** Calculate $(S_{1i})^{\eta} \leftarrow \sqrt{(x_i{}^2 + y_i{}^2 + z_i{}^2)} \& (S_{2i})^{\beta}\sqrt{(x_i{}^2 + y_i{}^2 + z_i{}^2)}$ where $i \in [1, N]$

***Step4:*** $(\overline{\mu_1})^{\eta} \leftarrow \frac{1}{N}\sum_{i\leftarrow 1}^{N}(S_{1i})^{\eta}$ $and$ $(\overline{\mu_2})^{\beta} \leftarrow \frac{1}{N}\sum_{i\leftarrow 1}^{N}(S_{2i})^{\beta}$

***Step5:*** $Scale^{\eta} \leftarrow \sqrt{3}/(\overline{\mu_1})^{\eta}$ and $Scale^{\beta} \leftarrow \sqrt{3}/(\overline{\mu_2})^{\beta}$

***Step6:*** $T^{\eta} \leftarrow \begin{bmatrix} Scale^{\eta} & 0 & 0 & -Scale^{\eta} \times \mu'_{x} \\ 0 & Scale^{\eta} & 0 & -Scale^{\eta} \times \mu'_{y} \\ 0 & 0 & Scale^{\eta} & -Scale^{\eta} \times \mu'_{z} \\ 0 & 0 & 0 & 1 \end{bmatrix}$ and

$$T^{\beta} \leftarrow \begin{bmatrix} Scale^{\eta} & 0 & 0 & -Scale^{\eta} \times \mu_{x} \\ 0 & Scale^{\eta} & 0 & -Scale^{\eta} \times \mu_{y} \\ 0 & 0 & Scale^{\eta} & -Scale^{\eta} \times \mu_{z} \\ 0 & 0 & 0 & 1 \end{bmatrix}$$

***Step7:*** $(P^{new'}{}_{i})^{\eta} \leftarrow T^{\eta} \times (P'_{i})^{\eta}$ and $(P^{new}{}_{i})^{\beta} \leftarrow T^{\beta} \times (P_i)^{\beta}$

***Step8:*** $from\ equation\ (9)\ //\ A \times F \leftarrow 0$

$$[x_i x_i' \quad y_i x_i' \quad z_i x_i' \quad x_i' \quad x_i y_i' \quad y_i y_i' \quad z_i y_i' \quad y_i' \quad x_i z_i' \quad y_i z_i' \quad z_i z_i' \quad z_i' \quad x_i\ y_i] \times \mathrm{F} \leftarrow 0$$

***Step9:*** Calculate $(eigen\ value)\lambda_i \& (eigen\ vector)\ \Psi_i\ of\ A$.

***Step10:*** Find $\Psi\ corresponds\ to\ minimum\ \lambda$ and assign it as F.

***Step11:*** $F \leftarrow \Psi_{4\times 4}$ // reshape $\Psi\ to\ make\ it\ a\ matrix\ of\ dimension\ 4 \times 4$.

***Step12:*** $Apply\ singular\ value\ decomposition\ (SVD) on\ F\ to\ calculate\ upper\ tringular\ matrix\ (U), (eigen\ value)\lambda\ \& (eigen\ vector)\Psi$

***Step13****:* $\mathrm{F} \leftarrow U_{4\times 4} \times [\lambda_{1,1} \quad \lambda_{2,2} \quad \lambda_{3,3} \quad 0]^{t}{}_{4\times 1} \times \Psi^{t}{}_{1\times 4}$

***Step14:*** $\mathrm{F} \leftarrow T^{\beta} \times F \times T^{\eta}$

***End:***

Hartley 8 point calibration [217] is the heart of the estimation of the fundamental matrix. It starts with collecting "n" points from both the coordinate system and normalization of these points to achieve the same scale. So that the solution will be scale independent. Singular value decomposition (SVD) of the collected points helps to find the characteristic root of the equation. Further the selection of minimum Eigenvector and representation in the form of a 4X4 matrix reduces the solution size. The degree of solution should be 3 as we have the 3 dimensional data. Therefore, it is required to reduce the rank of the matrix and to make it three. SVD is applied on the 4X4 matrix in order to make the rank 3 as well as to find the final fundamental matrix. The algorithm is explained in detail below.

**Results Obtained after Calibration**

The results are summarized in the form of projection errors involved after projecting each test point over the NAO's body coordinate frame. We have calculated absolute error which shows how the projected points differ from the actual points along each coordinate axis. The statistical analysis of standard deviation and mean square error depicted in Table 4.3.3 shows the efficiency of the calibration matrix.

**Table 4.3.3 Statistical Analysis of absolute error involved along each axis**

| Measurements | X Axis | Y Axis | Z Axis |
|---|---|---|---|
| Mean | 1.4508 | 1.0057 | 1.3379 |
| Standard Deviation | 0.63264 | 0.23652 | 0.29779 |
| Variance | 0.40023 | 0.05594 | 0.08868 |
| Mean Square Error | 2.4984 | 1.0664 | 1.8771 |

All these three measurements shown in Table 4.3.3 are calculated on the basis of all 60 samples. The mean and mean square error shown and the error involved in projection while the standard deviation and variance shows how the error are distributed along the mean. As Y axis has the mean and mean square error less than other two axes, it has the good projection direction. The variance along Y axis is less, hence the transformation along the Y axis is worthy than other two dimensions. Figure 4.3.3 reflect the absolute error involved along each projection direction for every sample point.

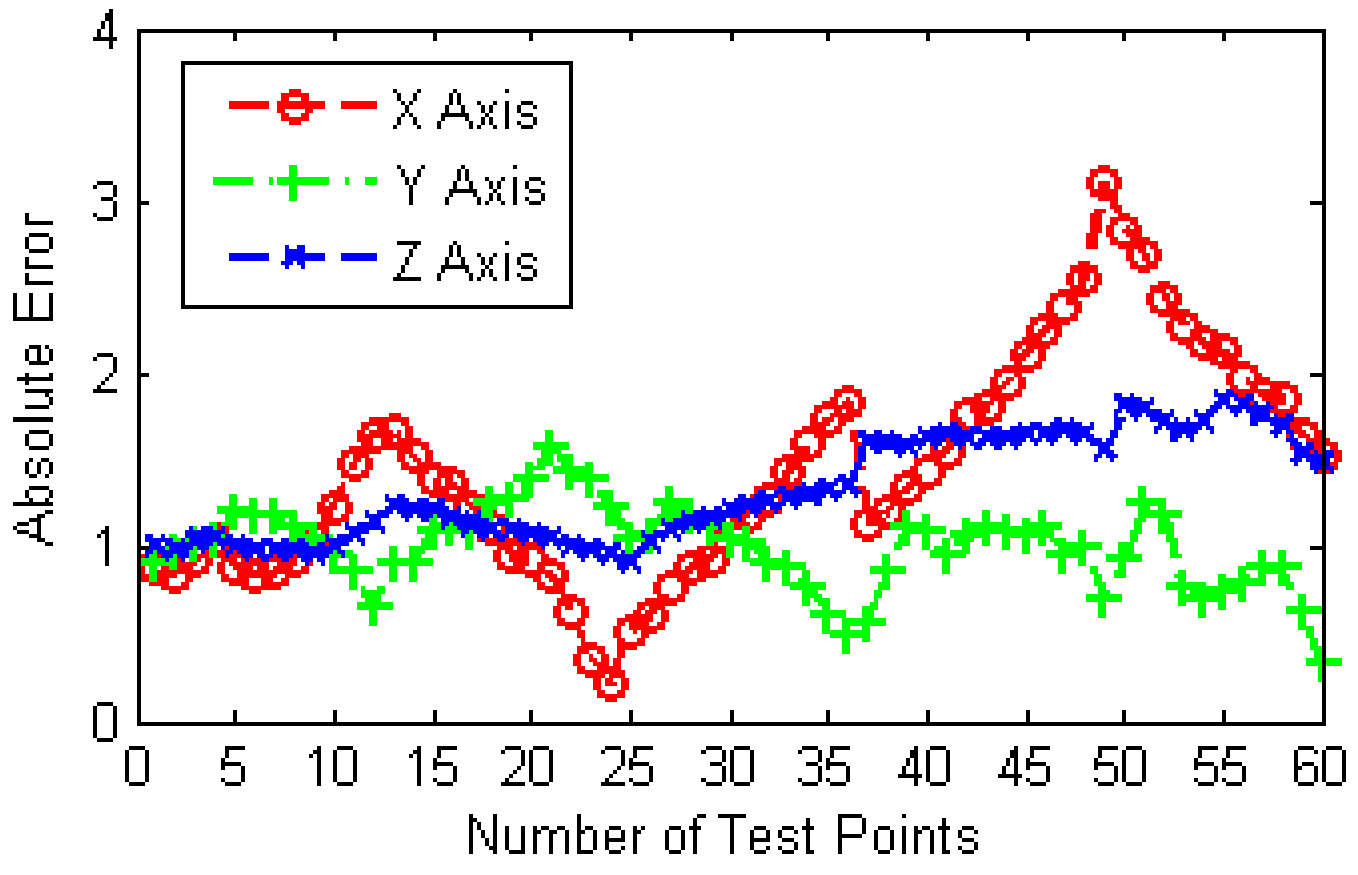

**Figure 4.3.3 Absolute error involved along X,Y, Z axis with respect to each sample point**

### 4.3.5 Four Point Calibration

The cost of transformation presented in equation (4.12) can be further estimated with the help of 4 point calibration. We need only 4 points of table and the corresponding points defined in NAO's body coordinate system. If we expand the equation (4.12) we will get following equations.

$$x' = f1 * x_i + f2 * y_i + f3 * z_i + f4 \qquad (4.15)$$

$$y' = f5 * x_i + f6 * y_i + f7 * z_i + f8 \qquad (4.16)$$

$$z' = f9 * x_i + f10 * y_i + f11 * z_i + f12 \qquad (4.17)$$

Here $i \in [1, n]$, n is the number of calibration points (we kept n=4). After expanding equation (4.15, 4.16 and 4.17) for all four calibration points, we will get.

$$\begin{bmatrix} x'_1 \\ x'_2 \\ x'_3 \\ x'_4 \end{bmatrix} = \begin{bmatrix} x_1 & y_1 & z_1 & 1 \\ x_2 & y_2 & z_2 & 1 \\ x_3 & y_3 & z_3 & 1 \\ x_4 & y_4 & z_4 & 1 \end{bmatrix} \times \begin{bmatrix} f1 \\ f2 \\ f3 \\ f4 \end{bmatrix} \qquad (4.18)$$

$$A = T * B$$

If we extend this solution for the rest of the two axes, we will see that the transformation matrix contents are same. This shows that there is no need to estimate the transformation matrix for each independent axis. The solution of equation (18) can be obtained by simply solving for $\mathrm{B} = \mathrm{T}^{-1} * \mathrm{A}$.

### Results Obtained after Calibration

The beauty of four point calibration is its least number of calibration points. Only four points are enough to establish the relation between these two coordinate frames. However, the validation of the transformation matrix is being evaluated with the help of four measurement units presented in Table 4.3.4.

**Table 4.3.4 Statistical Analysis of absolute error involved along each axis**

| *Measurements* | *X Axis* | *Y Axis* | *Z Axis* |
|---|---|---|---|
| Mean | 0.49288 | 0.38818 | 0.089551 |
| Standard Deviation | 0.44217 | 0.32692 | 0.061755 |
| Variance | 0.19552 | 0.10688 | 0.0038137 |
| Mean Square Error | 0.4352 | 0.2558 | 0.0118 |

The mean and other measurement units of the absolute errors along each axis show the strength of the projection matrix along these directions. As the mean, and variance along the z axis is very less, hence the projection error involved along z axis is very less for each sample point. The absolute error for each axis is displayed in Figure 4.3.4.

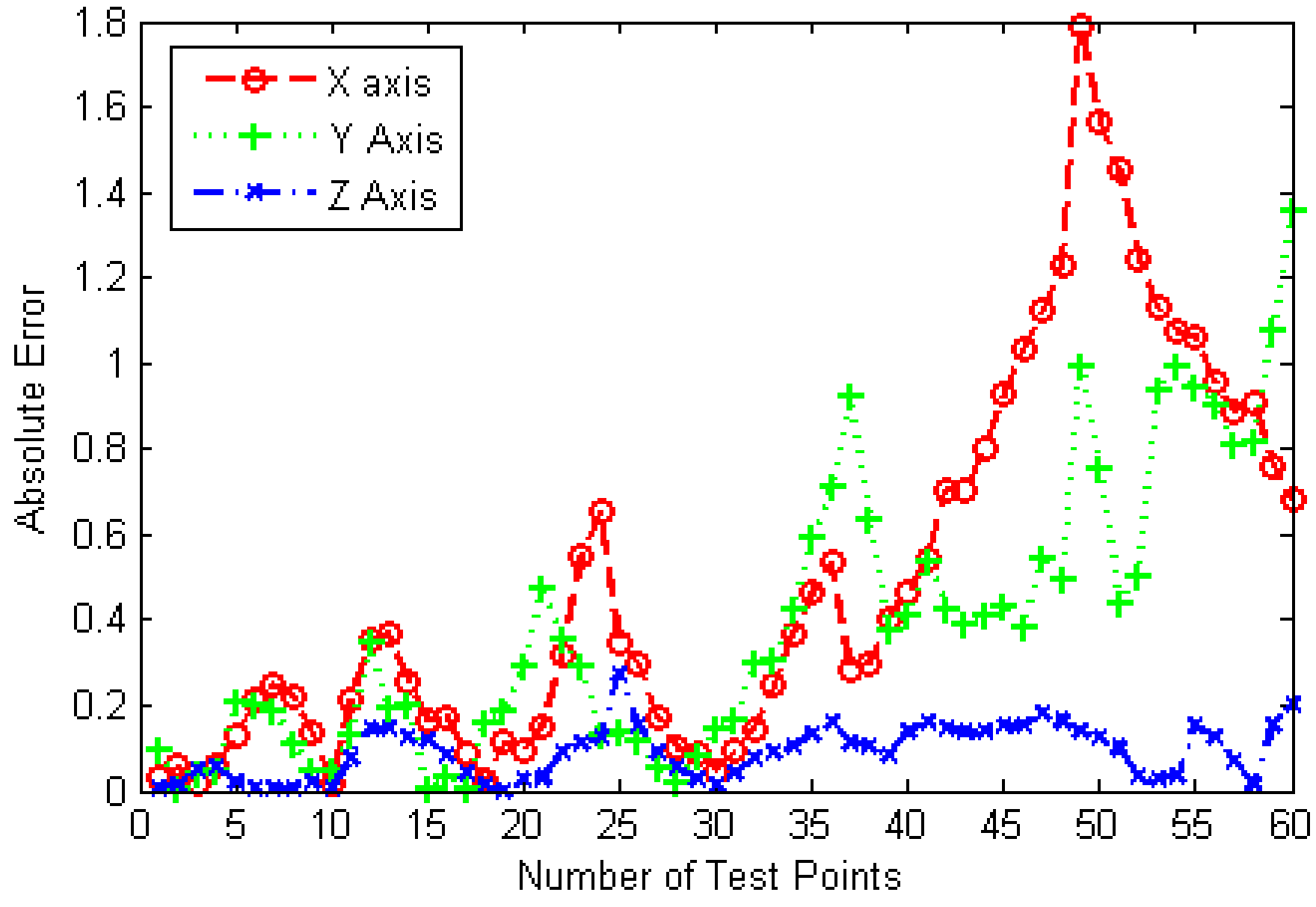

**Figure 4.3.4 Absolute error involved along X,Y, Z axis with respect to each sample point**

### 4.3.6 Regression Analysis with Artificial Neural Network

A linear regression analysis has been used to find out the predicted value of any of the table coordinate in terms of NAO's body coordinate. This is the case of linear regression analysis having one independent and one dependent variable. Table's coordinates are considered to be the independent coordinates whereas the NAO's coordinate system is assumed to be the dependent. If we use the regression analysis to estimate the value of the dependent variable, a prediction function always exists which transforms the value of the independent variable to the predicted value. As this is the case of linear regression, the prediction function would be the equation of a line. Given as: $y = m * x + c$, here y is the predicted value of the input variable x, m is the magnitude and c is the coefficient of the line and it is a constant value. We took help of artificial neural network to find the coefficient of the line. Artificial Neural Network is widely known as the abbreviation of ANN, it has the similar analogical structure which we are having in our brain. It is basically three layer architecture shown in Figure 4.3.5, wherein the first layer we supply input to the system, the second layer is the hidden layer where learning takes place and at the third layer output is defined.

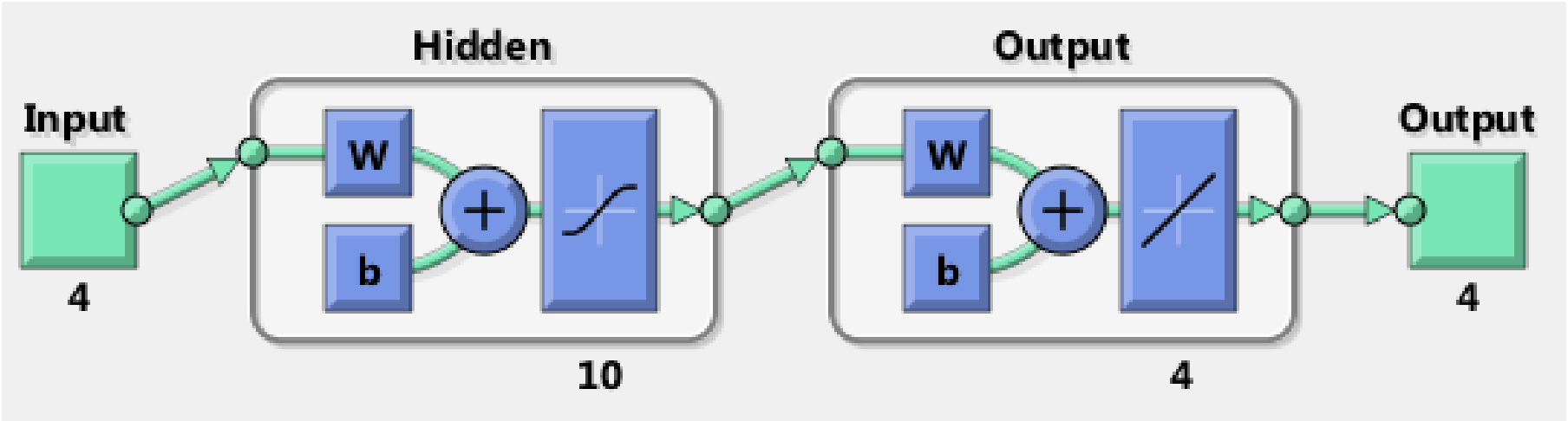

**Figure 4.3.5 Artificial Neural Network Structure**

**Table 4.3.5 ANN Specification**

| | | | |
|---|---|---|---|
| Input | 4 dim | Stopping Criteria | MSE |
| Hidden Layer Neurons | 10 | Training Population | 75% |
| output | 4 dim | Test Population | 15% |
| Training Function | trainlm | Validation | 15% |
| Learning Constant | 0.001 | | |

The structure of neural network consists of one input layer which accepts a four dimensional table coordinates as an input vector. One hidden layer having 10 neurons, one output layer having four neurons in the output corresponds to NAO's body coordinate frame. The specification of the Artificial Neural network Structure is expressed in Table 4.3.5. There are 60 samples of Tables and of NAO's body coordinates, we kept 75% of the population as training while 15% for both the testing and validation. We have used random function to split the population over training, testing and validation population. We have used Lovenberg Marquardt back propagation [218] algorithm in order to train the network. Sigmoidal function ($F(X) = 1/(1 + e^{-x})$) is used as the transfer function form input to hidden layer, while linear transfer function ($f(x) = a * x + b$) is used as the activation function from hidden layer to output layer. Here a & b are the coefficient of the line. The learning constant is 0.001 and the stopping criteria is the mean square error, if the mean square error is less than the threshold, we will stop the weight updating as well as training phase.

**Design of Neural Network Learning Algorithm for estimating Calibration matrix**

The primary aim of training algorithm is to update the weight matrix and biases involved in the network. As soon as the weight and biases are updated and there is no further change in them, we believe that the network is trained. The training section has several sub modules define below.

**Step 1: Feed Forward (Input to output):** Let "$\underline{p}$" be the input to the network and $a^k$is the output at layer $k, k + 1, .. k + 2, ... M$. The computation performed for each neuron at each layer will be:

$$\underline{a^{k+1}} = \underline{f^{k+1}}(w^{k+1}\underline{a^k} + \underline{b^{k+1}}) \text{ where } k = 0.1.2 \ldots., M - 1 \quad (4.19)$$

Here "w"is the weight matrix, M represents the number of layer in the neural network, f" denotes the transfer function and "b" represents the bias corresponding to layer "k". At the first layer output will be equal to the input vector: ($\underline{a^0} = \underline{p}$). The principle of learning algorithm is to minimize the gap between the input vector ($\underline{p}$)and the target vector($\underline{t}$).The training of the network is evaluated based on the fitness or performance function. In this case the performance function is evaluated with

$$V = \frac{1}{2}\sum_{i=1}^{Q}(\underline{t}_q - \underline{a}^M{}_q)^T(\underline{t}_q - \underline{a}^M{}_q) = \frac{1}{2}\sum_{q=1}^{Q}\underline{e}_q{}^T\underline{e}_q \quad (4.20)$$

Here $\underline{a}^M{}_q$ is the output at the given input $\underline{p}_q$while the error is evaluated based on the difference between the target and the desired. For the $q^{th}$ input it is calculated by: $\underline{e}_q = \underline{t}_q - \underline{a}_q{}^M$.

**Step 2: Weight update:** The error could be minimized by taking the derivative of the error the same way as we do in case of gradient descent learning algorithm. The change in weight ($\Delta$w) and bias ($\Delta$b) will be calculated by:

$$\Delta w^k(i,j) = -\alpha\frac{\partial\hat{V}}{\partial w^k(i)} \text{ and } \Delta b^k(i,j) = -\alpha\frac{\partial\hat{V}}{\partial b^k(i)} \quad (4.21)$$

Here $\alpha$ is the learning constant defined by:

$$\partial^k(i) \equiv= \frac{\partial\hat{\hat{V}}}{n^k(i)} \quad (4.22)$$

Similarly the change with respect to each layer with respect to the performance function V is:

$$\frac{\partial\hat{V}}{\partial w^k(i,j)} = \frac{\partial\hat{V}}{\partial n^k(i)}\frac{\partial n^k(i)}{\partial w^k(i,j)} = \partial^k(i)a^{k-1}(j) \quad (4.23)$$

$$\frac{\partial\hat{V}}{\partial b^k(i)} = \frac{\partial\hat{V}}{\partial n^k(i)}\frac{\partial n^k(i)}{\partial b^k(i)} = \partial^k(i) \quad (4.24)$$

**Step 3: Back Propagation (Output to Input):** Let $\underline{x}$ be the input vector and V is the fitness function. In order to fit a best curve it is required that the $V(\underline{x})$ should be minimum. If we apply the Newton's method it would be:

$$\Delta\underline{x} = -[\nabla^2 V(\underline{x})]^{-1}\nabla V(\underline{x}) \quad (4.25)$$

Where $\nabla^2 V(\underline{x})$is the Hessian matrix and $\nabla V(\underline{x})$ is the gradient. As required the intention of the fitness function is to minimize the sum of square error between the target and the desired and it can be calculated by:

$V(\underline{x}) = \sum_{i=1}^{N} e^2{}_i(\underline{x})$where"e" represent the error for the corresponding input "i". The error can also be minimized by calculating the first and second derivative of the error and then by compensating it by sending it back to the network.

$$\nabla V(\underline{x}) = J^T(\underline{x})\underline{e}(\underline{x}) \quad (4.26)$$

$$\nabla^2 V(\underline{x}) = J^T(\underline{x})J(\underline{x}) + S(\underline{x}) \quad (4.27)$$

Here $J(\underline{x})$ represents the Jacobian matrix for the input vector $(\underline{x})$.

$$\underline{(x)} = \begin{bmatrix} \frac{\partial e_1(x)}{\partial x_1} & \frac{\partial e_1(x)}{\partial x_2} & \cdots & \frac{\partial e_1(x)}{\partial x_n} \\ \frac{\partial e_2(x)}{\partial x_1} & \frac{\partial e_2(x)}{\partial x_2} & \cdots & \frac{\partial e_2(x)}{\partial x_n} \\ \vdots & \vdots & \ddots & \vdots \\ \frac{\partial e_N(x)}{\partial x_1} & \frac{\partial e_N(x)}{\partial x_2} & \cdots & \frac{\partial e_N(x)}{\partial x_n} \end{bmatrix} \quad (4.28)$$

In this case the sum of square can also be represented in the form of:

$$S(\underline{x}) = \sum_{i=1}^{N} e_i(\underline{x})\nabla^2 e_i(\underline{x}) \quad (4.29)$$

If the sum of square is almost zero or tending to be zero ($S(\underline{x}) \cong 0$) then equation 16 will become (when $S(\underline{x}) \cong 0$, this is the special case of Gauss − Newton method).

$$\Delta\underline{x} = [J^T(\underline{x})J(\underline{x})]^{-1}J^T(\underline{x})\underline{e}(\underline{x}) \quad (4.30)$$

But when the $S(\underline{x}) \cong 0$, the problem can be modified and solved by using Marquardt-Levenberg [218].

$$\Delta\underline{x} = [J^T(\underline{x})J(\underline{x}) + \mu\beta]^{-1}J^T(\underline{x})\underline{e}(\underline{x}) \quad (4.31)$$

Here $\mu$ is the learning constant multiplied to a compensating factor $\beta$ whenever sum of square error increase and divided whenever the sum of square error reduces.

**Performance Evaluation:** The network has converged very fast, just within 12 epochs. The sum of square error is minimized and the training stopped when it reached to 0.0040 of mean square error. Figure 4.3.6 show the convergence rate of the network. A cross validation technique which divides the total population into their respective proportionate of training, testing and validation has been put to control over learning. Usually the sample points are distributed randomly over these three sets. Each set has an individual role in the design of neural net like training, taking care of the weight updating, validation is responsible for the controlled learning (to avoid the over fitting) and testing module ensures the reliability of prediction/approximation. The quality of the approximation can be evaluated by:

**Mean Square Error (MSE):** Mean Square Error is a parameter to evaluate the error in each epoch. In total, 75% of the total population has been feed forwarded in each epoch. A definte error is involved against each sample point. The MSE of these error shows the global error of that particular epoch. If the MSE approaches to zero, we believe that the error is going to minimize and soon it will converge. The MSE can be calculated using the following expression

$$\mathrm{MSE} = \sqrt[2]{\mathrm{e1}^2 + \mathrm{e2}^2 + \mathrm{e3}^2 + \mathrm{e4}^2 + \cdots . + \mathrm{en}^2} \qquad (4.32)$$

Here $\mathrm{e1}^2 + \mathrm{e2}^2 + \mathrm{e3}^2 + \mathrm{e4}^2 + \cdots . + \mathrm{en}^2$ are the error with respect to each sample point and each can be calculated by $(\mathrm{error} = \mathrm{Output}_{\mathrm{desired}} - \mathrm{Output}_{\mathrm{actual}})$. In our case the desire output is the NAO end effector output while the actual output is what we are getting from the network at the output layer.

**Pearson's Correlation Coefficient (R):** The second evaluation criteria is the correlation between the points of NAO end effector and table points. If the correlation between these two are approaching to one, this means their differences are turning to zero. It means the errors are going to minizes and network is going to converge. The Pearson correlation can be calculated based on the variance of Table Points ($\sigma_{\mathrm{table}}$), variance of respective NAO's end effector points ($\sigma_{\mathrm{NAO}}$) and the co-variance between the Table points and NAO end effector points ($\Psi_{\mathrm{table,NAO}}$). $\sigma_{\mathrm{table}} = \sum(\mathrm{x_i} - \bar{\mathrm{x}})^2$here x is the point belonging to table and $\bar{\mathrm{x}}$ is the mean of all x $\in$ table, similarly $\sigma_{\mathrm{NAO}} = \sum(\mathrm{y_i} - \bar{\mathrm{y}})^2$here y is the point belonging to NAO and $\bar{\mathrm{y}}$ is the mean of all y $\in$ NAO. For both the populations$\mathrm{i} = 1$ to n, after expanding both the equations we will get $\sigma_{\mathrm{table}} = \sum \mathrm{x}^2 - \mathrm{n}\bar{\mathrm{x}}^2$and $\sigma_{\mathrm{NAO}} = \sum \mathrm{y}^2 - \mathrm{n}\bar{\mathrm{y}}^2$, now the co-variance between these two can be calculated as:

$$\Psi_{table,Nao} = \sum(x_i - \bar{x})(y_i - \bar{y}) \qquad (4.33)$$

$$\Psi_{table,Nao} = \sum xy - n\bar{x}\bar{y} \qquad (4.34)$$

$$R = {}^{\Psi_{table,Nao}}\!/_{\sigma_{table} \times \sigma_{Nao}} \qquad (4.35)$$

$$R = {}^{\sum xy - n\bar{x}\bar{y}}\!\Big/_{\sum x^2 - n\bar{x}^2 \times \sum y^2 - n\bar{y}^2} \qquad (4.36)$$

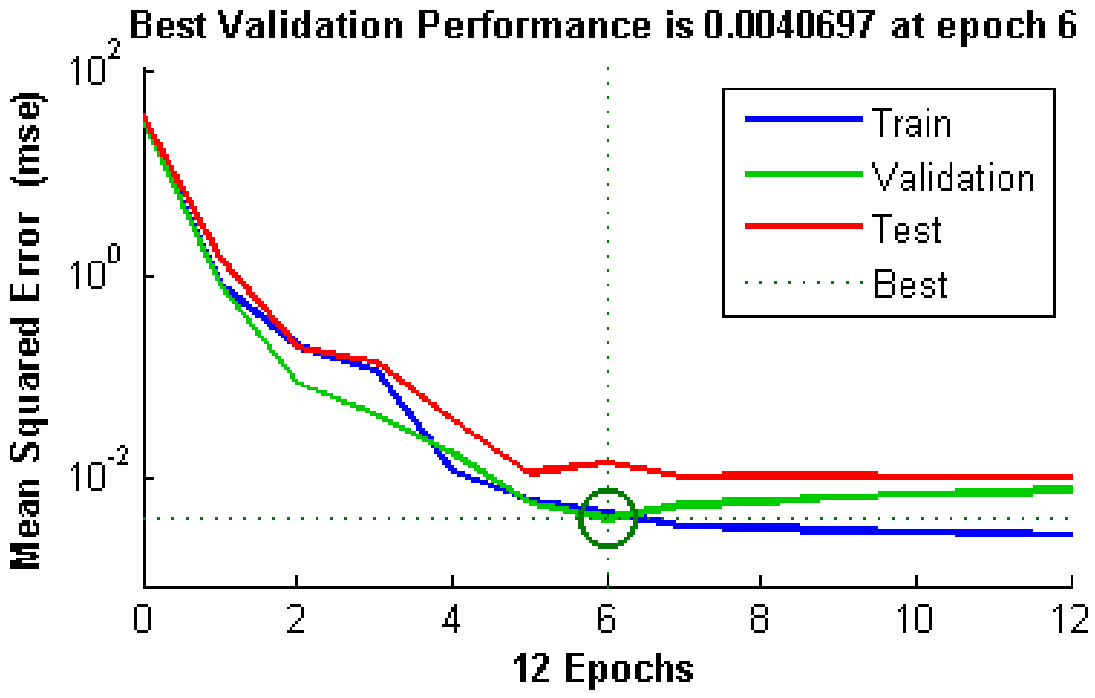

**Figure 4.3.6 Convergence of the network for training, testing and validation points (number of epochs VS mean square error)**

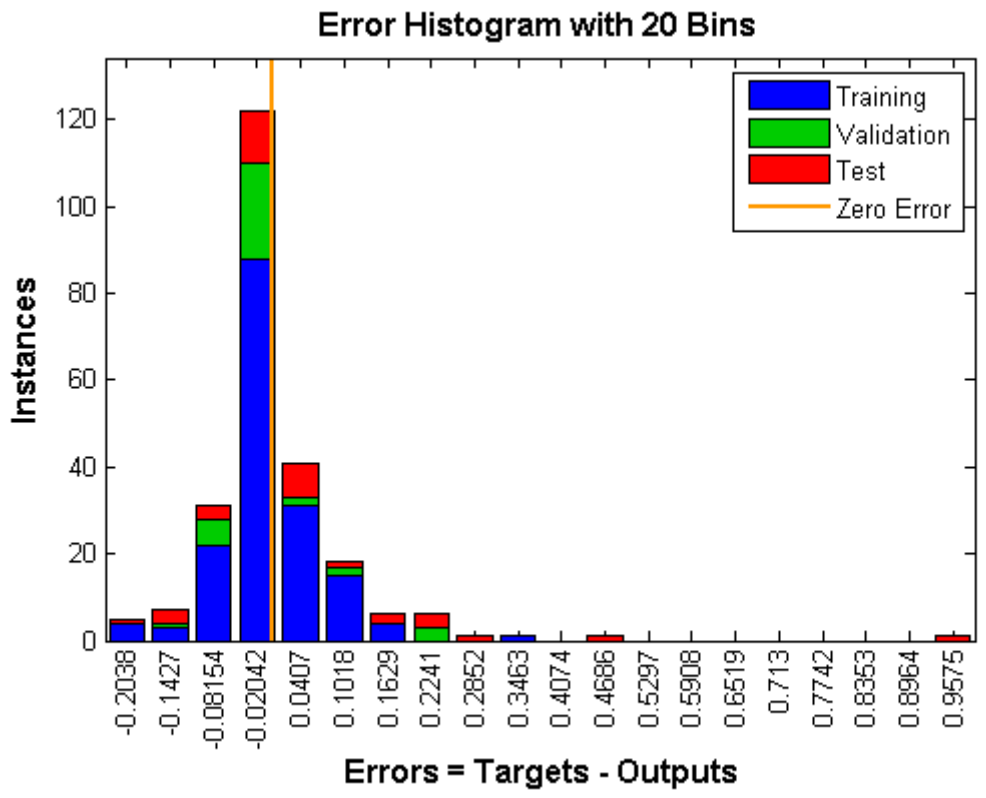

**Figure 4.3.7 Error Histogram (window size 0.0612) for the training, testing and validation points.**

Figure 4.3.7 shows the error involved in the different stages of training, testing and validation. Each one is symbolized with a different color, i.e. blue is representing the training, green for validation and red for testing. The error involved in the approximation is grouped at the interval of .0612. There are 20 bins used to represent the error histogram where most of the errors lie between the -0.2038 to 0.2241. Apart from showing the distribution of error, the histogram also helps to identify the outliers, for example the testing point with an error of 0.9575 and 0.4686 and training point with an error of 0.3463 can be considered as outliers. It is always better to remove the outer from the training set, otherwise it can mislead the network, and enforce it to fit the outlier by extrapolating the given input points. The regression plots of each of the phases (test, train and validation) are also depicted in Figure 4.3.9, Figure 4.3.8, Figure 4.3.10 and Figure 4.3.11 which confirms that the error is minimized. All the input points are falling on the regression line, which proves that the variance between the points are less and it has been minimized. There are 60 points corresponding to each axis, further these 60 points are modulated between the training, testing and validation. We have considered 75% of the population as training while the 15% and 10% population is used for validation and testing. In total 45 number of points is being used for training, 9 points used for validation and 6 points are used for testing the network. The Pearson correlation coefficient "R" is almost 1 in all the cases which ensures the efficiency of approximation.

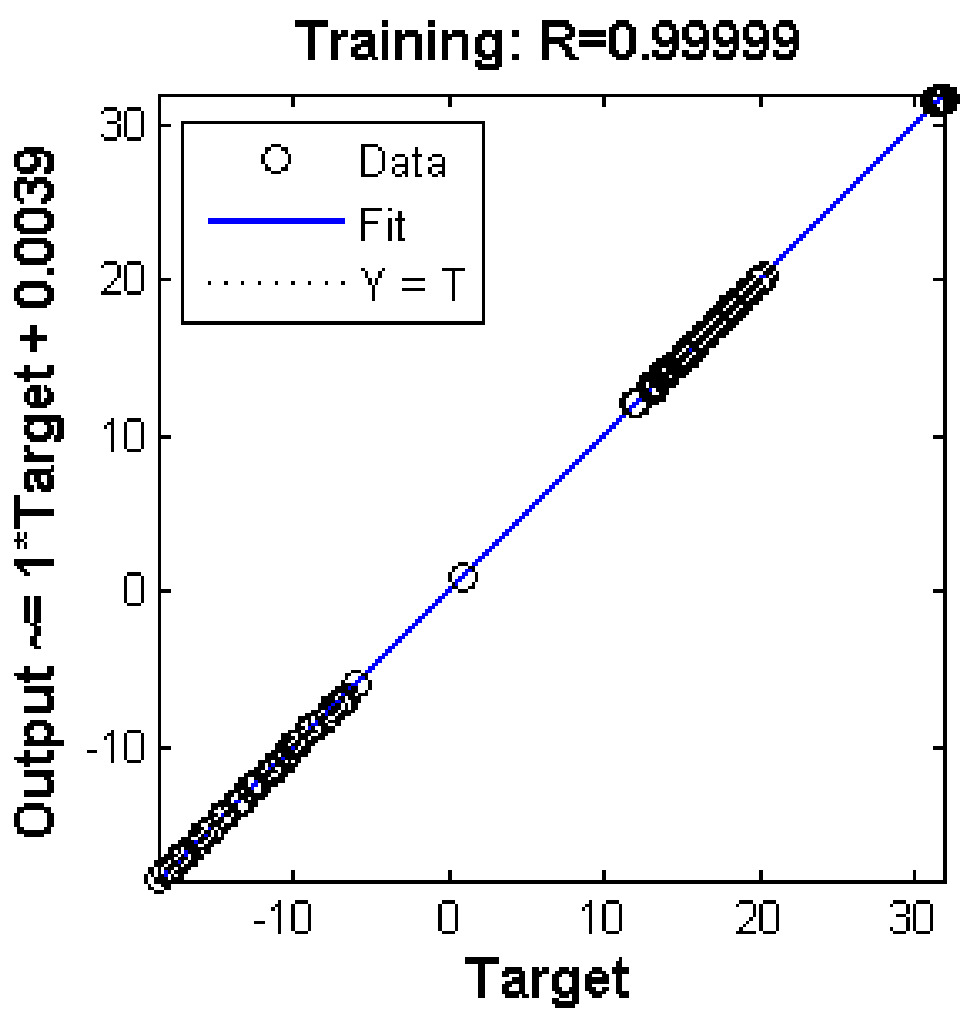

**Figure 4.3.9 Regression plot between target and actual output for the training points**

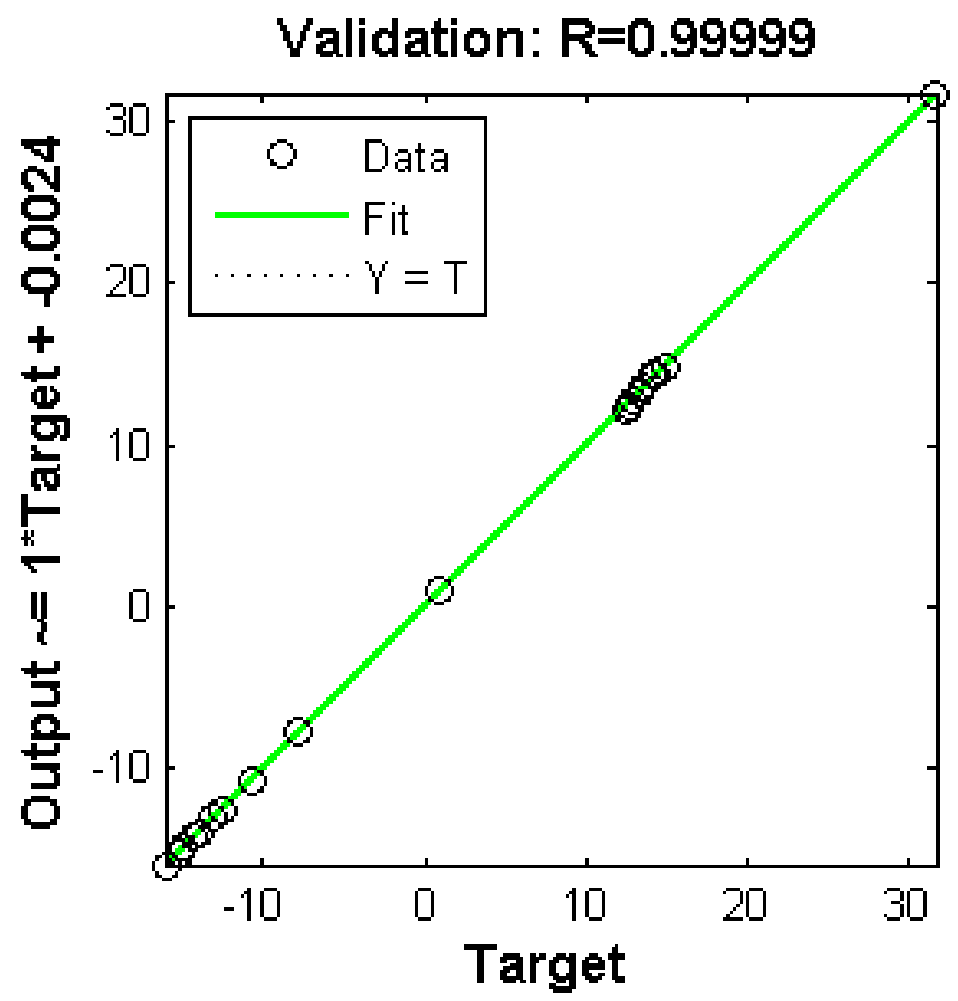

**Figure 4.3.8 Regression plot between target and actual output for the validation points**

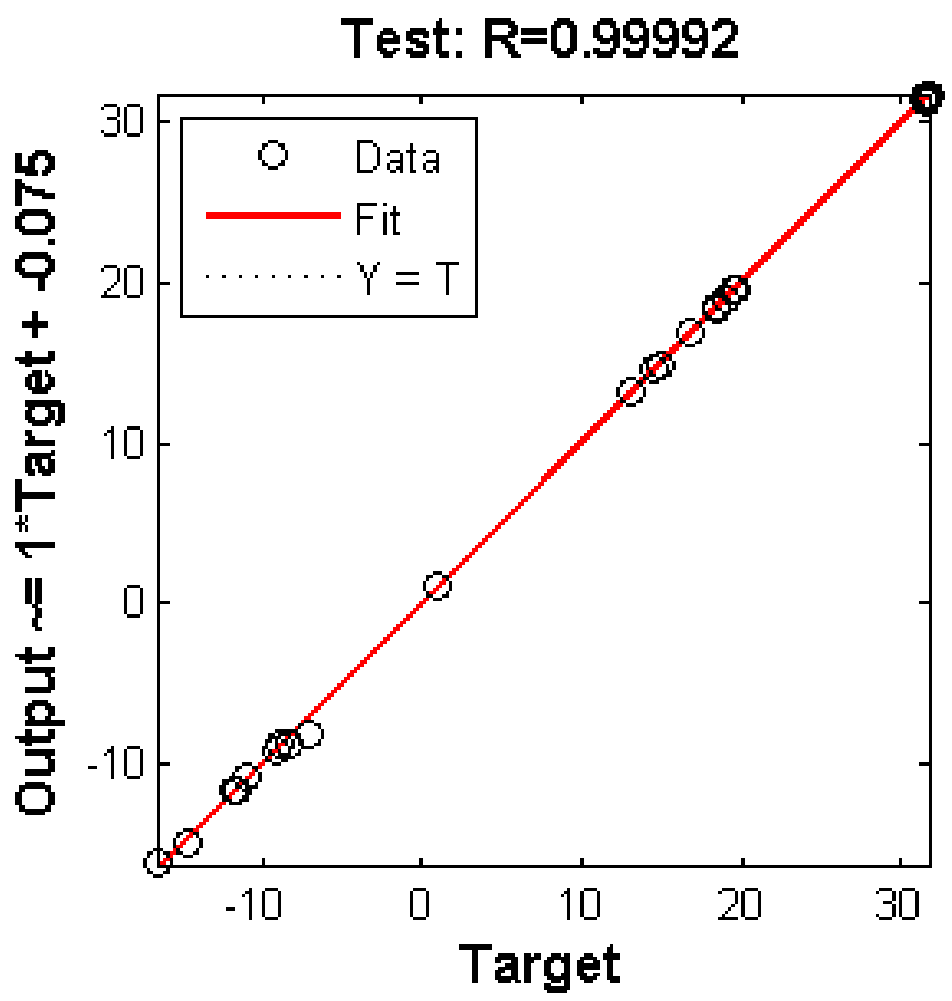

**Figure 4.3.11 Regression plot between target and actual output for the testing points**

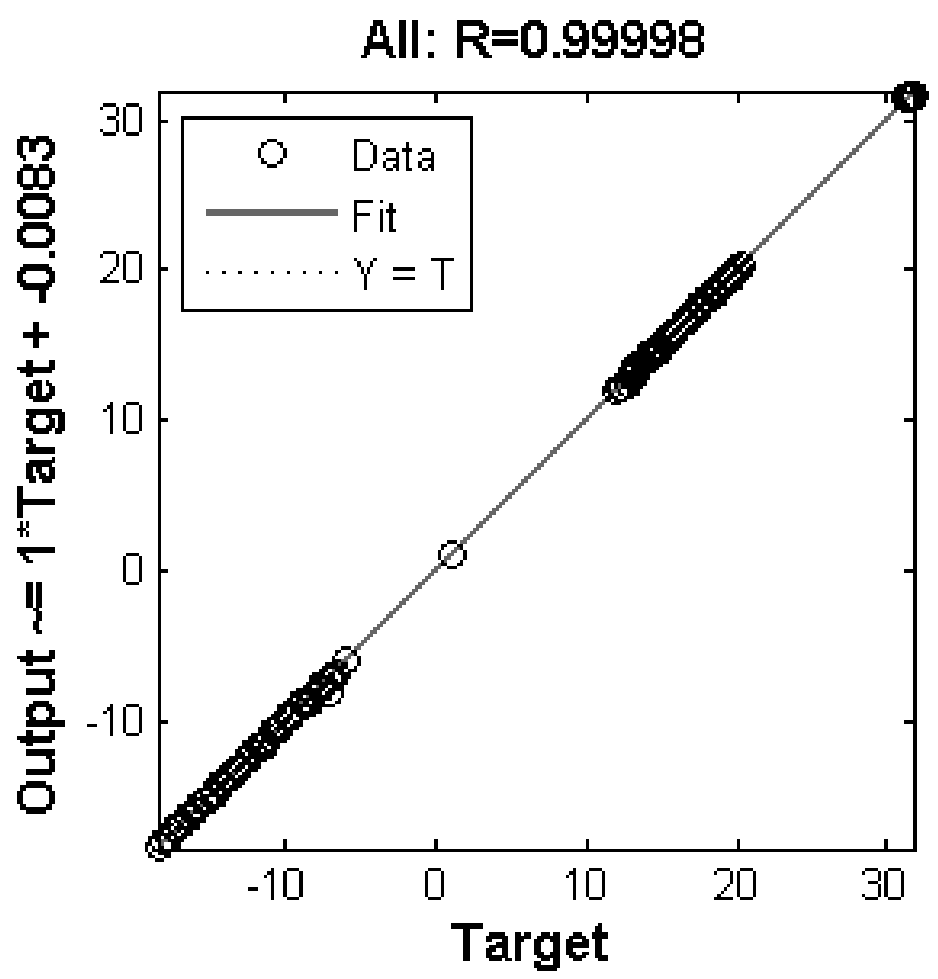

**Figure 4.3.10 Regression plot between target and actual output for the all training, validation and testing points**

The regression plot shown in Figure 4.3.9, Figure 4.3.8, Figure 4.3.10 and Figure 4.3.11 reflects the error of projection involved in training, testing and validation phases. A generalized regression line has been developed which can fit to every test point corresponding to all x, y and z axes. In total we have 27 validation points, 18 testing points and 15 training points corresponding to x, y and z axis.

**Results Obtained after Calibration**

After applying the regression analysis with the help of Artificial Neural Network, we have achieved very good approximation results. The gap between the projected table points over the NAO's body coordinate frame and the NAO's actual points are very less. The mean square error which was also the stopping criteria for ANN has the minimum values, which ensures that the variations between these two are very less. Rest of the statistical performance metrics are summarized in Table 4.3.6.

**Table 4.3.6 Statistical performance matrices**

| Measurements | X Axis | Y Axis | Z Axis |
|---|---|---|---|
| Mean | 0.0079 | -0.0089 | -0.0025 |
| Standard Deviation | 0.0922 | 0.1175 | 0.0636 |
| Variance | 0.0085 | 0.0138 | 0.0040 |
| Mean Square Error | 0.0084 | 0.0137 | 0.0040 |

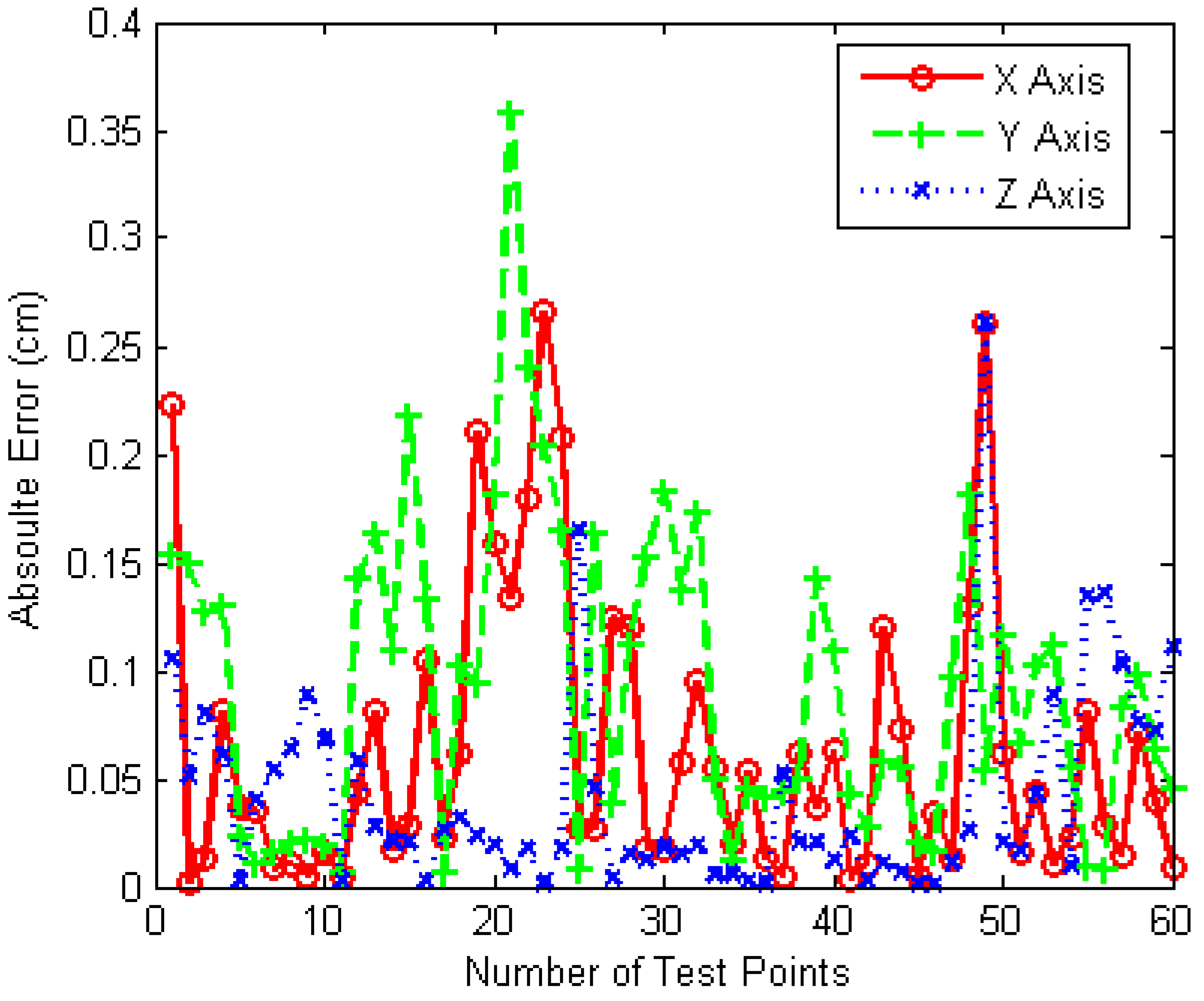

**Figure 4.3.12 Absolute error involved along X,Y, Z axis with respect to each sample point**

The mean and variance along each direction is nearer to zero, which ensures that the error in transformation is very less (almost zero). The projection towards Z axis, we can see that for every test point (we have assumed all points as test) the difference between the approximated point (actual) and desired point is very less. Desired points are the points described with respect to NAO end effector position extracted from the inertial encoders reading. Actual points are the approximated point defined with the help of ANN with respect to the given table point. The absolute error involved against each projection direction is expressed in Figure 4.3.12. The absolute error along Z axis is less in comparison to X and Y axes. The absolute error for each axis is very less and due to this we can see an overlapping between the actual and desires for all the test points.

### 4.3.7 Comparative Analysis of Proposed Solution

We have considered three parameters (a) mean square error involved in projection and (b) time complexity analysis. The mean square error reported for each technique along their X, Y and Z projection directions are depicted in Figure 4.3.16. If the mean square error is less and closure to zero, the projection along that direction is considered to be good. It is clearly visible from the comparative analysis of all the techniques that ANN based regression analysis has a minimum mean square error. Hence it could be considered as the best method to calibrate.

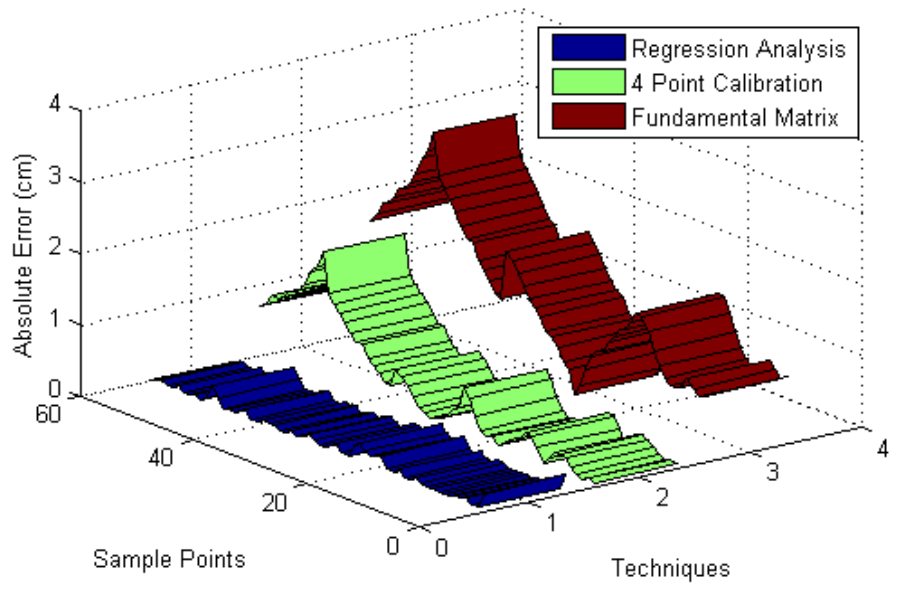

**Figure 4.3.14 Absolute error along X axis for each sample point with respect to each transformation technique**

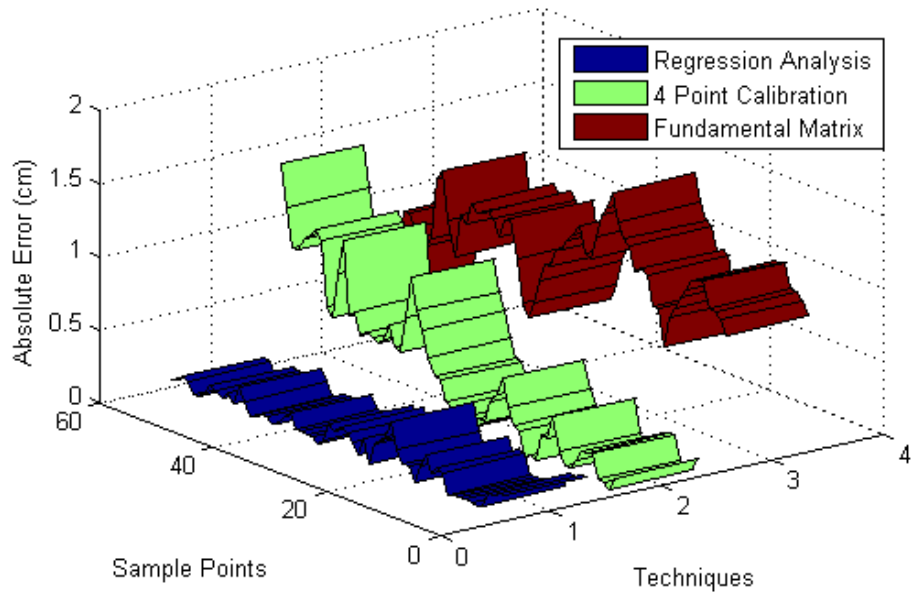

**Figure 4.3.13 Absolute error along Y axis for each sample point with respect to each transformation technique**

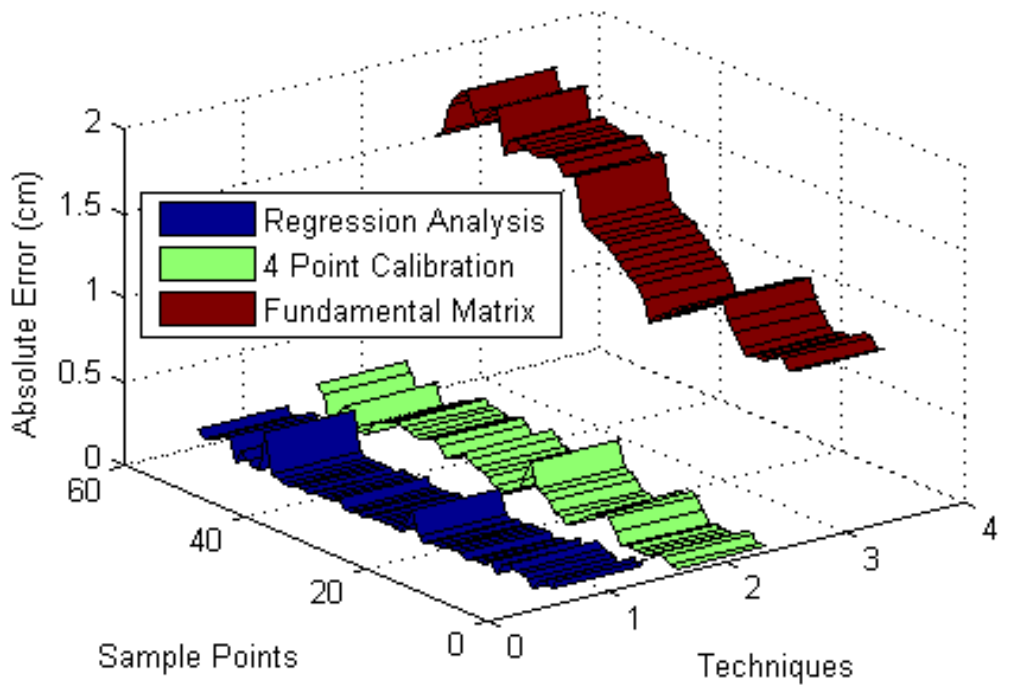

**Figure 4.3.15 Absolute error along Z axis for each sample point with respect to each transformation technique**

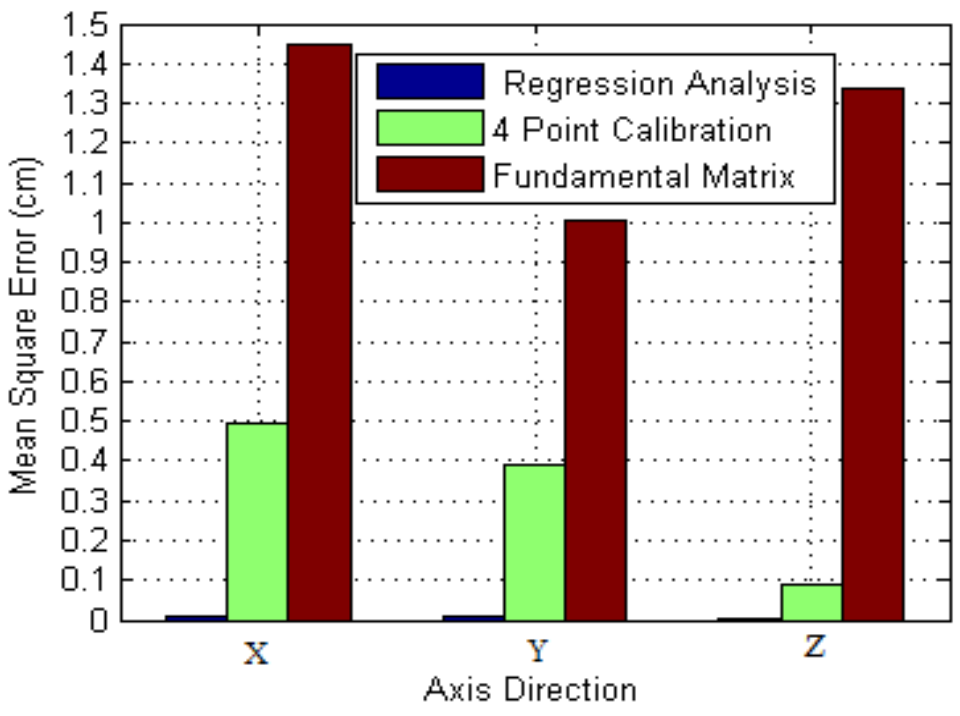

**Figure 4.3.16 Mean Square Error of each transformation matrix along each projection direction**

The absolute error along X direction is depicted in Figure 4.3.14 which shows for each sample point how much projection error is involved in regression analysis, 4 point calibration and fundamental matrix. Almost for every point the absolute error of regression analysis is less tending to zero; in comparison to other two approaches. The same has been reported along Y and Z direction in Figure 4.3.13 and Figure 4.3.15. The performance of all three techniques can be ranked on the basis of mean square error and their absolute error for every point. These two parameters indicate that regression analysis is the best way to estimate the calibration matrix followed by the 4 point calibration and the fundamental matrix approaches. The other two performance measurement techniques are discussed below.

## Time Complexity Analysis

The time complexity of each technique is evaluated based on the time involved in establishing the homogeneous transformation matrix. The time complexity of the algorithm is estimated based on its each individual step. These algorithms are less iterative in nature while more mathematical. The time taken to execute the mathematical equation is usually less than the iterative approaches. Therefore the time complexity of these algorithms is not very high. Number of sample data points play a major role in estimating the algorithm complexity. If the number of sample points is less, it will require less time to estimate the contents of calibration matrix. Therefore the time complexity of the 4 point calibration is less than the other two techniques. The time complexity is estimated only on the basis of steps required for establishing the calibration matrix. It does not consider the time involved in projecting the input points to the output.

- **Fundamental Matrix Approach**

  The total complexity of the algorithm is evaluated after summing up all the individual steps listed in Table 4.3.7. The algorithm is composed in 14 steps. The complexity of most of the steps are constant while for step 8,9 and 10 it is variable and dependent on the number of points. As the number of sample points are known to us, the total time complexity is also assumed to be somewhat constant. Total Running Time: $T_1 + T_2 + T_3 + T_4 + T_5 + T_6 + T_7 + T_8 + T_9 + T_{10} + T_{11} + T_{12} + T_{13} + T_{14} + n + n + n + n^p + n + T_{15} + n^p + T_{16} + T_{17}$.

**Table 4.3.7 Time Complexity of Fundamental matrix**

| Steps | Complexity |
|---|---|
| **1** | $T_1 + T_2 + T_3$ // $meancalculation$ |
| **2** | $T_4 + T_5 + T_6$ // $meanalignment$ |
| **3** | $T_7 + T_8$ // $magnitudealongboardandnao$ |
| **4** | $T_9 + T_{10}$ // $meanofmagnitudes$ |
| **5** | $T_{11} + T_{12}$ // $scallingforeach$ |
| **6** | $T_{13} + T_{14}$ //$ScaleTransformation$ |
| **7** | $n + n$ //$pointtransformation$ |
| **8** | $n$ //$pointmultiplication$ |
| **9** | $n^3$ // $svd\ of\ nX16\ matrix$ |
| **10** | $n$ // $findingminimum$ |
| **11** | $T_{15}$ //$reshaping$ |
| **12** | $4^3$ // $svd\ of\ 4X4\ matrix$ |
| **13** | $T_{16}$ //$4X4\ matrixmultiplication$ |
| **14** | $T_{17}$ //$assignment$ |

$T_1, T_2, T_3, \ldots, T_{17}$are having the unit constant time O(1). So we can eliminate them as they have little effect on the overall complexity. If we leave these unitary constants, then we will be left with: $4n + 4^3 + n^3$. There are two terms involved in the complexity one is linear and the other one is a polynomial of degree 3. The degree of the polynomial depends on the number of unknown variables. The first term is linear, hence its complexity will be O (n), while the complexity of polynomial will be $n^3$. Therefore, the complexity of the fundamental matrix approach is $n^3$.

- **Four Point Calibration**

  There are two steps involved in the four point calibration. First step is two find the inverse of a 4X4 square matrix and second step consists of a simple matrix multiplication. Therefore the complexity of the four point calibration would be: $n^3 + O(1)$. $n^3$is the complexity of the n×n dimension matrix interspersion and O(1) is the complexity of matrix multiplication. Our matrix is having the dimension 4, so the above complexity will reduce to $4^3$. This is also a constant term, hence the total time complexity of the four point calibration will be constant.

- **Artificial neural network based regression analysis**

  A multilayer neural network consists of two major building blocks (1) feed forward and (b) back propagation. The feed forward module feeds inputs from the input layer to the output layer and the back propagation modules propagate the errors from the output layer to the input layer. Hence the complexity of both the modules are evaluated separately and combined at the end.

  a) **Complexity of Feed Forward:** There are 4 inputs, 10 hidden neurons and 4 output neurons in the output layer. The neurons present in the hidden layer as well as the output layer are the replicas of each other. The complexity of the feed forward network would be:

  $$\text{For each neuron at hidden layer (y)} = sigmoid\left[\sum_{i=1}^{n} x_i \cdot w_{ij} - \theta_j\right]$$

  $$\text{For each neuron in the output layer (output)} =$$

$$a \times \left[ \sum_{j=1}^{k} y_{jk} \cdot w_{jk} - \theta_k \right] + b$$

In the hidden layer there are 10 neurons, so $j \in [1,10]$, similarly i is the number of inputs which is a four dimensional vector, therefore $i \in [1,4]$, $\theta$ is a representation of bias at the hidden layer. So the complexity of the hidden layer will be $10 \times O(n^2)$, where n=4. The complexity of the output layer would be measured as: $4 \times O(k^2)$, where k is the number of output neurons, a and b are the coefficient of the line (predictor). So the total complexity of the feed forward network is: $10 \times O(n^2)$+ $4 \times O(k^2)$, which will be a constant as n and k are fixed.

b) **Complexity of Back Propagation:** We have used Levenberg-Marquardt [218] as a back propagation algorithm which helps in weight updating and training the network. Levenberg-Marquardt is a heuristic based optimization method widely known as a L-M optimization method. It is a mixture of Gauss-Newton and gradient descent algorithm. The complexity of the Levenberg-Marquardt algorithm based on the number of weights assigned in the network. If the numbers of weights are less it will converge faster otherwise it can take longer time. The main problem in convergence lies in finding the inverse of the Jacobian matrix of the error function, the inverse of the Hessian matrix which scale up to $N^3$, where N is the number of weights. Hence the complexity of the back propagation will be: $O(N^3)$.The total time complexity of Artificial Neural Network will be: $O(n^2)$+ $4 \times O(k^2)$+ $O(N^3)$. As $O(N^3)$ supersedes the behavior of rest two elements. The time complexity of ANN will be treated as $O(N^3)$. After analyzing the theoretical computation complexity of all the above techniques we can conclude that four point calibration will take less time in comparison to ANN and fundamental matrix approach. In order to confirm our theoretical analysis, we have conducted a practical test. In our pen test we have executed each algorithm and recorded the CPU time for each algorithm. Total 50 test cases have been generated for each algorithm. A plot between the number of test cases and time taken by each algorithm is presented in Figure 4.3.17. The average time has also been calculated based on the given 50 test cases shown in Figure 4.3.18. The average running time of four point calibration is 0.3441 sec which is very less in comparison to fundamental matrix of 0.3559 sec and ANN of 1.1256 sec.

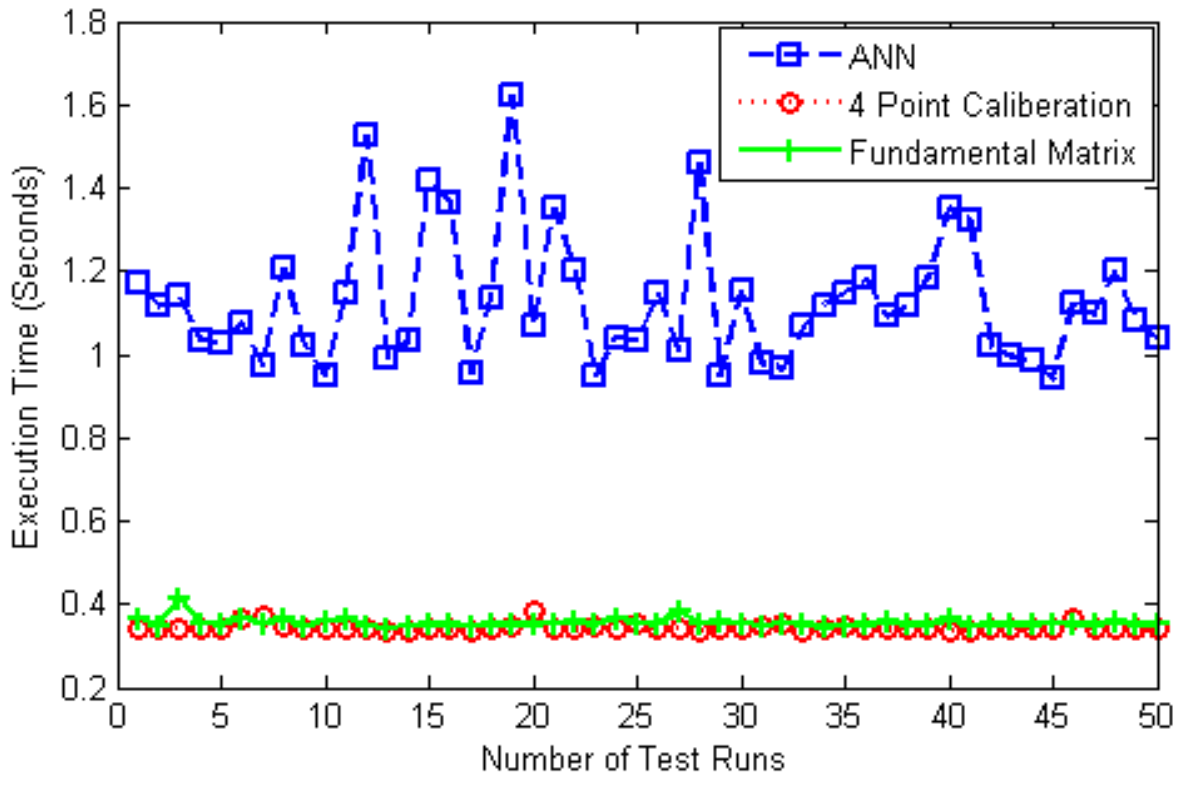

**Figure 4.3.18 Execution time of each test**

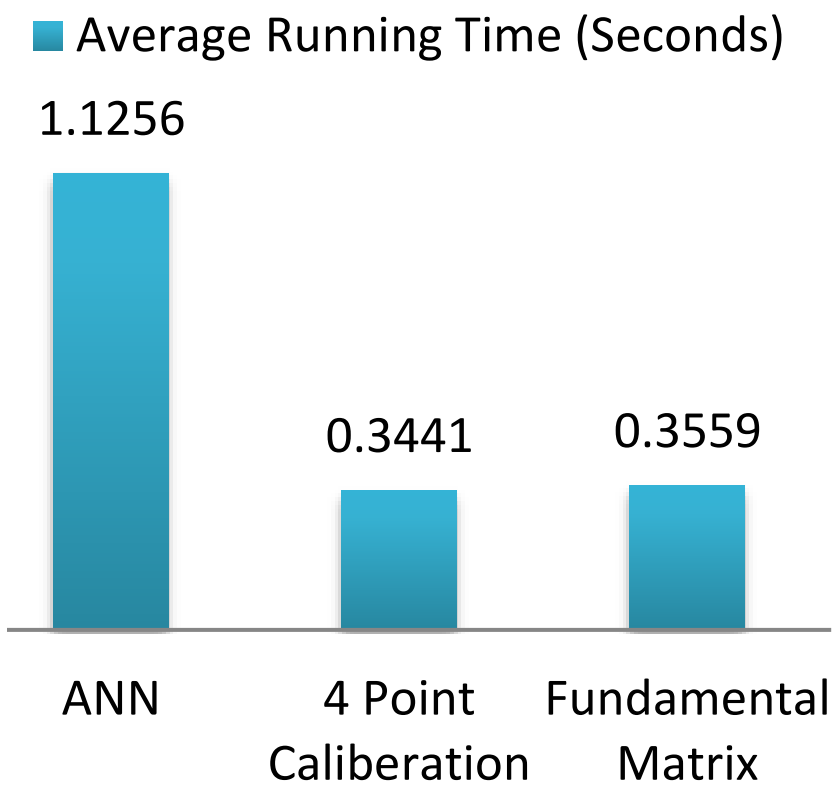

**Figure 4.3.17 Average running (different test runs)**

The test runs are executed on a 32 bit (Windows 7) machine having Intel Core I-5 processor with 2.50GHZ CPU speed. The machine has 8GB of Internal Memory (RAM).

## 4.4 Solution-II: Inverse Kinematic Solution

The calibration problem discussed in the above sections defines each point of the image plane with respect to the NAO body coordinate system. We have kept the orientation $(\alpha, \beta, \gamma)$ as fixed. The above calibration matrix helps to define end effector Cartesian position (X,Y,Z) for each image point (X,Y), while the inverse kinematic solution discussed in next section helps to estimate the joint values given the location (X,Y,Z) and orientation $(\alpha, \beta, \gamma)$. As NAO has five degrees of freedom in both of his hand there are five joint variables $\theta_1, \theta_2, \theta_3, \theta_4, \theta_5$) that we have to compute. Later these joint values plugged into the forward kinematic solution to reach to those positions lying on the table. In order to define the forward and inverse kinematic solution, the first estimate is to assign frame and define the Denavit-Hartenberg (DH) parameter for NAO right hand. No DH parameter and its frame assignment are discussed below.

### 4.4.1 Denavit Hartinberg (DH) Parameters

Jacques Denavit and Richard Hartenberg in 1955 have given the method which is widely known as DH principle to establish the relation of one link to another link by a 4×4 transformation [219].If there are different links connected in a sequential chain function and assume that any of the link is moving then it will be applying its effect on the rest of the joints connected with this. DH principles can help us to build a relation between these connected links in understanding their relative motion. We are using DH principles to define the end effector position with respect to the base. There are four parameters defined by Denavit and Hartenberg which can fully describe the content of the 4×4 transformation matrix. These four parameters are link length (a), twist angle ($\alpha$), link offset (d) and joint angle ($\theta$). The descriptions of all four parameters are given below.

- $a_i$ = the distance between $Z_i$ to $Z_{i+1}$ measured along $X_i$, it is called common normal.
- $\alpha_i$ = the angle from $Z_i$ to $Z_{i+1}$ measured about $X_i$ .
- $d_i$ = the distance between $x_{i-1}$ to $x_i$ measured along $z_i$.
- $\Theta_i$ = the angle from $x_{i-1}$ to $x_i$ measured about $Z_i$ .

The first step in DH principle is to attach the coordinate frame for each axis. There are two mandatory conditions which should be followed at the time of coordinate attachment.

1. The $Z_i$-axis is set to the direction of the joint axis (the rotation direction).
2. The $X_i$-axis is parallel to the common normal between $Z_i$and $Z_{i-1}$.

### 4.4.2 Estimation of NAO DH Parameters

NAO has five degrees of freedom in his right as well as left hand. The specification of right and left hand, both are same except their joint's angle of rotation [18]. For making it simple, we have only configured the DH parameter table for NAO right hand. In order to establish the DH parameter for NAO, it is necessary to define the different degree of freedom associated with each link. Figure 4.4.1 and Figure 4.4.2 briefly describe different joints and their rotation axis. Figure 4.4.1 is the sketch representation of Figure 4.4.2 for giving more insight to the problem. We have added link length and their axis of rotation additionally for better explanation of the NAO DH parameters. We have used two different colours yellow and white in Figure 4.4.2 and blue and orange in Figure 4.4.1 for symbolizing two different joints at the same place. The DH parameters estimated based on above specification are summarized in Table 4.4.1.

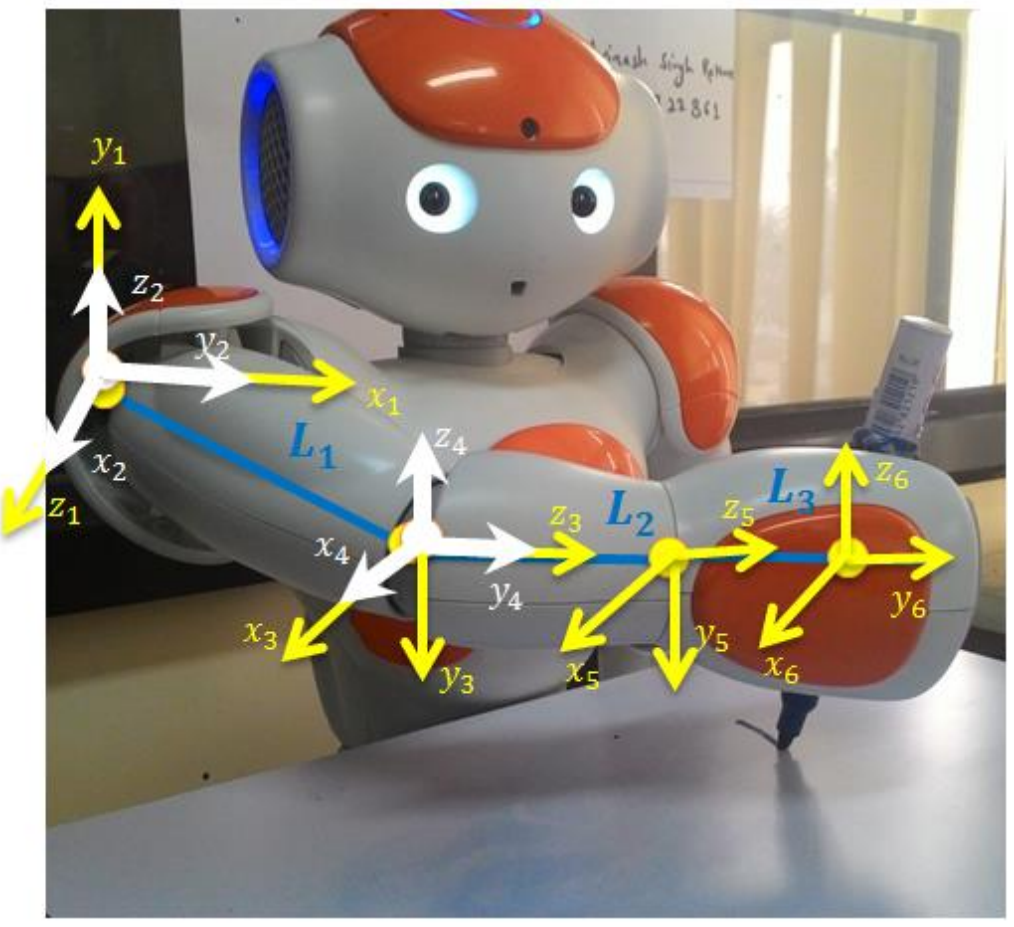

**Figure 4.4.2 Coordinate Frame attached on NAO's right hand**

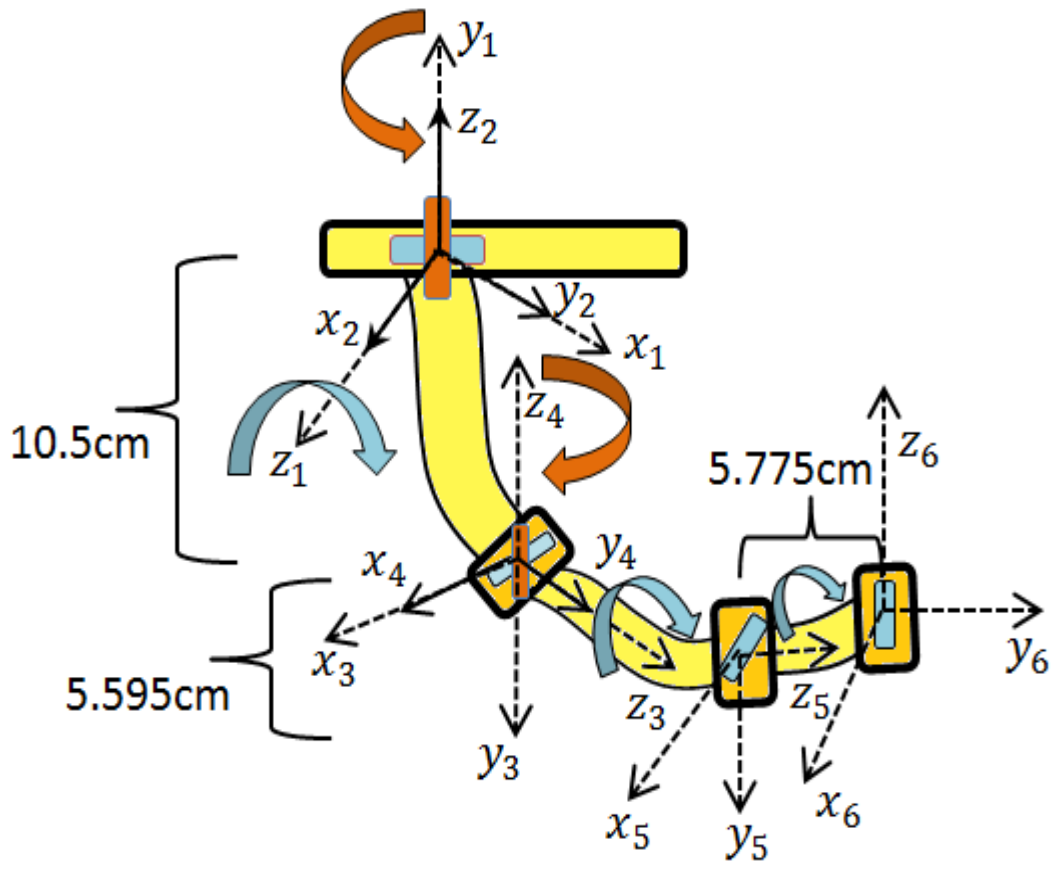

**Figure 4.4.1 Sketch representation of NAO's Right hand with axis of rotation and link specifications**

**Table 4.4.1 NAO's Right Hand DH parameters**

| Link $i$ | Twist angle ($\alpha_{i-1}$) | Common Normal ($a_{i-1}$) | Offset Length ($d_i$) | Joint angle ($\theta_i$) |
|---|---|---|---|---|
| 1 | 0 | 0 | 0 | $\theta_1$ |
| 2 | $-90^{\circ}$ | 0 | 0 | $\theta_2 - 90$ |
| 3 | $-90^{\circ}$ | 0 | 10.5cm | $\theta_3$ |
| 4 | $+90^{\circ}$ | 0 | 0 | $\theta_4$ |
| 5 | $-90^{\circ}$ | 0 | 5.595cm | $\theta_5$ |
| 6 | $+90^{\circ}$ | 0 | 5.775cm | 0 |

The twist angle is measured between $Z_i$ to $Z_{i+1}$, we have assumed $Z_0$ and $Z_1$ at the same location, therefore the twist angle for first link is set to 0. The other links have their twist angle as $\mp 90^\circ$ based on their orientation with respect to frames. The common normal is also set to zero for all links because the z axis of all consecutive links are orthogonal to each other. The link offset is measured along $x_{i-1}$ to $x_i$. Therefore, it is set to zero for all links that have a common origin. Joint angle is variable because each joint has its own rotation range. The transformation between links $i-1$ to i is defined as:

$$T_1^0 = \begin{bmatrix} cos\theta_1 & -sin\theta_1 & 0 & 0 \\ sin\theta_1 & cos\theta_1 & 0 & 0 \\ 0 & 0 & 1 & 0 \\ 0 & 0 & 0 & 1 \end{bmatrix}; T_2^1 = \begin{bmatrix} sin\theta_2 & -cos\theta_2 & 0 & 0 \\ 0 & 0 & 1 & 0 \\ -cos\theta_2 & -sin\theta_2 & 0 & 0 \\ 0 & 0 & 0 & 1 \end{bmatrix};$$

$$T_3^2 = \begin{bmatrix} cos\theta_3 & -sin\theta_3 & 0 & 0 \\ 0 & 0 & 1 & l_1 \\ -sin\theta_3 & -cos\theta_3 & 0 & 0 \\ 0 & 0 & 0 & 1 \end{bmatrix}$$

$$T_4^3 = \begin{bmatrix} cos\theta_4 & -sin\theta_4 & 0 & 0 \\ 0 & 0 & -1 & 0 \\ sin\theta_4 & cos\theta_4 & 0 & 0 \\ 0 & 0 & 0 & 1 \end{bmatrix}; T_5^4 = \begin{bmatrix} cos\theta_5 & -sin\theta_5 & 0 & 0 \\ 0 & 0 & 1 & l_2 \\ -sin\theta_5 & -cos\theta_5 & 0 & 0 \\ 0 & 0 & 0 & 1 \end{bmatrix};$$

$$T_6^5 = \begin{bmatrix} 1 & 0 & 0 & 0 \\ 0 & 0 & -1 & -l_3 \\ 0 & 1 & 0 & 0 \\ 0 & 0 & 0 & 1 \end{bmatrix}$$

There are six transformation matrixes in existence, which are capable of deriving the relation between the end effector position with respect to the first link of right hand. We can define a relation of 6 links to 1st link using the equation (4.37).

$$T_{endEffector}^{1st\ link} = T_1^0 \times T_2^1 \times T_3^2 \times T_4^3 \times T_5^4 \times T_6^5 \qquad (4.37)$$

These transformation matrixes are the concise representation of the four individual transformation matrixes described below in equation (4.38). These transformations are defined as:

$$ {}^{i-1}_{i}T = \text{(rotation angle } \alpha_{i-1} \text{ about } x_i) * \text{(translation distance } a_{i-1} \text{ along } x_i) * $$
$$ \text{(translation distance } d_i \text{ along } z_i) * \text{(rotation angle } \theta_i \text{about } z_i) \qquad (4.38)$$

For example the content of the

$T_5^4$ transformation matrix is dervied by using the equation below:

$$ {}^{i-1}_{i}T = \begin{bmatrix} 1 & 0 & 0 & 0 \\ 0 & cos\alpha_{i-1} & -sin\alpha_{i-1} & 0 \\ 0 & sin\alpha_{i-1} & cos\alpha_{i-1} & 0 \\ 0 & 0 & 0 & 1 \end{bmatrix} \times \begin{bmatrix} 1 & 0 & 0 & a_{i-1} \\ 0 & 1 & 0 & 0 \\ 0 & 0 & 1 & 0 \\ 0 & 0 & 0 & 1 \end{bmatrix} \times \begin{bmatrix} 1 & 0 & 0 & 0 \\ 0 & 1 & 0 & 0 \\ 0 & 0 & 1 & d_i \\ 0 & 0 & 0 & 1 \end{bmatrix} $$

$$ \times \begin{bmatrix} cos\theta_i & -sin\theta_i & 0 & 0 \\ sin\theta_i & cos\theta_i & 0 & 0 \\ 0 & 0 & 1 & 0 \\ 0 & 0 & 0 & 1 \end{bmatrix} $$

Further simplification can deduce it into

$$ {}^{i-1}_{i}T = \begin{bmatrix} 1 & 0 & 0 & a_{i-1} \\ 0 & cos\alpha_{i-1} & -sin\alpha_{i-1} & 0 \\ 0 & sin\alpha_{i-1} & cos\alpha_{i-1} & 0 \\ 0 & 0 & 0 & 1 \end{bmatrix} \times \begin{bmatrix} cos\theta_i & -sin\theta_i & 0 & 0 \\ sin\theta_i & cos\theta_i & 0 & 0 \\ 0 & 0 & 1 & d_i \\ 0 & 0 & 0 & 1 \end{bmatrix} $$

$$ {}^{i-1}_{i}T = \begin{bmatrix} cos\theta_i & -sin\theta_i & 0 & a_{i-1} \\ sin\theta_i cos\alpha_{i-1} & cos\theta_i cos\alpha_{i-1} & -sin\alpha_{i-1} & -d_i sin\alpha_{i-1} \\ sin\theta_i sin\alpha_{i-1} & cos\theta_i sin\alpha_{i-1} & cos\alpha_{i-1} & d_i cos\alpha_{i-1} \\ 0 & 0 & 0 & 1 \end{bmatrix} $$

After putting values from the DH table as $\alpha_{i-1}$=-90,$a_{i-1}$=0,$d_i$=$l_2$,$\theta_i$ =$\theta_5$the resultant matrix will be:

$$ {}^{4}_{5}T = \begin{bmatrix} cos\theta_5 & -sin\theta_5 & 0 & 0 \\ 0 & 0 & 1 & l_2 \\ -sin\theta_i & -cos\theta_i & 0 & 0 \\ 0 & 0 & 0 & 1 \end{bmatrix} $$

### 4.4.3 Inverse Kinematic Solution: A gradient descent based numerical method

To control the movement of the end effector, it is desired to plug the joint values into the forward kinematic equation described in previous section. But getting the joint values from the defined position and orientation is a nonlinear problem. The problem becomes more complex when the number of joints in the system increases. According to [220] two category of solution exists. The first category of solutions known as closed form solutions are useful when the number of joints are less, while the second category of solutions (numerical solutions) are used when closed form solutions are difficult to achieve. NAO has five degrees of freedom in both hands. Therefore, it is difficult to achieve a closed form solution for it. The closed form solution is also very difficult to achieve as the mechanical design of NAO's hand does not follow the Pieper's recommendation [219]. Pieper's solution for six link manipulators states that the last three joints should be attached at the same point (intersecting each other). By placing these joint together at the same place the common normal will become zero, this reduces some of the difficulty in solving the inverse kinematics problem. We have looked into different numerical methods to establish an inverse kinematic solution for NAO 5DOF right hand. Existing literatures [221-223] are the witness of three kinds of numerical solutions i.e (a) iterative procedure based inverse transformation, (b) combined solution (iterative + closed form) and (c) optimization methods (Newton's method, Genetic Algorithm etc). Iterative procedure feedback the error and update the joint angle values till the mean square error is less than the threshold whereas in combined approach iterative procedure is used to find some of the joint values while the closed form solution is used to determine the rest ones. Optimization methods require a close initial guess to the exact solution. As in our case the input trajectory is continuous and each via point is dependent on its predecessor value, iterative solution could be the best suitable option. We have used gradient descent method to calculate the inverse kinematic. The process of using the gradient descent is summarized in the flow chart presented in Figure 4.4.3. The forward kinematics, which define the position ($\mathrm{P} = [\mathrm{P_x}, \mathrm{P_y}, \mathrm{P_z}, \mathrm{W_x}, \mathrm{W_y}, \mathrm{W_z}]$) given the joint values ($\theta = [\theta_1, \theta_2, \theta_3, \theta_4, \theta_5]$) is expressed in (4.38).

Here P and θ represents the Position and joint Vector. FK is the forward kinematics symbol. If the time stamp (ΔT) between two via points is very less then there exists a linear relation which holds true for a small displacement in the Cartesian as well as joint space. The relationship is given by (4.39).

$$\Delta P = FK(\Delta\theta) \tag{4.39}$$

Also the relation holds true for the rate of change for Cartesian as well as for angular displacement shown in (4.40).

$$\frac{\delta(\Delta P)}{\delta(\Delta T)} = \frac{\delta(FK)}{\delta(\Delta T)} \times \frac{\delta(\Delta\theta)}{\delta(\Delta T)} \tag{4.40}$$

$\frac{\delta(\mathrm{FK})}{\delta(\Delta \mathrm{T})}$ is known as Jacobian (J), $\frac{\delta(\Delta \mathrm{P})}{\delta(\Delta \mathrm{T})}$ and $\frac{\delta(\Delta\theta)}{\delta(\Delta \mathrm{T})}$ are velocities $(\Delta \mathrm{P}, (\Delta\theta)$ in Cartesian as well as joint space. After replacing the values in (4.40) the modified expression is given by (4.41).

$$\Delta P = J(\Delta\theta) \tag{4.41}$$

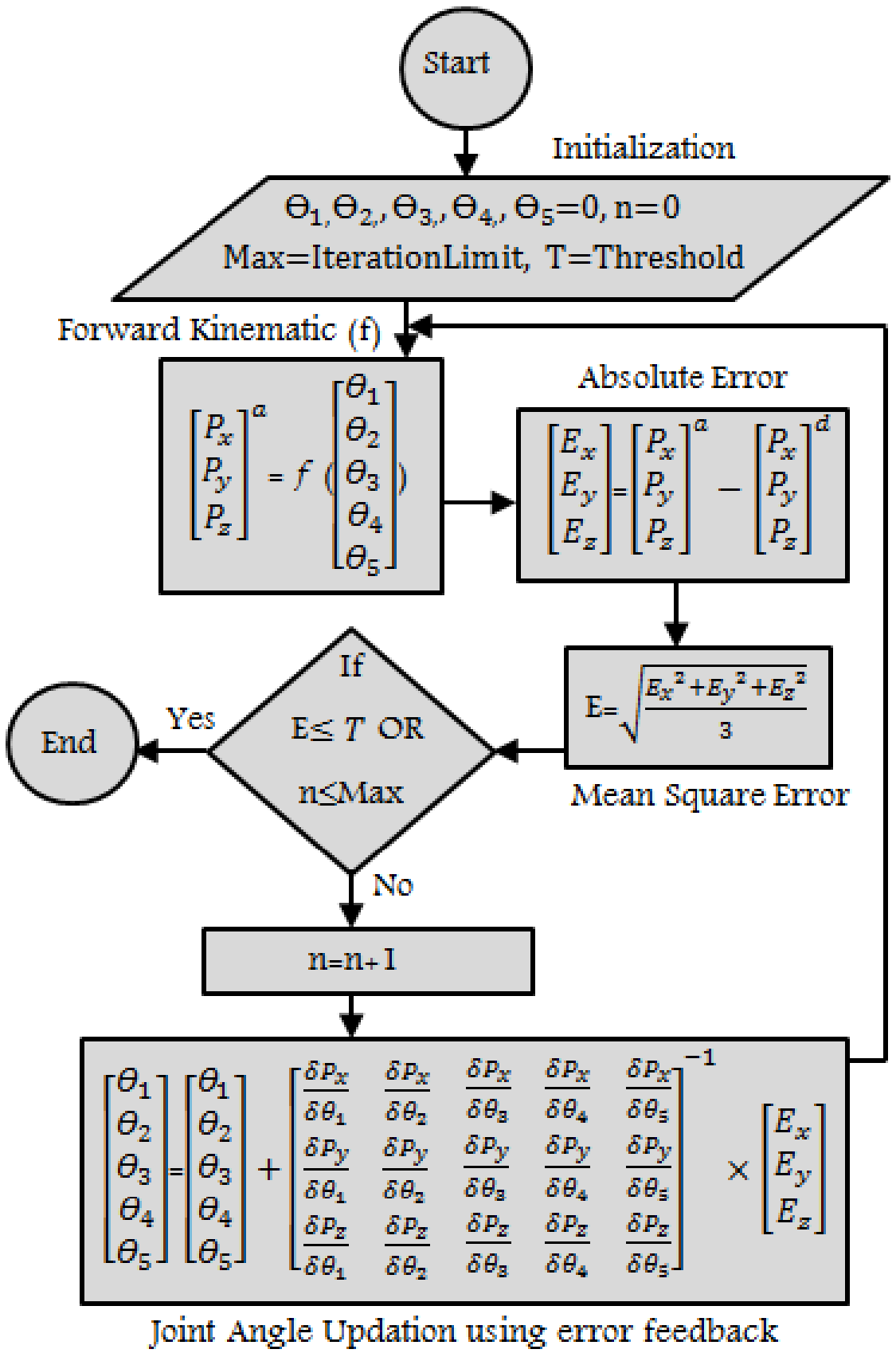

**Figure 4.4.3 Gradient Descent based Inverse Kinematic Solution**

From (4.41), joint values are calculated after doing the adjustment as given in (4.42).

$$\Delta\theta = J^{-1}(\Delta P) \qquad (4.42)$$

$J^{-1}$denotes the inverse of the Jacobian. The content of the Jacobian matrix is calculated as:

$$J = \begin{bmatrix} \frac{\delta(P_x)}{\delta(\theta_1)} & \frac{\delta(P_x)}{\delta(\theta_2)} & \frac{\delta(P_x)}{\delta(\theta_3)} & \frac{\delta(P_x)}{\delta(\theta_4)} & \frac{\delta(P_x)}{\delta(\theta_5)} \\ \frac{\delta(P_y)}{\delta(\theta_1)} & \frac{\delta(P_y)}{\delta(\theta_2)} & \frac{\delta(P_y)}{\delta(\theta_3)} & \frac{\delta(P_y)}{\delta(\theta_4)} & \frac{\delta(P_y)}{\delta(\theta_5)} \\ \frac{\delta(P_y)}{\delta(\theta_1)} & \frac{\delta(P_y)}{\delta(\theta_2)} & \frac{\delta(P_y)}{\delta(\theta_3)} & \frac{\delta(P_y)}{\delta(\theta_4)} & \frac{\delta(P_y)}{\delta(\theta_5)} \end{bmatrix} \qquad (4.43)$$

We cannot calculate the inverse of the Jacobian directly as it is not a perfect square matrix. Therefore, we need to calculate its pseudo inverse. Initialized all joint values are set to random values and in each iteration they are updated using (4.44).

$$\Delta\theta_{T+1} = \Delta\theta_T + J(\theta_T)^{-1}(\Delta P) \qquad (4.44)$$

Where $\Delta P$ will be calculated as:

$$\Delta P = P^{desired} - P^{actual} \text{ Where } P^{actual} = FK(\theta_T) \qquad (4.45)$$

$P^{desired}$ is the given position of the end effector where it has to reach. After solving the inverse kinematics for the first via point, the set of joint values will be used as the input for the next via points. We can use the previous joint values because the via point has a very little deviation from its just previous via point.

### 4.4.4 Result Analysis

There are two factors used by us to evaluate the result obtained from inverse kinematic solution and the drawing made by NAO humanoid robot. The inverse kinematic solution is represented in terms of their joint angle calculated for each given object, while the error involved for each drawn object shows the significance of the solution. In total we used 70 different objects drawn by NAO.

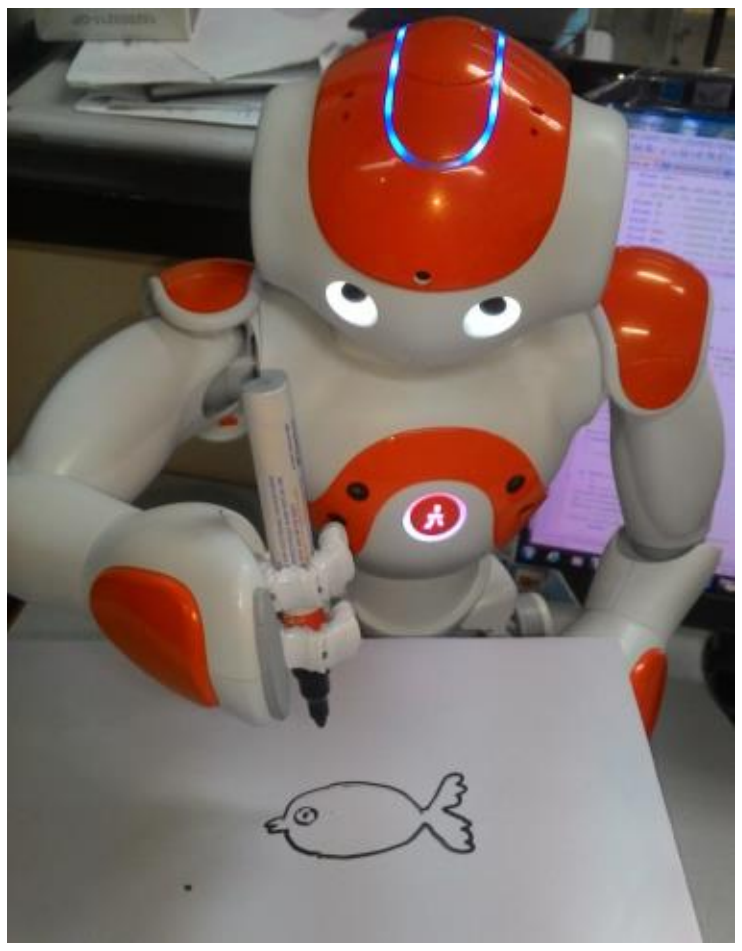

**Figure 4.4.4 Object Drawn by NAO**

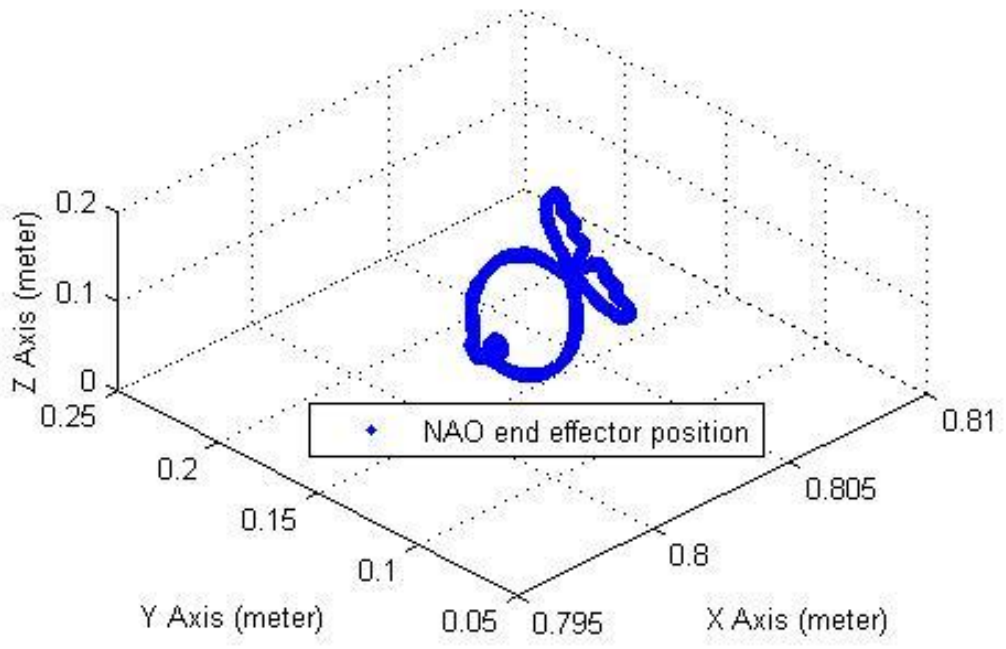

**Figure 4.4.5 NAO end effector position w.r.t Figure 4.4.4.**

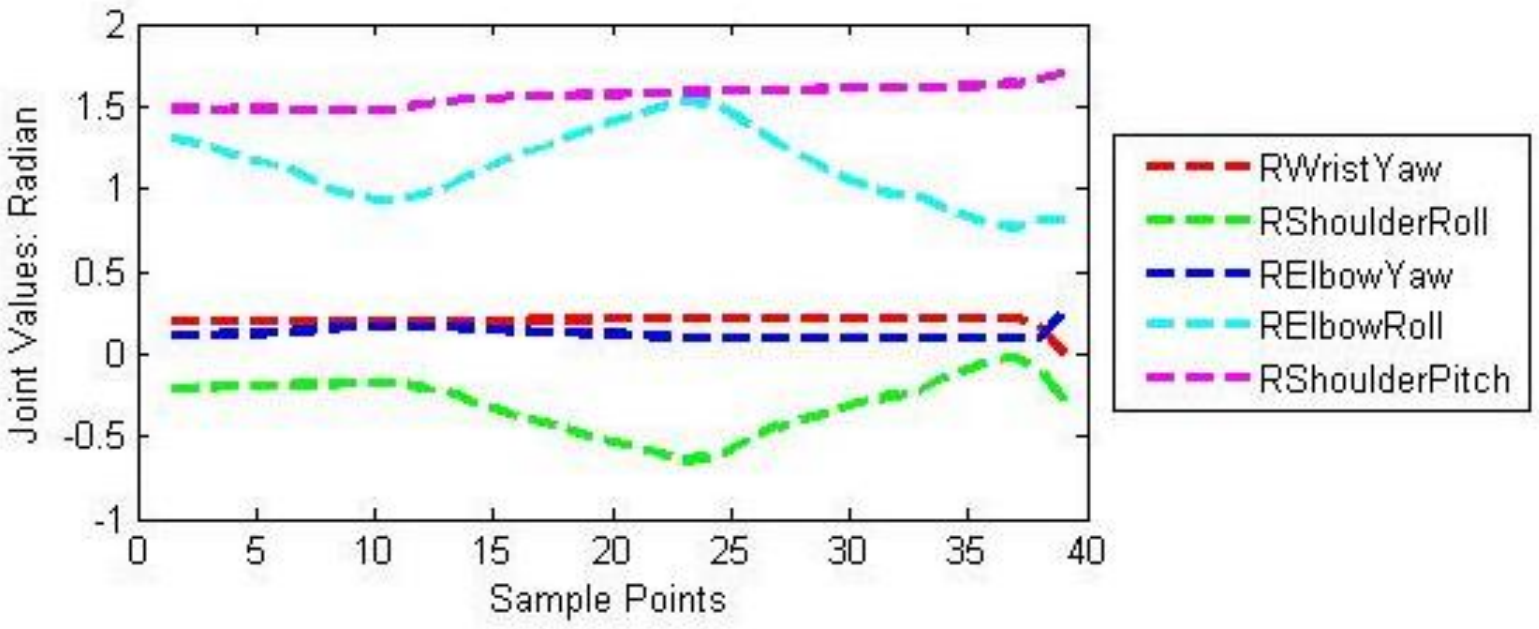

**Figure 4.4.6 NAO inverse kinematic solution w.r.t Figure 4.4.5.**

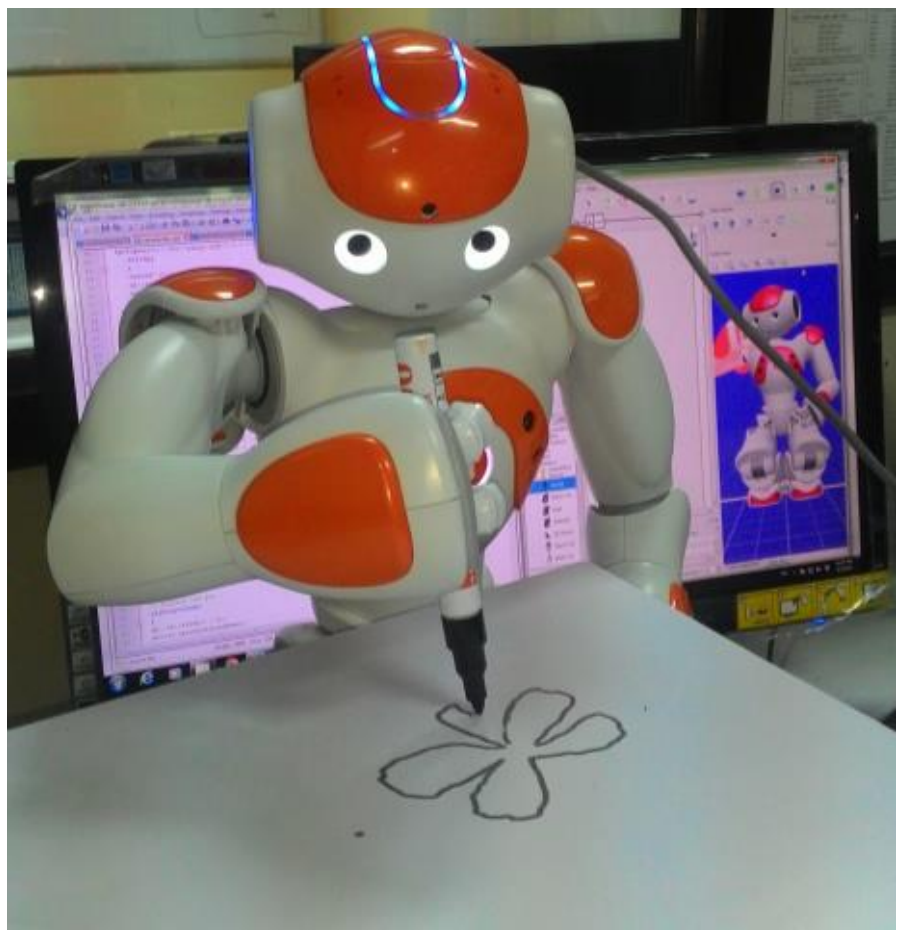

Figure 4.4.7 Object drawn by NAO

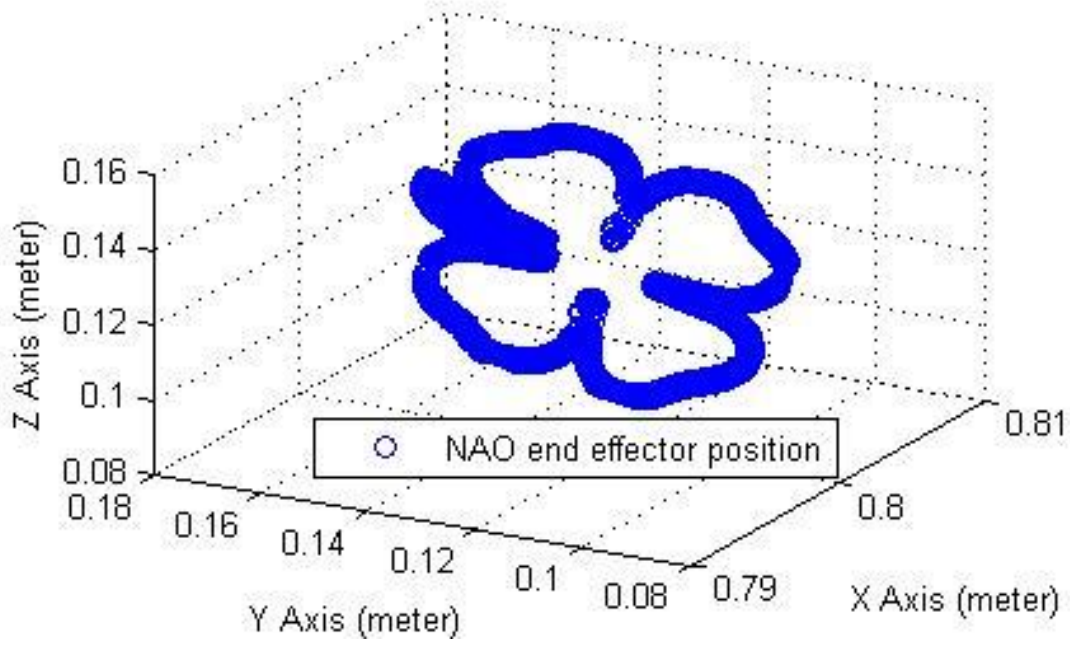

**Figure 4.4.8 NAO end effector position w.r.t Figure 4.4.7**

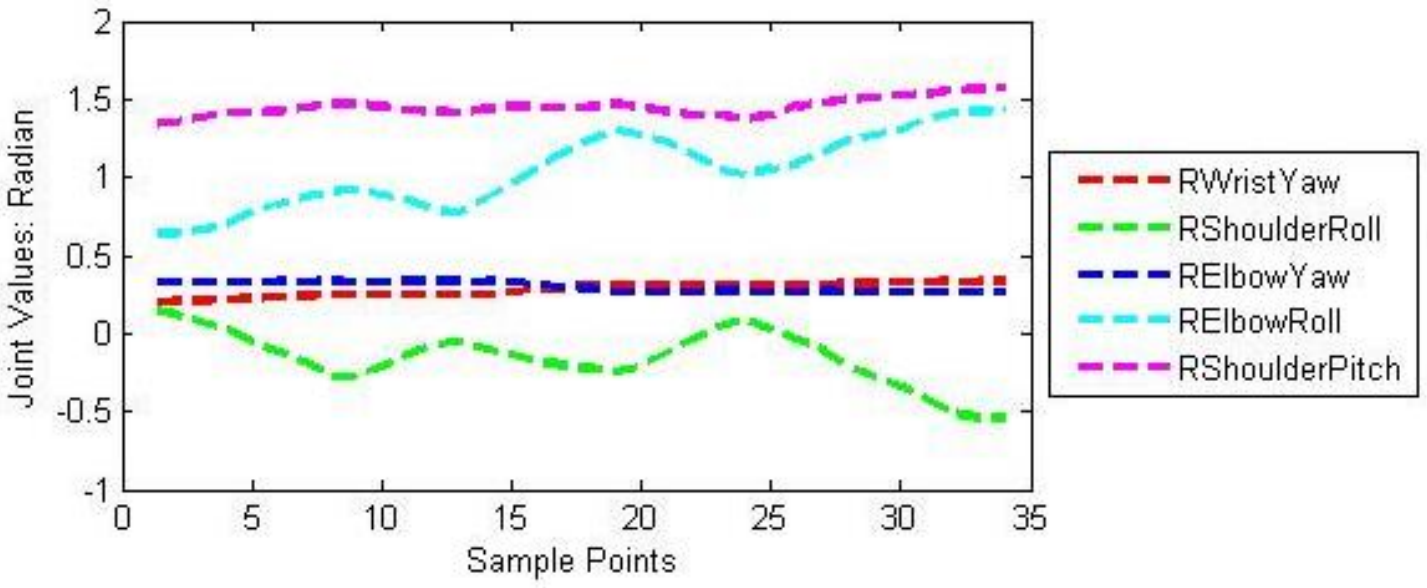

**Figure 4.4.9 Inverse Kinematic solution w.r.t Figure 4.4.8.**

There are 26 English alphabets (A-Z), 10 numeric characters (0-9), 15 random objects like fish, cat, dog, apple, etc. and 20 human faces. The inverse kinematic solution for each category is depicted in Figure 4.4.6 and Figure 4.4.7. Figure 4.4.4 and Figure 4.4.8 reflects the outcome of steps carried out to design and development of this experiment, while Figure 4.4.5 and Figure 4.4.9 reflects the Cartesian space geometry of the image points with respect to NAO body coordinate system. At last, Figure 4.4.6 and Figure 4.4.7 depicts the inverse kinematic solution obtained for drawn objects. From the above Figure 4.4.6 and Figure 4.4.7, it is analyzed that the out of five joints, two joints RWristYaw and RElbowYaw have little contribution in sketch drawing. The value of these two joints is almost constant for the entire sample points. The rest of the object drawn by NAO is depicted in Figure 4.4.10.

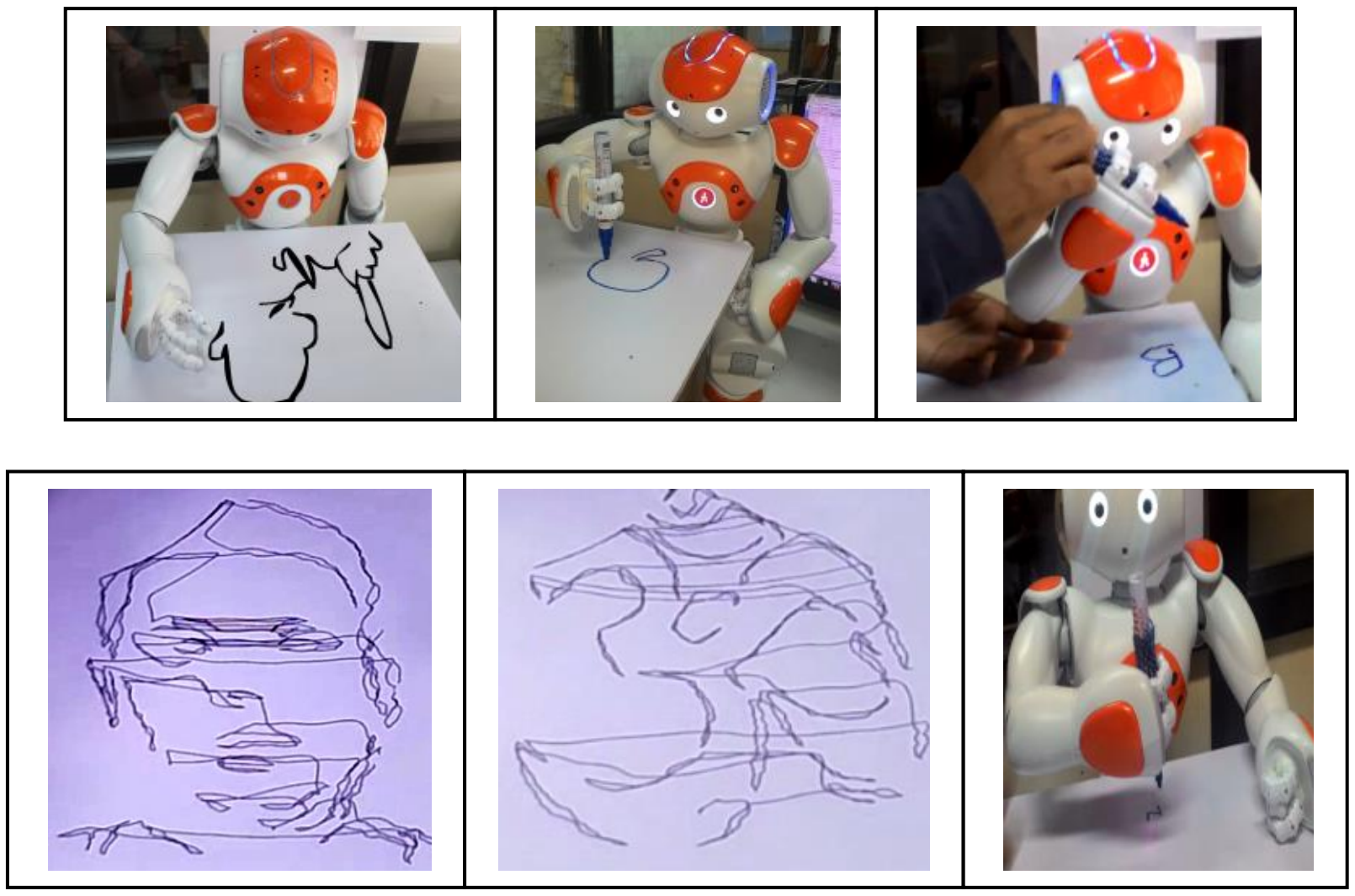

**Figure 4.4.10 Various Objects Drawn by NAO Humanoid Robot**

We have used mean square error to evaluate the performance of the NAO sketch drawing. The error is computed based on the difference between the given trajectory points in the Cartesian space and the actual points drawn by the NAO. The given points are computed for each single image based on the transformation discussed in section 4.3.1, while the actual points are calculated after applying the forward kinematics on the retrieved joint values. The forward kinematic of NAO is defined in section 4.3.2 in equation (4.19). The mean square error exists because there is a deviation between the given and actual (X,Y,Z) points of no end effector, involved for each point. The mean square computed for all 70 objects is depicted in Figure 4.4.11. The average mean square error is measured in meter and it is not more than 3cm. The mean square error depends on the number of points “n” involved in the drawing and the absolute error. The expression is given below.

$$MSE(DrawnObject) = \frac{\sum_{i=1}^{n}(p_i - p_i')^2}{n} \qquad (4.46)$$

Here “n” is the number of points for the particular objects. $p_i$is the given point and $p_i'$is the drawn point.

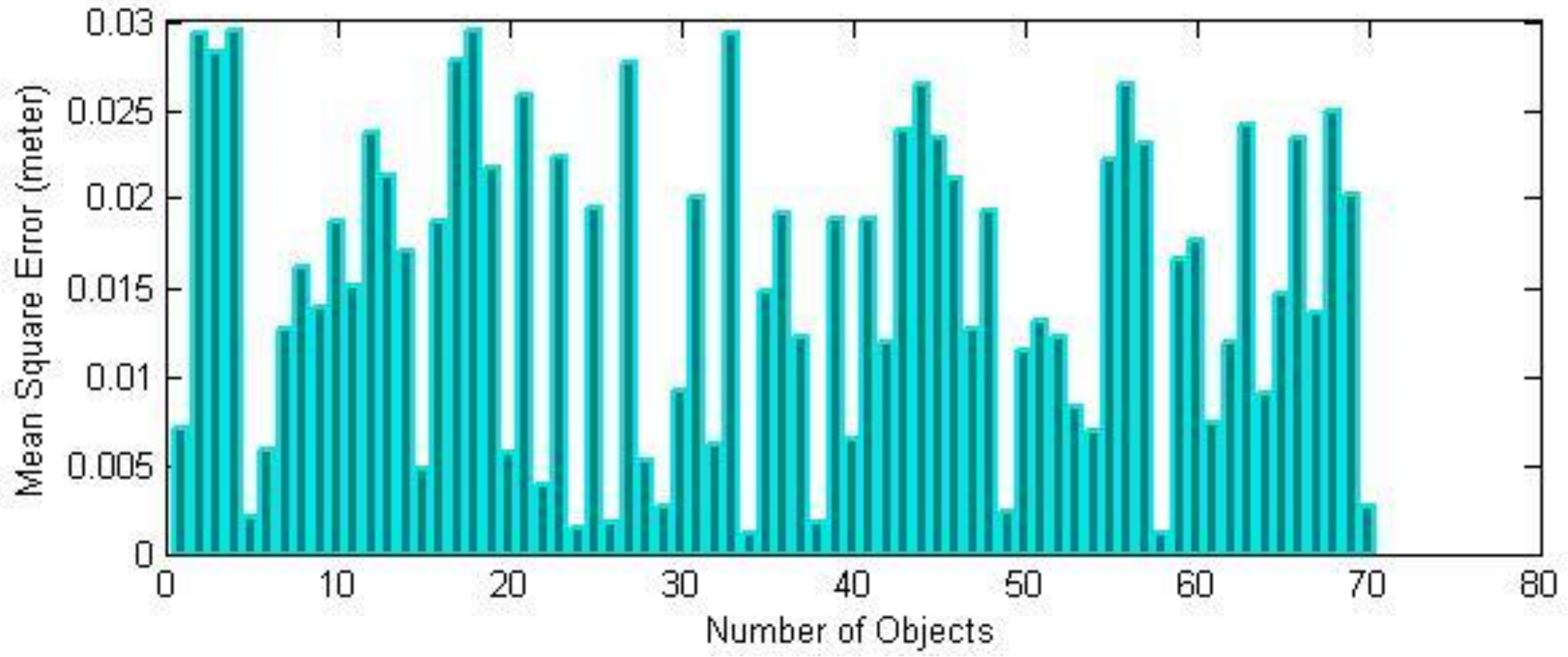

**Figure 4.4.11 Mean Square error (meter) for every drawn object**

## 4.5 Extension of Solution-II: A close form Inverse Kinematic Solution

It has been seen and analyzed from the previous section that active joints which are participating in sketch drawing are RShoulderPitch, RShoulderRoll, RElbowRoll. The rest two joint joints RElbowYaw and RWristYaw have less contribution in drawing; therefore we can keep these two joints as constant. After losing two degrees of freedom (one along the elbow and one along the wrist), from the right hand, we are left with only three degrees of freedom. Keeping the wrist as constant, the link length is also reduced to two links (shoulder and wrist combined together and represent one link). The modified coordinate system attached to the NAO right hand is represented in Figure 4.4.11 and its description is presented in Figure 4.4.12. Frames are attached keeping DH principle in mind, which results difference in DH table presented in Table 4.5.1.

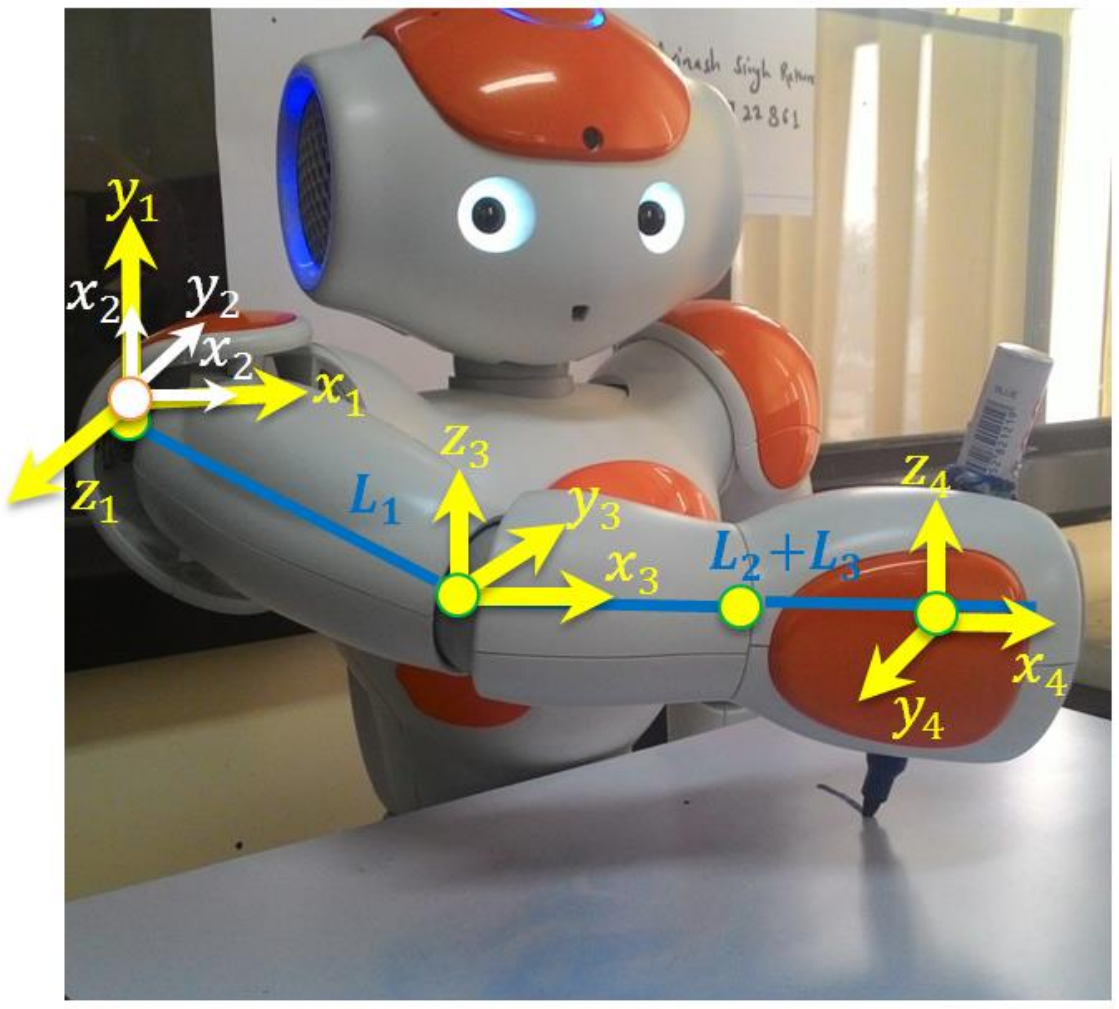

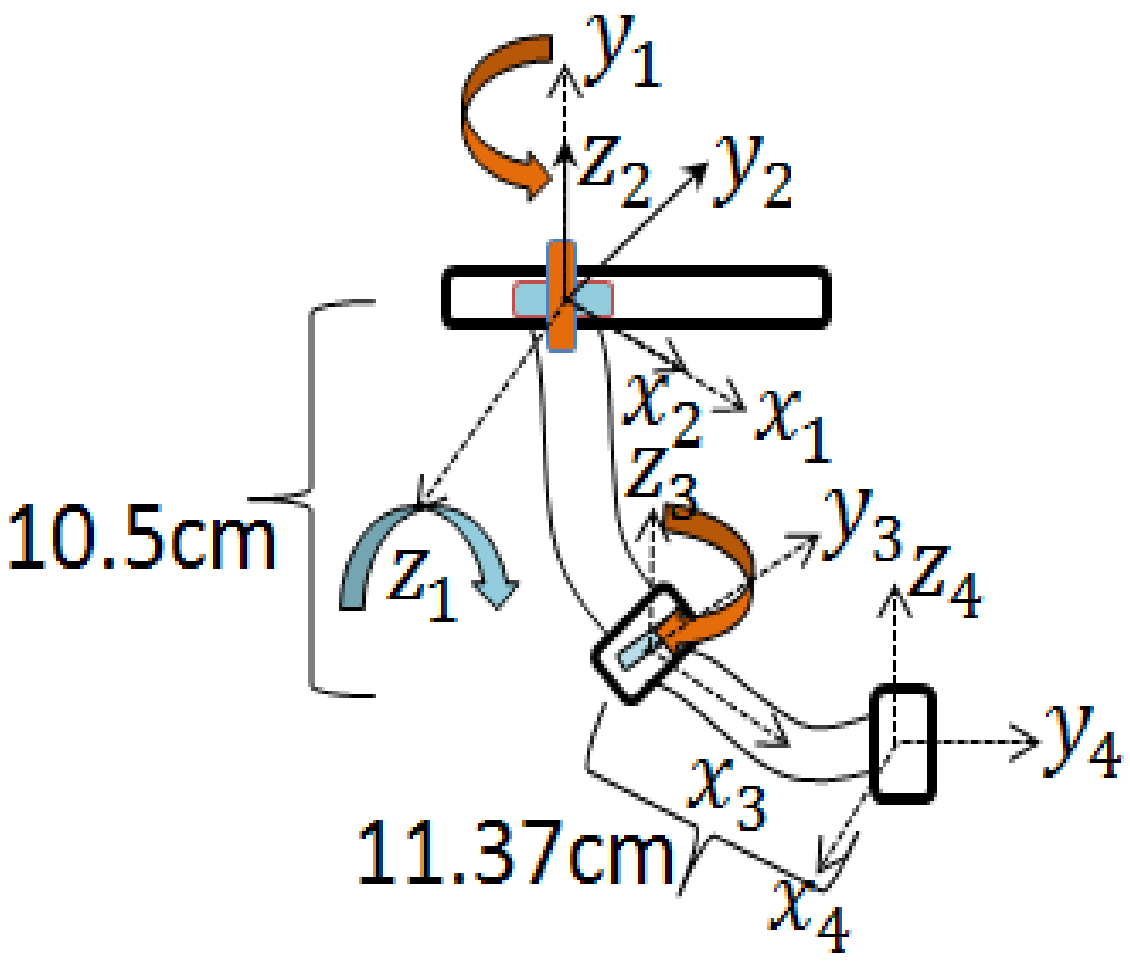

**Figure 4.5.1 Specification of modified coordinate system right hand with 2 links**

**Figure 4.5.2 Coordinate Frame attached on NAO's right hand with only 2 links**

The coordinate's frames attached in the Figure 4.5.1 and Figure 4.5.2 is different from Figure 4.5.1 and Figure 4.5.2 due to the change in the coordinate axis. The second joint (RShoulderRoll) in the modified coordinate system shown in Figure 4.5.1 is parallel to the next joint (RElbowRoll) which enables us to assign the X axis perpendicular to the successive z axis. Link length and their rotation axis are described in Figure 4.5.2. The coordinate system of the last link (fingers) and RElbowRoll has been kept same to reduce the complexity of the system. The transformation between joint "i-1" to "I" is reacted in following transformation matrices. These transformation matrices are different from the specification due to the removal of two degrees of freedom

**Table 4.5.1 NAO's Right Hand DH parameters for only 3 joints.**

| Joint $i$ | Twist angle $\alpha_{i-1}$ | Common Normal $a_{i-1}$ | Offset Length $d_i$ | Joint angle $\theta_i$ |
|---|---|---|---|---|
| 1 | 0 | 0 | 0 | $\theta_1$ |
| 2 | $-90^{\circ}$ | 0 | 0 | $\theta_2$ |
| 3 | 0 | 10.50cm | 0 | $\theta_3$ |
| 4 | 0 | 11.37cm | 0 | $-90^{\circ}$ |

The transformation between joint $i-1\ to\ i$ is reflected in following transformation matrixes. These transformation matrixes are different from the specification due to the removal of two degree of freedom form the system.

$$T_1^0=\begin{bmatrix} cos\theta_1 & -sin\theta_1 & 0 & 0 \\ sin\theta_1 & cos\theta_1 & 0 & 0 \\ 0 & 0 & 1 & 0 \\ 0 & 0 & 0 & 1 \end{bmatrix}; T_2^1=\begin{bmatrix} cos\theta_2 & -sin\theta_2 & 0 & 0 \\ 0 & 0 & 1 & 0 \\ -sin\theta_2 & -cos\theta_2 & 0 & 0 \\ 0 & 0 & 0 & 1 \end{bmatrix} T_3^2=$$

$$\begin{bmatrix} cos\theta_3 & -sin\theta_3 & 0 & 0 \\ sin\theta_3 & cos\theta_3 & 1 & l_1 \\ 0 & 0 & 0 & 0 \\ 0 & 0 & 0 & 1 \end{bmatrix}; T_4^3=\begin{bmatrix} 0 & 1 & 0 & l2 \\ -1 & 0 & 0 & 0 \\ 0 & 0 & 1 & 0 \\ 0 & 0 & 0 & 1 \end{bmatrix}$$

As the modified system have only four transformation matrices. The end effector position with respect to torso would be defined as:

$$T_{endEffector}^{torso} = T_0^{torso} \times T_1^0 \times T_2^1 \times T_3^2 \times T_4^3 \quad (4.47)$$

After losing two degrees of freedom and combining link2 and link3 into one link, the specification of the right hand has been modified. The updated specification is provided below in Table 4.5.2.

**Table 4.5.2 NAO's Right Hand Updated Specifications.**

| Link | DOF | Range (radian) | Length (cm) |
|---|---|---|---|
| Link 1 (shoulder) | RShoulderPitch | -2.0857 to 2.0857 | 10.50 |
| Link 1 (shoulder) | RShoulderYaw | -1.3265 to 0.3142 | 10.50 |
| Link 2 (Elbow) | RElbowRoll | 0.0349 to 1.5446 | 11.37 |

From the forward kinematic equation (4.26) of the modified coordinate system, we know the end effector position with respect to NAO torso position is

$$\mathrm{T}_{\mathrm{endEffector}}^{\mathrm{torso}} = \mathrm{T}_0^{\mathrm{torso}} \times \mathrm{T}_1^0 \times \mathrm{T}_2^1 \times \mathrm{T}_3^2 \times \mathrm{T}_4^3 \quad (4.48)$$

Putting all transformation together and solving for $\mathrm{T}_{\mathrm{endEffector}}^{\mathrm{torso}}$ the resultant Cartesian position of the end effector will be:

$$P_x=(l_1+l_2c_3)c_1c_2-l_2c_1s_2s_3 \quad (4.49)$$

$$P_y=(l_1+l_2c_3)s_1c_2-l_2s_1s_2s_3 \quad (4.50)$$

$$P_z=-(l_1+l_2c_3)s_2-l_2c_2s_3 \quad (4.51)$$

where $c_1 = \cos(\theta_1)$, $c_2 = \cos(\theta_2)$, $c_3 = \cos(\theta_3)$ and $s_1 = \sin(\theta_1)$, $s_2 = \sin(\theta_2)$, $s_3 = \sin(\theta_3)$ and $l_1, l_2$ represents link1 and link 2. Divide equation (4.51) by equation (4.50) we will get:$P_y/P_x=s_1/c_1$ then $\tan(\theta_1)= P_y/P_x$ and $\theta_1 = \text{atan}(P_y,P_x)$Squaring and summing equation (4.49), (4.50) and (4.51) we will get:$P_x^2 + P_y^2 + P_z^2 = l_1^2+l_2^2+2*l_1*l_2*c_3$Then $c_3=(P_x^2 + P_y^2 + P_z^2-l_1^2-l_2^2)/2*l_1*l_2$ and $s_3=\sqrt{1-c_3^2}$ and $\theta_3=\text{atan}(s_3, c_3)$. Let $k_1=l_1+l_2c_3 = \cos\alpha$ and $k_2=l_2s_3 = \sin\alpha$, then$\tan(\alpha) = \frac{k_2}{k_1}$ further we can conclude; $\alpha=\text{atan}(k_2,k_1)$. Putting the value of $k_1$&$k_2$ into equation (4.51), We will get; $-P_z = \cos\alpha * \sin\theta_2 + \cos\theta_2 * \sin\alpha$.

We also know that; sin(a+b)=sin(a)*cos(b)+cos(a)*sin(b)

Then; $-P_z=\sin(\theta_2 + \alpha)$ and $\cos(\theta_2 + \alpha)=\sqrt{1-P_z^2}$

Then; $\tan(\theta_2 + \alpha)=-P_z/\sqrt{1-P_z^2}$

Further; $\theta_2 + \alpha=\text{atan}(-P_z,\sqrt{1-P_z^2})$, $\theta_2= \text{atan}(-P_z,\sqrt{1-P_z^2}) -\alpha$

After putting the value of α, will be get;

$\theta_2 = \text{atan}(-P_z,\sqrt{1-P_z^2}) -\text{atan}(k_2,k_1)$

### 4.5.1 Result Analysis

The main motive of reducing the degrees of freedom from NAO right hand is to get the close form solution. The closed form solution discussed in the previous section is very fast which saves lot of computation time. The time taken by the system to calculate the inverse kinematic solution depends on the number of points and the method used for computation. As the numerical method, calculates the inverse kinematics using the iterative it consumes more time in comparison to close form solution. The time difference between the both the method is depicted on Figure 4.5.3. Here, GD represents the gradient descent and 3DOF represents 3 degrees of freedom closed form solution approach. The average time taken to compute the inverse kinematic for each and every point of the object for gradient descent is 7-15sec while the modified solution is taking only 3-9sec.

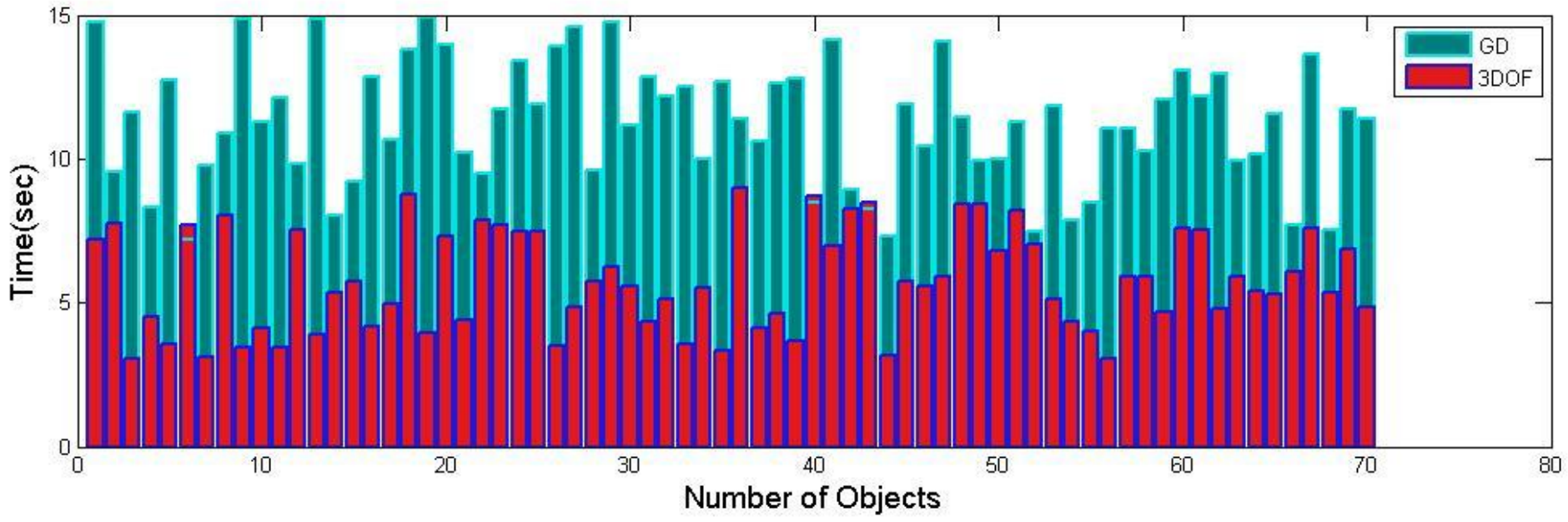

**Figure 4.5.3 Time Complexity: Gradient Descent Vs 3DOF closed form solution**

The inverse kinematic solution is also depicted in Figure 4.5.4. Here we kept the two joints RElbowYaw and RWristYaw as constant. The mean square error totally depends on the controller, hence we have almost the same error for both the approaches. In both the solution, we have used Proportional Integral Derivative (PID) controller for each joint which feedbacks the error computed between the set point and the measured point and control the trajectory. Therefore, we receive a smooth drawing trajectory.

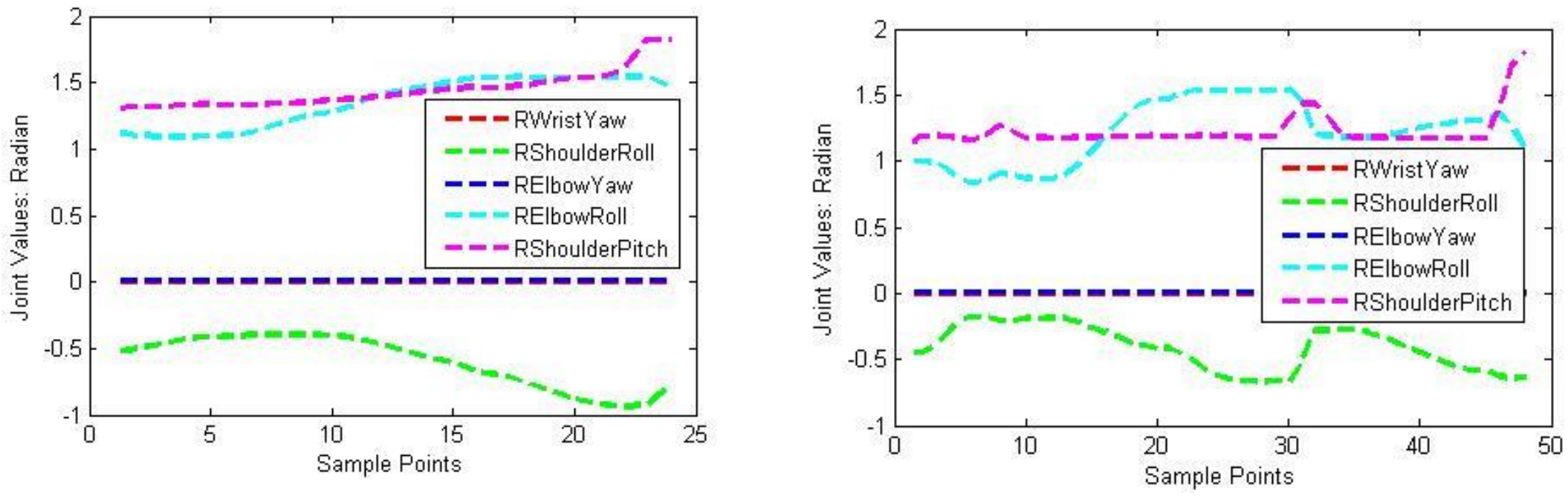

**Figure 4.5.4 Closed Inverse Kinematic solution for NAO right hand**

## 4.6 Conclusion

In this project, different calibration techniques have been used to establish a correlation between the image plane of NAO robot and its end effector position. These techniques help to measure the orientation of the NAO end effector and to solve the inverse kinematic problem. Three different approaches (a) Fundamental matrix (b) 4 point calibration and (c) ANN based regression analysis have been introduced to find the coefficient of calibration matrix. The experiment has been performed with 60 data points collected from the NAO end effector and from its image plane. The projection validity has been ensured using four statistical metrics a) error mean, b) standard deviation, c) variance and d) mean square error. The minimum mean square error and variance along each projection direction improves the strength of these calibration techniques. The regression analysis technique produces 0.0084cm, 0.0137cm, and 0.0040cm of mean square error (MSE) along X, Y and Z direction which is less than the MSE value (2.4984cm, 1.0664cm, 1.8771cm) of fundamental matrix approach and MSE value (0.4352cm, 0.2558cm, 0.0118cm) of 4 point calibration approach. Two other metrics such as time complexity and transformation reliability are also introduced to evaluate the performance of these techniques. We have obtained a computational complexity of O(1) for 4 point calibration which is minimum with respect to computational complexity of fundamental matrix (O(n3)) and regression analysis techniques (O(n3)). The average computational time for all the three techniques have been verified on 50 iterations where regression analysis takes 1.1256 sec, 4 point and fundamental matrix takes 0.3441sec and 0.3559sec to establish the coefficient of the calibration matrix. On the basis of all three performance metrics, it is concluded that regression analysis is a better way to estimate calibration matrix with less mean square error while 4 point calibration takes less time than others. The inverse kinematics problem is solved using the gradient descent method. This is a numerical method which requires comparatively larger time in finding the solution. Also Two degrees of freedom along the elbow and wrist have been omitted due to their less contribution in the drawing. Hence, they are sacrificed to obtain the closed form solution.

# Chapter 5:

# Face Biometric based Criminal Identification

*The preceding two chapters predict and re-synthesis the criminal face based on the given facial description while in this chapter, two novel face recognition techniques have been proposed to identify criminals given the face images. In the real time scenario the face photograph of the criminal has low resolution, taken from a wide distance and occluded, making the face recognition problem more challenging. The first recognition framework uses the facial bilateral symmetry to recognize criminals while the second model processes only a subset of the face to disclose the identity of the criminal.*

## 5.1 Introduction

The process of criminal identification has already been discussed in the first chapter where we have proposed two models to identify criminals based on the vague perception of eyewitness. In the second chapter, we gave the liberty to the constraint that even if the person is not enrolled in the database or committing the crime at the first time, we could also create the sketch of that person using our proposed portrait drawing module. In the previous two chapters, we have used the perception of eyewitness to either match with the existing criminal records or to recreate the face. In this chapter, we assume that the face of the criminal is accessible to us. Given the face image of the criminal, we have to identify the identity of the criminal. Although, this problem looks like a straightforward problem of face recognition, but the addition of the domain criminal identification adds some new challenges to it. These challenges are like, the quality of the face image is bad resolution, taken from the far distance, face is not completely visible, only the partial information of the face is accessible, etc. Some of these challenges are listed in Figure 5.1.1.

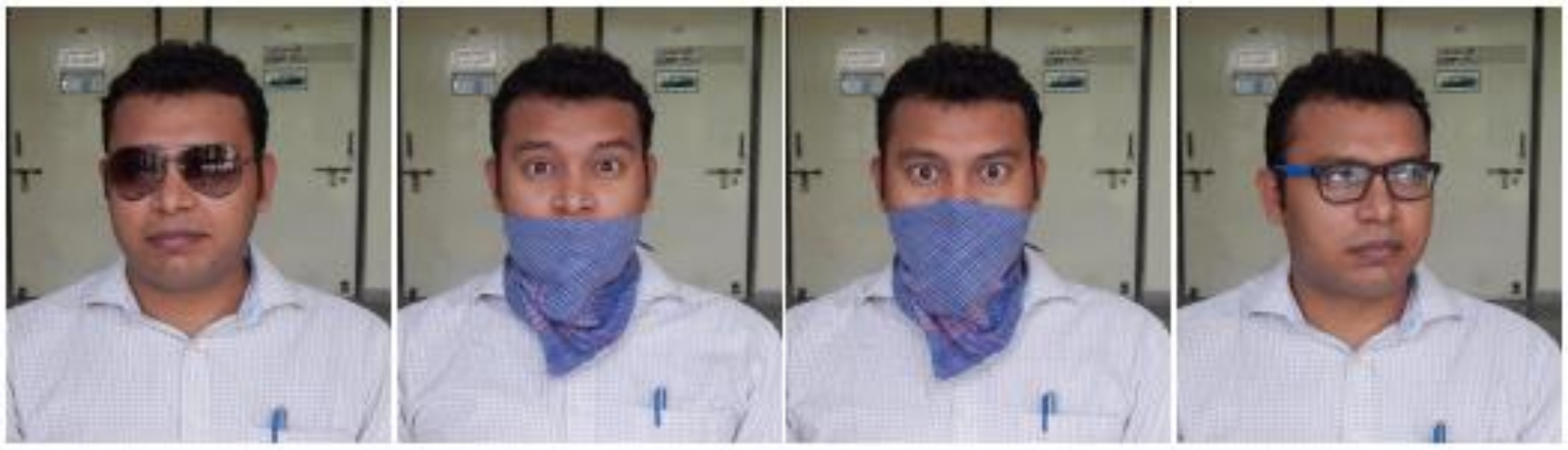

**Figure 5.1.1 Criminal Identification Challenges**

Here in this chapter, we have confined our domain only to the occluded face recognition. The problem statement of this chapter is to identify the criminal, given the partial information of the face. In order to solve this problem, we have proposed two methods in this chapter. The first model described in section 5.2 uses human face facial symmetry to recognize a person. Human face has bilateral symmetry, so even if the first half information is not visible or not in good quality, we can use the other half to recognize the person. The experimental proof on 40 persons ORL face database [224] shows the reliability and significance of the proposed model. An experiment is conducted on full faces as well as half faces confirms that half faces are having the same level of accuracy as of the full faces. On the other hand their computation complexity is just the half time of the full faces. The strength of this approach lies in the fact that the recognition accuracy is not sacrificed while the computation complexity is minimized.

The limitation of facial symmetry based face recognition is, it uses vertical bisection of the face (left half or right half). Therefore, the approach would not be effective when the face is occluded by the goggles/spectacles or by scarf. For handling this occlusion problem we have further proposed component based face recognition. In this model, we have divided the face into its most discriminating constituents like eyes, nose, mouth, forehead, cheek and chin. Each component is processed individually and in parallel to recognize a person. It is quoted in previous literature that, in the face all components do not have equal weightage. Based on their individual discriminative power, some of the components have more weightage while other have less. Considering these facts, we have first established a proof on 360 male and female faces which ensures that in face the most important features lies under eye, nose and mouth component. We have used density estimation model to evaluate the feature density within these regions. The result confirms that in face majority of facial features belong to only eye, nose and mouth regions. Based on the feature density a weightage factor is also assigned to these regions which helps in classification. We have further extended this proof to evaluate the performance with respect to full face approach. ORL benchmark face database [224] is used for this purpose; the results are significant and ensures the same level of accuracy to full faces. A person can be recognized only on the basis of these features or the minimum combination of these features. Each component has its own recognition accuracy, and sometimes a person can be identified only using the eyes, and sometimes only with the combination of nose and mouth or eyes and nose.

The rest of this chapter is organized as follows: section 5.2 discusses the facial symmetry based face recognition, which addresses the questions why the half face is sufficient to recognize a person, why the time complexity of the algorithm is just reduced and reach up to half of the full faces. In section 5.3, the component based face recognition approach is discussed with its proposed framework and experiment results. At the end, we conclude this chapter in section 5.4 with its contribution towards the occlusion based person identification/criminal identification.

## 5.2 Facial Symmetry based Face Recognition

The facial symmetry approach is being used as long back as in 1990s in digital imaging by David O'Mara and Robyn Owens [225]. They proposed a concept of bilateral symmetry in digital images. With the help of a dominant hyper plane they divided the objects in the image into two parts, so that each one looked like a mirror image of the other. Later, this approach of facial symmetry was also used by Xin Chen et.al [226]. They designed a fully automatic tool based on the gray level difference histogram to define the symmetry in faces, but the disadvantage of their algorithm is that it cannot define the symmetry when faces are not perfectly aligned in front of the camera. Human faces are symmetrical or not is still a topic of debate, because the answer varies from person to person. Due to some environmental issues or due to some diseases like craniofacial deformity, some people do not have perfect bilateral symmetry. Some Children are born with these deformities, while others acquire it by trauma or other diseases. The same question in this regard has been well addressed in [227]. But the successful implementation of facial symmetry in 3D faces to handle pose variations by Georgios Passalis et al. [228] motivated us to proceed in that direction. Therefore, we first define the bilateral symmetry on ORL database [224], and on the basis of symmetry of the full faces, we have extracted the half faces. Later, these half faces are used for the training and testing purpose.

### 5.2.1 Generating Symmetrical Half Faces

We have followed a very simple mathematical model for finding the symmetry. We have assumed face as a rectangle/square depending on the dimension. In our experiment we have taken them as rectangles. If we vertically bisect the rectangle, it will be divided into two identical parts. Though face is not a rectangle, but if the face is symmetrical and perfectly aligned, we can apply this concept. The approach to finding the symmetry is given below.

**Algorithm 1:** Generating Symmetrical Faces

1 Detect Front Face in the given input Image using HAAR Classifier (Frontal Face Detection)

2 Extract the region of Interest (ROI) i.e. Face and make them in fixed size

3 Divide the face vertically from the middle

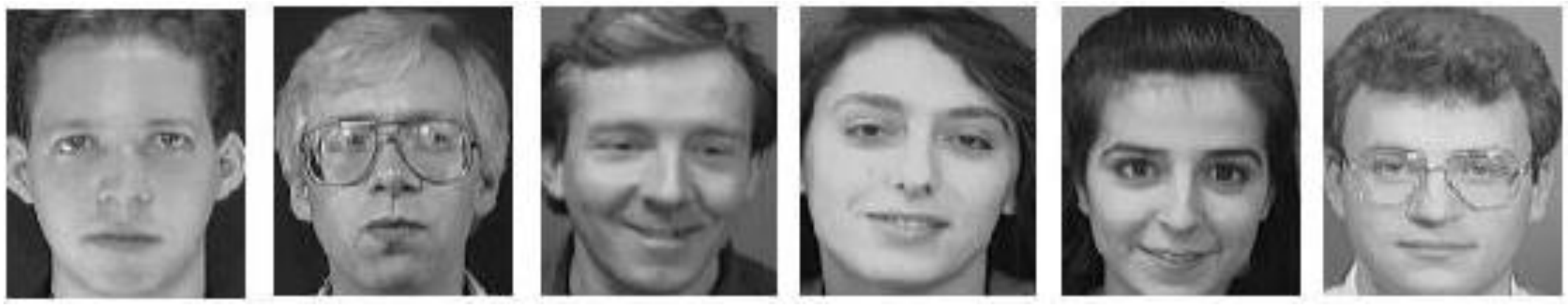

**Figure 5.2.1 Samples of Full Faces from the ORL Database**

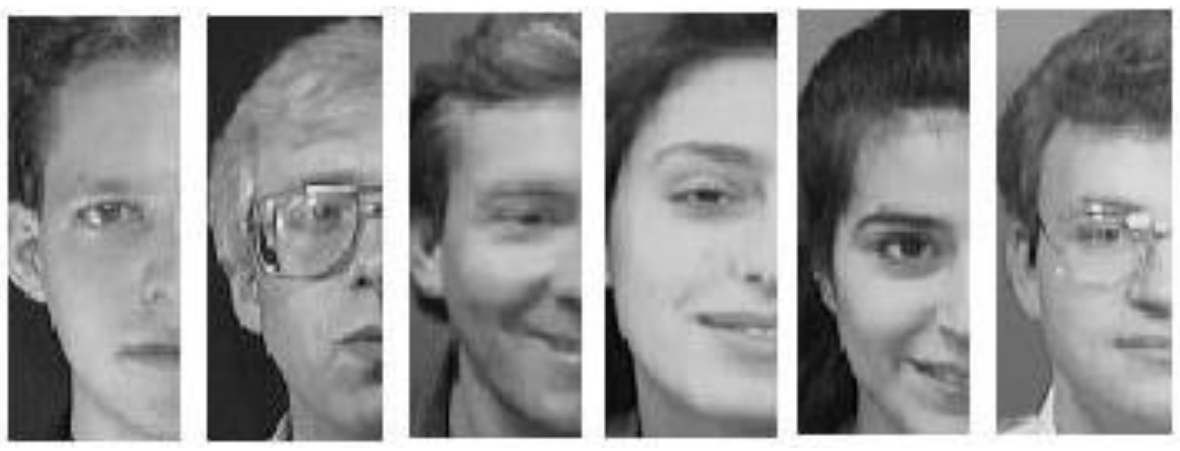

**Figure 5.2.2 Samples of Half Faces Corresponding to**

### 5.2.2 Experimental Setup

Our hypothesis is based on the assumption that the human face has the property of vertical symmetry. This symmetry can be defined over the system S where S'=T (S), Here T is a transformation applied to S. To evaluate the symmetrical property of an object in an image Dava O'Mara [225] has given a very significant equation (5.1). In the given equation, S values lie between 0 and 1. If the value of S is zero, then the object is having a perfect symmetry. As the value of S increases the symmetry ratio decreases.

$$S = 1 - \frac{\sum_{\vec{p} \in \text{Object}} f(\vec{p})}{\sum_{\vec{p} \in \text{Object}} \text{maxdiff}} \qquad (5.1)$$

$$\text{Where } f(\vec{p}) = \begin{cases} \text{abs}(I(\vec{p}) - I(\vec{p'}) & \text{if } (\vec{p'}) \in \text{Object} \\ \quad maxdiff & if\ (\vec{p'}) \notin Object \end{cases}$$

In the above equation $\vec{p}$ represents any point in the image and$\vec{p}$' is the reflection through the candidate hyper plane of bilateral symmetry. *I (n)* represent the intensity value of point n, and maxdiff is the difference between the maximum intensity values. From the above example, it has been proved that Symmetry is defined along a reference point. Here we have considered a reference point is the mid distance between two eyes. The Actual face images and the bisected face images (cropped on with respect to symmetric axis) are depicted in Figure 5.2.1 and Figure 5.2.2 respectively. Hence we have vertically bisected the face along the reference point. Both the two faces are almost identical in nature. The database we have taken here is the ORL database available at Cambridge University Computer Laboratory [224]. The database consists of 400 grayscale images of 40 persons. Each person has 10 images. Training and testing is performed in the ratio of 70% and 30%, hence total 280 images are used for training and 120 images is used for the testing purpose. For the validation of our hypothesis, we have divided the overall experiment into two parts.

### 5.2.3 Do Half Faces are Sufficient for Recognition

The human face is having bilateral symmetry. This redundant information is minimized during the time of feature extraction. Our hypothesis is, if human face have equal weightage for both left and right half, will it be useful to recognize a person only by using half of the face. In order to establish the proof, we trained the system with the half faces of each person and did recognition of them. The results produced by the system were very good and we got almost similar recognition accuracy as we experienced on full face, while time factor (time taken by system to recognize these identities) is reduced by half as it was on full faces. We have used ORL database for experiments. A theoretical proof has been established for why we will get almost similar accuracy on full and half faces? The basic idea in face recognition is to first reduce the redundancy that we have in human face.

Let a human face be represented by M features like $X = \{x_1, x_2, x_3, x_4, \ldots\ldots, M\}$

Let's assume that face is having bilateral symmetry.

Then we will have two sets

Let $X_1 = \{x_1, x_2, x_3, x_4, \ldots\ldots, x_k\}$ and $X_2 = \left\{x_{(k+1)}, x_{(k+2)}, x_{(k+3)}, x_{(k+4)}, \ldots\ldots, x_M\right\}$

$= X = X_1 \cup X_2$ And $X_1 \cap X_2 = \phi$

From above assumption we can say that $X_1 \sim X_2$ We applied PCA [87] [89] [99] on both X and X1 to find out the principal components, the idea of implementing PCA is to find out those k (best) features which can represent the whole face data. In other words, we can say that we are going to find the best direction where variance is maximized. The variance will be maximized when features are independent in nature. In face data we have $\{x_1, x_2, x_3, x_4, \ldots\ldots, x_m\}$Features those are independent and $\{x_1, x_2, x_3, x_4, \ldots\ldots, x_n\}$ features those can be expressed in terms of a linear combination of other features such that $m \cup n = M$. We can minimize this linear dependency and end up with only independent features (best features). Therefore, we can draw a conclusion here that k feature in half faces and m features in full faces are almost similar k$k \sim m$. Hence it proves that if we do the classification based on PCA we will get almost similar results.

Once we know that half and full faces both will give the similar level of accuracy. The theoretical analysis of the computation time involves in both full face and half face is summarized in the last of this section. In most of the steps carried out in PCA, the half faces require just a half computation in comparison to full faces. In our pen test on AT & T bell lab dataset of 40 people where each person has 10 images. We kept 7 images for training and 3 images for testing. The time taken for full faces is 10.2 Sec while in case of half faces it is 5.3 Sec, which is almost half of the full faces. For the verification of our hypothesis, we have applied PCA to both, the left half faces and the right half faces. Recognition is performed by measuring the Euclidian distance between the gallery image and the probe image. We have performed the training and testing in the groups of 10, 20, 30 and 40 people, the results of which have been summarized in Table 5.2.1 and Table 5.2.2.

**Table 5.2.1 Experiment Performed on Left Half Face**

| No. of Persons | Training Population | Testing Population | Accuracy | Time (Sec) |
|---|---|---|---|---|
| 10 | 70 | 30 | 96% | 1.3 sec |
| 20 | 140 | 60 | 93% | 2.6 sec |
| 30 | 210 | 90 | 94% | 4.0 sec |
| 40 | 280 | 120 | 93% | 5.3 sec |

**Table 5.2.2 Experiment Performed on Right Half Face**

| No. of Persons | Training Population | Testing Population | Accuracy | Time (Sec) |
|---|---|---|---|---|
| 10 | 70 | 30 | 93% | 1.3 |
| 20 | 140 | 60 | 91% | 2.6 |
| 30 | 210 | 90 | 94% | 3.9 |
| 40 | 280 | 120 | 91% | 5.2 |

These tables also show the statistics of the experiment. Accuracy and the recognition rate are the two parameters of our experiment. By observing the result in Table 5.2.1 and Table 5.2.2, we conclude that in terms of recognition rate, they are same. But when measured in terms of accuracy, they have slight difference in the result which is due to the alignment problem. Figure 5.2.3 shows the comparison of the symmetry of the left face and the right face. From the figure below it is proved that left and right faces of a person are almost identical. This hypothesis also confirm that if human faces are symmetric in nature, then for the recognition purpose, we may use half faces instead of full faces. We proceed with this idea and in the next section, verify it with our experimental results.

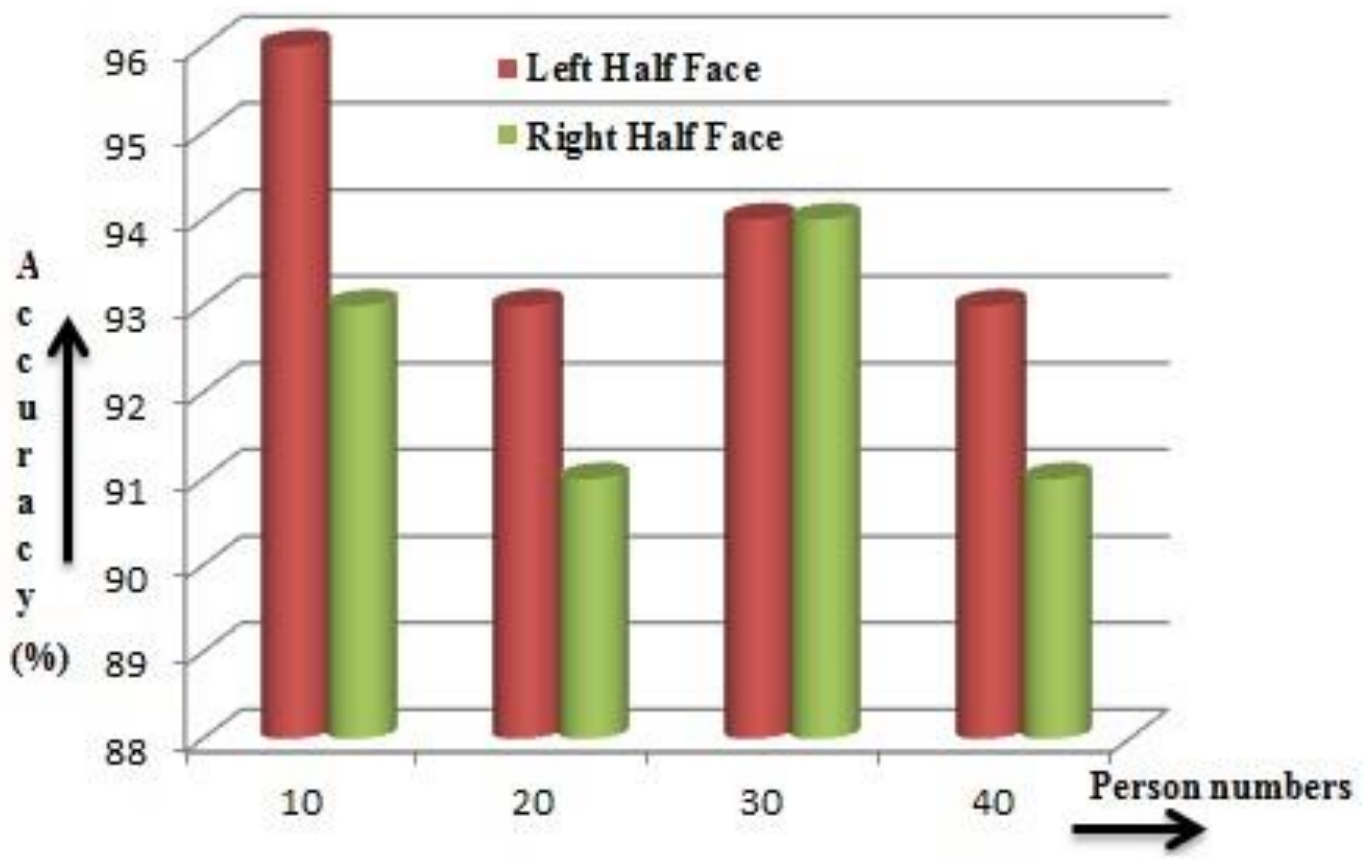

**Figure 5.2.3 Accuracy Graph of both the Left and Right Half faces based on PCA evaluation.**

| **PCA on Full Faces** | **PCA on Half Faces** |
|---|---|
| **1.** ***Training Set Generation:***<br>Let we have N sample each of size $\mathrm{I}'_{1*mn}$<br>Then face database will have $\mathrm{T}_{N*mn}$<br>$(\mathbf{StepsRequiredN} * \mathbf{mn})$<br><br>**2.** ***Calculation of Mean:***<br>$(\mu_z)_{mn*1} = \frac{\sum_{z=1}^{mn}\sum_{i=1}^{N}T(z,i)}{N}$<br>$(\mathbf{StepsRequiredN} * \mathbf{mn})$<br><br>**3.** ***Variance Calculation***<br>$(\sigma_z)_{mn*N} = \frac{\sum_{z=1}^{mn}\sum_{i=1}^{N}\left(T(z,i)-\mu_z\right)*\left(T(z,i)-\right.}{N}$<br>**(Steps required N * mn)**<br><br>**4.** ***Co-Variance Calculation*** | **1.** ***Training Set Generation***<br>Let we have N sample each of size $\mathrm{I}'_{1*mn/2}$<br>Then face database will have $\mathrm{T}_{N*mn/2}$<br>$(\mathbf{StepsRequiredN} * \mathbf{mn/2})$<br><br>**2.** ***Calculation of Mean:***<br>$(\mu_z)_{mn*1} = \frac{\sum_{z=1}^{mn/2}\sum_{i=1}^{N}T(z,i)}{N}$<br>$(\mathbf{StepsRequiredN} * \mathbf{mn/2}$<br>**3.** ***Variance Calculation***<br>$(\sigma_z)_{\frac{mn}{2}*N} = \frac{\sum_{z=1}^{\frac{mn}{2}}\sum_{i=1}^{N}\left(T(z,i)-\mu_z\right)*\left(T(z,i)-\right.}{N}$<br>**(Steps required N * mn/2)**<br><br>**4.** ***Co-Variance Calculation*** |
| $(\Sigma_z)_{mn*mn} = \frac{\sum_{z=1}^{mn}\sum_{y=1}^{mn}\sum_{i=1}^{N}(T(z,i)-\mu_z)*(T(y}{N}$<br>**(Steps required mn * mn)**<br><br>5. ***Eigen Value Eigen Vector Decomposition***<br>$EigenValue\ (\lambda_{N*N})\ and$<br>$EigenVector\ (\Omega_{N*N})$<br>**(Steps required N * N)**<br><br>6. ***Calculation of Eigen Faces (full)***<br>$(\text{Eigenfaces})_{mn*N} = (\sigma)_{mn*N} * (\Omega_k)_{N*N}$<br>**(Steps required N * mn)**<br>7. **Calculating Training Faces**<br>$(\Psi_i)_{k*1} = (\text{Eigenfaces})^t * \sigma(i)$<br>**(Steps required N * mn)** | $(\Sigma_z)_{mn/2*mn/2} = \frac{\sum_{z=1}^{mn/2}\sum_{y=1}^{mn/2}\sum_{i=1}^{N}(T(z,i)-\mu_z)*(T}{N}$<br>**(Steps required mn/2 * mn/2)**<br><br>5. **Eigen Value Eigen Vector Decomposition**<br>$EigenValue\ (\lambda_{N*N})\ and$<br>$EigenVector\ (\Omega_{N*N})$<br>**Steps required N * N)**<br><br>6. ***Calculation of Eigen Faces (half)***<br>$(\text{Eigenfaces})_{mn/2*N} = (\sigma)_{mn/2*N} * (\Omega_k)_{N*N}$<br>**(Steps required mn/2 * mn/2)**<br>7. **Calculating Training Faces**<br>$(\Psi_i)_{k*1} = (\text{Eigenfaces})^t * \sigma(i)$<br>**(Steps required N * mn/2)** |

### 5.2.4 Comparative analysis between Full and Half Face Recognition

For verifying the efficiency of both the aspects we again implemented PCA on both the full faces and the half faces. The experimental results given below are very much promising. By comparing the accuracy and recognition from Table 5.2.1 and Table 5.2.3 we conclude that the recognition rate is almost similar in both the cases, while the recognition time of half faces is nearly half of that of the full faces. While Figure 5.2.4 is showing the accuracy over full and half faces Figure 5.2.5 shows the curve of recognition rate.

**Table 5.2.3 Experiment Performed on Full Face**

| No. of Persons | Training Population | Testing Population | Accuracy | Time (Sec) |
|---|---|---|---|---|
| 10 | 70 | 30 | 96% | 2.6 |
| 20 | 140 | 60 | 93% | 5.0 |
| 30 | 210 | 90 | 94% | 7.5 |
| 40 | 280 | 120 | 95% | 10.2 |

Time taken by system to recognize a person is directly proportional to the number of samples and size of each sample. As numbers of samples are same for both half and full faces, hence the complexity of the program is only dependent on the size of the face. Half faces are just the half in size of full faces, so its time complexity will also be half of the full faces. We can measure the complexity of the program by analyzing each step that we carried through in PCA [87].

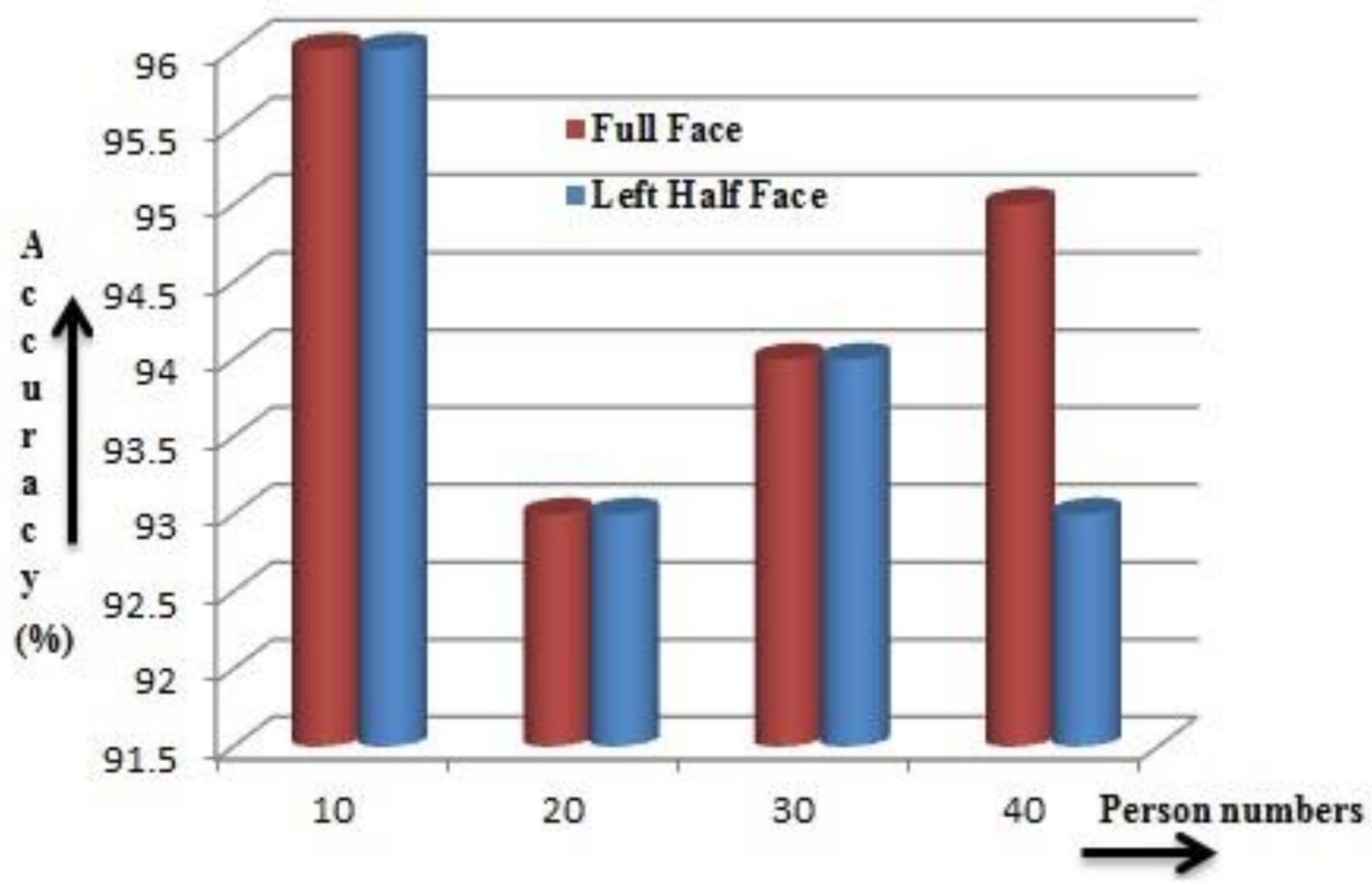

**Figure 5.2.4 Accuracy Comparison Graph of Full Faces and Left Half faces based on PCA evaluation**

Analyzing above all steps we can say that in every step complexity of the half faces reduced just half of the full faces, therefore the overall complexity of the program will also be almost half of the full faces. Hence it verified that our assumption was right.

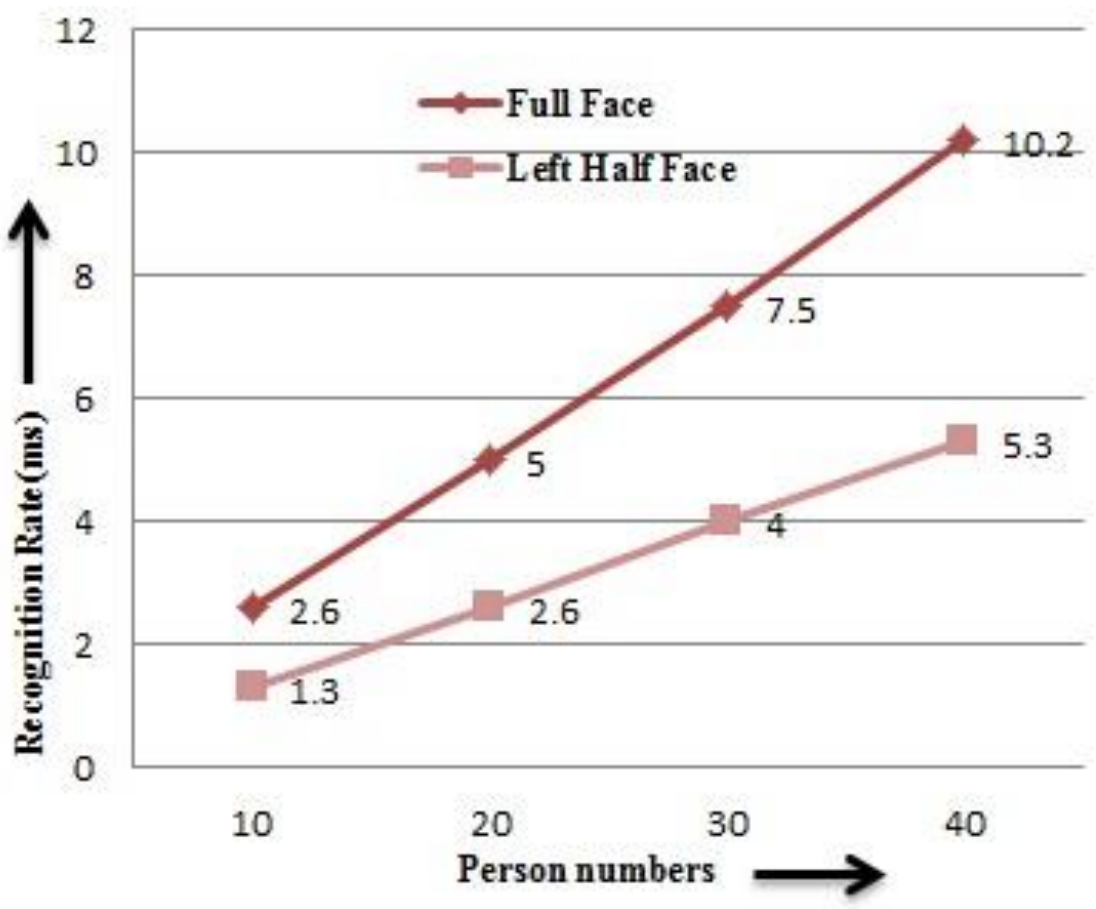

**Figure 5.2.5 Recognition Rate Comparison Graph between Full Face and Left Half faces.**

## 5.3 Component based face Recognition

Face recognition literature is very old and rich. It has very glorious stories to tell, with unsolved challenges. It is basically a pattern recognition problem where one needs to first extract unique features of the face by using a feature extraction technique, then selecting the best features which are later used for classifying the face data over one of the classes. There are various feature extraction techniques which face recognition literature has in his umbrella like Principal Component Analysis (PCA: Eigen faces approach) [87][89][99], Linear Discriminant Analysis (LDA: Fisher faces approach) [90-92], Independent Component Analysis (ICA) [93], Scale Invariant Feature Transform (SIFT) [106-108], etc. Feature selection is as much important as feature extraction one can use Fisher score, sequential forward selection, sequential backward selection, etc. to select the best features, among those extracted features. Classifier design is the main and final task of the pattern recognition problem; we can use Bayesian classifier, Artificial Neural Network (ANN), Support Vector Machines (SVM) etc.

Other group of researchers utilizes eye, nose and mouth components in order to recognize the person. However, they did not provide reasons why ones face is only divided into these components and why not to other components. They have also not tested the reliability of their approaches under these challenges. Therefore, we have presented a theoretical as well as experimental proof for why we consider eye, nose and mouth components. We have also tested this approach under the constant illumination, but over the variable distance from the source. The System is also tested in different facial expression and in the presence of partial occlusion. Rotation and Scale, the system are invariant to these disturbances.

The main agenda of this chapter is to show that most of the invariant features of the face belong to eye, nose and mouth region. In this section we are proving this hypothesis theoretically as well as experimentally by using the SIFT algorithm. A hybrid database is used which consists of 30 faces from California university face database [230], 50 from the University of Essex Face94 dataset [230], 25 from our own Robotics and Artificial Intelligence Laboratory Dataset and 20 faces randomly selected from the internet. Total 125 subject's frontal faces are used in this experiment. The database includes both male and female faces.

### 5.3.1 Problem Definition

For the given face images$\mathrm{I(x,y)}$, and its component images $\mathrm{Eye(x,y)}$, $\mathrm{Nose(x,y)}$ and $\mathrm{Mouth(x,y)}$ prove that most of the principal features Extracted from face images and its component images are same. If $\mathrm{f_p}$ is the feature set of Full Face extracted from I, and similarly $\mathrm{f_u, f_v, f_w}$ are the feature sets of Eye, Nose and Mouth then it should be proven that $\{\{\mathrm{f_u} \cup \mathrm{f_v} \cup \mathrm{f_w}\} \cap \mathrm{f_p}\}$ should be maximum.

### 5.3.2 Proof of the Concept

We can represent face as the combination of its various regions like forehead, eyes, nose, cheek, mouth, and chin. If the face is a superset which consists of all these subsets, then, it can be represented as:

$$I = \{\{forehead\}, \{eye\}, \{nose\}, \{mouth\}, \{chick\}, \{chin\}\} \quad (5.2)$$

If we combine these three subsets eyes, nose, and mouth (i.e.$Eye \cup Nose \cup Mouth$) and find the difference from the superset $I$, we will end up with$Forehead \cup Chick \cup Chin$. In order to show that most of the face features preserved by $Eye \cup Nose \cup Mouth$ rather than $Forehead \cup Cheek \cup Chin$ we have conducted onew test described below.

- **Face Alignment:** Every face should be properly aligned for the reference eye and mouth coordinates. In order to incorporate proper alignment the problem of rotation and resizing should be solved. We have used affine transformation shown in equation (5.3) in order to rotate the image. The rotation angle is measured by the angle formed between left eye and right eye while the translation is fixed to half of the image. If $(x1, y1)$ is the location of any point in the original image and let $(x11, y11)$ is the new transformed location of $(x1, y1)$ then this will be calculated as:

$$\begin{bmatrix} x^1{}_1 \\ y^1{}_1 \end{bmatrix} = \begin{bmatrix} \cos(\theta) & \sin(\theta) & \text{ImgWid} * 0.5 \\ -\sin(\theta) & \cos(\theta) & \text{ImgHgt} * 0.5 \end{bmatrix} \times \begin{bmatrix} x_1 \\ y_1 \end{bmatrix} \quad (5.3)$$

Here "ImgWid" represents image width and "ImgHgt" shows image height. The angle θ can be calculated by using the center of left eye and right eye. If their center coordinates are (x1, y1) and (x2, y2). Then the angle θ can be calculated as: $\theta = \text{aTan2}^{-1}(\frac{y_2 - y_1}{x_2 - x_1})$

The following Figure 5.3.1 (a), (b) and (c) shows the input images aligned by the geometric transformation and the aligned face database.

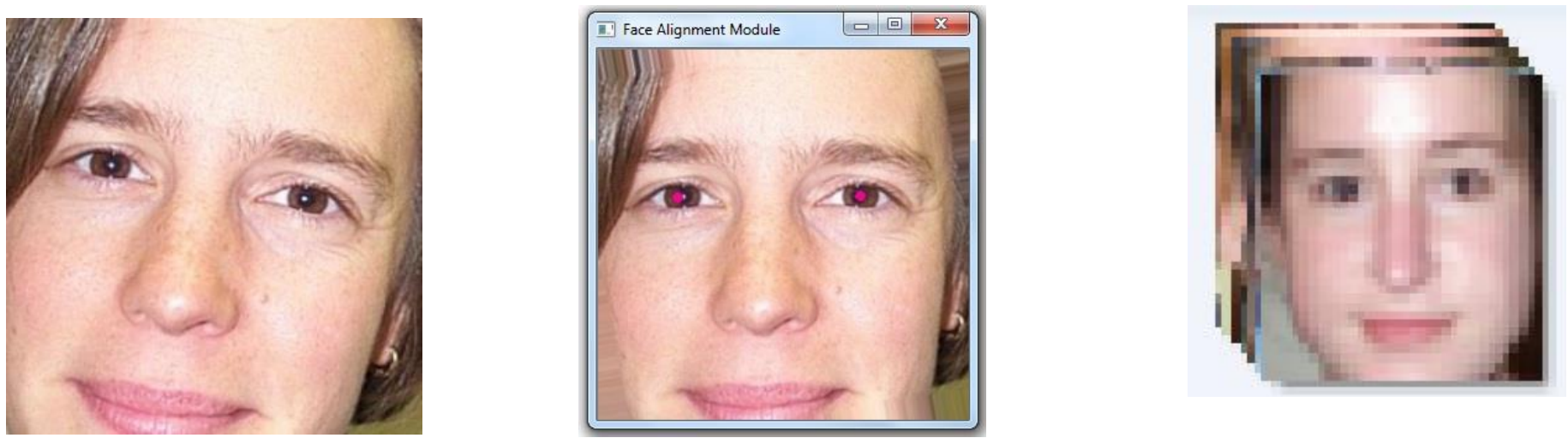

**(a) Input Face** **(b) Aligned Face** **(c) Aligned Face Database**

**Figure 5.3.1 Face Alignment using geometric transformation**

- **Feature Extraction:** Scale Invariant Feature Transform (SIFT) [65] is used to extract unique features of the face. The basic three steps of SIFT algorithm used to discover the feature point location while the last fourth step is to create a feature vector (feature descriptor). If we see the second and third of SIFT algorithm, it calculates the difference of Gaussian (DoG) of each octave images.

The difference of Gaussian gives us the edges and corners. In image processing paradigm these Edges and corners are treated as features of an image so this might be possible that these features could be considered as the key features of the face. As edges are highly sensitive to noise, some of the edges are discarded in the next step. The Harris corner detector is used in the third step to get rid of bad points. In Figure 5.3.2 left image is the given input face image and the left image is the face image with it's SIFT feature locations. Here "+" represents the (x, y) coordinates of the key features of the face, "rectangle" shows the size of features in the image and the "red line" represents as orientations.

With respect to face there is a higher probability to appear are corners of eyes, nose and mouth regions. Therefore, most of the features should fall into these regions.

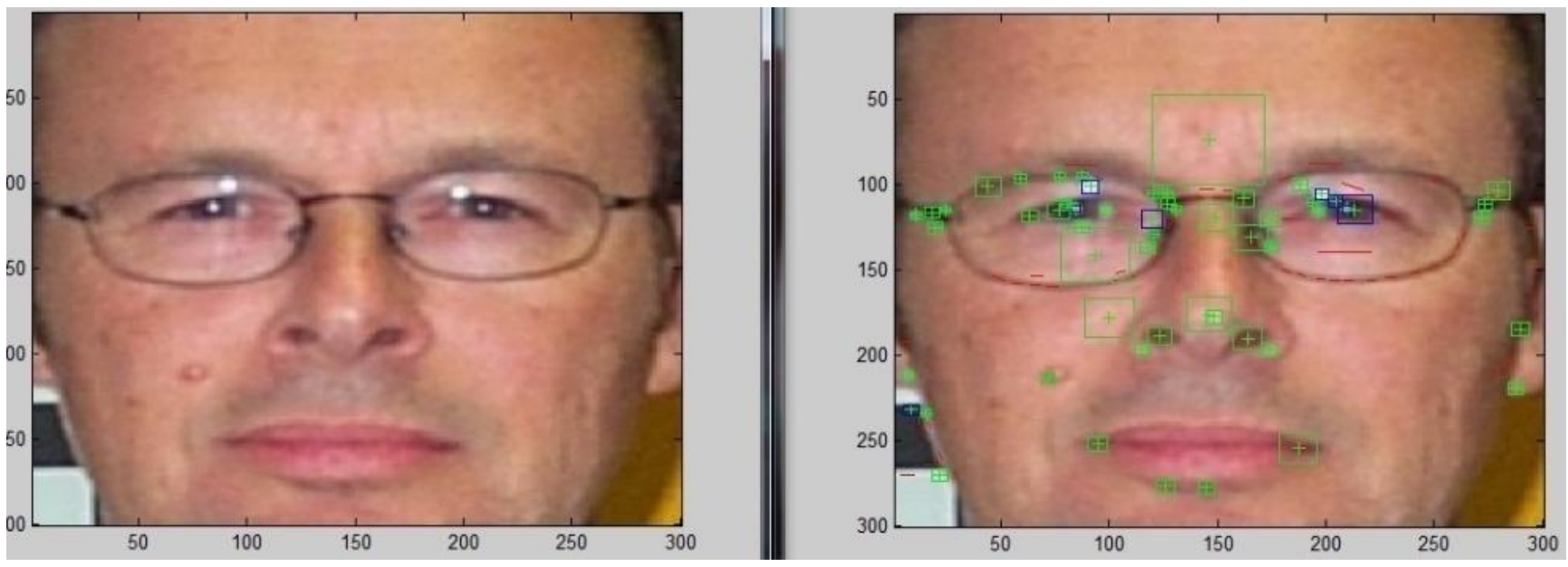

**Figure 5.3.2 Input Face with SIFT features**

- **Feature Location Database Generation:** The feature extracted from each face image are stored in a matrix call face feature matrix shown in Figure 5.3.3 having only 0/1 for each location (x, y). Here 1 represent that at that particular location (x, y) there is a presence of a feature and 0 represents absence.

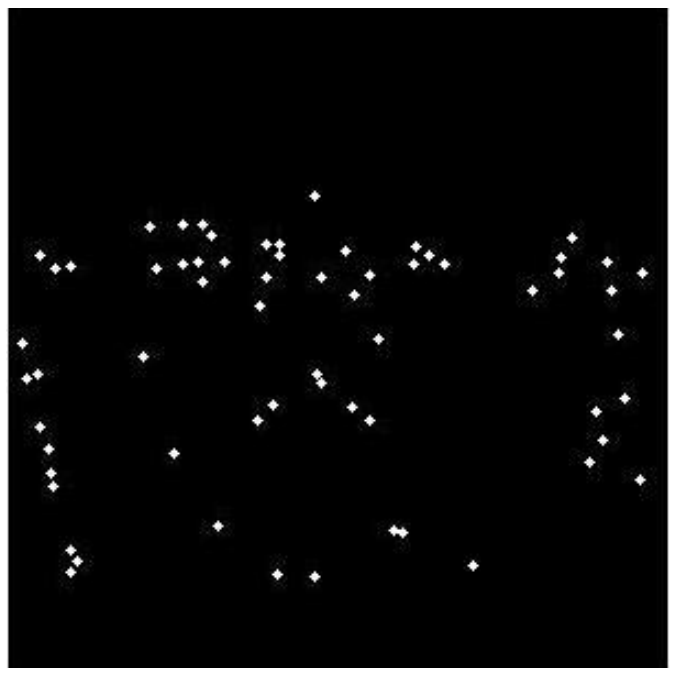

**Figure 5.3.3 Face Feature Matrix w.r.t Figure 5.3.2**

A face feature, database shown in Figure 5.3.4 is created which consists combine features of all the face feature matrices. The database is created by:

$$DB = Ff_1|Ff_2|Ff_3| \dots |Ff_p \qquad (5.4)$$

Here $Ff1$ represents face feature matrix 1, similarly 2, 3 till $p$ ($p$ is the number of images used to conduct this test) and $(j)$ represents the OR logical operator. Figure 5.3.4 shows a constant face window of size 200x200 which is used for all the 360 faces used in the experiment. The face feature database is a binary image where for each pixel (x, y), if there exists a feature point, then a value 1 is assigned to it otherwise 0 is assigned.

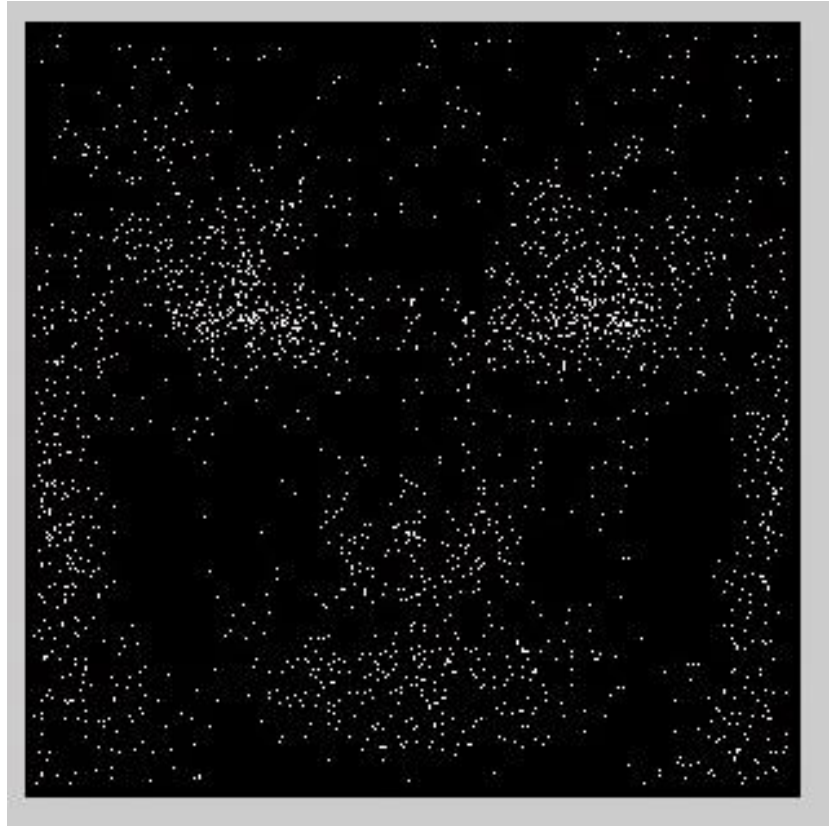

**Figure 5.3.4 Face Feature Database**

- **Density Calculation:** We calculate the density along eyes, nose and mouth region as shown in Figure 5.3.5. Usually the density (d) is calculated on the basis of mass$(m)$ and volume $(v)$ (say$d = \frac{m}{v}$). Here we are applying the same concept for images, hence higher volume is replaced by the total number of features present in the face feature database and mass is calculated by the number of features present in these components. The volume of face $Vf$ is calculated by:

$$V_f = \sum_{i=1}^{a} \sum_{j=1}^{b} \begin{cases} 1 & if F(i,j) == 1 \\ 0 & otherwise \end{cases} \qquad (5.5)$$

Here Vf is the number of features belonging to face database, a and b are the height and width of the face. Similarly, we can calculate the mass of eye (Me), nose (Mn) and mouth (Mm) region. Let $de$, dn, and dm are the density of eye, nose and mouth regions then they can be computed as:

$$d_e = \frac{M_e}{V_f}, d_n = \frac{M_n}{V_f}, d_m = \frac{M_m}{V_f} \qquad (5.6)$$

If $de + dn + dm$>= 0.8 then we can say that these regions preserve almost all the features of the full face image. In this experiment we have used a fixed window size of Face (200x200), Eyes (140X50), Nose (40X40) and Mouth (90X40). Volume of face feature, database $Vf$ is 3382 and masses of eyes ($Me$), nose ($Mn$) and mouth ($Mm$) are 1319, 778, 947. Their respective densities are $de$=0.39, $dn$=0.23, $dm$=0.28 and $de + dn + dm$=.9. It confirms that the combination of Eyes, Nose and Mouth preserve most of the features of the face.

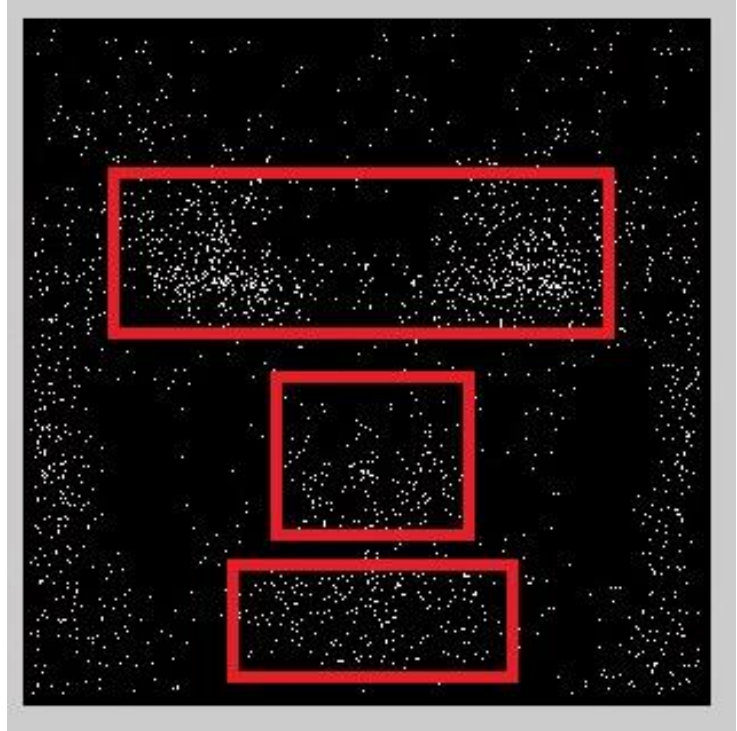

**Figure 5.3.5 Estimated Density wrt. Each Face Component**

The total number of features present in the fixed face window is 3382, which is calculated on the basis of all 360 images. The number of features present in each individual components is: eye (1319), nose (778), mouth (947). The combined features of eyes, nose and mouth is (3044). If we compute the difference between the total numbers of facial features and combined facial features of these components, it is 378 features which are very less in number compared to the full face features. Hence we can draw a conclusion that the combination of these components preserves most of the discernment features of the face.

### 5.3.3 Proposed Framework

The proposed framework shown in Figure 5.3.6 consists of three major modules (a) Landmark Localization based Haar Classifier [116], (b) SIFT feature extraction and the last one (c) Matching of feature point. All these modules are briefly discussed below.

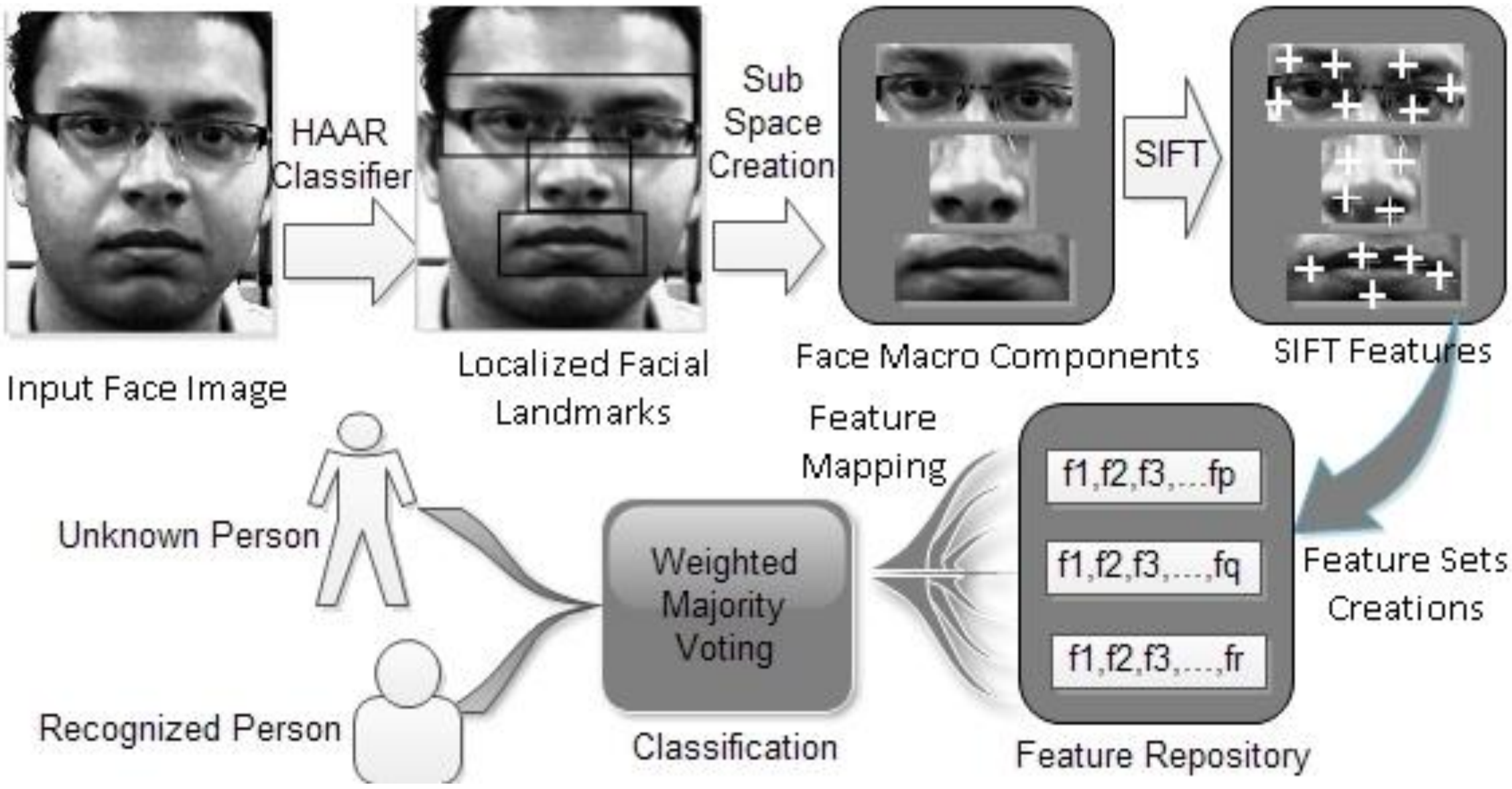

**Figure 5.3.6 Proposed Framework**

**Description of the Proposed Framework**

- **Face Landmark Localization:** Haar Classifier devised by Viola and Jones [116] used Haar like features to detect the objects. These features used change in contrast values to detect the object rather than using the intensity values of the image. For detecting the faces and facial features, it is required to first train the classifier. Viola and Jones used AdaBoost and Haar feature algorithms to train the system. Some images that do not contain any object (called negative images) and some images having the objects (called positive images) are used to train the classifier. They used 1500 images, taken at different angles varying from 0 to 45 degrees from a frontal view to train each facial feature. All the three classifiers that were trained for eye, nose and mouth have shown a good positive hit ratio, which shows the robustness of their approach. Once the face is detected, the rest of the facial features are detected by measuring the probability of the area where they most likely existed.

- **SIFT Feature Descriptor:** SIFT feature descriptors [65] are the 128 dimensional vectors. In order to make these features rotation invariant, a grid of 4X4 is taken along the center of key point suggested by Lowe [65]. For each grid Gradient and magnitude of neighboring pixels are calculated shown in expression (6) and (7).Total angle $360^{\circ}$ is divided into 8 equal parts called bins of$45^{\circ}$. Each grid has 16 pixels and for each pixel we have particular orientation. All the orientation values are assigned to their respective beans. Therefore, each grid gives us 8 dimensional features, while we have overall 4X4 window grid. A Total of 4X4X8=128 dimensional feature vector is generated. This will be done for all key points extracted from the previous step.

$$m(x,y) = \sqrt{(O(x+1,y) - O(x-1,y))^2 + (O(x,y+1) - O(x,y-1))^2} \quad (5.7)$$

$$\theta(x,y) = tan^{-1}(\frac{O(x,y+1)-O(x,y-1)}{O(x+1,y)-O(x-1,y)}) \quad (5.8)$$

- **Feature Mapping:** Minimum distance pair technique [230] has been used for feature matching. Let we have 3 feature set databases for eyes, nose and mouth say gallery feature set.

$$Eye_{gallery} = \{u^1{}_{c1}, u^2{}_{c1}, u^3{}_{c1}, u^4{}_{c2}, u^5{}_{c2}, u^6{}_{c2}, u^7{}_{c3}, u^8{}_{c3}, u^9{}_{c4},, \dots, u^i{}_{cl}\} \quad (5.9)$$

$$Nose_{gallery} = \{v^1{}_{c1}, v^2{}_{c1}, v^3{}_{c1}, v^4{}_{c2}, v^5{}_{c2}, v^6{}_{c2}, v^7{}_{c3}, v^8{}_{c3}, v^9{}_{c4},, \dots, v^i{}_{cm}\} \quad (5.10)$$

$$Mouth_{gallery} = \{w^1{}_{c1}, w^2{}_{c1}, w^3{}_{c1}, w^4{}_{c2}, w^5{}_{c2}, w^6{}_{c2}, w^7{}_{c3}, w^8{}_{c3}, w^9{}_{c4},, \dots, w^i{}_{cn}\} \quad (5.11)$$

Here $u^i{}_{cl}, v^i{}_{cm}, w^1{}_{cn}$ represents that the $i^{th}$ sample belongs to l, m, and n classes respectively. The test set also involves the similar component's features set.

$$Eye_{test} = \{a^1, a^2, a^3, a^4, a^5, \dots, a^p\} \quad (5.12)$$

$$Nose_{test} = \{b^1, b^2, b^3, b^4, b^5, \dots, b^q \quad (5.13)$$

$$Mouth_{test} = \{c^1, c^2, c^3, c^4, c^5, \dots, a^r\} \quad (5.14)$$

For each feature 1 to n matching are performed. On the basis of the minimum distance from each class features, it will be assigned to the respective class. A graphical representation is shown in the Figure 5.3.7. It has been taken care that Eye component will match the eye and nose and mouth will match to their respective components. From Figure 5.3.7 we can see that if $d1 < d2 < d3 < d4 < d5$, it shows that d1 has the minimum distance among all pairs. Therefore, test feature $a^1 \in C_1$. The same process is applied to all features. At the end composite score on the basis of each component is computed.

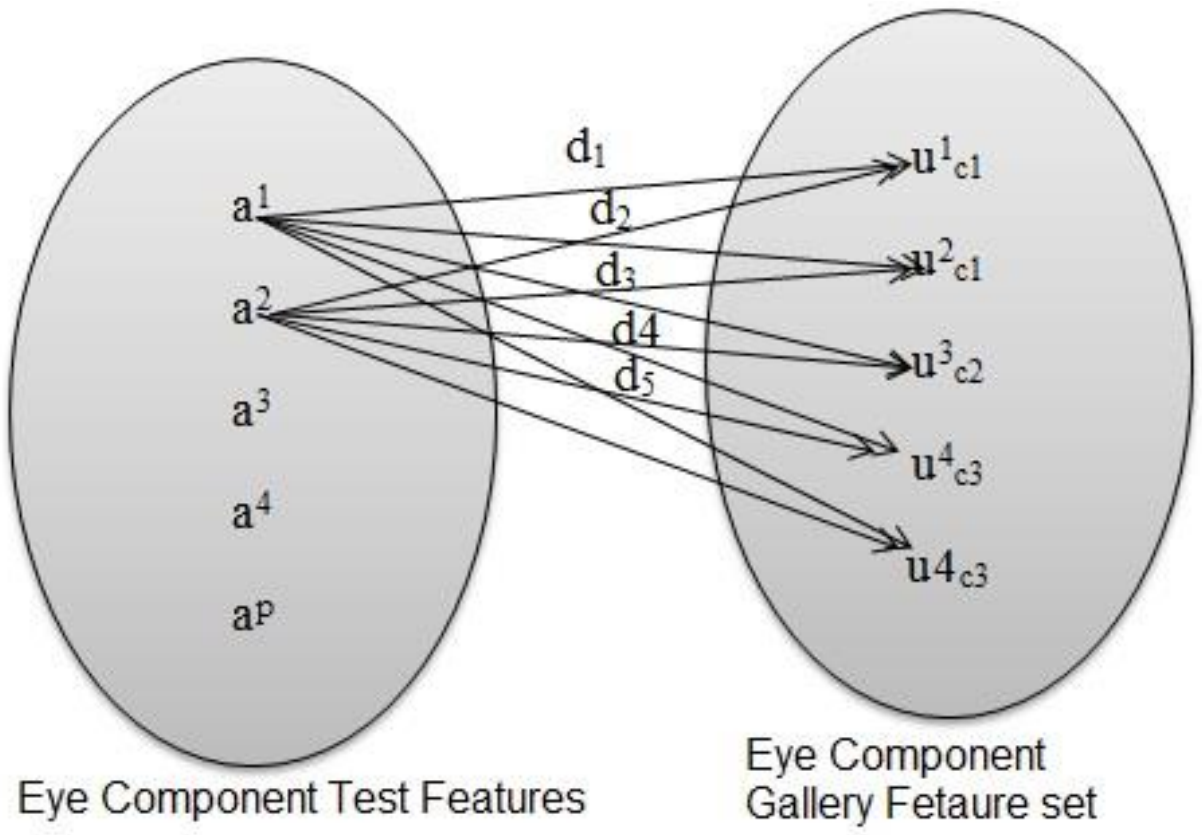

**Figure 5.3.7 An example of feature mapping**

### 5.3.4 Results and Discussion

It has been confirmed from the test conducted in section 5.3.2 that most of the features of human face lies in eyes, nose and mouth regions. Hence, by dividing the face for these components most of the discriminant features are still preserved. In order to prove this, we have applied face recognition on both the full faces and on their components. We have compared the accuracy of these components with respect to full faces. A public ORL Face dataset [229] has been used. The dataset has 40 subjects including male and female, each subject has 10 samples. Three different feature extraction techniques (PCA [87], LDA [90], and SIFT [65]) discussed in chapter 2, are used for feature extraction, whereas minimum distance classifier (Euclidean distance) used for face classification. But in case of component based approach we have used weighted majority voting as classifier over the minimum distance classifier shown in Figure 5.3.8. Here we have assigned weights for each component on the basis of the features it has. As eye component has the maximum number of features followed by the mouth and nose components their weights are fixed as 0.5, 0.25, and 0.25. The final score is computed with the combination of all three components as given follows.

$$FinalScore = 0.5 * (EyeScore) + 0.25 * (NoseScore) + 0.25 * (MouthScore) if (FinalScore \geq 0.5) then Face is Recognized \quad otherwise Not Recognized$$

If any sample belongs to their correct class, then its score is assigned as 1 otherwise 0. Figure 5.3.8 shows an example of voted majority voting. If the eye component predicts the test sample belongs to class C1 and if the test sample actually belongs to class C1. Its value will be 1, similarly for nose and mouth components their values will be 0 and 1. Their Final Score will be calculated as:

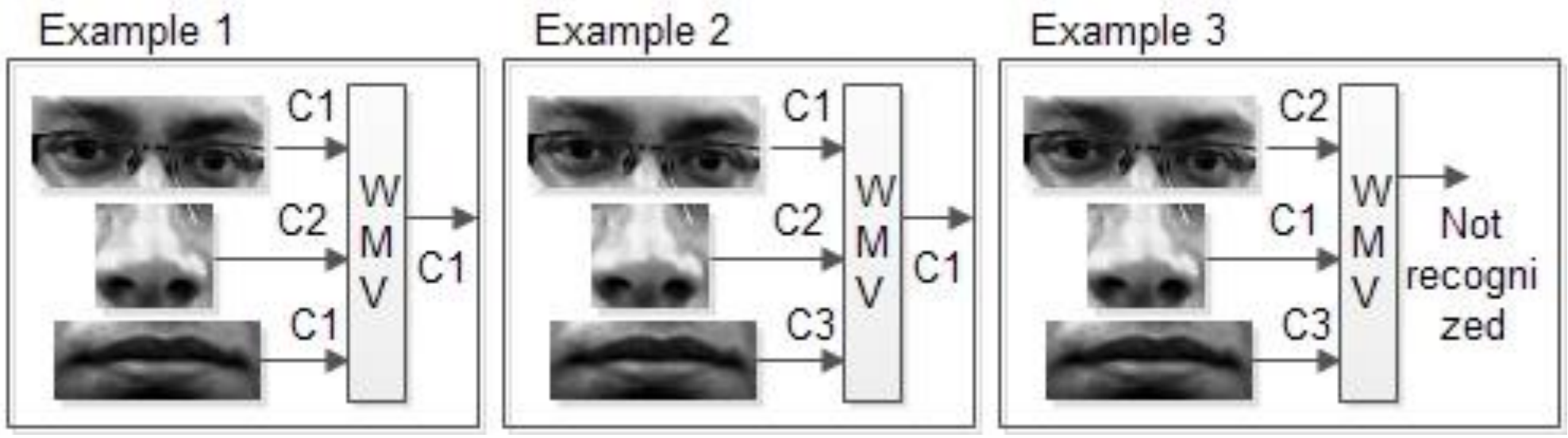

**Figure 5.3.8 Weighted Majority Voting (WMV) Classifier on Component Based Approach**

For example (1)

$$FinalScore = 0.5 * (1) + 0.25 * (0) + 0.25 * (1)$$
$$FinalScore = 0.75$$

For example (2)

$$FinalScore = 0.5 * (1) + 0.25 * (0) + 0.25 * (0)$$
$$FinalScore = 0.50$$

For example (3)

$$FinalScore = 0.5 * (0) + 0.25 * (0) + 0.25 * (1)$$
$$FinalScore = 0.25$$

In the example (1) and (2) the test sample belongs to class $Cl$, while in the example (3) it results as unrecognized person. Table 5.3.1 shows the advantage of weighted majority voting over simple majority voting. It shows that the average misclassification rate decreases for each case in comparison to simple majority voting, whereas it is almost same in case of SIFT. 10-fold cross validation has been applied with respect to all feature extraction techniques. The average classification on these classifiers shown in Figure 5.3.9, while misclassification has been presented in Table 5.3.1.

**Table 5.3.1 Average Misclassification (%) on ORL Dataset**

| Techniques | Full Faces | Component based Approach with simple majority voting | Component based Approach with weighted majority voting |
|---|---|---|---|
| PCA | 5.75 | 25.50 | 8.5 |
| LDA | 2.50 | 20.75 | 7.5 |
| SIFT | 1.25 | 2.25 | 1.25 |

As we can see from Table 5.3.1 that component based methods are giving good results for SIFT (SIFfComp) while in case of PCA and LDA their accuracy suffers. The above results also confirm that local feature analysis of these components will provide better results than global approaches. Experimental results in both the data sets confirmed that we can even recognize the face with only these components. We have extended this and try to see what are the optimal number of components that would result in the same number of accuracy, as we get with the combination of these three components. We have achieved very promising results in case of $Eyes \cup Mouth$ where we are having the same level of accuracy as in case of $Eye \cup Nose \cup Mouth$. The idea of using different combination is to further reduce the feature space by selecting the subset of $Eye \cup Nose \cup Mouth$. We have summarized the recognition accuracy of individual component and their possible combinations in Table 5.3.2. By selecting the subset, the time complexity is also reduced as can be seen in Figure 5.3.10. It shows that how the time complexity of fragmented face components are reduced to half the time required by the SIFT on full faces.

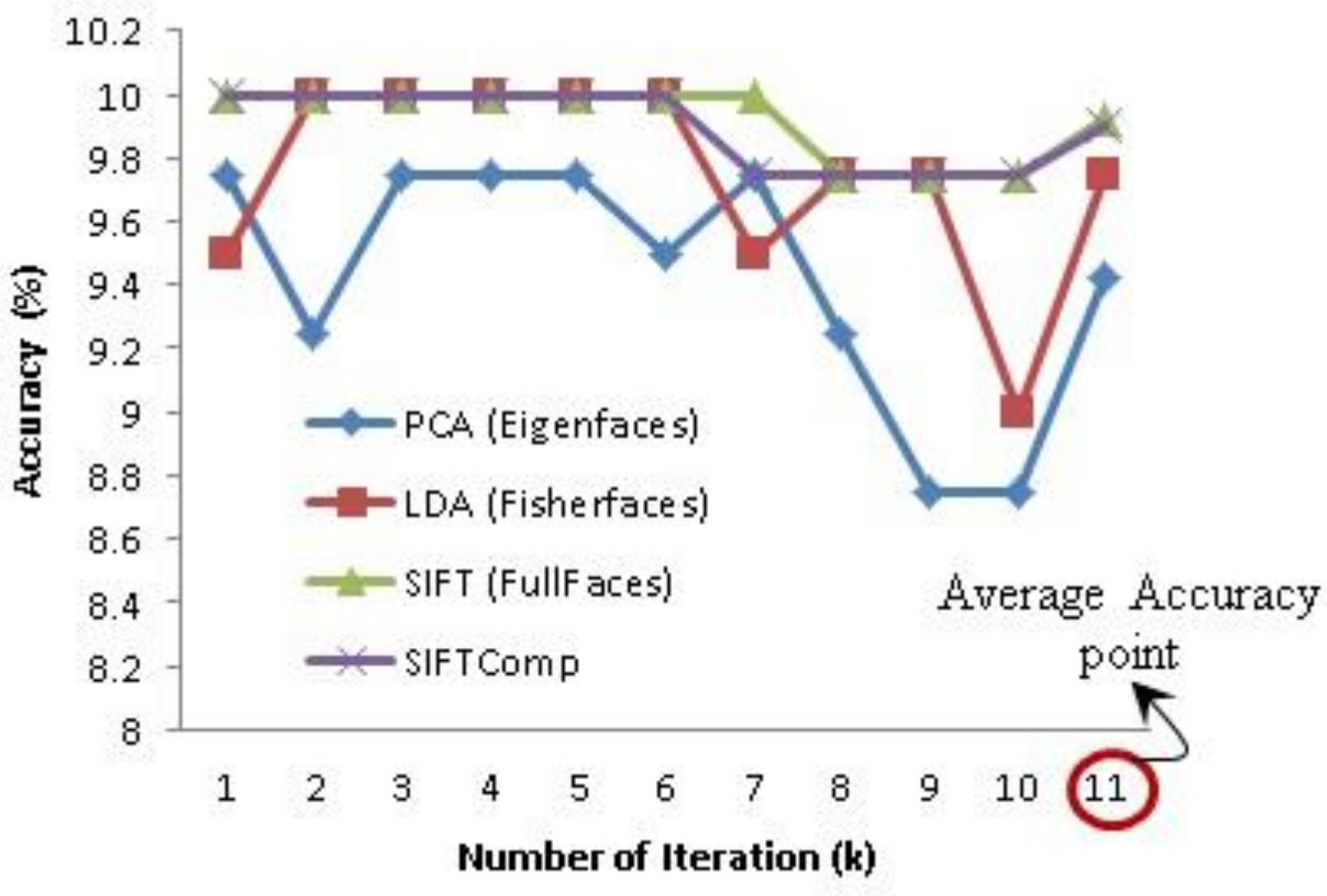

**Figure 5.3.9 10 fold cross validation classification accuracy**

**Table 5.3.2 Recognition accuracy of individual components/their combinations**

| Face Components | Accuracy (%) with SIFT | Face Components | Accuracy (%) with SIFT |
|---|---|---|---|
| Eye | 92.50 | Eye ∪ Nose | 92.50 |
| Nose | 75.00 | Nose ∪ Mouth | 92.50 |
| Mouth | 90.00 | Mouth ∪ Eye | 97.50 |

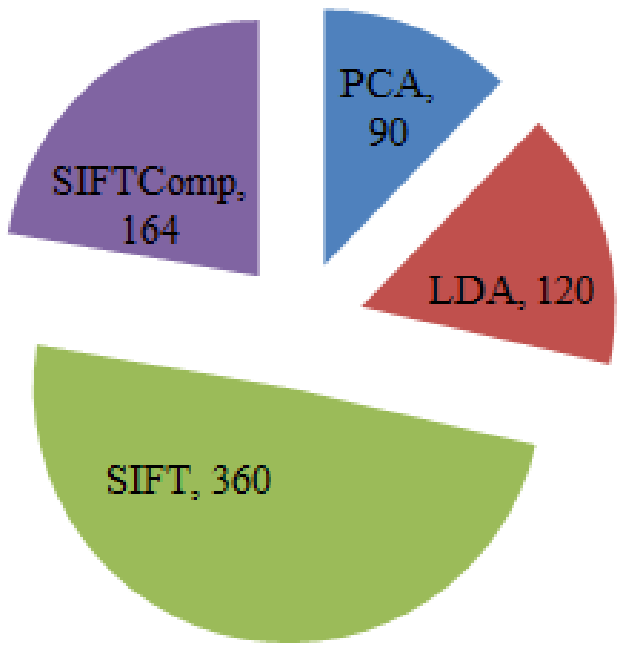

**Figure 5.3.10 Time Comparison Analysis: Recognition Time (ms) on ORL Dataset**

## 5.4 Conclusion

In general, recognition in real time has several challenges; time is also a big factor in this regard. Time is directly proportional to the size (dimension) of the data (images). As human faces are almost bilaterally identical in nature, they have the vertical symmetry. Hence we have used only left half of the human faces for training and testing. We have tested our hypothesis on an ORL database of 40 people having 400 images. Symmetry is verified on the basis of the accuracy of the classification rate between the left and right portion of the face. Experimental results of full faces and a half faces shows that they are almost equal in accuracy, but the recognition rate (time) based on half faces is nearly half of the time taken by the full faces. Hence it is verified that half face is also sufficient for the face recognition paradigm. This approach saves both time and resources (computation and memory). Another advantage of using the half face approach is that they are invariant to facial expressions. For the better classification, faces should be aligned perfectly.

We have also tested the component based face recognition framework religiously in order to establish our hypothesis which is already supported by existing literature. An attempt has been made to define a mapping between the feature space and the input space. The density estimation among spatial features confirms that most of the features belong to fragmented face components. The independent features from the segmented regions are extracted using Principal Component Analysis (PCA), Linear Discriminant Analysis (LDA) and Scale Invariant Feature Transform (SIFT). The extracted features are later engaged in classification purposes. The K-fold cross validation technique (where k=10) has been applied on these features in order to make the diagnosis the average misclassification error between full face and fragmented face approach. The average recognition rate of 98.75% on SIFT algorithm demonstrates the strength of the extracted local features over Eigenfaces and Fisherfaces. We have further extended this approach to explore best possible combination of these fragmented components. This process will ensure us to obtain optimal fragmented face components which reduce feature metric with less comparison time. In case of Eye ∪ Mouth we have achieved 97.50% recognition accuracy which is very close to initial fragmented face recognition approach.

Both of the techniques emphasize on using the component of the face over the full face for the person identification. These facial components can handle the uncertainties present in the recognition such as (i) it can minimize the effect of the change in facial expression over the face, which indirectly helps in better recognition, (ii) using only the subset of the face, we can nullify the effect of different types of occlusion and (iii) the last most important thing which has been proven through the rigorous experiment that, they speed up the recognition rate while perverse the recognition accuracy. Face recognition literature is vast and has various techniques for feature extraction, feature representation, feature selection and classification. The proposed technique neither helps in feature extraction, representation, selection nor in classification. It just suggest another way of looking into the problem of face recognition. We have investigated and diagnosed some of the basic challenges of face recognition and presented a supplementary method which can minimize the uncertainties lies in identification.

# Chapter 6:

# Face Anti-Spoofing

*The preceding chapter discuss the face recognition framework to identify criminals, however the current system can be fooled by placing a face spoofing attack. face anti-spoofing is a safe guard against face spoofing attacks performed in order to get unauthorized access of robot neural schema or to misuse it or to bypass the identification process. The attack is very simple to perform, an attacker need to acquire photo or video of the enrolled user. The collection of valid user photo/video is very simple, it is mostly available on World Wide Web / social platform. This chapter proposed two anti-face spoofing mechanisms to detect and prevent these kinds of attacks. The first method analyzes facial movement while the second approach utilizes 3D facial structure (disparity map) of both real and imposter to distinguish between these two classes.*

## 6.1 Introduction

Previous chapter discusses the problem of criminal identification of the given face image of the criminal. In most of the situations the given face image is occluded, partial and blurred in nature. Therefore, to address these issues, facial symmetry and component based face recognition techniques are proposed as described in the previous chapter. The face recognition problem is twofold, in the first step the face detection module distinguishes between the face and other objects. Once the face is localized and extracted from the given frame, the second module is of reliable matching of that extracted face image to one of the classes (person) enrolled in the database. The overall scenario of face recognition does not bother about the reliability of the medium. A robot having the capability of face recognition can identify the criminal, but it can also be easily fooled. Face Spoofing is an attack which can bypass the face recognition (authentication) process, just by placing photograph or video of the genuine user in front of the camera (robot eye). A robot is a programmable machine having various sophisticated technologies and tools embedded in it. A robot can also be misused. Therefore, in order to protect these humanoid robots and to prevent them from unauthorized access, it is mandatory to train them so that they could differentiate between the real and the imposter. This chapter proposes two anti-face spoofing mechanism to block these kinds of attacks.

Spoofing is an attack which is used to impersonate the identity of someone else. As face is the medium of authentication and recognition, face spoofing is an attack to get unauthorized access by displaying the photo/video/mask of the enrolled user. The existing literature suggests liveness detection test to detect the face spoofing attacks. Liveness is the measurement of life in the present-day source which is used for the recognition, i.e. face is used to recognize the person, therefore the liveness of the face is tested before being processed for recognition. This chapter proposes two different techniques to test the liveness of the source based on the principle of movement in facial features and the 3D face structure analysis of both imposter and real face. The first solution is built on the challenge and response protocol of cryptography. In the challenge and response protocol the system challenges the second party to present some token or key or asks some questions to reveal the identity. The challenges are thrown in such a way that the challenger can verify the second party. We have mounted our facial movement tracker model over this protocol to test the liveness of the user. The robot asks to perform some movement in the eyes such as eye blink or close or open eyes, similarly challenges to perform some movements of the mouth such as the open / close mouth. The response of each challenge is then tracked by robot eyes. We have used Haar eye and mouth cascade to first localize these macro components. We have tested the efficiency of the system against photo and video attacks. The second solution exploits the facial structure to detect and prevent face spoofing attacks. Due to the odd and even surface of the face, we have 3 dimensional geometry of the face, while the medium used for placing the face spoofing attack is a 2D medium (photo/video). As Humanoid robot has stereo vision, we can utilize them to generate the disparity map of each face image. Disparity at any point of the face image is inversely proportional to the distance of that pixel from the center of the camera. As face is the 3D medium, we will get variation in real face disparity, but the imposter face has fixed distance from the camera, hence it has same disparity at each point of the face images.

We have generated the disparity image of both the classes and applied three machine learning algorithm Linear Discriminant Analysis, Principal Component Analysis and K-Means to classify the test disparity to one of the classes (Imposter Vs Real).

The rest of the chapter is summarized as follows: Section 6.2 proposed the challenge and response method to detect liveness of the user, this section describes the framework first and then in sub section it describes and validate the experimental setup. In section 6.3 introduce a novel depth perception based face liveness detection. 3D faces are generated using the stereo vision and three machine learning algorithms are used to classify the 3D face generated from both real and imposter classes. In the end, section 6.4 summarizes the chapter with a conclusion and future work.

## 6.2 Face Recognition with Liveness Detection using Eye and Mouth Movements

Face spoofing is an attack where the attacker tries to bypass the face recognition system by placing a photo/video/mask of the enrolled persons in front of the camera. The problem lies within the working principle of Face Recognition. The principle of face recognition is unconcerned for who is submitting the credentials. It is only concerned whether the person is enrolled or not. Hence, whatever effort is done for making the classification good and effort in using a good resolution camera will not be effective. Besides this, it helps the attacker to perform their attack more accurately. This problem leads to the question where one can think about the significance of the accuracy and efficiency of the system, when the reliability is not assured. Liveness detection of the user could be a way to deal with this problem.

Researchers had observed the need of security mechanism and proposed various ways to deal with this problem. On the basis of literature we have grouped possible solution in three main categories (1) liveness detection by using the challenge and response method (2) liveness detection by utilizing face texture (image quality), and (3) liveness detection by combining two or more biometrics (multi-modal approach). Challenge and response system throws some challenge in terms of the eyes and mouth movement which can only be performed by real user not by photo/video, and analyses their response on account of the given challenges. In this regard, most of the researchers [147-148] [150] have utilized eye blinking, while in type-II researcher’s exploited texture information (smoothness/roughness, edges, etc.) to distinguish between real and imposter [127][129].Multimodal approach mostly uses speech and face as the combination to deal with this attack [162]. These spoofing techniques are going to be more complex day by day, from a simple photograph to painted contact lenses and polymeric face, fingers. Hence a list of modern approaches to deal with all these circumstances is mentioned by [34][149]. They have suggested in their report that for liveness detection, maximum utilization of face macro features is the best way.

Therefore, here we utilize both eye and mouth movements to detect the liveness with the constraint of challenge and response. Previous techniques discussed in the literature are mostly based on the eye blinking, which can be easily forged by the video imposter attack. Therefore on the basis of existing literature, we are proposing here a challenge and response technique in such a way that an imposter will not be able to forge it. The technique is briefly discussed in the next section (challenge generation) and described in Figure 6.2.1. This adds more difficulty to bypass the system, and if the attacker tries to forge, it will result in wrong verification, resulting in authentication failure. For testing our hypothesis, we have designed five types of attack by using (1) Photo Imposter Attack, (2) Eye imposter Attack, (3) Mouth Imposter Attack, (4) Eye and Mouth Imposter Attack, (5) Video Imposter Attack. The system successfully prevented attacks in all these conditions without any False Rejection Rate (FRR). Experimental results show that his system is able to detect the liveness when subjected to all these attacks except the eye & mouth imposter attack. This attack is able to bypass the liveness test, but it creates massive changes in facial structure. Therefore the resultant is unrecognized or misclassified by the face recognition module. An experimental test conducted on 65 persons on university of Essex face database confirms that removal of eye and nose components results 75% misclassification.

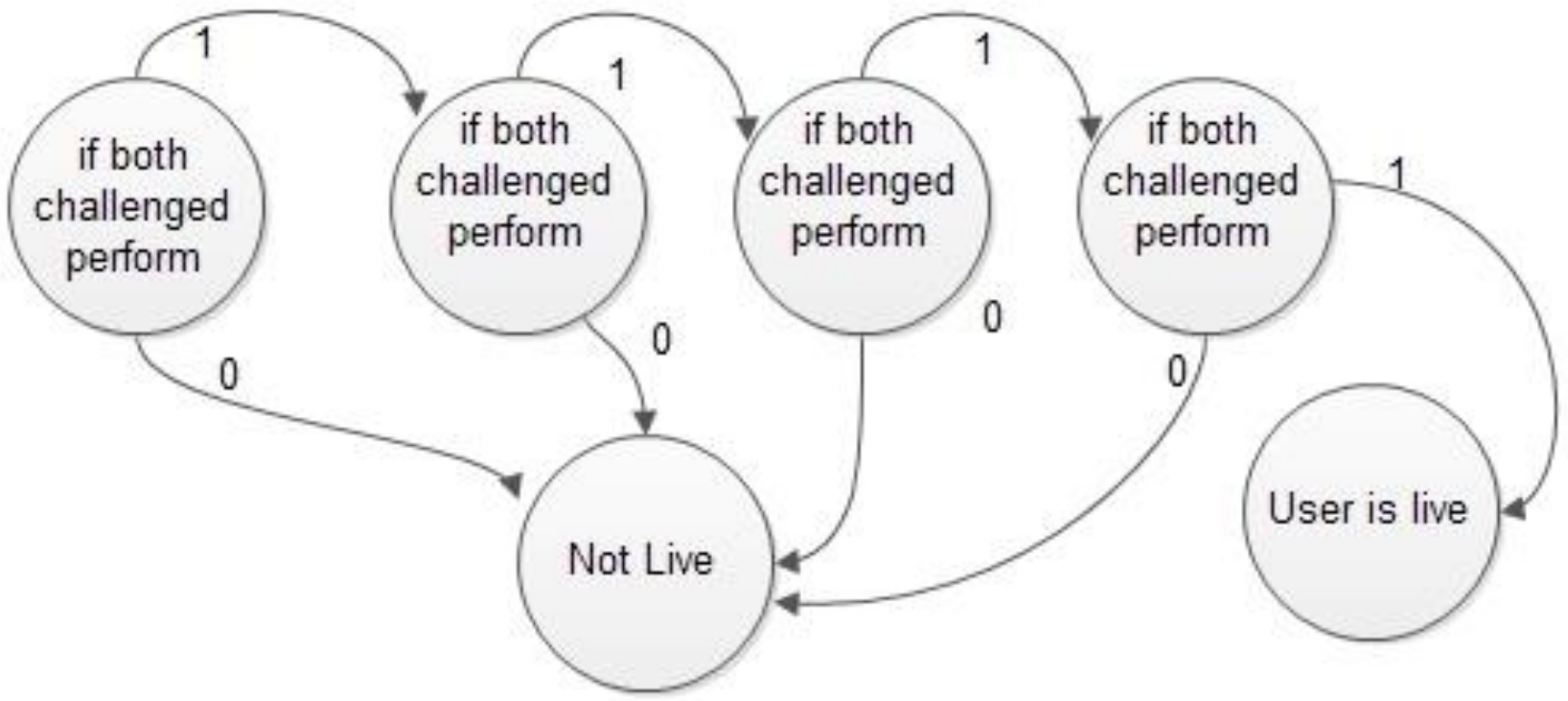

**Figure 6.2.1 Liveness Detection and Challenge Representation**

## 6.2.1 Proposed Framework

The work flow of the proposed framework is presented in Figure 6.2.2. Here we have divided the approach in two parts (a) liveness detection and (b) face recognition. First, we are testing for liveness of the user, and if the person is live then the system will recognize the identity otherwise not. The human face is dynamic in nature and comes in many forms and colors. Hence the detection of human faces among different objects is even more problematic than recognizing any other object. Here, for detecting the faces and facial features, we are using Haar Classifier devised by Viola and Jones [116].

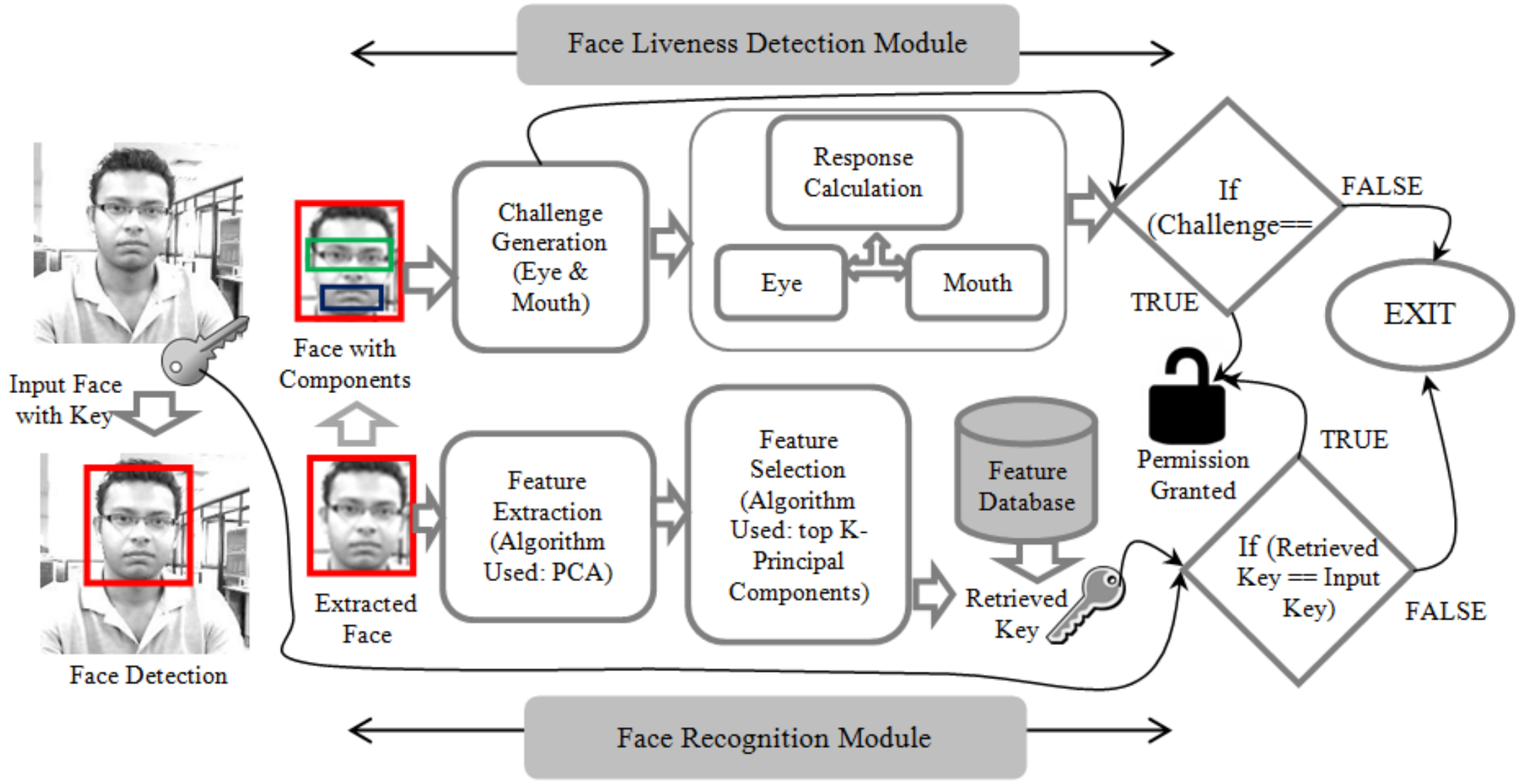

**Figure 6.2.2 Proposed Framework: Face Recognition with Liveness Detection**

#### 6.2.1.1 Challenge Generation

Challenges are generated for testing whether the person is live or not. We assumed that if the person is live then he/she can move her/his face and facial feature too, but it is not true for all cases. This assumption will fail when attacker plays any recorded video of the genuine user, hence here we are generating the challenges in such a way that only the person who is live, can only respond to those. Challenges are presented in terms of their eye and mouth movement (openness/closeness) in a sequence. These sequences are generated at random, so that no one can make a prior estimation about the challenges. The designed system is able to calculate the movement by measuring the teeth HSV value (Hue Saturation Value). Both of these challenges are generated at random, and presented at the same time (dependent on each other). This dependency is also due to the fact that the attacker will not be able to forge it, even if they try to forge, it will lead to misclassification.

#### 6.2.1.2 Response Calculation

Responses are calculated by counting the movements (eye and mouth movement). Eye openness and closeness is calculated by searching the eye in the region where the eye should exist shown in Figure 6.2.3, if found we assume that eye is open otherwise it is closed. Similarly mouth openness and closeness is calculated by searching teeth (symbolized by the HSV value of the teeth) in the mouth region where it's likely to exist, if noted mouth is open otherwise close. A number of challenges thrown by the system acted as the threshold and if the sum of the responses is equal to the threshold, then the system will recognize the person as live is not. We have used the human face convention here, in which we have fragmented human face into different regions shown in Figure 6.2.3. Eye found in eye region i.e 65% to 75% of the face height, Nose and Mouth found in their respective regions (nose region 35% to 65% of face height, mouth region 25% to 35% of face height). The idea of fragmenting the face in this way is that it reduces the search complexity. Eye, nose and mouth all are searched in their respective region.

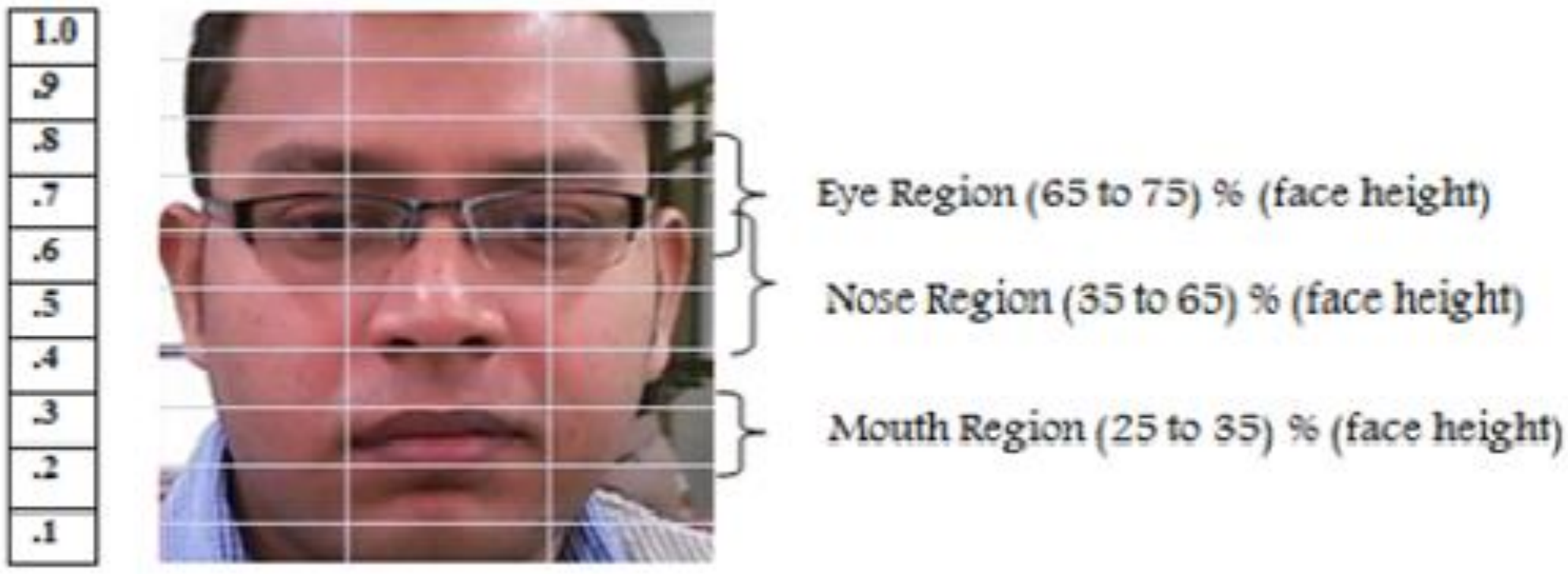

**Figure 6.2.3 Human Face Anatomical Structure**

As an attacker tries to impersonate the identity, it will result in different classification (True or false Classification). All these results are stored in the file and if the face is live then whoever scored highest, their name will be fetched from the list. If the recognized name and the name presented by the user is matched, permission will be granted access the system otherwise not. Here user name is treated as a key which is assigned at the time of user enrollment. If the user is genuine (not an attacker), the presented name and the matched name will be always same.

### 6.2.2 Experimental Setup

Experimental setup is divided into two units, (a) Attack generation and (b) Liveness detection and Recognition. In attack generation, we have shown all five types of spoofing attacks and in prevention unit we have demonstrated how our system identifies these attacks.

**Attack Generation:** Five types of attacks listed below are used in this experiment shown in Figure 6.2.4 (a),(b),(c) and (d).

1. Photo Imposter Attack: A simple photograph of the genuine user is enough to bypass the authentication, if liveness measure has not been taken care of.
2. Eye Imposter Attack: If liveness measure is taken care by the user, like blinking or eye closet is detected by the system, it could be possible that by using this, attackers bypass this attack.

3. Mouth Imposter Attack: If only mouth movement is used in detecting the liveness then it could be possible that by using this kind of attack, attacker fools the system. This is same as Eye imposter, only difference is that the attacker only removes eye region.
4. Eye and Mouth Imposter Attack: These kinds of attack are where both the features are being forged by the attacker. This kind of attack sometimes bypasses the liveness test.
5. Video Imposter Attack: Attacker can use any recorded video of the genuine user to forge the system; video is played reversely many times till system accepts this as a genuine threat.

**Liveness Detection and Recognition:** Recognition and liveness detection both are taking place simultaneously. Same frame is used for testing the liveness as well as recognition. The designed system first blocks the attack in its primary phase by detecting the liveness. If the system ensures that the person is live, then only it verifies the respective identity. A user has to be first enrolled in the system with his key (i.e. name) and it will be stored with its face data. Whenever a person is recognized by the system, respective key will be drawn and stored in a "recognized identity key database". At the time of authentication, the system asks to submit the key and then throws random challenges of the user to test liveness. If the valid responses are given by the user, then only submitted key will be verified with the maximum occurred key drawn from the "recognized identity key database". Examples of spoof detection and genuine user recognition are presented here to show the effectiveness of the approach. Here we have used 4 symbols to generate the challenges.

Eye open 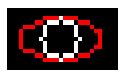 Eye close

Mouth Open 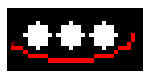 Mouth Close

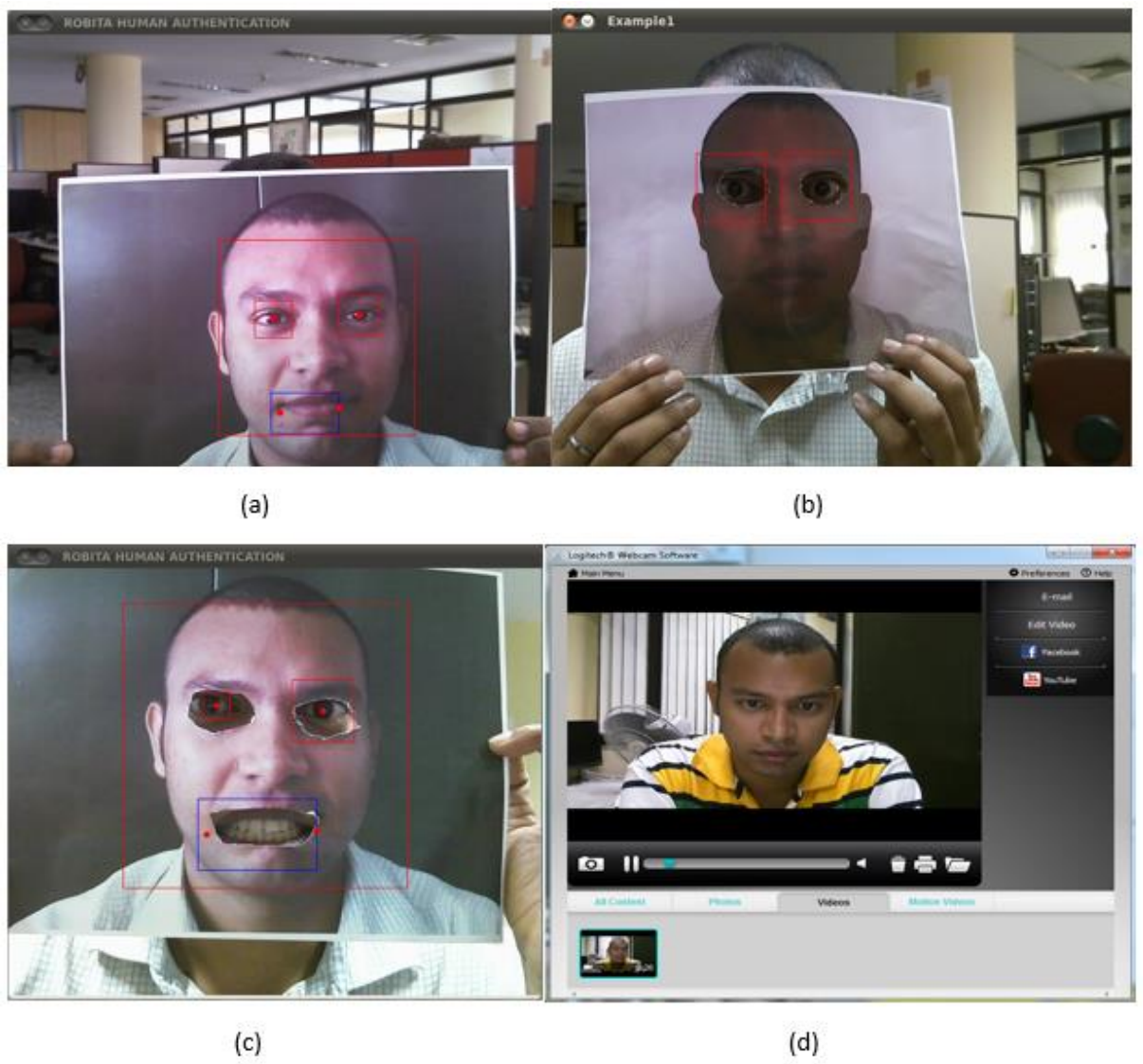

**Figure 6.2.4 Top First, Image Imposter Attack. (b): Top Second, Eye Imposter Attack. (c): Bottom First, Eye and Mouth Imposter Attack. d): Bottom Second, Video Imposter Attack.**

If the attacker has passed the liveness test, the system blocks it because of the failure of its verification (Recognition module) [87]. To pass the liveness test, the attacker has to forge both the eye and mouth region of the genuine user which will result in major changes in the face and hence result in misclassification. The key presented by the attacker and that drawn from the "recognized identity key database" will not match shown in Figure 6.2.5. Figure 6.2.6 shows the recognition of a genuine user, if the user follows instructions given by the system, it will pass the liveness test.

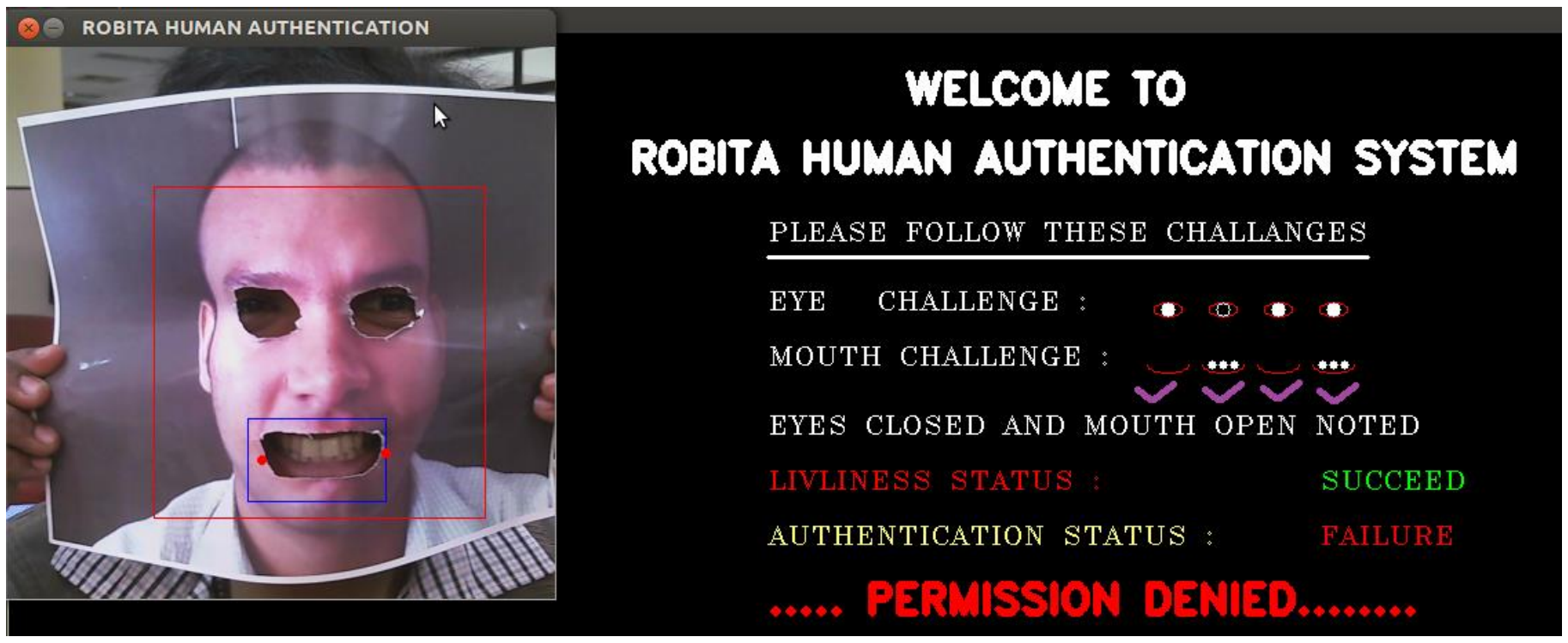

**Figure 6.2.5 A Spoof Attack detected by the System**

The system will verify the user by matching its presented key and the key drawn from the "recognized identity key database". Here, the person is already registered in the system, hence, the system is showing successful authentication. Results obtained by us are summarized in Table 6.2.1. It shows how the system behaves when attacks are placed. The report is very convincing. Only one attack bypasses the liveness test, but it is further blocked by verification module. When the attacker forges both the eye and mouth, it will result in misclassification as shown in Figure 6.2.5.

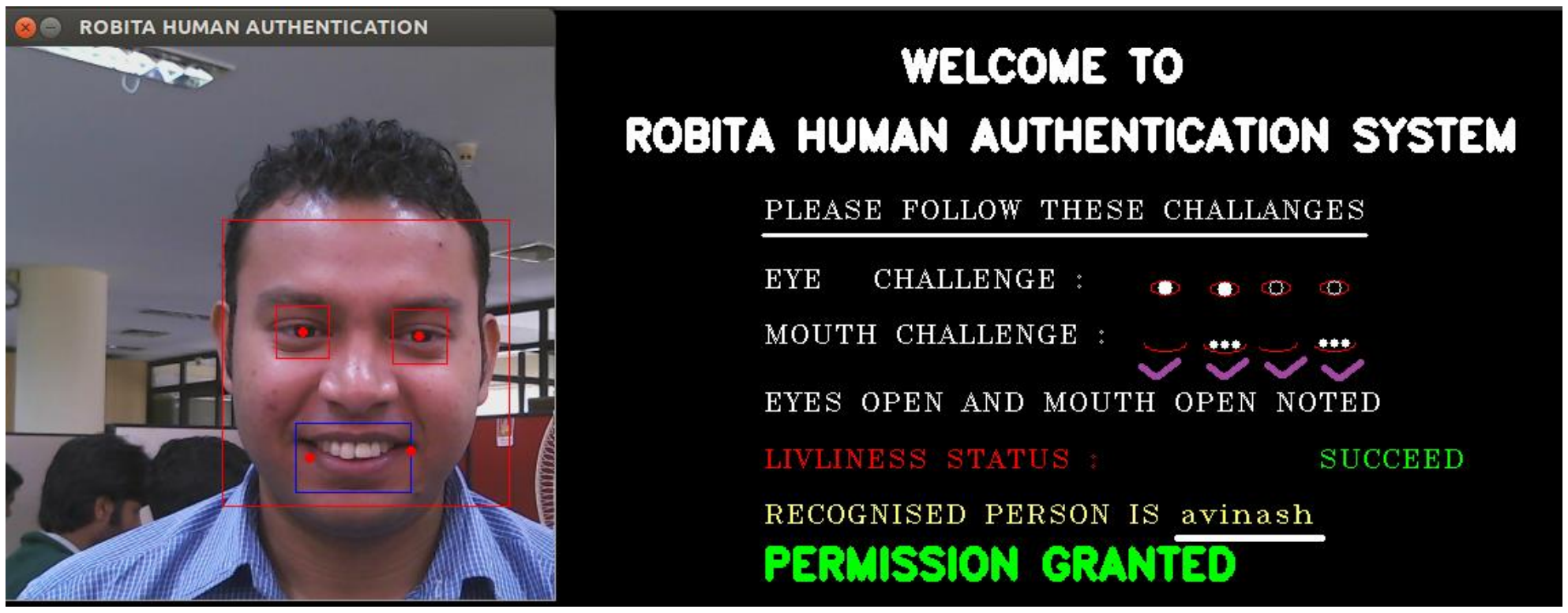

**Figure 6.2.6 A Genuine User Recognized by the System**

### 6.2.3 Results and Discussion

Results that we have got are summarized in Table 6.2.1. It shows how the system behaves when attacks are placed. The report is very convincing. Only one attack bypasses the liveness test, but it is further blocked by verification module. When the attacker forges both the eye and mouth, it will result in misclassification as shown in Figure 6.2.7.

**Table 6.2.1 Summarized Report of System is undergoing these attacks**

| Attack Module | Liveness Detection Module | Verification Module |
|---|---|---|
| Photo Imposter Attack | No | No |
| Eye Imposter Attack | No | No |
| Mouth Imposter Attack | No | No |
| Eye & Mouth Imposter Attack | Yes | No |
| Video Imposter Attack | No | No |

**Why Eye and Mouth Removal Effect the Misclassification more than any other:**

For verifying this concept we have used two assumptions, which support our hypothesis.
**Support 1:** Face recognition literature shows that, in face most important features are of eyes, nose and mouth. They are also named as T features. From the given literature, we can say that these T features have a major contribution to those m features. Therefore the absence of anyone of these can affect the classification.
**Support 2:** The human perception also ensures that these features of the face make the person unique from others. So absence of any of these features could lead to misclassification.

For verifying these statements, we have used face images of the University of Essex, UK (face94) [229] shown in Figure 6.2.7. In our experiment we removed eye, mouth and eye-mouth both region of the face described in Figure 6.2.7 (b) and calculated the efficiency of the PCA on (a) and (b). There are 65 people in the database, both Male and Female. Each person has 15 images; we have used 9 images for training and 6 images for testing (60% training and 40% testing). For removing the eye and mouth region of the face we first detected these regions in the face using the Haar classifier [116], and placed an ellipse, filled up with black color on these regions The black color having bit 0, shows that this feature is absent from the face. The reason for choosing ellipse from the available shapes (circle, rectangle, etc.) is that it provides the optimal region covered by that feature. Misclassifications of these experiments are summarized in Figure 6.2.8, which shows the misclassification ratio when attacks are performed. Misclassification shows that how many times system generated false key.

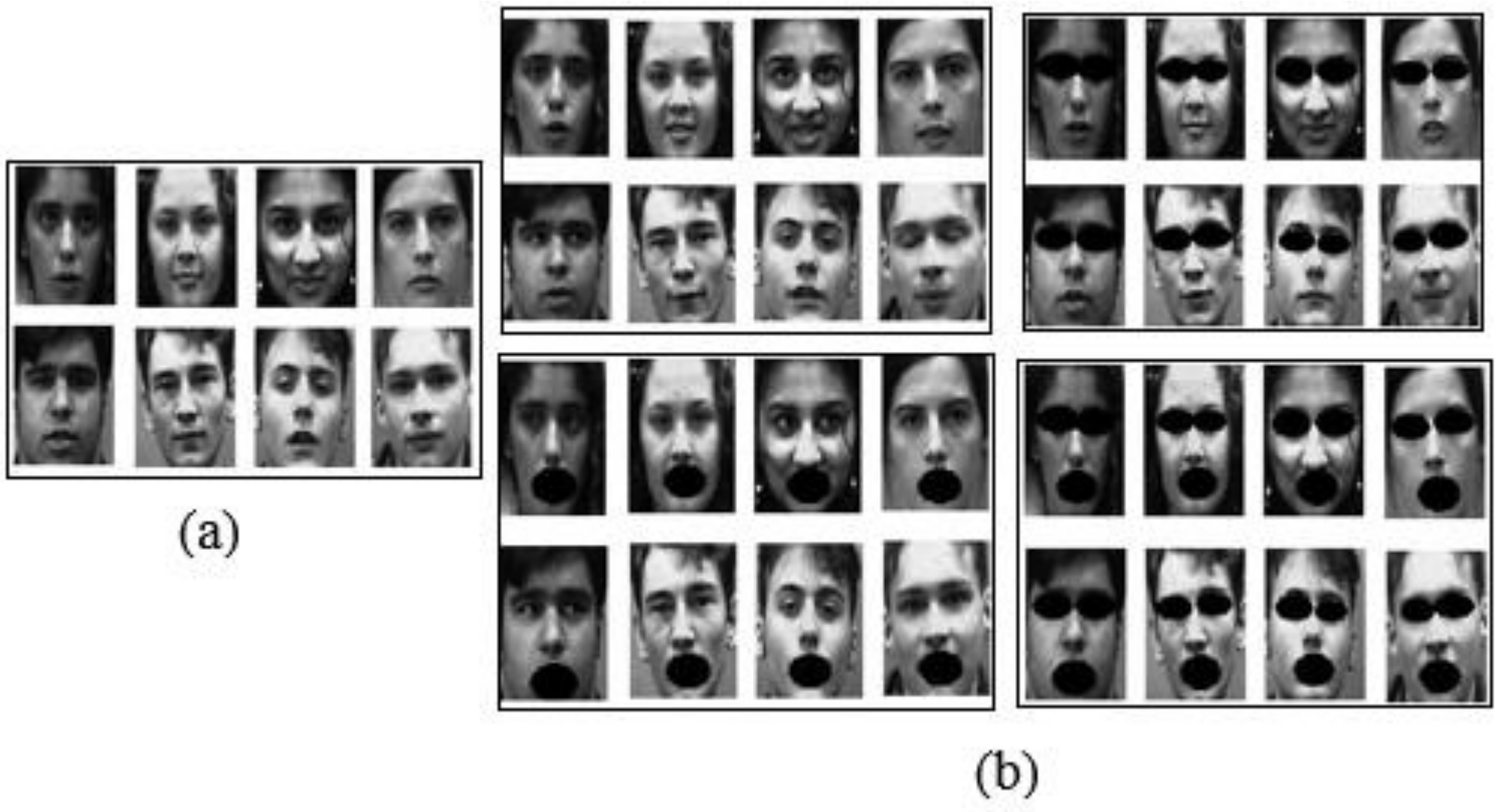

**Figure 6.2.7 (a)Shows Face samples from train (b) First row shows simple test faces which are same as train with changes in facial expression and Eye removal attack, Second Row shows the Mouth and eye and mouth both removal attack**

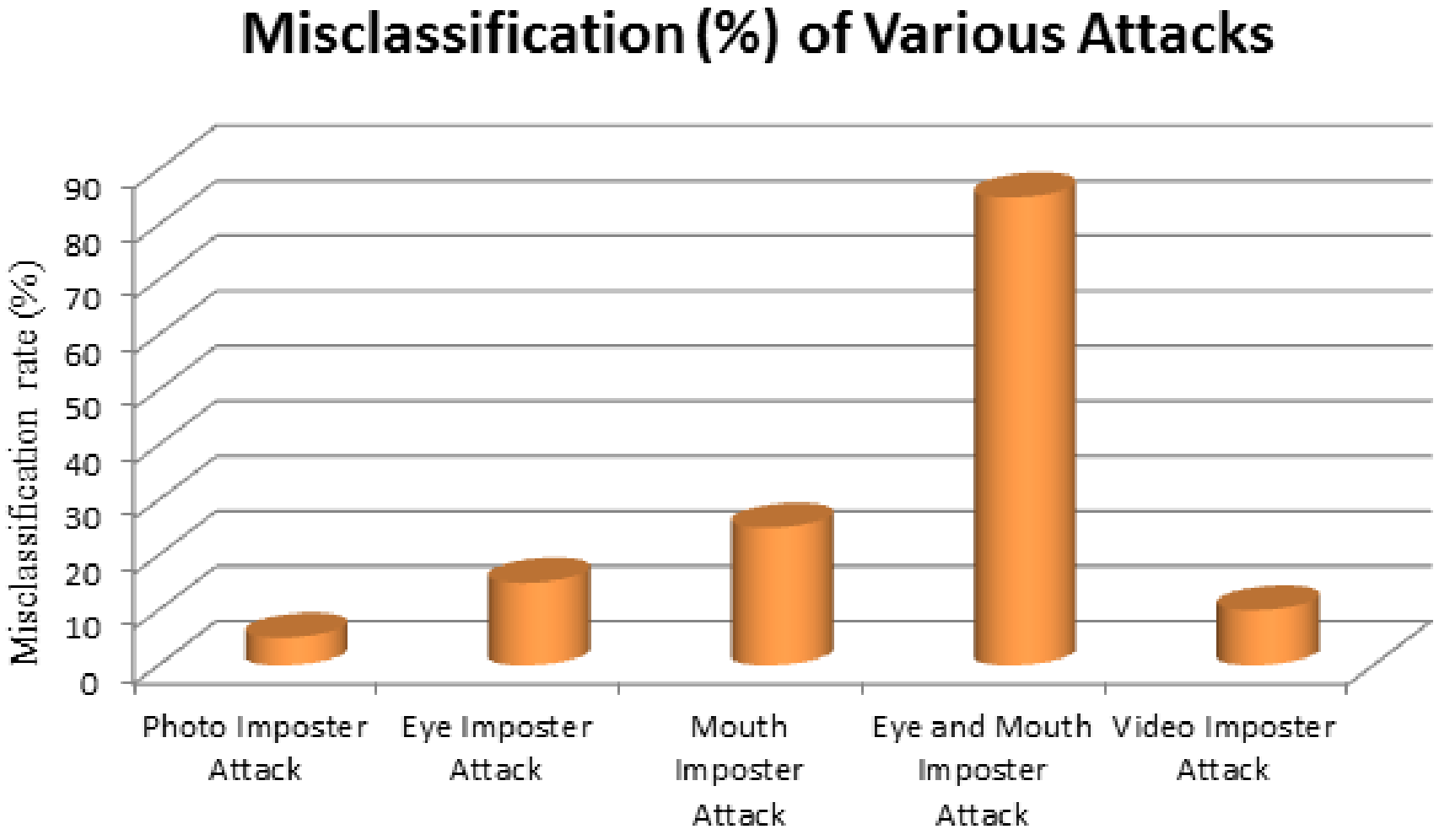

**Figure 6.2.8 Misclassification rate of various imposter Attacks**

## 6.3 Face Liveness Detection through Face Structure Analysis

For human beings, it is very simple to detect that the person is live or not by observing the person’s face structure. It is very simple to make a difference between the two, because humans have 3D perception. We have binocular vision, with the help of which we can generate the 3D structure of any object and hence we can easily analyze about their height, width, and depth. Hence here, we have proposed a liveliness detection scheme based on the structural analysis of the face. Our hypothesis is based on the fact that human face is a 3D object, while the medium used by attacker is a 2D object/2D planner surface. So if somehow we are able to model this difference we can distinguish between the real and the imposter. Therefore, we have set up the stereo setup to capture two images of the same object in order to calculate the depth information of the face. Later these depths (disparity) images are used to train the system by using the machine learning algorithms like Principal Component Analysis (PCA) [87][89][99], Linear Discriminant Analysis (LDA)[90-92] and C-Mean [119] and differentiate between spoof and real. Once the system is trained, it will be able to project the data into their respective classes either in real class or in spoof class. A gradient based 8 neighbour feature extraction technique is also proposed to extract unique features from these disparity images. These 8 neighbours (left eye, forehead, right eye, right cheek, right mouth, chin, left mouth and left cheek**)** are the 8 vibrant features of the face where nose is the center.  Gradient toward each direction is our features; hence a feature vector for a face consists of 8 features

### 6.3.1 Proposed Framework

There are two modules in our framework shown in Figure 6.3.1, one used in generating the disparity map and extracting features from them, while the second consists of machine learning algorithms (PCA, LDA and C-mean) to classify between real and fake faces. The idea of using stereo vision is inspired from the human perception. We humans have binocular vision which basically helps us to perceive the things more accurately. This gives us the third dimension (depth) apart from the height and width which is very helpful to visualize the things in 3D. We are utilizing this third dimension to discriminate between the real and spoof. The setup is briefly described below.

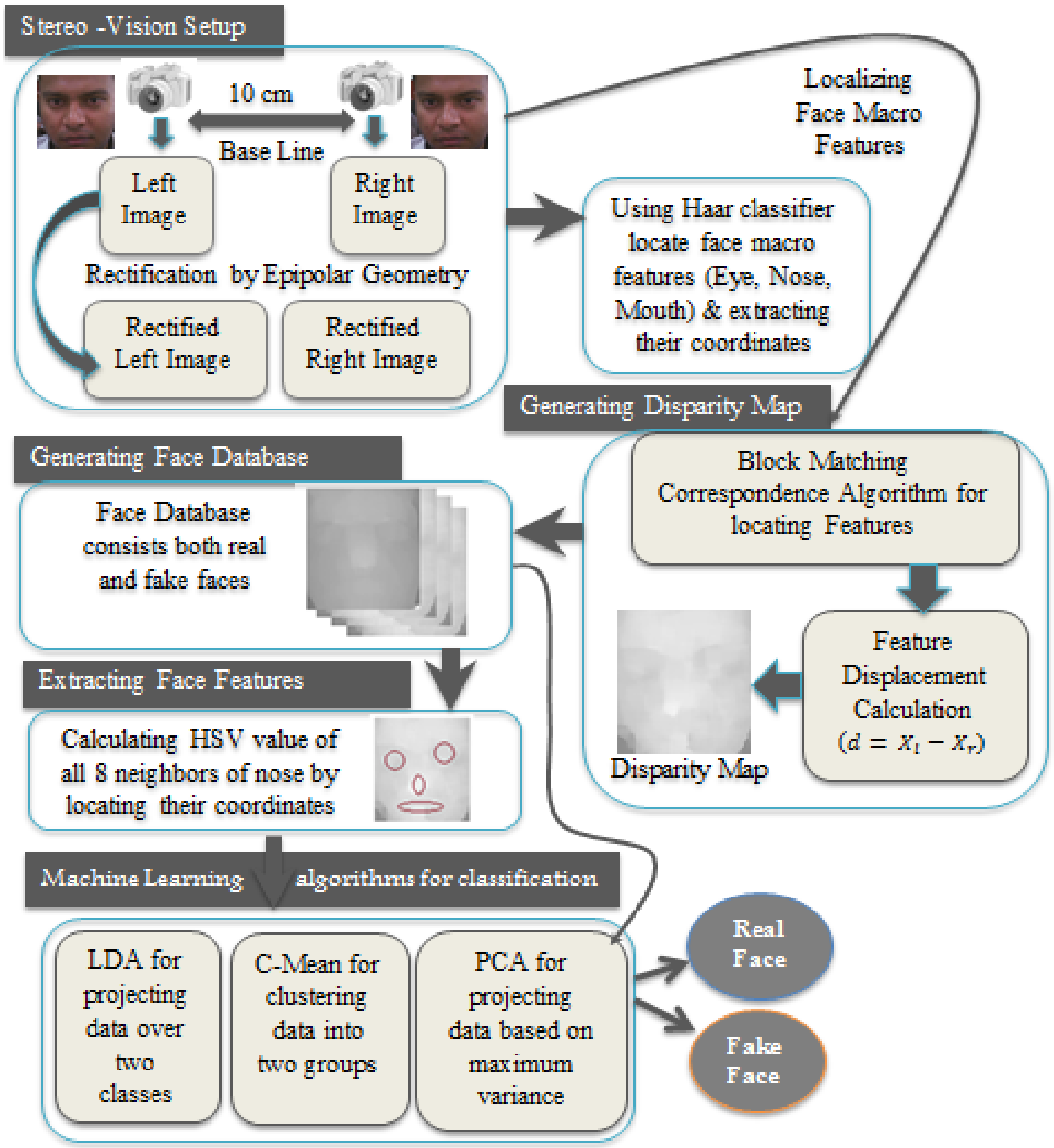

**Figure 6.3.1 Proposed framework for detecting real and fake faces**

### 6.3.1.1 Stereo Vision Setup and Generation of Disparity Database

Stereo vision proposed by Kanade and Okutomi [231] is a way of extracting 3D information from 2D images. They came up with the idea that by comparing information of an object captured from two vantage points and examine the relative positions of objects in both the images (taken by two cameras), we can get 3D information about that object. For analyzing the 3D structure of the face, we are implementing the same idea here. We are using two cameras placed horizontally to each other, capturing two different views of a scene just like human binocular vision. Now for particular object, by comparing it's positioning in both images, we can conclude about disparity which is inversely proportional to the depth of the object.

Our stereo setup shown in Figure 6.3.2 has two cameras with resolution of 1,280 × 720 pixels and having 10 cm baseline distance. It captures 30 frames per second. Its video resolution is 1,600 × 1,200. For more information we can refer to Setup Specification (http://www.amazon.com/Logitech-Webcam-C600-BuiltMicrophone/dp/B002GP7ZT6). For generating the efficient disparity map, we need to know the basic geometry of stereo imaging system briefly described in [232]. This geometry is known as epipolar geometry [233]. We are using two pre-calibrated cameras with popular configuration for our experiment. In order to achieve epipolar configuration we need to rectify stereo images (left and right images) with the help of intrinsic and extrinsic parameters of both the cameras [234]. Epipolar configuration transforms both images so that they will be row aligned to each other. It helps in finding the particular feature in the same row in both the images. Hence, it reduces search time by rejecting many points that could lead to false correspondence. We used block matching algorithm proposed by Konolige [235] to match a particular feature in left to right image. For finding the features sum of absolute difference (SAD) windows is used. By sliding SAD window correspondence is computed.

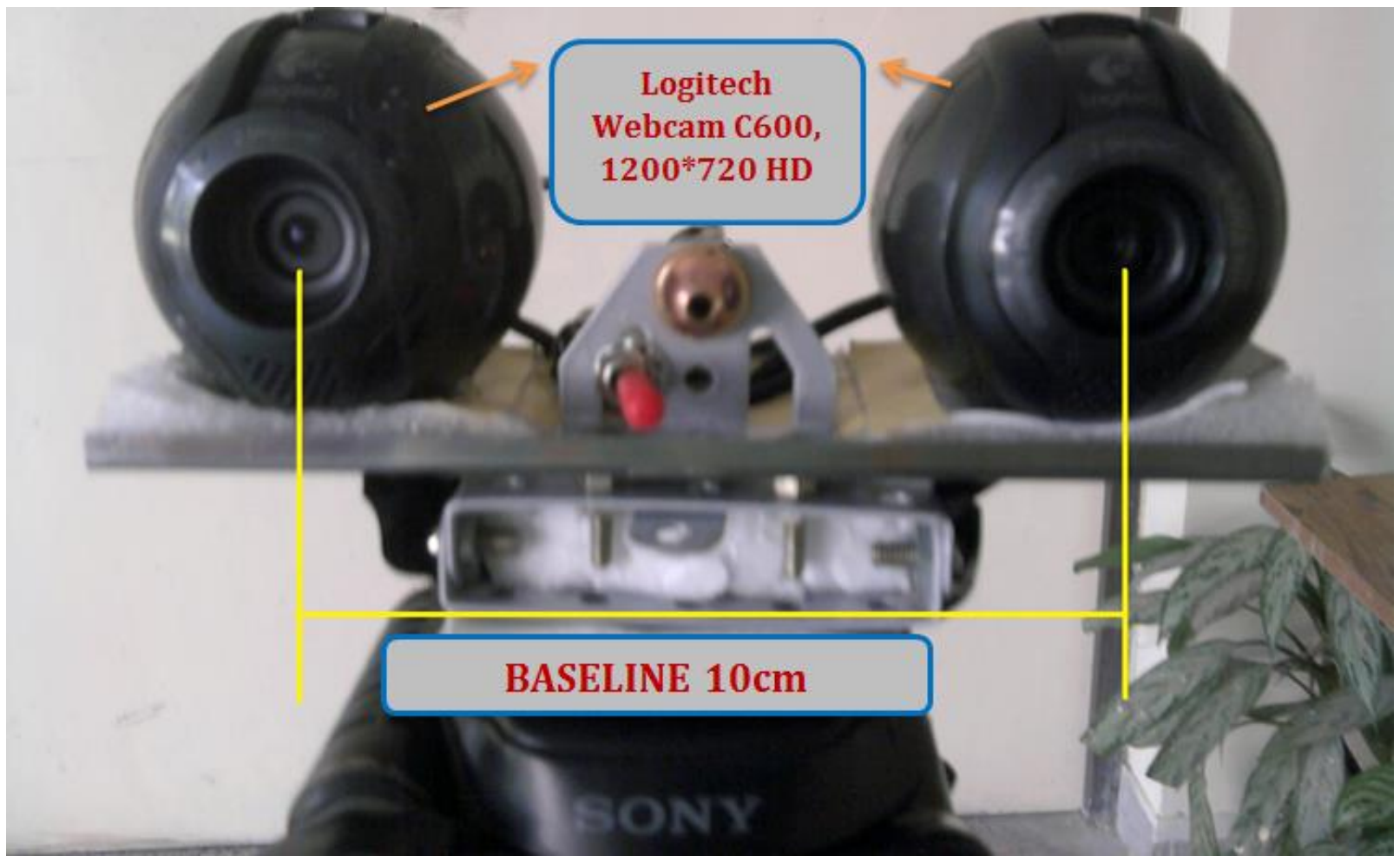

**Figure 6.3.2 Stereo vision setup details**

We are having two images taken from the left and right cameras say $I_l(x,y)$and $I_r(x,y)$respectively. Disparity (d) is computed as the horizontal displacement between corresponding pixels $P_l(X_l, Y_l)$ of the image $I_l(x,y)$ and $P_r(X_r, Y_r)$ of the image$I_r(x,y)$. Therefore, we can write this as: $d = X_l - X_r$. On the basis of this setup we have generated a disparity map of both the classes (real and imposter) shown in Figure 6.3.3 (a) and (b). In Figure 6.3.3 it is clearly visible that when the disparity (d) is more the pixel is more bright and if the disparity (d) is less pixel is relatively dark.

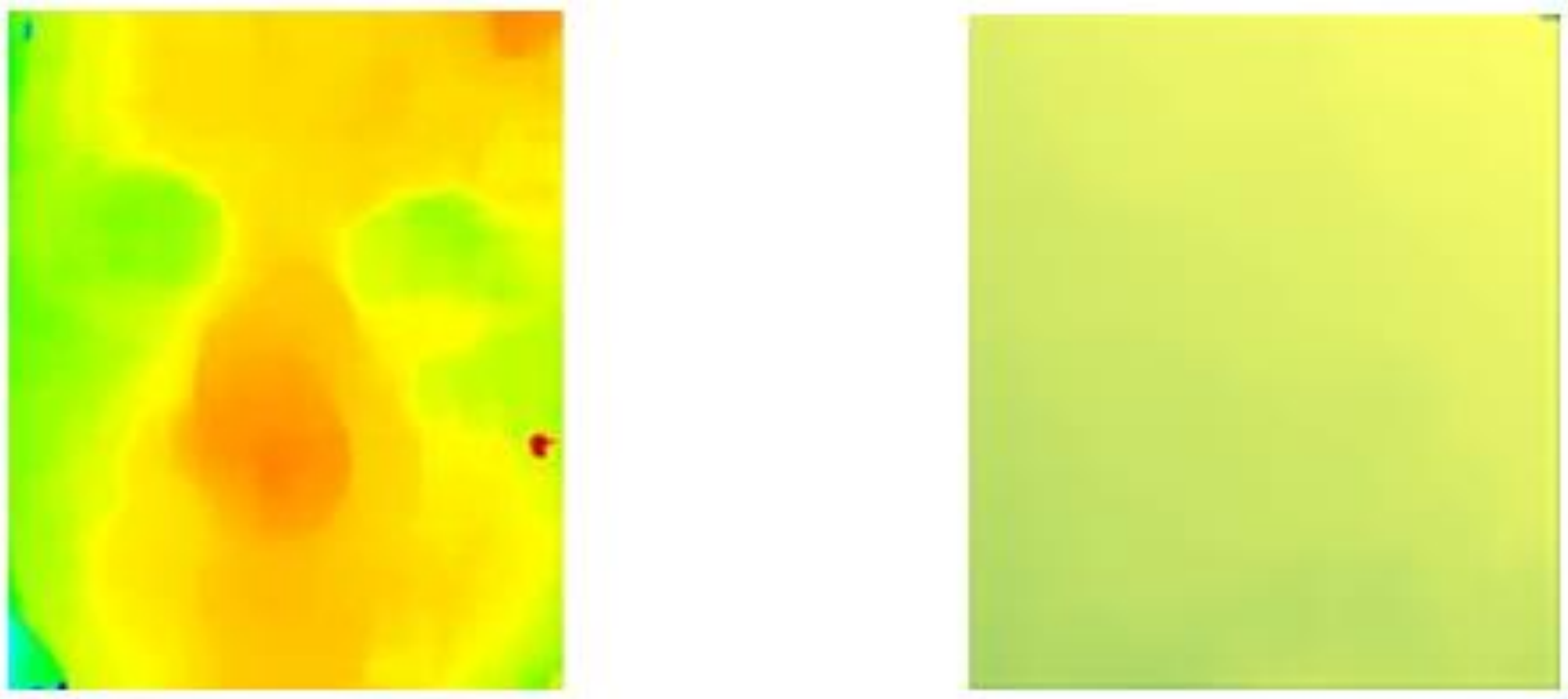

**Figure 6.3.3 Disparity Map: (a) real face pseudo image (b) spoof face pseudo image**

### 6.3.1.2 Generating Face Database

Haar Classifier devised by Viola and Jones [116] used Haar like features to detect the objects. These features used change in contrast values to detect the object rather than using the intensity values of the image. For detecting the faces and facial features, it is required to first train the classifier. Viola and Jones used AdaBoost and Haar feature algorithms to train the system. Some images that do not contain any object (called negative images) and some images having the objects (called positive images) are used to train the classifier. They used 1500 images, taken at different angles varying from zero to forty five degrees from a frontal view to train each facial feature. All the three classifiers that were trained for eye, nose and mouth have shown a good positive hit ratio, which shows the robustness of their approach. Once the face is detected, the rest of the facial features are detected by measuring the probability of the area where they likely exist.

We have extracted faces with the help of Haar classifier and generated disparity map discussed in section A. After generating the disparity map of each image (Real and Fake) we have saved it for future use and named it as face database.

#### 6.3.1.3 Proposed 8 Neighbour based Feature Extraction

If we follow the facial anatomy discussed in [218], we will see that these facial features (eyes, nose, mouth, etc.) have their unique geometrical representation. In all we can say that they have different shape and size. Hence, when we generate the disparity map of face, on each point we get different depth information as shown in Figure 6.3.4 (a). As the medium attacker use is the 2D medium where these features are all in the same level. Therefore, all points have the same depth; resulting no disparity for the spoof faces, clearly visible in Figure 6.3.4 (b).

By analyzing both the disparate images of real and fake user, we can draw one conclusion that the disparity map of fake faces has more homogeneous region than the disparity map of real faces. In order to calculate the homogeneous region in both the images we can calculate the gradient of both the images. Gradient of these images will give us the idea about ups and down in the image which is a clear representation of facial structure shown in Figure 6.3.3 (a) and Figure 6.3.3.

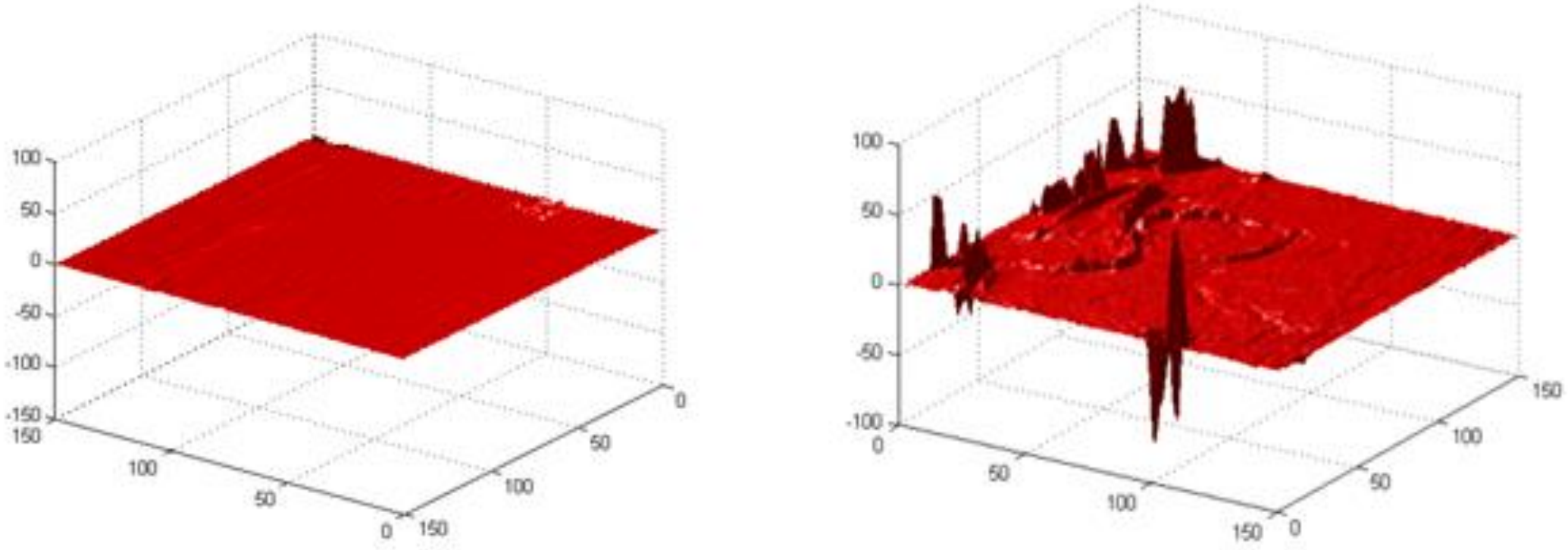

**Figure 6.3.4 (a) Real face gradient and (b) spoof face gradient**

In X and Y axes, we are representing rows and columns of the image and in Z axis magnitude of gradient is represented. High picks displayed in Figure 6.3.4 (b) are the variation in the intensity levels and these are due to the different features present at different level in the face. As we are getting high peaks over these regions we are detecting gradient only in these directions.

Disparity image D could be visualized as a function of x and y, say D(x, y), where$x \in row, y \in column$. Gradient of this image can be calculated by:

$$\nabla f = \begin{bmatrix} \frac{\partial D}{\partial x} \\ \frac{\partial D}{\partial y} \end{bmatrix} \quad (6.1)$$

Here $\frac{\partial D}{\partial x}, and \frac{\partial D}{\partial y}$are the image gradients in x and y direction. We can calculate these as the equation given below.

$$\frac{\partial D}{\partial x} = \frac{D(x+\Delta x)-D(x)}{\Delta x} \quad (6.2)$$

$$\frac{\partial D}{\partial y} = \frac{D(x+\Delta y)-D(y)}{\Delta y} \quad (6.3)$$

Here $\Delta x \ and \ \Delta y$ represent the small displacement in x and y, in our experiment we kept these as 1. While the magnitude ($\nabla f$) towards these directions can be calculated by:

$$(\nabla f) = \frac{\partial D}{\partial z} =[ (\frac{\partial D}{\partial x})^2 + (\frac{\partial D}{\partial y})^2]^{1/2} \quad (6.4)$$

These high picks are nothing but the facial features represented in the form of vertices and edges shown in Figure 6.3.5 (a) and Figure 6.3.5 (b). Each vertex has coordinates x and y. V0 is the center vertex and V1, V2, V3, V4, V5, V6, V7, V8 are its neighbors known as by the left eye, forehead, right eye, right cheek, right mouth, chin, left mouth and left cheek. For each vertex, we are calculating gradient from the center. Hence, for each disparity image we end up with 8 features**.** The algorithm for generating the feature set of each disparity image is mentioned below. Here N is the neighbor vertices (here 8), V (0,0) are the x coordinate of center vertex, V (0,1) is the y coordinate of center vertex. Start_x and start_y are the starting while end_x and end_y are the ending point which basically give you the path to calculate the gradient.

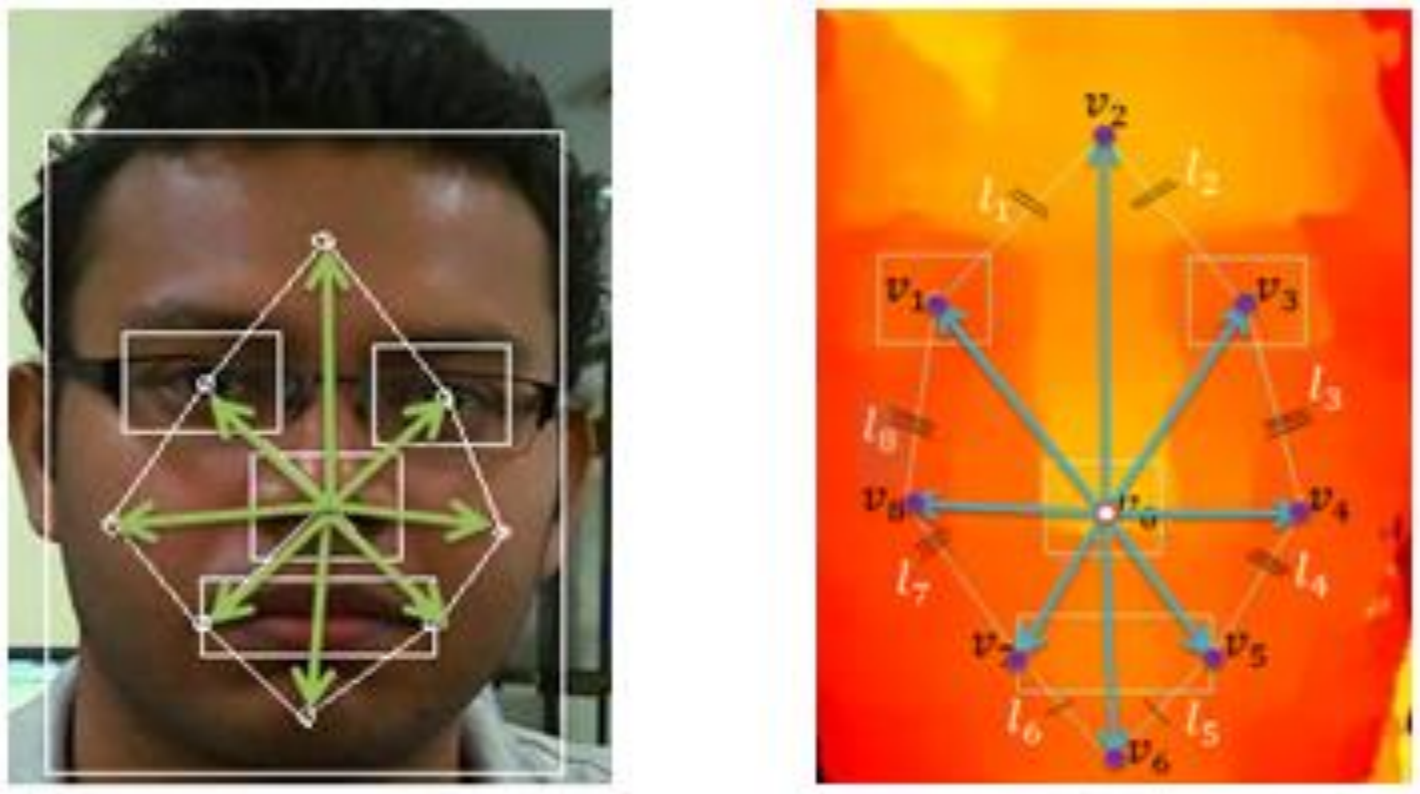

**Figure 6.3.5 (a) Localization of facial features (b) graphical**

---

**Algorithm:** Pseudo code for neighbours based feature extraction technique

---

**Input:**

1. *Disparity image of size m*n* ($Disp_img_{m*n}$)
2. *Set of feature vertices V* ($V=[V_0, V_1, V_2, \ldots \ldots, V_N)$)

*//* $V_0$ *is the center vertex and for each vertices* $V_i \in V \exists$ *coordinate set (x,y).*

**Output**: Total neighbour gradient (TNG_V) for each feature vertex

**Begin:**

$i \leftarrow 1$ *;*

**Step1:** *while i<= N*

*do*

**Step2:** *if (V(0,0)<=V(i,0))*

*then*

*start_x*← $V(0,0)$ *;*

*end_x*← $V(i, 0)$ *;*

*endif*

**Step3:** *if (V(0,1)<=V(i,1))*

*then*

*start_y*← $V(0,1)$ *;*

*end_y*$V(i, 1)$ *;*

---

*endif*

*total_gradient*← 0 *;*

**Step4**: *for ( j*← $star_xtoend_x$*)*

*sum_dz*← 0 *;*

**Step5:** *for ( k* ← $start_y toend_y$)

$d_x$ ← *Disp_img( j,k ) – Disp_img( j+1, k ) ;*

$d_y$ ← *Disp_img( j,k ) – Disp_img( j,k+1 ) ;*

$d_z \leftarrow ( {d_x}^2 + {d_y}^2)^{1/2}$ *;*

*sum_dz*← *sum_dz* + $d_z$ *;*

*endfor*

*total_gradient*← *total_gradient + sum_dz ;*

*endfor*

*TNG_V( i )* ←*total_gradient ;*

*end while*

***End:***

### 6.3.1.4 Time Complexity Analysis

The proposed algorithm consists of 5 major steps; hence in order to calculate the overall complexity of the algorithm we can divide it into five different modules. By summing and finding the most dominated term in the expression, we can decide the total time complexity. Table 6.3.1 shows the steps performed for complexity analysis: Here pq is the polynomial of degree 2 which is the most dominating term. Therefore the time complexity of the algorithm is O(pq) where p and q are the subset of the face image.

**Table 6.3.1 Time Complexity Analysis of the Algorithm**

| Steps | Running Time |
|---|---|
| Step 1: | $8 * T_1$ // 8 times |
| Step2: | $8 * T_2$ // Maximum 8 times (T2<=8) |
| Step3: | $8 * T_3$ // Maximum 8 times (T3<=8) |
| Step4: | $T_4 * 8$*(p+1) // distance in x coordinates |
| Step5: | $T_5$*8*$\{(p + 1) * (q + 1)\}$ // distance in x and y both |
| Summation: | $= 8T_1+8T_2+8T_3+(p + 1)*8T_4+T_5 * 8 * ((p + 1)(q + 1))$ |
| | $= 8T_1+8T_2+8T_3+(p + 1)* 8T_4+8T_5(pq + p + q + 1)$ |
| | $= 8T_1+8T_2+8T_3+ 8pT_4 + 8T_4+ 8pq T_5+8 pT_5+ qT_5+T_5$ |
| | = 8pq $T_5$ // dominating term (polynomial degree 2) |
| | = O(pq) |

### 6.3.2 Classification using Machine Learning Algorithm

Machine Learning is a way of making machine intelligent so that it can take decision like as we human do. As we can distinguish between the fake and real, we are trying here to implement same level of intelligence by using some machine learning algorithms.

#### 6.3.2.1 Linear Discriminant Analysis (LDA)

The aim of LDA is to find the best projection vector/matrix so that data would be linearly separable [90-92]. It can be represented by $Y = W^T X$. Where Y is the resultant direction, X is the input and $W^T$ is the transformation matrix. This $W^T$ (Transformation matrix) is based on two factors, within a class scatter matrix and between class scatter matrix, denoted by SW and SB. Within class scatter matrix (SW) is shown how the data is distributed along its mean within a class while the between class scatter matrix (SB) represents about how data of one class is distinguished from the other class. In the objective function J = $SW^{-1} * SB$ , suggested by Fisher, we aim to minimize the within class scatter matrix and try to maximize the between class scatter matrix. The algorithm is discussed below.

A. ***Creating Class Labels (C1, and C2)***

We are having 90 samples from the fake faces and 30 samples from the real faces. Fake faces are represented by class C1, while real faces are represented by class C2. Each sample

(X) is represented by 8 feature vectors. We have represented these feature vectors as a column vector say $(X_{1,1})_{8*1}$, where $X_{11}$ represents sample number first belongs to first class (C1). As the C1 class is having 90 samples we can represent it by combing all samples together say; $,(X_{1,1,}X_{1,2,}X_{1,3},X_{1,4,\dots,}X_{1,N})$ here N=90. Similarly, we can combine all samples of second class C2 say; $(X_{2,1,}X_{2,2,}X_{2,3},X_{2,4,\dots,}X_{2,M})$ here M=30. At the end, these classes have population$C1_{8*90}$ and $C2_{8*30}$. We have divided these samples over training and testing having the ratio 90% and 10% respectively.

### B. *Calculating within a class scatter matrix (SW):*

We can calculate the within class scatter matrix by:

$$SW = S_1 + S_2 \qquad (6.5)$$

Where $S_1$ is the within class scatter matrix of C1, and $S_2$ is the within class scatter matrix of C2.

$$S_1 = \sum_{i=1}^{N}(X_{1i} - \mu_1) * (X_{1i} - \mu_1)^t \qquad (6.6)$$

$$S_2 = \sum_{i=1}^{M}(X_{2i} - \mu_2) * (X_{2i} - \mu_2)^t \qquad (6.7)$$

Where n is the number of samples belonging to C1, $X_i$denotes each sample in C1 class, $\mu_1$is the mean of C1 class, which can be calculated by

$$\mu_1 = \frac{1}{N}\sum_{i=1}^{N} X_{1i} \qquad (6.8)$$

$$\mu_2 = \frac{1}{M}\sum_{i=1}^{M} X_{2i} \qquad (6.9)$$

By putting the value of equation (6.6) and equation (6.7), SW will be obtained

### *C. Calculating between class scatter matrixes (SB):*

SB is measured by calculating the difference between the mean of each class. Therefore SB will be: $SB = (\mu_1 - \mu_2) * (\mu_1 - \mu_2)^T$ (6.10)

### *D. Calculating the Transformation Matrix (W):*

On the basis of SW and SB we can calculate W, which is the combination of w's (W= $[w_1, w_2, w_3, \dots, w_k]$), or we can select any direction from these. Transformation matrix (W) is calculated as:

$$\frac{SB}{SW} * w = \lambda w \quad (6.11)$$

$$\text{OR} \quad (SW^{-1} * SB) * w = \lambda w \quad (6.12)$$

Where w is the eigenvector and (is the eigenvalues, these w's shows the direction and (determines which direction is significant, hence on the basis of (we can select w's. k is the number of direction that we have selected. We can calculate all possible characteristic root ($\lambda$) of $(SW^{-1} * SB)$ by using the equation$[(SW^{-1} * SB) - \lambda I] = 0$. Where I is the identity matrix.

***E. Decision Generation:***

With the help of transformation matrix W, we can project, both the training classes and testing samples.

$Y1 = W^T * C1$ // Y1 is the projection of first class

$Y2 = W^T * C2$ // Y2 is the projection of the second class

$Z = W^T * \text{Test Samples}$// projection of test samples

Now by calculating the Euclidean distance between test and train we will get two distances as:

$$d1 = \|Z_i, Y1_j\| \text{ and } d2 = \|Z_i, Y2_k\|$$

Where, $i \in 1,2,\dots P, P+1, P+2,..Q$ ( P: number of testing samples $\in$ C1, Q: number of testing samples $\in$ C2), $j \in 1,2, \dots$ N-9 (samples belong to class C1 who are not participating in training) and $k \in 1,2,\dots$ M-3 (samples belongs to class C1 who are not participating in training)

If ((d1>=d2) &&( $i \in [1,P]$)) Then $i \in C1$

elseIf ((d2>=d1) &&( $i \in [P+1, Q]$)) Then $i \in C2$

### 6.3.2.2 K-Means

C- Mean is proposed by J.MACQUEEN in 1967 beautifully discussed in [236], is a clustering mechanism used to partition n observation into C clusters. Here the numbers of clusters are two, one for real faces and second for fake faces. The algorithm is discussed below with the abbreviation used (follow). The output of C-Means algorithm provides us two updated means (centers) of class C1 and C2 along which data is maximum distributed.

| Algorithm: Pseudo code for C Means Algorithm, where C=2 |
|---|
| **Input:** *We have given S={ x1,x2,x3,………, Xn)* $\in R^m$<br>**Output:** *Total neighbour gradient (TNG_V) for each feature vertex*<br>**Begin:**<br>**Step1:** *Choose c points y11,y12,y13,… y1C* $\in R^m$<br>**Step2:** $A_{2i} = \{ x \in S: d(\underline{x}, y_{1i}) <= d(\underline{x}, y_{1j}) \ \forall j \neq i \}$<br>**Step3:** $y_{2i}$ *= Mean* ($A_{2i}$)*, i=1,2,3,… c*<br>**Step4:** *if* $\|y_{1i} - y_{2i}\| <= \varepsilon$<br>*then*<br>*STOP with the output* $A_{2i}$<br>*else*<br>$A_{2i} = \emptyset$<br>$y_{1i} = y_{2i}$<br>*GO TO Step2*<br>*endif*<br>***End:*** |

**Table 6.3.2 Abbreviations used in the Algorithms**

| Symbol | Meaning |
|---|---|
| S | Denotes the Set of overall samples including fake and real faces |
| $x_1, x_2, x_3, \ldots x_N$ | Individual Samples |
| N | Number of Samples |
| $y_1$ | Seed Points (Mean) of Fake Faces (First Class) |
| $y_2$ | Seed Points (Mean) of Real Faces (Second Class) |
| C | Number of seed points, here C=2 |
| A2 | Partition For Real faces |
| A1 | Partition For Fake Faces |
| $\|\| \|\|$, d(, ) | Euclidean norm |
| ∅ | Null Set |
| ε | Threshold, here ε=0 |
| $R^m$ | Real Space (M dimensional) |

#### 6.3.2.3 Principal Component Analysis (PCA)

Disparity images of each sample can be represented as a matrix $I(x, y)$ having $m * n$ dimension, where $x \in m$, and $y \in n$. For simplicity, we can also use this as a row or column vector like I'(z) having $mn * 1$ dimension, where $z \in mn$. In disparity image each pixel is a feature and hence we have mn features, which is very hard to process due to time and space complexity of the program. Therefore, in PCA we aim to find out what are those important features on behalf of which we can processed for the classification and where the variance is more [87-89]. In other words, we find out those k features, such as $k <= mn$ which are useful for representing others (linear combination). These k features are basically the principal components (Eigen value and Eigen vector). In our experiment we have fixed k as 10 (top 10 principal components which have maximum values than others). For finding out the best features several steps have been carried out which are discussed in chapter 5, section 5.3.2. In order to perform the testing we have performed several steps which are listed below.

A. Let us have a test image $(\alpha)_{mn*1}$
B. Calculate mean aligned image $(\beta)_{mn*1} = \sum_{z=1}^{mn}(\alpha_z - \mu_z)$
C. Calculate projected test images $(\gamma)_{k*1} = ((\text{Eigenfaces})^t)_{k*mn} * (\beta)_{mn*1}$
D. Calculate the distance (Euclidean distance) between γ and Ψ, and whatever the minimum distance that will consider as the recognized identity.

### 6.3.3 Result Analysis

When we applied these three machine learning algorithms to train the system, we ended up with good results. We have used 90 fake faces and 30 real face data in order to train and test the system. Laptop (LCD 14" inches) and photograph (on photo paper) are used as the spoof medium. We applied Principal Component Analysis (PCA) on raw disparity images of both the classes, while Linear Discriminant Analysis (LDA) and C-Means algorithms are used on extracted 8 neighboring features. The maximum misclassification achieved on these techniques, their training and test set proportion all are shown in

Table 6.3.3, while the accuracy of these algorithms is presented in Figure 6.3.6. The maximum misclassification reported by the PCA is 8.4%, and for LDA and C-Means it is 2.5% and 1.7%. To ensure the effectiveness of our feature extraction techniques we did 10 fold cross validation analysis on both the PCA and LDA. We have also performed a spreadsheet analysis of the classification results briefly summarized in

Table 6.3.4. The results were surprising in case of LDA, we encounter only 2.5% misclassification. For every k we have 12 test samples and 108 train samples. Out of 12 test samples, 9 belong to fake and 3 belong to real users. In the same way, out of 108 train samples 81 belong to fake and 27 belong to real users.

**Table 6.3.3 Misclassification results of PCA, LDA and C-Means**

| Algorithm used | Maximum Misclassification | Training Set | Testing Set | Input Data Format |
|---|---|---|---|---|
| PCA | 8.4% | 90% | 10% | Raw Disparity Images |
| LDA | 2.5% | 90% | 10% | Extracted 8 Neighbouring features |
| C-Means | 1.7% | 0% | 100% | Extracted 8 Neighbouring features |

**Table 6.3.4 Spreadsheet Analysis of the classification results**

| | **PCA** | **LDA** | **C-Means** |
|---|---|---|---|
| Classified | 91.6% (110 images) | 97.5% (117 images) | 98.3% (118 images) |
| Misclassified | 8.4% (10 images) | 2.5% (3 images) | 1.7% (2 images) |
| Misclassified | 10 images | 3 images | 2 images |
| Misclassified | 7 Fake 3 Real | 0 Fake 3 Real | 0 Fake 2 Real |
| False Acceptance Rate | 7/90=.078 or (7.8%) | 0/90=0 or (0%) | 0/90=0 or (0%) |
| False Rejection Rate | 3/30=.1 or (10%) | 3/30=.1 or (10%) | 2/30=.067 or (6.67%) |

We have also compared our approach with Lagorio et al [237] which is summarized in point selection plays an important role. The conversion rate of the algorithm depends on these seed points.

Table 6.3.5. In their approach they have used an optoelectronic stereo system (VECTRA 3D CRT) in order to capture the 3D shape of objects (faces). The setup they used is very costly around 450$ per print (for one face scan), while our setup is too chip briefly described in previous sections. Apart from this, their setup is also too big, having two high resolution cameras for texture acquisition, two projectors and four frontal calibrated cameras. The environment is highly structured. Only 1000 watt halogen lamps associated with the setup are acting for the homogeneous distribution of light; no other light source is needed. They have used PCA in order to calculate the curvature value at every point of the 3D scan. For every point p∈mn, where mn is the size of face scan, curvature value C is calculated by: $\mathbf{C}=\frac{(\boldsymbol{p}-\boldsymbol{b})*\boldsymbol{v}}{\boldsymbol{d}^{2}}$ .Here, b is the bary center of the coordinates belongs to the spherical neighborhood $\Omega_r$of radius r at center p. v is the eigenvector corresponding to the smallest eigenvalue, and d is the mean distance of center to all points within the radius r. There are some parameters which required proper tuning, like here it is radius r in case of Logorio et. al [237] and in our case it is the Projection direction in case of LDA and initial seed points in case of C-Means algorithm. Projection direction is used to transform/project data's over the direction/s where they are linearly separable. At $w_3$ (third direction) they are mostly separable while in other direction they are not much. In case of C-Means seed

**Table 6.3.5 Comparison details of the proposed algorithm with Logorio et. al**

| Author | Technique used | Dataset used | Training Set | Testing Set | FRR | Variable Parameter |
|---|---|---|---|---|---|---|
| Logorio et. al | PCA based curvature determination | Only 70 (Real Face Scans) | - | - | 0% | Radius r=6 |
| Logorio et. al | PCA based curvature determination | 70 Real Face Scans + 70 Fake Faces | - | - | 10% | Radius r=4 to 20 |
| Present Study | 8 Neighbour based feature extraction with LDA | 90 Fake + 30 Real Frontal Faces | 90% | 10% | 10% | Projection direction w =($w_3$) |
| Present Study | 8 Neighbour based feature extraction with C-Means | 90 Fake + 30 Real Frontal Faces | 90% | 10% | 6.7 % | Seed point selection $Y1 = S_{13}$ and $Y2 = S_{97}$ |

Apart from these if we consider the time complexity of both the algorithm Logorio et al have time complexity $O(n * m)$ , where $n \in$ width and $m \in$ height of the image. They are computing face curvature at every point of the image while in our approach we are calculating the gradient only for the 8 neighbours of the nose. The order of the time complexity is same such as $O(p * q)$, where p and q are the small subset ($p \subset n$ and $q \subset m$) of the overall image. Therefore, its execution time is always less than the existing approach.

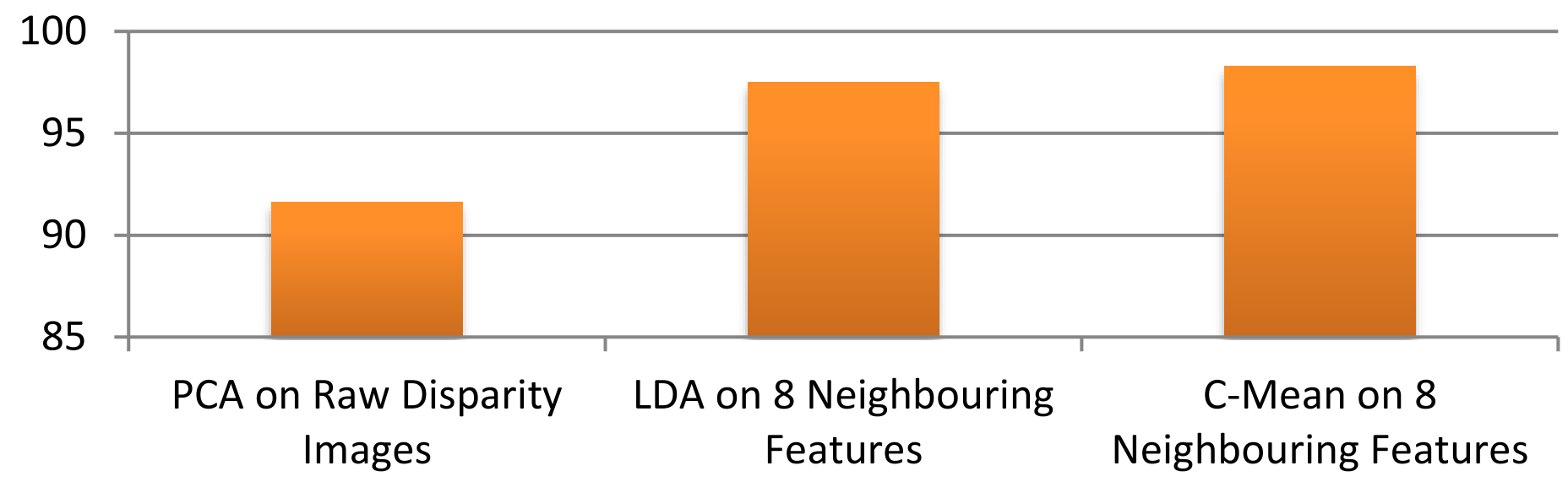

**Figure 6.3.6 Classification accuracy of different Machine Learning Algorithm**

**Discussions:**

In case of LDA and C-Mean Algorithms we are getting promising results. This is because after extracting the relevant features from both the classes we have less non-linearity. This was further removed by the transformations used in LDA and by selecting the best centers along which the data are highly distributed in case of the C - Mean. When we applied PCA directly on raw disparity images we got misclassifications because in the real domain they are non-linear. This non-linearity is due to various kinds of noise associated with the data viz. calibration noise, environmental noise, illumination and it could be the motion noise. Due to these noises when we applied block matching algorithm to these images we got black holes. These holes are created when the features present in the left image is not present in the right image. Apart from this, if we go far from the setup (Camera) it will not be able to create good disparity, because if the distance is more we will not be able to extract face 3D structure clear from the captured images. If the attacker has distorted the photo or wore it over the face, wrinkled it, then it could create disparity at some places on the overall disparity map. Hence, due to these issues, we will get non-linearity between these two classes, resulting in some misclassification. Therefore, we are extracting features by using 8 neighbouring techniques, because if we follow the face anatomy as discussed by Peter M. Prendergast [238] and Boris Bentsianov et al [239] shown in Figure 6.3.7. The human face is structured in such a way that all macro features of the face are arranged in a proportion; hence we will get the clear disparity in the same proportion as features are structured. While in fake faces due to the surface, it has 2D structure, hence the disparity is very vague, almost similar everywhere shown in Figure 6.3.3 (a) and Figure 6.3.3 (b).

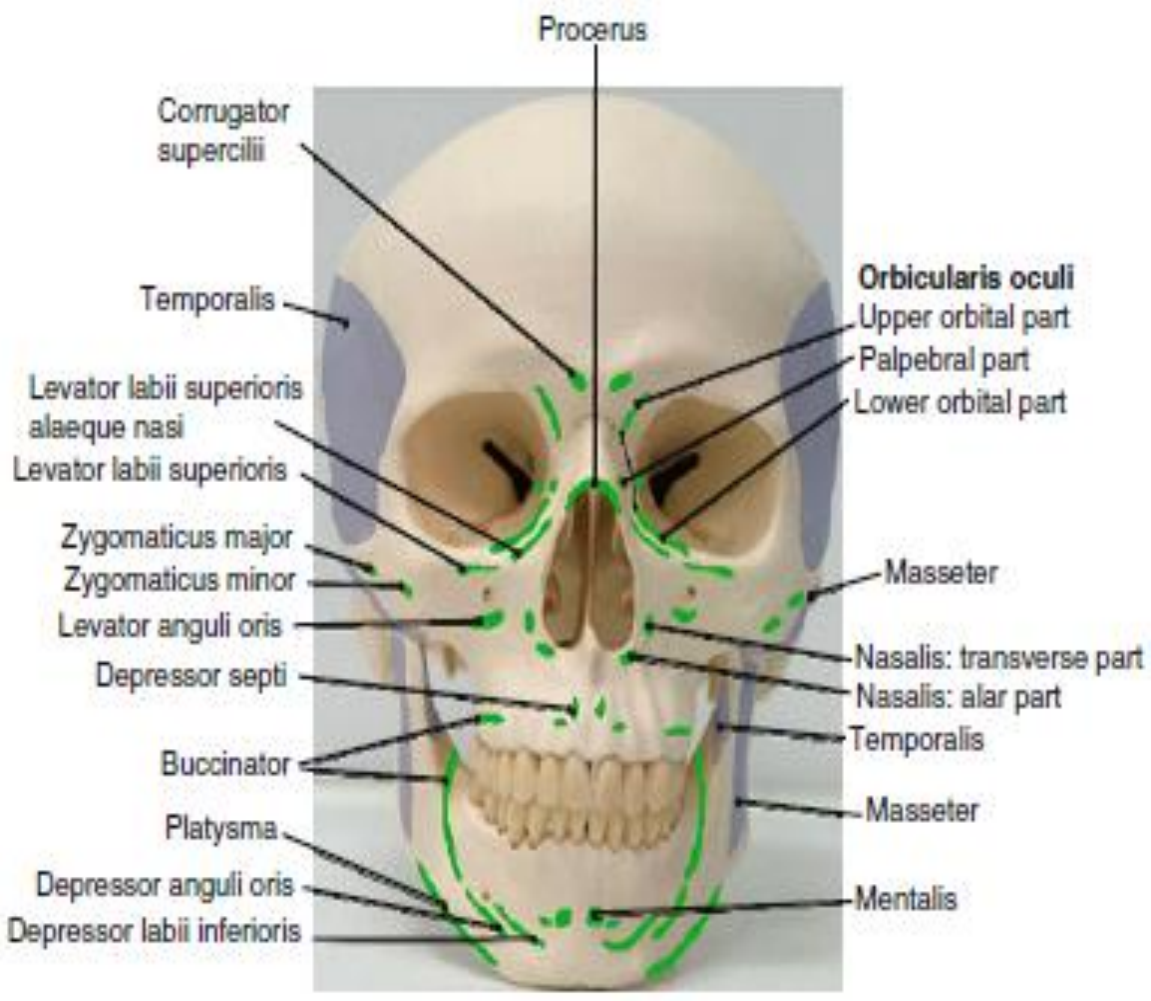

**Figure 6.3.7 Areas of muscle attachment to the facial Skelton [238]**

## 6.4 Conclusion

Liveness detection is necessary because it ensures whether the person is physically present or not. As no two persons can have the same biometric traits, the system will identify that a genuine user is presenting their biometrics. Experimentally it is proven that, to bypass the liveness test, the attacker has to remove eye and mouth region of the photograph and has to place his eye and mouth instead. These attempts bring about drastic changes in the structure of genuine user's face, hence resulting in misclassification. Eye and mouth movements are identified by using the Haar classifier, Eye openness and closeness in a definite time interval shows the eye liveness while the teeth HSV value estimation shows the movement in the mouth region. The HSV value of the teeth is calculated by estimating the mouth Region of Interest (ROI). We have used a challenge and response method here to test the liveness while Principal Component Analysis (PCA) has been used effectively for recognizing the person. PCA is a good tool for dimensionality reduction; therefore we used this to prove our hypothesis. We have tested our approach on 40 people each having 20 images. The system has shown a good accuracy ratio and has successfully identified all the trained persons.

Analysis of the facial anatomical structure to detect liveness is a new technique which can easily discriminate real faces from the imposters. This technique gives us a depth perception to visualize the problem in such a way that these two classes can be linearly separable. We have extracted 3D face structure by calculating the disparity map of both the classes. Stereo setup has been used in this research work for disparity map creation. The disparity map of face images is affected by several factors such as illumination, noise, calibration noise, motion disturbance and a variety of distances from the camera. It becomes the problem more non-linear with the distorted images of the attackers. In order to solve these challenges, we have applied machine learning techniques such as PCA, LDA, C-means to tackle this problem in a linearized way. Initially we have applied PCA directly on the disparate images and obtained FAR of 7.78% and FRR of 10%. In order to minimize the FAR and FRR an attempt has been investigated with our proposed feature extraction techniques applied to face disparity images. This proposed eight neighbour technique provides extremely important features based on the face geometrical structure. We have defined these eight gradients based features from the centroid (nose) of the disparity map images to eight different directions (left eye, forehead, right eye, right cheek, right mouth, chin, left mouth and left cheek). LDA and C-means are applied on these features to validate the prominence in classification purposes. We have achieved encouraging amount of classification results of 97.5% and 98.3% using LDA and C-means. There is no FAR reported for LDA as well as C-Means, while the FRR is reduced up to 10% and 6.67%. This proposed approach has a time complexity of O $(p * q)$ where p and q are the small subset of the overall image. Hence, running time of the algorithm is also much less. The proposed system is being tested on the photo imposter attack, video imposter attack and the medium used was photo paper and laptop. The system works well and classified all the attacks. The future work is intended to test this setup and the proposed feature extraction technique on 3D mask attack.

# Chapter 7:

# Conclusion and Recommendations

## 7.1 Conclusion of the thesis

In this thesis, a novel human perception based criminal identification framework has been proposed which acts as a supplementary of the mugshot detection approach. The proposed approach enables humanoid robots to help police in arresting the criminals. Due to their aesthetic appeal, these humanoid robots are able to make a friendly connection with the eyewitness during the interrogation. The earlier mechanism of criminal identification involve sketch creation of the criminal by the sketch artist and then releasing that sketch on the television, newspaper or any other public media. Later the computerized programs were used to perform the sketch to photo matching in order to find its previous criminal record. The matching between the sketch to photo is a nonlinear process (non linearity due to the boarder and shading of sketches) and requires a large processing, while the time complexity is O(n)*t, where n is the number of criminal records and "t" is the time taken to match with the single face image. The proposed criminal identification framework captures the same knowledge which is used in sketch creation with the help of the dialogue based system. In the proposed system a humanoid robot asks questions about the facial description of the criminal such as eye type (deep set, monolid, Hooded, etc.), face type (oblong, oval, rectangle, etc.), face colour etc. The acquired description is then transformed to get the crisp value of each feature type. The query based and rough set based approaches are then used to find the correct match. Total 10 feature (3 anthropomorphic and 7 facial features) have been used to represent each criminal information. The experiment conducted on the benchmark mugshot dataset of Chinese University of Hong Kong of 360 people showed efficiency of the proposed approach over the existing approaches. Although the accuracy of the proposed system is less than the previous approaches, but its recognition time is very less. Moreover the proposed system is supplicatory of the mugshot detection approach not supplant. It could be placed before the mugshot detection module, hence can increase the speed of criminal tracking more efficiently.

To make the robot self-reliant for criminal identification, we have further trained them with the sketch drawing capability. We have setup the prototype of each facial component, as soon as the eyewitness gave the description about any facial attribute, Robot picked that shape from the feature attribute gallery. In a face each attribute has particular and fixed location, by placing these attributes at their respective location, the humanoid robot virtually creates the face in its camera plane. A calibration between the robot camera plane and its body coordinates, enables it to perceive that virtual image in terms of its configuration space. This transformation represents each pixel of the face image in the form of robot end effector location (X,Y,Z). Further, the presented gradient descent based inverse kinematic helps it to reach those points and to perform drawing. The proposed framework is generalized and it can be applied to any humanoid robot. The computation complexity of the gradient descent algorithm is high, hence we have also defined the closed form inverse kinematic solution for the humanoid robot.

The above two contributions enriched humanoid robots to identify criminal based on the inference gained from the eyewitness, but it could not trace the criminal/person by just seeing them. Therefore, a facial symmetry and component based face recognition system has been proposed. The proposed system recognizes the person even in the presence of a partial face information (using either half of the face or using eye, nose and mouth component). Only subset of the face has been used for processing therefore, the computation complexity is also minimized. The theoretical proof to support that half face could also produce the same level of accuracy is driven by applying the principal component analysis on both the full faces and half faces. The time comparison between the full and half faces show that the time complexity of half faces are just the half of the full faces. In the component based face recognition, first the face is fragmented over eye, nose and mouth components, then the Scale Invariant Feature Transform (SIFT) has been applied on the full face and its components. The geometrical location of the extracted features is then estimated. The density estimation has been applied on both face and its components to find feature's distribution. The experiment is performed over 345 public faces which shows that 80% of the face features belong to the union of eye, nose and mouth region. Further the recognition based on these components and full faces shows that the component based face recognition produces the same level of accuracy. Both of the approaches help in fighting against variable illumination condition, change in expression and occlusion.

The face recognition system in robots helps them to recognize persons/criminals and at the same time it also protects them from getting unauthorized access. The face recognition can be intruded by placing face spoofing attacks against them. Hence, it has become mandatory to prevent face recognition system from face spoofing attacks. In the last contribution of this thesis, we have proposed two innovative ideas about anti-face spoofing techniques to detect these attacks at the very initial level. Liveness is a parameter to test the face spoofing attacks. If the genuine person is present while submitting her/his biometric for recognition, the liveness co-efficient will be true otherwise false. In the first approach we have generated some challenges like open eyes, close eyes, open mouth close mouth, etc. and on account of the given challenges we have analyzed the responses. If all the challenges are performed the person is treated as live. Haar Eye and mouth cascade classifier have been used to localize eye and mouth in the face. The second approach exploited face depth information on the face to detect liveness. The attacks presented to spoof a face are either by using video or photo. But all these spoofs have no depth information, while due to the odd even surface of face, there is a different depth value at each point. We have utilized robot's stereo-vision and performed depth analysis on both sources. Later the classification is performed using Principal Component Analysis, Linear Discriminant Analysis and K-Means algorithms.

In a nutshell all the above presented techniques help humanoid to achieve social attention. Each proposed framework is established with rigorous experimental as well theoretical proofs.

## 7.2 Outcomes derived from the thesis

- The visual perception of the eyewitness has been utilized to design the human-robot interaction based framework for criminal identification.
  - This is the first time where prototype of each facial feature has been summarized under one umbrella to design the symbolic facial feature database. Any human face can be re-synthesized using these facial prototypes.
  - The query based model builds a decision tree to club similar facial prototypes based on their attribute values, it helps to localize the criminal face over the similar cluster.
  - The rough set based framework gives its prediction based on their rule library. The rule library is derived from the dataset, therefore, the proposed framework has the ability to handle the uncertainty and the imprecise present in the given visual perception of eyewitness.
- A generalized solution for estimating the calibration matrix between the humanoid camera plane and its end effector has been presented. Three different techniques for estimating the coefficient of calibration matrix has been discussed. The efficiency of each technique has been compared with respect to the number of points used in calibration, time complexity and the error involved in the projection.
- A novel facial symmetry based face recognition system has been proposed to identify the criminal/person only using the half face. A theoretical as well as experimental proof has been established for the proposed system.
- Another experimental analysis on 345 public faces shows that almost 80% of the face features belongs to eye, nose and mouth regions. Hence, a person/criminal can be recognized only on these feature sets. This helps in identifying the criminals in all those cases where their face is occluded with scarf/goggles/beard/mustache.
- The challenge and response based anti-face spoofing detection prevent these humanoid robots from illegal access. The addition of anti-face spoofing mechanism over face recognition system removes the possibility of intrusion in these system.

- Imitating responses against the given challenges requires person involvement, further this dependency has been eliminated using the proposed depth perception based anti-face spoofing detection technique.

### 7.3 Assumptions and Limitations of our Research

The solution proposed in this thesis can be utilized for the criminal identification, sketch drawing, face recognition and anti-face spoofing problems. However, they are based on some assumption and each technique has its limitation and constraints. The assumption kept in designing the solution and the limitation of each technique is discussed below.

- Human perception based criminal identification starts with an assumption that the criminal should have enrolled in the system. If the criminal is not enrolled in the system, then the proposed method would not provide wanted results. The proposed system grouped the matched records in terms of Top 10, Top 20, Top 30 and so on based on the similarity matching between the probe and gallery symbolic database. In designing the system we have also not included the False Positive (FP) into the considered as false positive results are discarded by the user, but the False Negatives (FN) are very crucial. The false negatives are those cases where the criminal is enrolled but the system has not listed her/him in the search results. The false negative could arise when the given perception about the criminal is very much deviated from the actual one. The variance between the probe feature vector and the gallery feature vector is too high.
- The sketch drawing capability of the humanoid robots are based on two things (a) the proper calibration and second (b) the efficient inverse kinematic solution. In order to design the sketch drawing framework we took following assumptions, such as (1) with respect to the end effector, there should be no movement in the z axis, the robot hand is only allowed to move on the planner surface (left, right and forward, backward), (2) end effector is only allowed for the Yaw and Pitch movement, the Roll movement is fixed. Moreover, the proposed approach only talks about the calibration and the inverse kinematics solution of the proposed system. No controller is added in the proposed system. The system utilized the default controller of NAO humanoid robot, while the controller is as important as the above two modules.

- In order to perform facial symmetry and component based face recognitions we made some assumptions. In both the solutions only frontal face information is utilized. In order to fragment the face into two equal regions, we assume that the person is free from any facial disease. The middle point between the eyes is selected as the bisection point, while Haar classifier is used in component based approach to localize the eye, nose and mouth regions. Although the proof is established theoretically and experimentally, the conclusion is drawn only on a single database. Secondly the feature extraction is performed using the principal component analysis and Scale Invariant Feature Transform, other feature extraction techniques should be utilized to generalize the proof.
- The anti-face spoofing mechanism is based on the liveness detection test and the liveness is decided based on the change in facial movement and 3D depth analysis of the face and imposter media. The challenge and response require a maximum change in the facial expression and the proper follow up of the given challenges. Both the approaches work well for the photo and video imposter attacks, however, their reliability is not tested against 3D mask attacks.

## 7.4 Recommendations for future researchers

- There are enormous opportunity in human perception based criminal identification, this thesis established and enlighten a new path of dealing with the criminal identification. The modelling has been performed only on the basis of the decision tree and rough set based approaches. In the near future the other approaches such as Dempster-Shafer theory and fuzzy inference system could be utilized to deal with the imprecision and uncertainty present in the knowledge acquisition module.

- The proper localization of facial components in the face for the component based face recognition is the prime concern. Haar classifier has been utilized to locate the face and facial components, but the Haar classifier suffers from occlusion and illumination effect for the local features. The Active Appearance Model, Active Shape Model, Robust Cascade Pose Regressor could also be utilized in the future to localize the facial macro features as they are robust against occlusion and face shape.

- The two methods challenge-response and 3D depth perception are utilized for the face spoofing detection. The proposed system only discards the photo and video imposter attack, it cannot prevent from the mask attack. The future research could be focused to design the mechanism to handle mask attack.